\documentclass[accepted]{uai2026} 
                        
\usepackage[american]{babel}

\usepackage{natbib} 

\usepackage{mathtools} 
\usepackage{amsthm}
\usepackage{amssymb}
\usepackage{algorithm}
\usepackage{algorithmic}
\usepackage[dvipsnames]{xcolor}
\usepackage{pdflscape}
\usepackage{booktabs} 
\usepackage{subcaption} 
\usepackage{graphicx}
\usepackage{hyperref}
\usepackage[capitalize,noabbrev]{cleveref}
\usepackage{soul}
\usepackage{float}

\theoremstyle{plain}

\theoremstyle{definition}

\theoremstyle{remark}

\newcommand{\algheader}[1]{\item[] \textit{#1}} 
\author[1]{Simon D. Nguyen}
\author[3]{Hayden McTavish}
\author[1]{Kentaro Hoffman}
\author[3]{Cynthia Rudin}
\author[1,2]{Tyler H. McCormick}
\affil[1]{Department of Statistics, University of Washington}
\affil[2]{Department of Sociology, University of Washington}
\affil[3]{Department of Computer Science, Duke University}
  
\begin{document}
\title{REALITrees: Rashomon Ensemble Active Learning for Interpretable Trees}
\maketitle

\begin{abstract}
    Active learning reduces labeling costs by selecting samples that maximize information gain. A dominant framework, Query-by-Committee (QBC), typically relies on \textit{perturbation-based diversity} by inducing model disagreement through random feature subsetting or data blinding. While this approximates one notion of epistemic uncertainty, it sacrifices direct characterization of the plausible hypothesis space. to do so. We propose the complementary approach: \textit{Rashomon Ensembled Active Learning} (\textit{REAL}) which constructs a committee by exhaustively enumerating the Rashomon Set of all near-optimal models. To address functional redundancy within this set, we adopt a PAC-Bayesian framework using a Gibbs posterior to weight committee members by their empirical risk. Leveraging recent algorithmic advances, we exactly enumerate this set for the class of sparse decision trees. Across synthetic and established active learning baselines, REAL outperforms randomized ensembles, particularly in moderately noisy environments where it strategically leverages expanded model multiplicity to achieve faster convergence.
\end{abstract}

\section{Introduction}
\label{sec:Intro}

Acquiring labeled data is a persistent bottleneck in high-stakes machine learning applications, from image labeling \citep{HuijserGemert2017} and sentence classification \citep{ALMQS_Schumann2019}, to verbal autopsy \citep{mccormick2016probabilistic,PPI_VerbalAutopsy}. Active Learning (AL) mitigates this cost by strategically querying samples that maximize information gain, thereby achieving high predictive performance with minimal human supervision \citep{Settles, CohnErrorBased}. A dominant framework for identifying samples to label is Query-by-Committee (QBC) \citep{QBC_Seung, freund1997selective}, which selects instances where a committee of competing hypotheses exhibits high disagreement.

However, the efficacy of QBC hinges on the models within the committee. Standard approaches, such as Random Forests \citep{RandomForest_Breiman2001}, generate committee members via \textit{perturbation-based diversity} by blinding models to specific features, or withholding data through bootstrapping \citep{ElementsStatisticalLearning}. While bootstrapping offers a principled approximation of one notion of epistemic uncertainty via the sampling variance of the empirical risk minimizer, it sacrifices direct characterization of the statistically plausible hypothesis space to do so. Random feature subsetting compounds this further; while it serves as a powerful decorrelation device to improve ensemble generalization, it acts as a stochastic approximation for diversity rather than a principled mechanism for exploring the hypothesis space. We propose the complementary approach: targeting the uncertainty inherent in model multiplicity by exhaustively enumerating near-optimal models and grounding committee disagreement in their structural ambiguity.

True epistemic uncertainty arises when multiple empirically equivalent hypotheses yield conflicting predictions. This is formalized by the \textit{Rashomon Effect} \citep{Breiman2001}, which posits that there exists a multitude of models, the \textit{Rashomon Set}, that explain the data with near-equal accuracy but rely on different structural explanations. Critically, the size of this set has been shown to increase in the presence of label noise \citep{RashomonRatioNoise}. Rather than viewing this expansion as a source of error, we interpret it as an enriched disagreement range that allows our committee to more precisely map the boundaries of the near-optimal hypothesis space.

We hypothesize that an effective active learning strategy should prioritize points that expose structural ambiguity among plausible models, rather than resolving perturbation-based disagreements. To this end, we propose \textbf{R}ashomon \textbf{E}nsembled \textbf{A}ctive \textbf{L}earning (\textbf{REAL}). Unlike standard ensembles, REAL constructs its committee strictly from the Rashomon set, utilizing a probabilistic weighting scheme to aggregate the competing structural hypotheses into a Rashomon-Weighted signal of entropy.

We specifically focus our work on the class of sparse decision trees. This focus is motivated by two factors. First, in the high-stakes domains where active learning is most valuable (e.g., medicine, financial credit reporting, or criminal recidivism prediction), interpretability is non-negotiable; users require models that are not only accurate but also transparent \citep{RashomonSimplerMLModels}. Second, recent algorithmic advances \citep{ExploringRashomonTrees, Arslan2024} allow for the exact and exhaustive enumeration of the Rashomon set for trees. 

Our contributions are as follows: (1) We introduce a querying methodology that replaces perturbation-based diversity with structural diversity, defining the committee strictly as members of the Rashomon set. (2) We integrate a PAC-Bayesian weighting scheme utilizing a Gibbs posterior to resolve functional redundancy within this near-optimal set. (3) We demonstrate that our framework thrives in moderately noisy environments by strategically leveraging expanded pattern diversity. Unlike randomized ensembles that may chase local inconsistencies, our method utilizes the expanded Rashomon set to isolate robust structural patterns, leading to faster and more reliable convergence under label noise. (4) We validate REAL across 20 real-world benchmarks
\footnote{To ensure reproducibility, our codebase is available at \href{https://github.com/thatswhatsimonsaid/RashomonActiveLearning}{https://github.com/thatswhatsimonsaid/RashomonActiveLearning}}. Our empirical results demonstrate that by grounding query selection in a PAC-Bayesian weighted version space of the Rashomon set, REAL offers a high-precision alternative to stochastic ensembles that is uniquely robust to the structural ambiguities of real-world data.

\section{Related Work}
\label{sec:RelatedWork}

The core of active learning lies in quantifying uncertainty to distinguish informative signal from uninformative redundancy. Established methods include uncertainty sampling (e.g., Entropy, Least Confidence) \citep{Uncertainty_LewisGale1994}, Expected Model Change (EMC) \citep{Cai2013}, and disagreement-based methods like Query-by-Committee (QBC) \citep{freund1997selective}. In the context of decision trees, QBC is typically implemented using Random Forests or Bagging \citep{DiverseEnsembleAL, QBB_Abe}. While effective, these strategies rely on stochastic randomization that can become unreliable under model misspecification as the committee fails to account for the learner's inherent bias \citep{Burbridge}.

To address this vulnerability, we propose a more principled foundation for committee construction. 
Rather than relying on the randomized bootstrap space, our work constrains the committee to the statistically plausible $\epsilon$-version space defined by the Rashomon set. This shift anchors QBC in the theoretical work of version spaces by \cite{Mitchell_1992_VersionSpace} as REAL queries prioritize resolving genuine model ambiguity over the variance of random perturbations. While the classical version space of \cite{Mitchell_1992_VersionSpace} is restricted to hypotheses perfectly consistent with the training data, we define the $\epsilon$-version space as the generalization of this concept to the Rashomon set to encompass all hypotheses within an $\epsilon$-radius of the empirical risk minimizer. This allows our learner to account for model uncertainty even in the presence of noise or misspecification.

The existence of multiple near-optimal models, termed the \textit{Rashomon Effect} by \citet{Breiman2001}, has gained renewed attention in the context of fairness and interpretability. Recent work on predictive multiplicity \citep{PredictiveMultiplicityMarx, PredictiveMultiplicity_WatsonDaniels2023} demonstrates that models with identical accuracy can assign vastly different predictions to individual users. While previous studies have studied predictive multiplicity as a source of algorithmic instability \citep{ModelMultiplicity_Black2022, RashomonGB}, REAL utilizes a Gibbs posterior supported by the Rashomon set to transform this multiplicity into a signal for information acquisition. By treating multiplicity as a signal rather than a nuisance, we utilize a PAC-Bayesian weighting scheme to identify regions of epistemic ambiguity where near-optimal hypotheses yield conflicting predictions. Critically, the size of the Rashomon set has been shown to increase in the presence of label noise \citep{RashomonRatioNoise}. Within the REAL framework, we treat this expansion as a source of pattern diversity for our QBC committee \citep{DiverseEnsembleAL}, using the PAC-Bayesian weighting to isolate the most robust structural signals from the resulting larger committee.

Enumerating the full Rashomon set is computationally intensive for many model classes. Pioneering approaches like TreeFarms \citep{ExploringRashomonTrees} established the methodology for extracting these sets using dynamic programming for sparse decision trees. However, active learning requires re-calculating this set iteratively as new labels are acquired. To facilitate this online setting, our work utilizes the recent SORTD algorithm \citep{Arslan2024}, which presents several computational speedups for sparse trees. This efficiency gain allows us to update the exact Rashomon set at every step of the active learning loop, scaling the robust analysis of TreeFarms to the iterative active learning setting. 

Our weighting strategy is grounded in the framework of PAC-Bayesian learning \citep{PAC_Bayesian1999, PAC_Bayesian_Cantoni} where we utilize a Gibbs posterior to navigate the structural uncertainty of the hypothesis space. Unlike traditional Bayesian Model Averaging \citep{BMAHoeting} which requires integrating over an often intractable hypothesis space, we adopt a PAC-Bayesian perspective. We utilize the $\epsilon$-Rashomon set as the empirical support for a Gibbs posterior, capturing the most significant risk-weighted mass of the $\epsilon$-version space without the computational burden of a full Bayesian treatment. By applying a Gibbs distribution to the members of the Rashomon set, REAL implements an approximated Bayesian query strategy. 

This approach aligns with the principle of ``Occam's Window'' of \cite{MadiganOccamWindow}, which suggests that models significantly less plausible than the best-supported model should be discarded to avoid ``blurring'' the structural signal \citep{AdamEddieKentaroPaper, AparaRPS}. By constraining our distribution to the Rashomon set, REAL avoids the dilution inherent in full BMA, focusing the learner's attention on the genuine structural contradictions between high-quality hypotheses that represent true epistemic uncertainty. In this regime, the weighted vote entropy reflects the posterior predictive uncertainty within the empirically plausible $\epsilon$-version space, allowing the learner to prioritize queries that resolve genuine structural contradictions rather than stochastic variance.

\section{Methodology}
\label{sec:Methodology}

We consider the standard pool-based active learning setting. Let $\mathcal{X} \in \mathbb{R}^p$ denote the feature space and $\mathcal{Y}$ the label space. We are given a small initial labeled dataset $\mathcal{D}_{tr} = \{(\mathbf{x}_i, y_i)\}_{i=1}^n$ and a large unlabeled pool $\mathcal{U} = \{\mathbf{x}_j\}_{j=1}^N$. Our objective is to learn a hypothesis $h: \mathcal{X} \to \mathcal{Y}$ from a hypothesis class $\mathcal{H}$ that minimizes generalization risk using a minimal number of queries from $\mathcal{U}$. In this work, $\mathcal{H}$ specifically denotes the class of sparse binary decision trees. While our framework and codebase is designed for multiclass labels, we focus our empirical evaluation on the class of sparse binary decision trees.

\subsection{The Rashomon Set of Sparse Trees}
\label{subsec:RashomonSet}

In discrete hypothesis spaces like decision trees, the empirical risk minimizer is rarely unique. Instead, there exists a multiplicity of models that fit the training data near-optimally. This phenomenon is formalized as the \textit{Rashomon Effect} \citep{Breiman2001}.

Following the formulation in the optimal sparse decision tree literature \citep{NEURIPS2019_ac52c626, Arslan2024}, we define the objective function for a tree $h$ as a weighted combination of misclassification loss and leaf-count complexity 
\begin{equation}
    \mathcal{L}(h, \mathcal{D}_{tr}) = \frac{1}{n} \sum_{i=1}^n \mathbb{I}(h(\mathbf{x}_i) \neq y_i) + \lambda \cdot \text{Leaves}(h)
\end{equation}
Let $h^*$ denote the optimal tree that minimizes this objective function: $h^* = \text{argmin}_{h \in \mathcal{H}} \mathcal{L}(h, \mathcal{D}_{tr})$.

We define the \textbf{$\epsilon$-Rashomon Set}, $\hat{\mathcal{R}}(\epsilon, \mathcal{D}_{tr})$, as the set of all trees with an objective value within a multiplicative factor $1 + \epsilon$ of the optimal model $h^*$ given the training dataset $\mathcal{D}_{tr}$:
\begin{equation}
    \hat{\mathcal{R}}(\epsilon, \mathcal{D}_{tr}) = \{h \in \mathcal{H} : \mathcal{L}(h, \mathcal{D}_{tr}) \leq (1 + \epsilon) \mathcal{L}(h^*, \mathcal{D}_{tr}) \}
\end{equation}
Here, $\epsilon \ge 0$ is the \textit{Rashomon Threshold}. When $\epsilon=0$, the set contains only the exact optimal trees, corresponding to the classical version space described by \cite{Mitchell_1992_VersionSpace}. As $\epsilon$ increases, the set expands to recover the effective $\epsilon$-version space of the problem, retaining diverse models that are indistinguishable based on training error alone. For brevity, we denote the set simply as $\hat{\mathcal{R}}(\epsilon)$ or $\hat{\mathcal{R}}$ when the training data $\mathcal{D}_{tr}$ is clear from context.

\subsection{Perturbed vs. Structural Diversity}\label{subsec:Diversity}

In the QBC literature, diversity is essential as a committee must represent the full range of disagreement that is actually possible given the current labeled data \citep{DiverseEnsembleAL}. A committee of identical or highly similar models will offer redundant predictions, failing to identify regions where other equally plausible hypotheses might disagree. The overwhelmingly common approach is to produce this diversity synthetically through resampling procedures \citep{QBC_Seung, QBB_Abe}. While these methods are theoretically grounded as approximations of the sampling distribution, this approach relies on the stochasticity of the (re)sampling process to induce and explore diverse committee members.

We take a different approach: rather than relying on stochastic perturbations to find diverse models, we enumerate the Rashomon set to exhaustively identify the entire space of hypotheses with high predictive performance. By leveraging the Rashomon effect, where many high-quality models rely on fundamentally different structural explanations for the same data, we provide a committee that represents the inherent model ambiguity of the problem.

\subsection{The REAL Algorithm}
\label{subsec:REAL}

To ground query selection in genuine model ambiguity, we propose \textbf{R}ashomon \textbf{E}nsembled \textbf{A}ctive \textbf{L}earning (\textbf{REAL}). Unlike standard ensembles that rely on a fixed number of randomized members, REAL utilizes the membership of the $\epsilon$-Rashomon set $\hat{\mathcal{R}}(\epsilon, \mathcal{D}_{tr})$ to characterize the $\epsilon$-version space. By requiring that every committee member satisfies the Rashomon bound, we ensure that any observed disagreement represents a conflict between two plausible structural hypotheses grounded in the same evidence.

While the Rashomon set provides a comprehensive view of the $\epsilon$-version space, naively treating every member with equal weight can lead to structural saturation, where minor structural variants of the same hypothesis dominate the ensemble vote. This leads to a loss of effective diversity, which is critical for the success of QBC as a diverse ensemble is necessary to accurately capture the boundaries of the $\epsilon$-version space \citep{DiverseEnsembleAL}.

Although there are many ways to prune a set of trees to a unique set of representative trees such as by agglomerative clustering or predictive equivalence \citep{PredictiveEquivalence}, pruning ignores the relative density of the $\epsilon$-version space. By reducing a cluster of similar models to a subset representative, the learner loses the strength of evidence signal that a weighted ensemble naturally preserves. In a PAC-Bayesian interpretation, the density of structurally distinct but functionally similar models suggests a region of significant empirical risk-weighted mass. A weighted ensemble preserves this density, whereas pruning treats all unique decision boundaries as equally likely regardless of how many near-optimal structures support them. Rather than discarding models, we adopt a PAC-Bayesian weighting scheme to ensure that the query strategy is informed by the structural multiplicity inherent in the Rashomon set.

We define a Gibbs posterior supported on the membership of $\hat{\mathcal{R}}(\epsilon, \mathcal{D}_{tr})$. Following the literature on PAC-Bayesian learning \citep{PAC_Bayesian_Cantoni}, we assign each tree $h \in \hat{\mathcal{R}}(\epsilon)$ a Gibbs weight $w_h$ based on its empirical risk:
\begin{equation}\label{eq:GibbsWeight}
    w_h = \frac{\exp(-\beta \cdot \mathcal{L}(h, \mathcal{D}_{tr}))}{\sum_{h' \in \hat{\mathcal{R}}} \exp(-\beta \cdot \mathcal{L}(h', \mathcal{D}_{tr}))}
\end{equation}
where $\beta \ge 0$ is an inverse temperature parameter that controls the ``peakedness'' of the distribution. This weighting scheme naturally addresses functional redundancy: trees that fit the training evidence more precisely have a greater influence on the query decision, while the threshold $\epsilon$ ensures that the ensemble maintains sufficient diversity to explore the $\epsilon$-version space.

Based on the choice of weight distribution, we define two specific implementations of the framework: (i) \textbf{UNREAL} (Uniform REAL), a baseline utilizing a maximum-entropy prior where $w_h = 1/|\hat{\mathcal{R}}|$ for all $h \in \hat{\mathcal{R}}$, and (ii) \textbf{BREAL} (Bayesian REAL), our primary methodology which leverages the Gibbs posterior to resolve functional redundancy and prioritize models with higher empirical support.

Given the weighted ensemble, we select the query instance $\mathbf{x}^*$ that maximizes the \textit{Weighted Vote Entropy}:
\begin{equation}
    P(y \mid \mathbf{x}, \hat{\mathcal{R}}) = \sum_{h \in \hat{\mathcal{R}}} w_h \cdot \mathbb{I}(h(\mathbf{x}) = y)
    \end{equation}
    \begin{equation}
    \mathbf{x}^* = \underset{\mathbf{x} \in \mathcal{U}}{\text{argmax}} \left[ - \sum_{y \in \mathcal{Y}} P(y \mid \mathbf{x}, \hat{\mathcal{R}}) \log P(y \mid \mathbf{x}, \hat{\mathcal{R}}) \right]
\end{equation}
This formulation allows for a continuous measure of disagreement that respects the relative likelihood of each structural hypothesis.

To quantify the richness of the $\epsilon$-version space during active learning, we calculate the Effective Committee Size (ECS):
\begin{equation}\label{eq:ECS}
    ECS(\hat{\mathcal{R}}) = \exp \left\{ -\sum_{h \in \hat{\mathcal{R}}} w_h \ln w_h \right\}
    \end{equation}
The ECS provides a diagnostic of model certainty: a high ECS indicates a broad, diverse $\epsilon$-version space where many theories are equally plausible, while an ECS approaching $1.0$ signals a collapse of the $\epsilon$-version space onto a single dominant hypothesis.

REAL targets points where the set of all near-optimal models exhibits maximal disagreement. As new labels $y^*$ are acquired, the Rashomon set is updated and the additional information prunes hypotheses inconsistent with the observed data. This process systematically narrows the $\epsilon$-version space, concentrating the learner's attention on the structural features of the hypothesis manifold that remain empirically plausible. For implementation details, the complete REAL procedure is formally defined in Algorithm \ref{alg:REAL} (Appendix \ref{Appendix:Methods}).

\section{The Boundaries of Precision: Misspecification and Noise}
\label{sec:StressTests}

To understand the robustness and limitations of Rashomon-based active learning, we systematically explore the impact of model misspecification and label noise.

\subsection{Experimental Design}
We utilize a data generating process based on a 20-dimensional binary feature space. The ground truth is constructed as a mixture of two labeling mechanisms: a tree-based XOR pattern $y_{\text{tree}} = X_0 \oplus X_1$ and a linear threshold model $y_{\text{linear}} = I[\mathbf{w}^T\mathbf{X} > 0]$, where $w_j \sim \mathcal{N}(0,1)$.

We control model misspecification via mixture parameter $\alpha$ for a Bernoulli formulation:
\begin{equation}
    Y_i = S_i \cdot y_{\text{linear},i} + (1 - S_i) \cdot y_{\text{tree},i}, \quad S_i \sim \text{Bernoulli}(\alpha)
\end{equation}
This setup creates a deliberate ``needle-in-a-haystack'' problem specifically designed to challenge the stochastic construction of standard ensembles. In a high-dimensional space where only two features are informative and the remaining 18 act as independent noise, the random feature subsetting and bootstrapping inherent to QBC-RF often produce committee members that miss the structural signal entirely by chance. In contrast, REAL utilizes an exhaustive search of the $\epsilon$-version space to identify the sparse set of models that remain consistent with the empirical evidence, allowing it to isolate the informative features more efficiently. As $\alpha \to 1$, however, the committee members of both REAL and QBC-RF become fundamentally misspecified as individual estimators struggle to approximate the diagonal linear boundary with axis-aligned splits.

In a separate experimental regime, we investigate the impact of stochastic corruption by applying symmetric label noise, flipping a fraction $\phi \in [0.05, 0.45]$ of labels. Our motivation for this study is grounded in the work of \citet{RashomonRatioNoise}, who demonstrate that increased noise fundamentally expands the Rashomon set and increases \textit{pattern diversity}, the average difference in predictions among near-optimal models. This experiment allows us to test whether the active learner can distinguish the underlying structural signal from the combinatorial explosion of noise-induced hypotheses that emerge as the $\epsilon$-version space expands.

\subsection{Experimental Protocol}
\label{subsec:Protocol}

\begin{figure*}[t]
    \centering
    \centerline{\small \textbf{Label Noise Sensitivity} (Fixed Misspecification $\alpha=0$)}
    \vspace{0.2em}
    \begin{subfigure}[b]{0.24\textwidth}
        \centering
        \includegraphics[width=\linewidth]{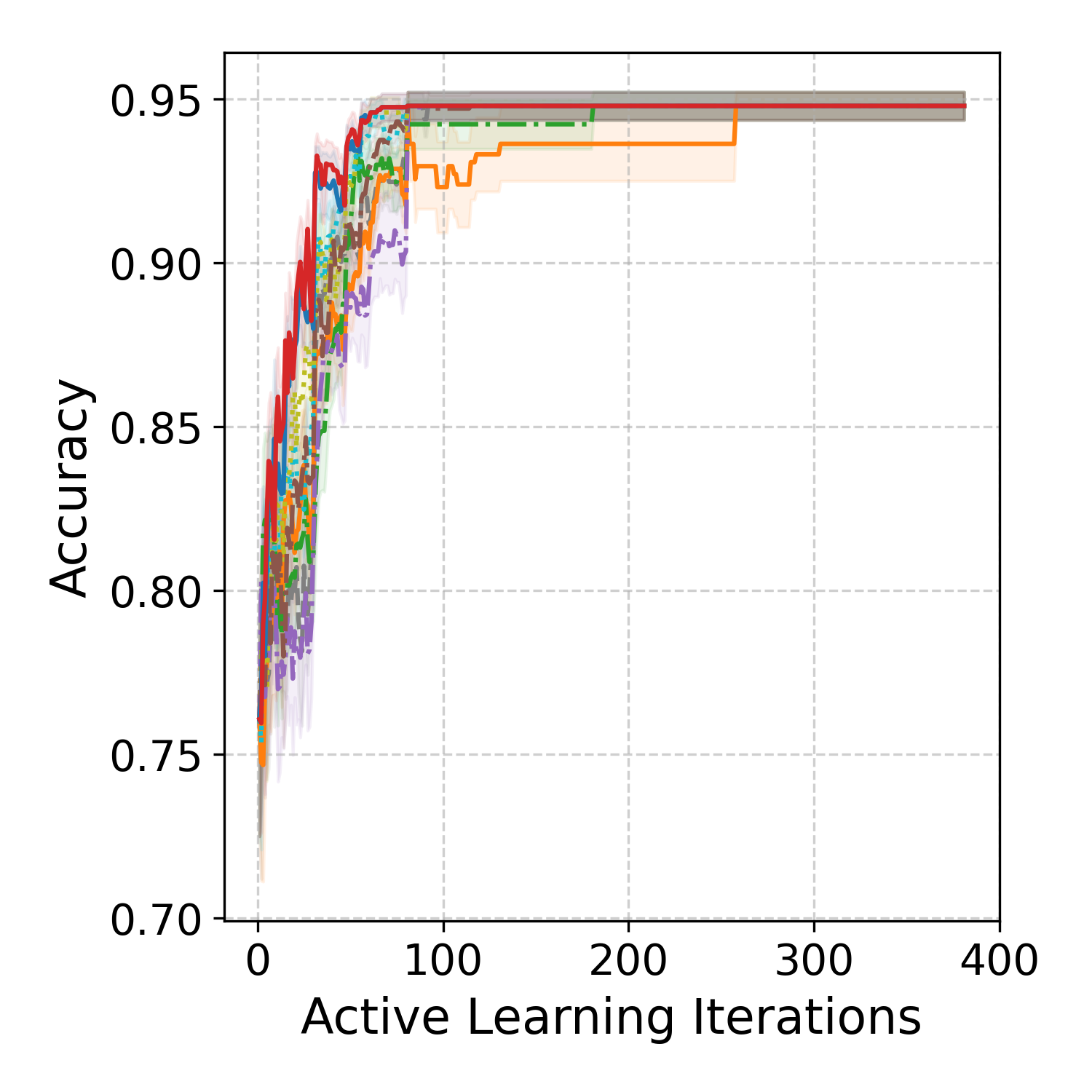}
        \caption{Noise ($\phi=0.05$)}
    \end{subfigure}
    \hfill
    \begin{subfigure}[b]{0.24\textwidth}
        \centering
        \includegraphics[width=\linewidth]{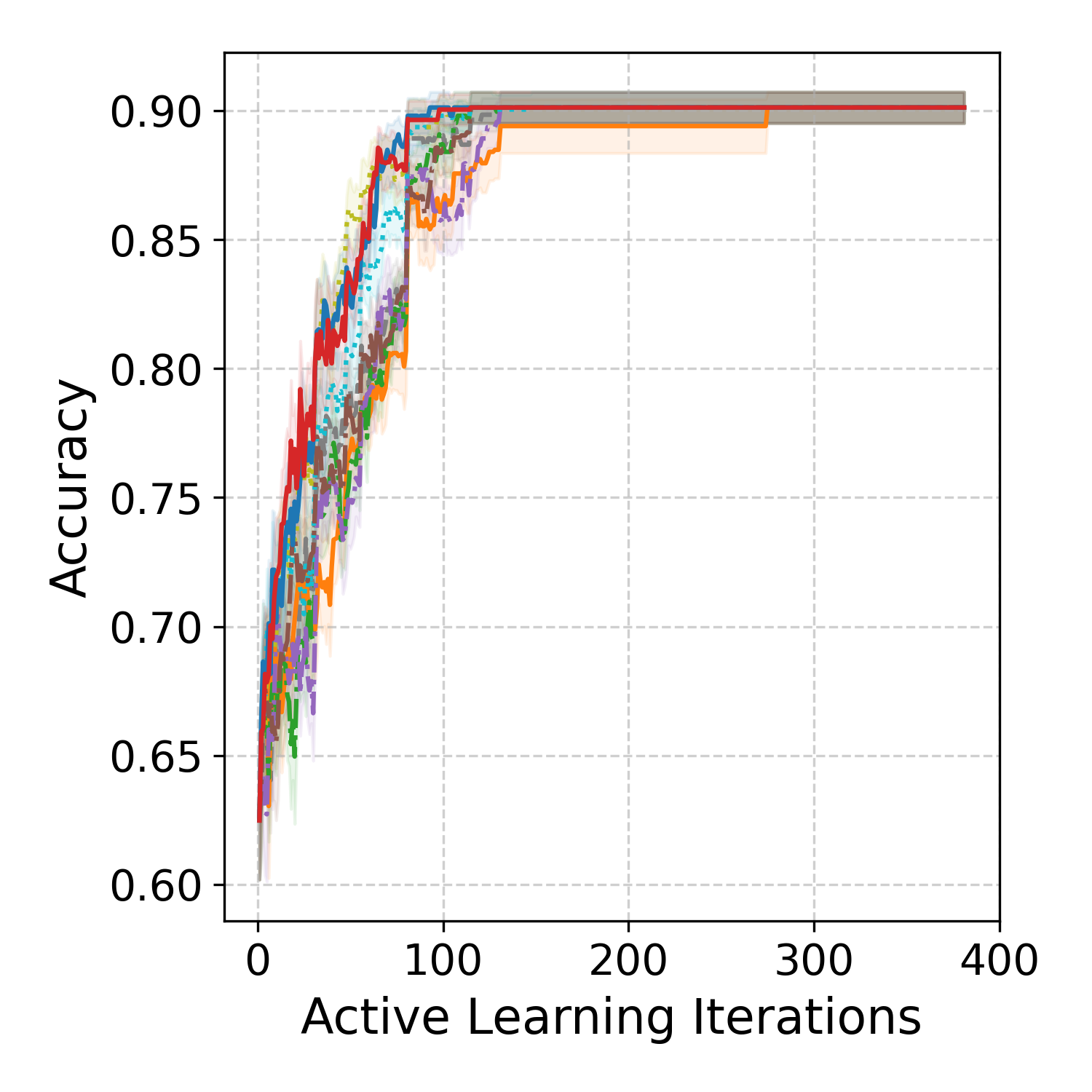}
        \caption{Noise $\phi=0.10$}
    \end{subfigure}
    \hfill
    \begin{subfigure}[b]{0.24\textwidth}
        \centering
        \includegraphics[width=\linewidth]{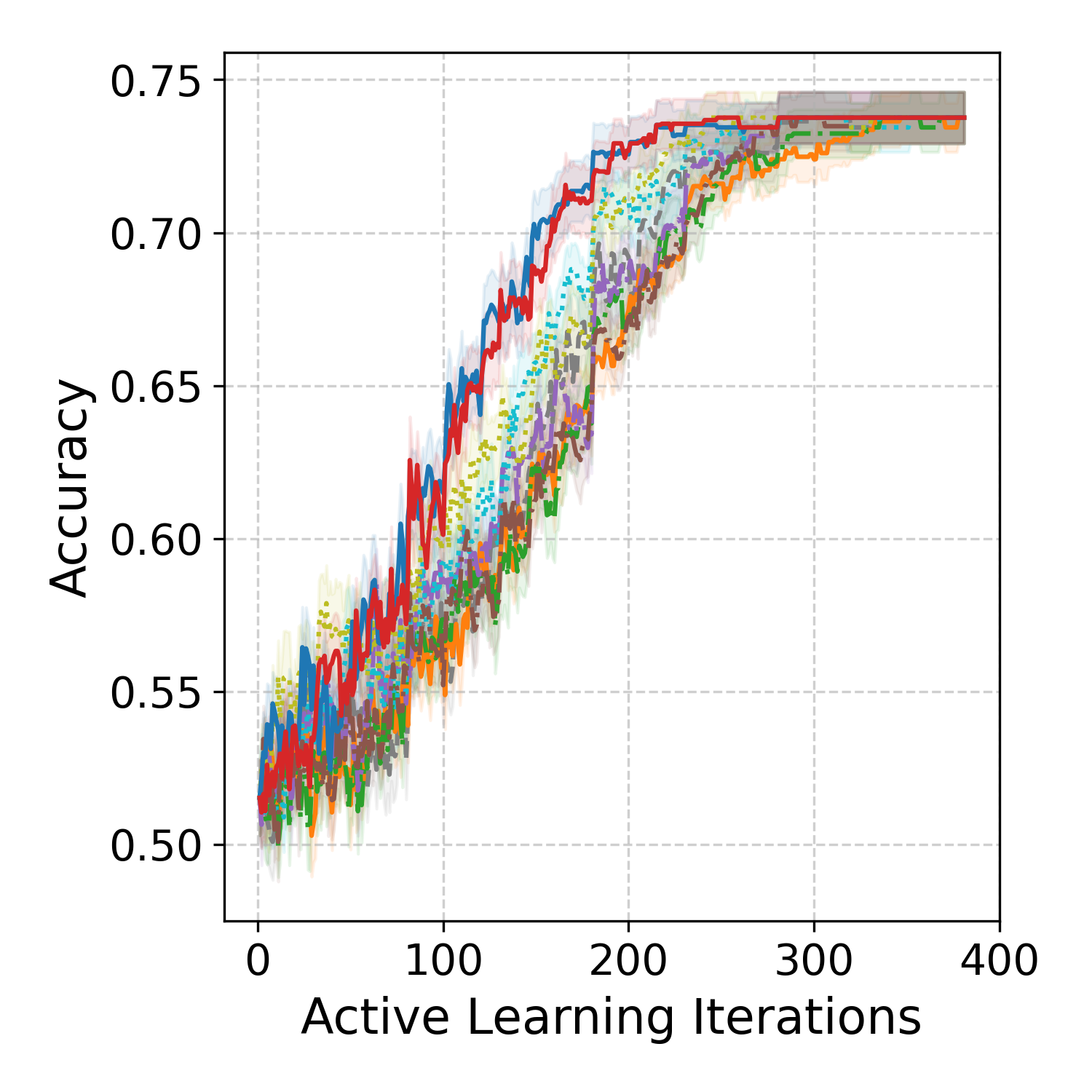}
        \caption{Noise $\phi=0.25$}
    \end{subfigure}
    \hfill
    \begin{subfigure}[b]{0.24\textwidth}
        \centering
        \includegraphics[width=\linewidth]{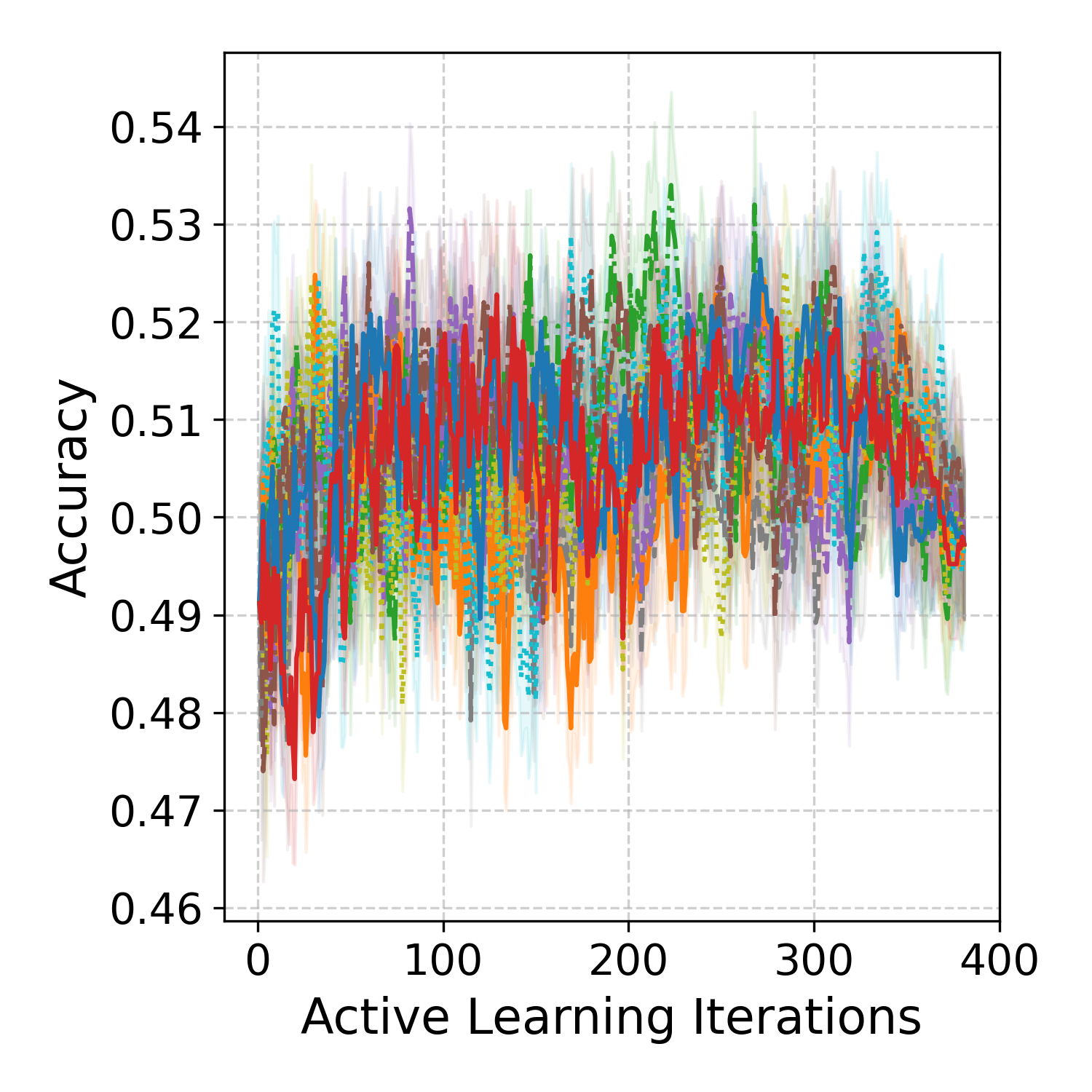}
        \caption{Noise $\phi=0.45$}
    \end{subfigure}

    \vspace{0.8em}

    \centerline{\small \textbf{Model Misspecification} (Fixed Noise $\phi=0$)}
    \vspace{0.2em}
    \begin{subfigure}[b]{0.24\textwidth}
        \centering
        \includegraphics[width=\linewidth]{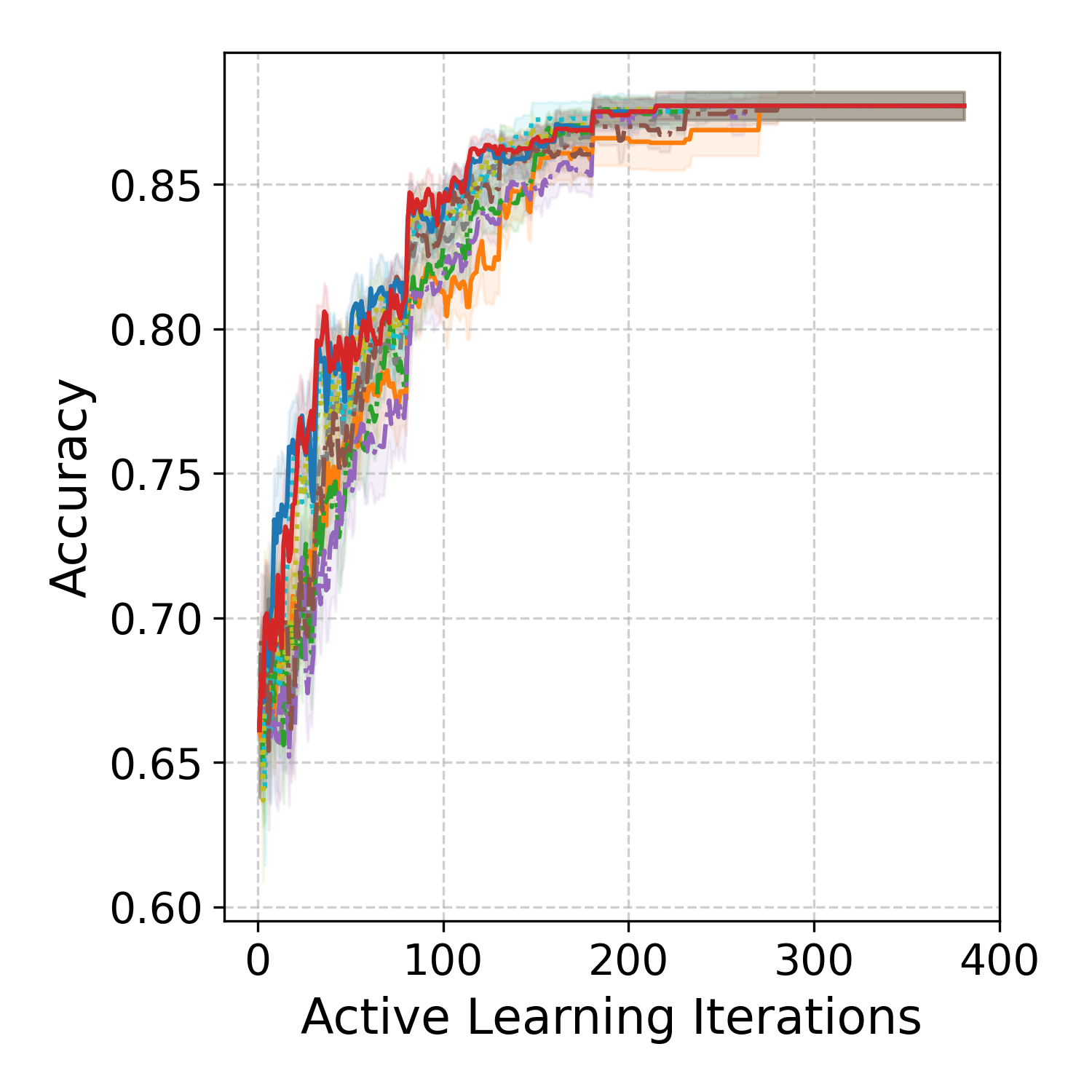}
        \caption{Misspecification $\alpha=0.25$}
    \end{subfigure}
    \hfill
    \begin{subfigure}[b]{0.24\textwidth}
        \centering
        \includegraphics[width=\linewidth]{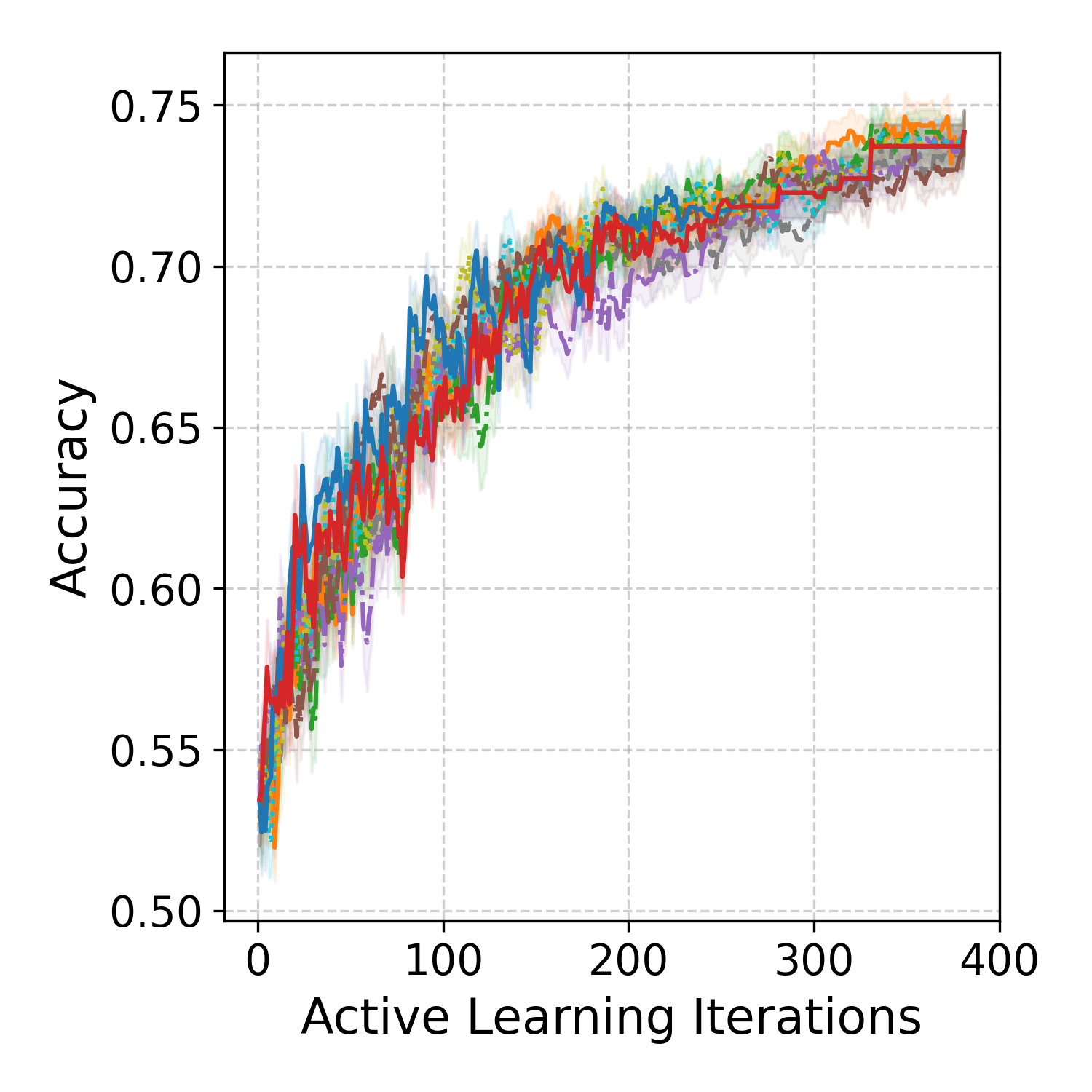}
        \caption{Misspecification $\alpha=0.50$}
    \end{subfigure}
    \hfill
    \begin{subfigure}[b]{0.24\textwidth}
        \centering
        \includegraphics[width=\linewidth]{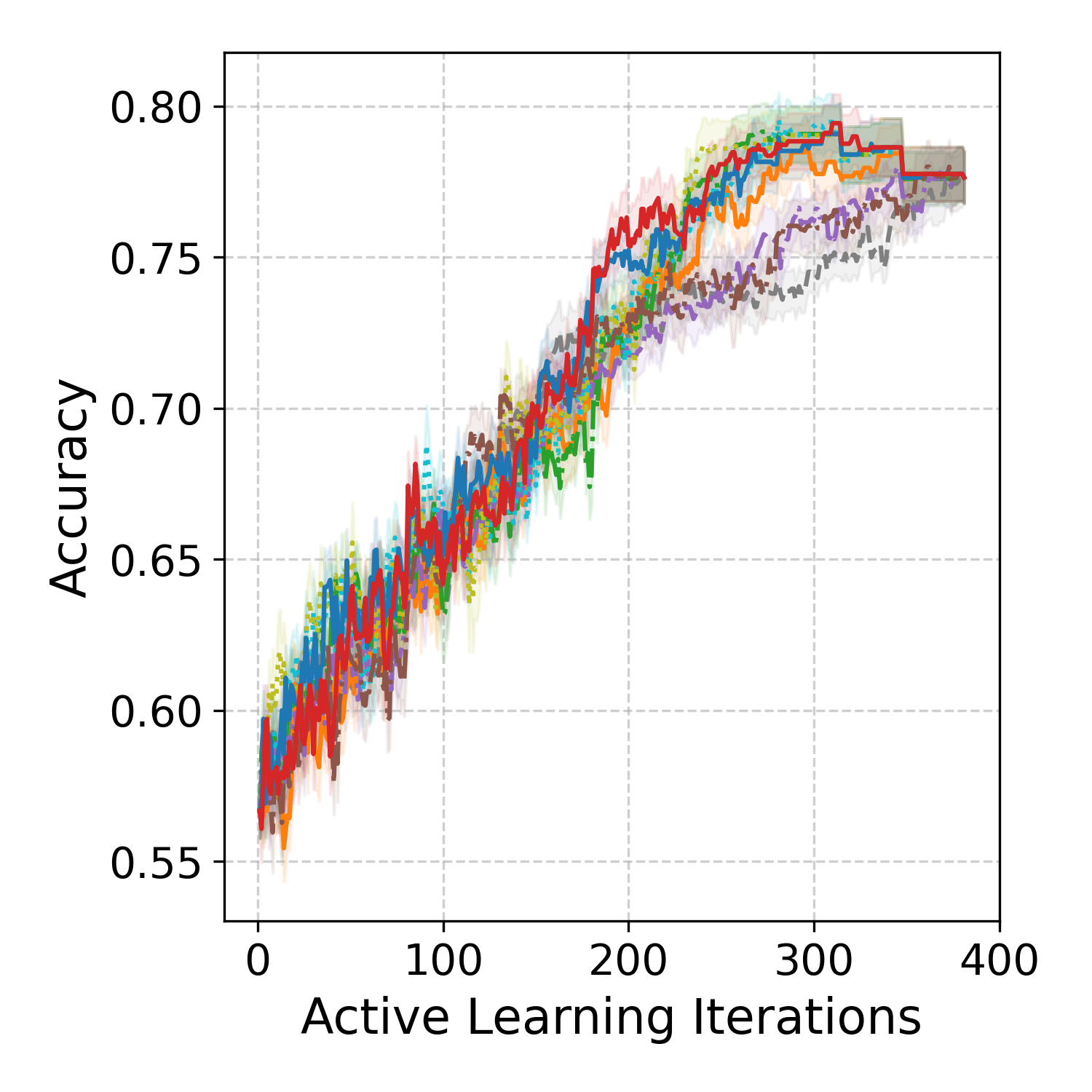}
        \caption{Misspecification $\alpha=0.75$}
    \end{subfigure}
    \hfill
    \begin{subfigure}[b]{0.24\textwidth}
        \centering
        \includegraphics[width=\linewidth]{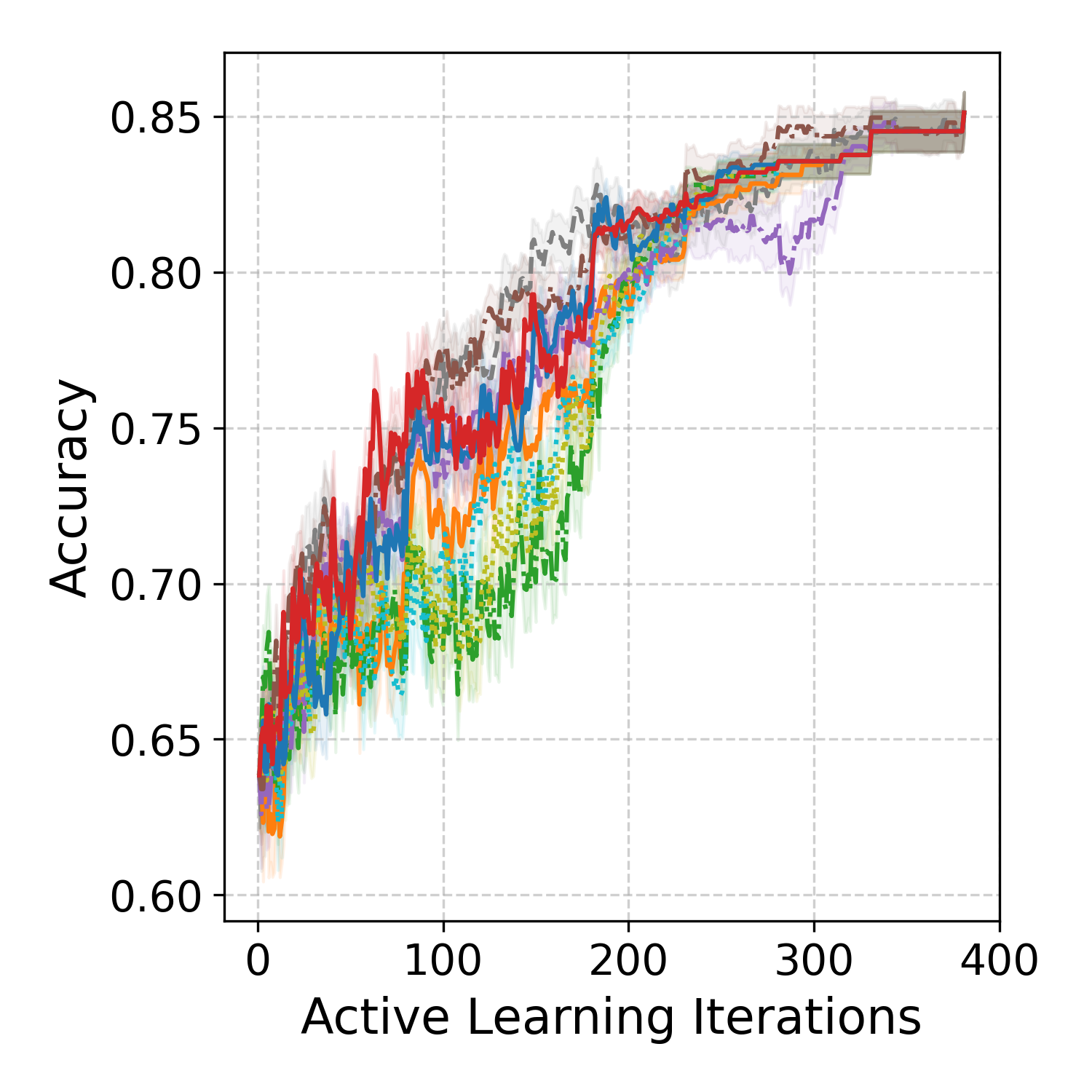}
        \caption{Misspecification $\alpha=1.00$}
    \end{subfigure}

    \vspace{0.8em}
    \centering
    \includegraphics[width=0.60\linewidth]{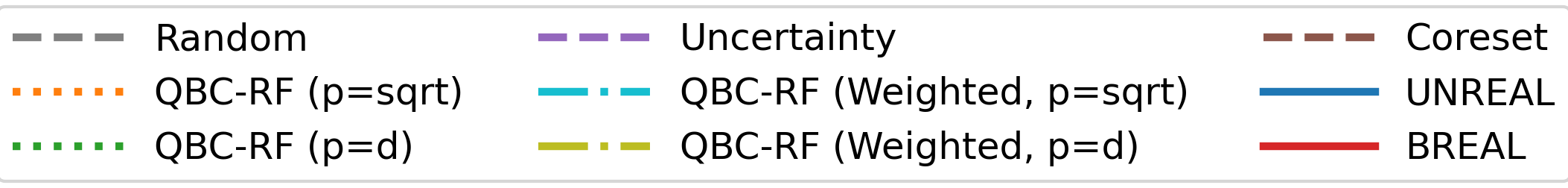}

    \caption{(Top) REAL leverages noise-induced version-space diversity to maintain superior performance, with its advantage widening until extreme noise floors ($\phi=0.45$) equalize all strategies. (Bottom) Under total misspecification ($\alpha=1.0$), all tree-based QBC strategies underperform relative to Passive Sampling. However, BREAL and UNREAL demonstrate unique resilience, outperforming the other QBC methods and maintaining early-stage efficiency despite the structural misalignment.}
    \label{fig:SyntheticGrid_Accuracy}
\end{figure*}

Experiments used $N=25$ independent replications with randomized initial seeds. For each run, we hold out $20\%$ of the observations as a static test set to evaluate predictive performance at each iteration. The remaining $80\%$ constitutes the working pool from which the learner begins with a stratified initial set of 20 samples.

To distinguish prediction from acquisition, we define two model roles: the \textit{Predictor} (maximum depth 5, $\lambda = 0.001$) is used for final test-set evaluation, while the \textit{Selector} characterizes the $\epsilon$-version space for query selection. We tune the \textit{Selector} over a restricted depth to prioritize robust structural patterns over aleatoric noise. By restricting depth ($d \in \{3, 5\}$) and increasing complexity penalties ($\lambda \in \{0.01, 0.001\}$), we create a structural bottleneck that forces the learner to resolve global topological ambiguities rather than exhausting the budget on local inconsistencies that deeper, unregularized models might overfit.

We calibrate the \textit{Selector} on the initial 20-sample pilot set $\mathcal{D}_{tr}^{(0)}$ to ground its parameters: (i) depth $d \in \{3, 5\}$ and complexity $\lambda \in \{0.01, 0.001\}$ are selected via leave-one-out cross-validation; (ii) the Rashomon radius $\epsilon$ is iteratively expanded in 0.05 increments until $|\hat{\mathcal{R}}| \ge 2$ (the floor for committee-based disagreement \citep{QBC_Seung}) and (iii) $\beta$ is optimized via grid search over $\{1.0, \dots, 500.0\}$ to maximize the Area Under the Precision-Recall Curve (AUPRC) between vote entropy and empirical classification errors. This ensures the uncertainty signal is a calibrated proxy for epistemic ambiguity. These optimal values are applied to all \textit{Selector} models to ensure a fair comparison.

Comparison methods and hyperparameters are selected to isolate version-space precision from perturbation-based diversity. We include two variants of QBC-RF utilizing feature subsets of size $m=\sqrt{d}$ and $m=d$ both with $100$ base learners. 
To control for the impact of weighting, we also apply both Uniform and PAC-Bayesian weighting to QBC-RF. This isolates whether the benefits of the Gibbs posterior are universal across ensemble types or specifically synergistic with the structural geometry of the Rashomon set. 

Within the REAL framework, we evaluate two distinct configurations to isolate the influence of version-space weighting from the underlying topology. Our primary methodology, BREAL, applies the Gibbs posterior (Equation \ref{eq:GibbsWeight}) to the committee, weighting each member's vote by its empirical risk to resolve functional redundancy within the $\epsilon$-version space. In contrast, UNREAL serves as a structural baseline that assumes a uniform distribution over the hypothesis space, treating all near-optimal hypotheses as equally informative. By assigning equal weight to all near-optimal hypotheses, it allows us to quantify the performance gains specifically attributable to the PAC-Bayesian weighting scheme.

While our primary objective is to evaluate REAL against QBC-RF, we include Uncertainty Sampling \cite{Settles} and Coreset Diversity \cite{CoresetPaper} to benchmark our results against two fundamental pillars of active learning. Uncertainty Sampling uses least-confidence scoring on an optimal tree. For Coreset, we implement the greedy $k$-center algorithm using Euclidean distance in the feature space to ensure geometric coverage. These baselines ensure that the Rashomon-based approach provides a meaningful improvement over standard industry heuristics. An overview of methods is provided in Table \ref{tab:strategy_reference} of Appendix \ref{Appendix:Methods}.

\subsection{Synthetic Results}

\begin{figure*}[!htp]
    \centering
    \includegraphics[width=\textwidth]{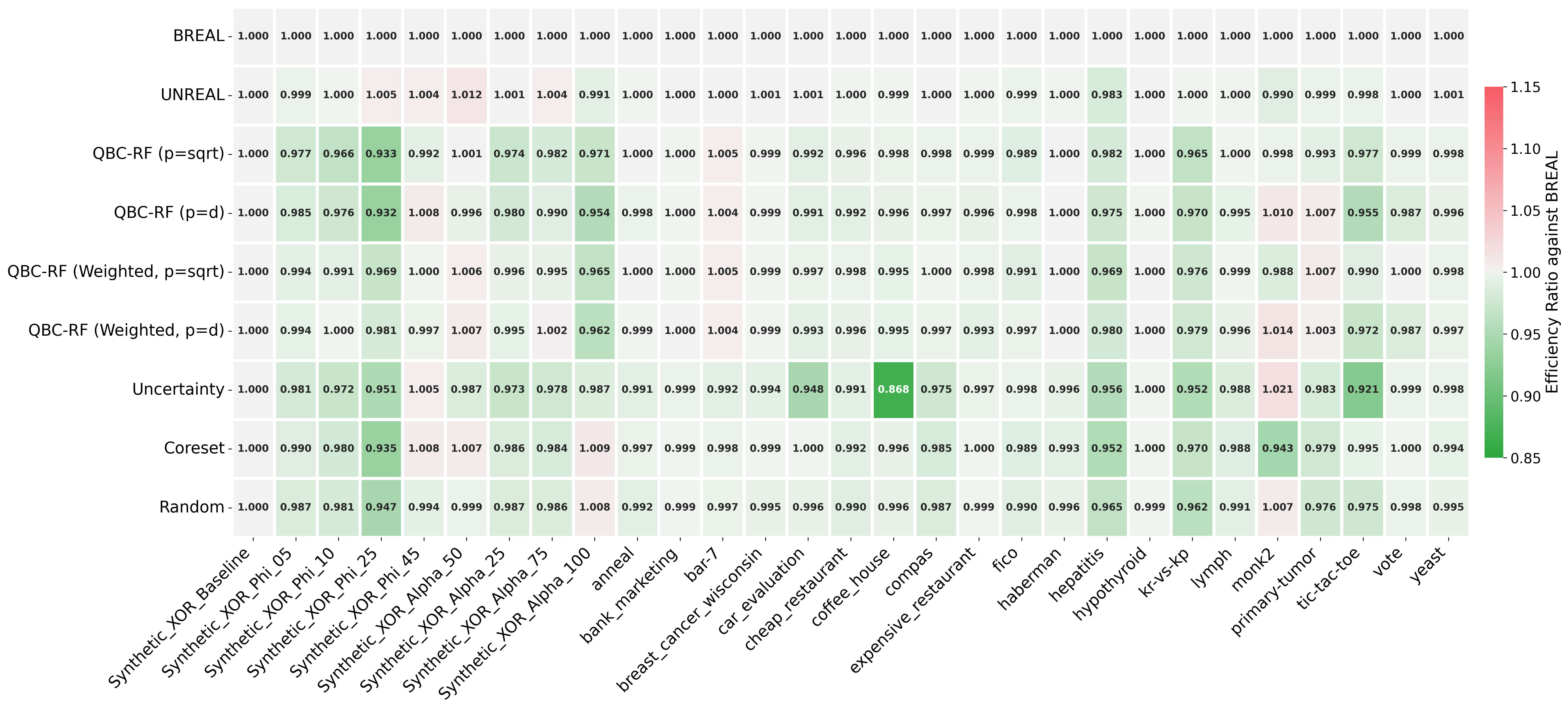}
    \caption{\textbf{Active Learning Efficiency Heatmap ($K=0.7$)}. This figure displays the Efficiency Ratio ($\rho$) of standard baselines relative to REAL, with values below $1.0$ (\textbf{Green}) indicating superior sample efficiency for our method. The results demonstrate that REAL reaches peak predictive performance significantly faster than randomized ensembles.}
    \label{fig:AUC_Accuracy}
\end{figure*}

\textbf{Misspecification Gap.} Our results reveal a nuanced relationship between version-space precision and model misspecification. In the settings of low misspecification ($\alpha =0.25$), REAL (both versions) demonstrates superiority (see Figures \ref{fig:SyntheticGrid_Accuracy} and \ref{fig:AUC_Accuracy}), reaching a higher accuracy ceiling while standard ensembles struggle to identify the sparse structural signal.

As we introduce misspecification ($\alpha = \{0.50,0.75, 1.00\}$), the performance gap narrows, yet BREAL and UNREAL demonstrate unique resilience. In these regimes, the Rashomon set effectively becomes a collection of "equally misspecified experts." While REAL’s committee becomes hypersensitive to local misalignments, treating every structural disagreement as a reason to query a new point, this exhaustive search ensures that the learner identifies the most plausible (albeit biased) solution faster than QBC-RF baselines. In contrast, the algorithmic variance of QBC-RF acts as a structural buffer. By subsetting features and bootstrapping data, individual trees become diverse across different subsets of noise, allowing the ensemble average to effectively smooth over the geometric confusion of the misspecification.

The advantage of this structural search is most apparent under total misspecification ($\alpha = 1.0$). While the smoothing effect of QBC-RF eventually stabilizes, it requires significantly more samples to resolve. In contrast, REAL (both BREAL and UNREAL) consistently outperforms QBC-RF baselines by identifying stable, near-optimal decision boundaries much earlier in the learning process. This indicates that even when the hypothesis space cannot perfectly represent the target function, an exhaustive search of the Rashomon set is a more efficient path to the accuracy ceiling than relying on the stochastic variance of feature subsetting or bootstrapping.

\textbf{Leveraging Noise.} REAL’s performance advantage widens with label noise ($\phi$). While standard randomized ensembles exhibit significant instability at $\phi = 0.25$, REAL reaches its accuracy plateau nearly 100 iterations earlier.

We attribute this to the fundamental relationship between noise and version-space geometry. As noted by \citet{RashomonRatioNoise},  label noise expands the Rashomon set, thereby increasing the pattern diversity of near-optimal models. REAL strategically leverages this expanded Rashomon set to construct a more diverse committee that captures a wider array of structural explanations. The Gibbs-weighted framework of BREAL then acts as a probabilistic filter for the expanding set of explanations, allowing the learner to explore the diversity of the expanding hypotheses space while simultaneously de-prioritizing those that overfit outliers. 

In this setting, moderate noise works in REAL's favor by providing a broader range of perspectives for the committee. At extreme noise ($\phi = 0.45$), however, the advantage evaporates as the structural signal becomes indistinguishable from aleatoric chaos. These results suggest that while stochastic ensembles treat noise as a source of instability, REAL transforms it into a signal of structural diversity that identifies the most robust patterns in the $\epsilon$-version space.

\section{Benchmark Experiments}
\label{sec:Benchmarks}

To evaluate the practical utility of REAL, we conduct an extensive study across 20 real-world datasets (Table \ref{tab:DatasetTable}). We maintain the exact experimental protocol of in Section \ref{subsec:Protocol}.

To provide a global assessment of performance across our 20 benchmarks, we utilize the Efficiency Ratio based on the Area Under the Curve (AUC). Since AL methods converge as budgets exhaust, total AUC dilutes the signal of early-stage efficiency. We therefore report AUC truncated at $70\%$ to focus on the window of maximal information discovery where labeling constraints are most acute.

To quantify cumulative performance, we calculate the Truncated AUC using the trapezoidal rule. For a budget fraction $K \in (0, 1]$, we define the truncation point $T = \lfloor K \cdot N \rfloor$. The AUC for method $m$ is given by:
\begin{equation}
    \text{AUC}_{m, K} = \sum_{t=1}^{T-1} \frac{A_m(t) + A_m(t+1)}{2}
\end{equation}
where $A_m(t)$ denotes the mean accuracy of method $m$ at iteration $t$. We then define the Efficiency Ratio ($\rho$) relative to the BREAL baseline:
\begin{equation}
    \rho_{m} = \frac{\text{AUC}_{m, K}}{\text{AUC}_{\text{BREAL}, K}}
\end{equation}
In this metric, a ratio $\rho < 1.0$ indicates that BREAL captures a greater area under the learning curve, demonstrating higher accuracy during the initial $70\%$ of the labeling budget.

\subsection{Benchmark Experiment Results}

\textbf{Accuracy.} Figure \ref{fig:AUC_Accuracy} presents the Efficiency Ratio of each method relative to BREAL (values $<1.0$ in green indicate that BREAL achieved superior sample efficiency). Results demonstrate that BREAL consistently outperforms Random Sampling and Uncertainty Sampling across a diverse range of geometries. This advantage is particularly pronounced in high-multiplicity benchmarks like \textit{Tic-Tac-Toe} and \textit{Hepatitis} (see Table \ref{tab:DatasetTable}), where REAL's exhaustive version-space characterization prevents the learner from being distracted by the algorithmic noise that hampers randomized ensembles.

While the truncated AUC provides a global summary of efficiency, the specific learning dynamics of the $\epsilon$-version space, including the rate of accuracy growth and the stabilization of the model structure, are best observed through individual learning curves. We provide the complete set of Accuracy along with other metrics (e.g. Tree Edit Distance) trace plots for all 20 benchmarks Appendix \ref{Appendix:BenchmarkDetails}.

We track the rate of version-space collapse via the ECS (Equation \ref{eq:ECS}). As evidenced in the trace plots of Appendix \ref{Appendix:BenchmarkDetails}, this metric visualizes how quickly the structural signal is resolved across different noise regimes.

\begin{figure*}[htbp]
    \centering
    \includegraphics[width=0.95\textwidth]{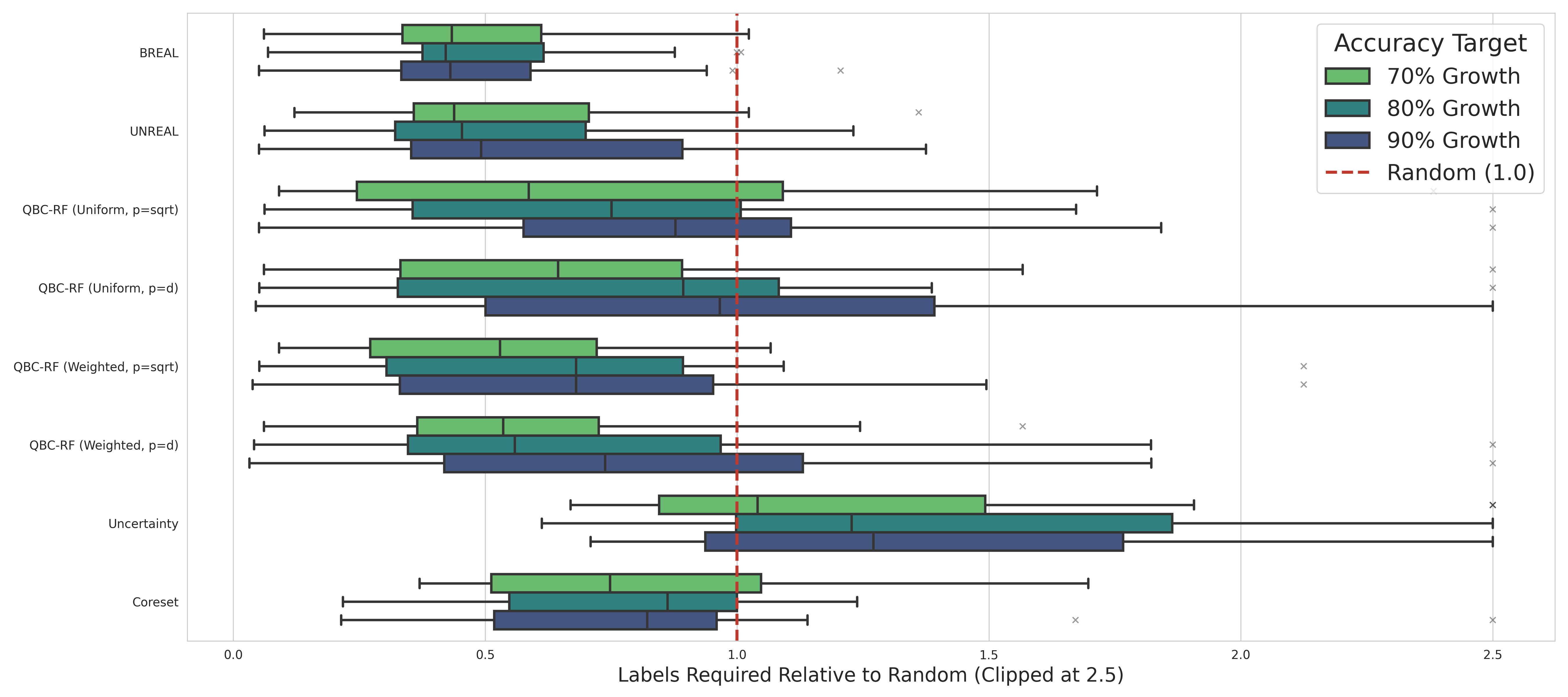}
    \caption{\textbf{Relative Label Efficiency ($N_{rel}$) across 20 datasets.} Each boxplot summarizes the labeling budget required to achieve $70\%$, $80\%$, and $90\%$ of the total possible accuracy gain. Values to the left of the dashed line ($<1.0$) indicate superior efficiency relative to Random Sampling. Both UNREAL and BREAL maintain a significantly tighter distribution and lower median than standard baselines, demonstrating that characterizing the Rashomon set provides a more reliable and robust path to cost reduction than randomized ensembles or greedy heuristics.}
    \label{fig:label_efficiency}
\end{figure*}

\textbf{Label Efficiency.} Practical deployment of active learning strategies often hinges on cost reduction. We quantify this using \textit{Relative Label Efficiency} ($N_{rel}$): the ratio of labels required by a method to reach a specific performance milestone relative to a Random Sampling baseline. We define these milestones as $70\%$, $80\%$, and $90\%$ of the total accuracy growth achieved over the labeling budget. An $N_{rel} < 1.0$ indicates superior efficiency. Figure \ref{fig:label_efficiency} aggregates these scores across the benchmark suite. Our results show BREAL frequently achieves the $70\%$ milestone with less than half the labels required by passive sampling.

The results reveal two critical advantages of REAL. Firstly, both REAL variants consistently shift the median efficiency distribution below $0.5$, implying a reduction in labeling costs of over $50\%$ on many datasets as opposed to Random Sampling. For BREAL, the median remains exceptionally stable across milestones, even as the labeling budget tightens for the $90\%$ growth target. 

Secondly, a major finding is the extreme volatility of advanced baselines that is not present in REAL. While QBC-RF methods and Uncertainty Sampling achieve competitive efficiency on select datasets (left whiskers extending toward $0.1$), they exhibit heavy tails of extreme inefficiency, with right whiskers and outliers extending beyond $2.0$. This indicates that these methods can often be twice as expensive as random sampling when their stochastic or greedy heuristics fail to capture the data geometry. In contrast, REAL maintains a tight, compact distribution with minimal right skew and no extreme outliers. This high-precision, low-variance performance makes REAL a safer and more predictable choice for real-world applications where labeling budgets are fixed and dataset characteristics are initially unknown.

\textbf{The PAC-Bayesian Advantage.} Our results reveal an improvement over standard QBC when incorporating PAC-Bayesian weighting into the active learning loop. As shown in Figure \ref{fig:AUC_Accuracy} and \ref{fig:label_efficiency}, and the individual trace plots in Appendix \ref{Appendix:BenchmarkDetails}, applying a Gibbs posterior to standard ensembles (QBC-RF Weighted) already improves efficiency compared to their uniform counterparts (QBC-RF Uniform).

However, the most significant performance gains are realized only when moving from randomized ensembles to the version-space architecture of the REAL framework. While a Gibbs posterior weighting refines randomized committees, the Rashomon set's structural diversity provides a more direct characterization of version-space uncertainty than synthetic RF diversity. BREAL consistently establishes a higher accuracy ceiling several iterations earlier than even the QBC-RF weighted baselines as evidenced in the as \textit{Anneal}, \textit{Hepatitis}, and \textit{Tic-Tac-Toe} datasets with accuracy gains of up to 0.045 over conventional QBC-RF and 0.031 over its PAC-Bayesian weighted counterpart.

\section{Conclusion}
\label{sec:Conclusion}

This work introduced REAL, an active learning framework that replaces the perturbation-based diversity of randomized ensembles with the principled diversity of the Rashomon set. By constructing committees from near-optimal models, REAL targets true epistemic uncertainty, allowing for superior sample efficiency in well-specified regimes.

Our empirical analysis disentangles the roles of version-space precision and algorithmic noise. We demonstrate that rather than being a vulnerability, moderate label noise acts as a driver of pattern diversity that REAL strategically leverages. By utilizing a Gibbs-weighted framework, the learner navigates the expanded $\epsilon$-version space to find a stable structural signal where randomized ensembles often falter. While the perturbation-based diversity of standard Random Forests provides a necessary structural buffer under severe geometric misspecification, REAL remains a superior instrument for high-precision structural discovery in well-aligned regimes.

Several directions remain open. While our implementation focused on sparse decision trees, the most immediate frontier is the extension of REAL to other hypothesis classes. The ongoing expansion of Rashomon enumeration to structures such as Rashomon Partitions \citep{AparaRPS} and Prototypical-Part Networks \citep{Hayden_RashomonCV} suggests that the paradigm of many near-optimal models can be adapted to more complex model classes. Future work will focus on developing efficient enumeration algorithms for these classes, allowing REAL to serve as a model-agnostic framework for resolving epistemic uncertainty across diverse architectural regimes.

By integrating near-optimal models rather than searching for the ``best model,'' we provide a more grounded path toward interpretable and efficient machine learning that directly addresses the challenge of quantifying epistemic uncertainty in the presence of unresolvable model ambiguity.

\bibliographystyle{plainnat}
\bibliography{bibliography}

\newpage
\onecolumn

\title{Supplementary Material}
\maketitle
\appendix

\section{Methods and Implementation Details}
\label{Appendix:Methods}
\begin{algorithm}[!htb]
   \caption{Rashomon Ensembled Active Learning (REAL)}
   \label{alg:REAL}
    \begin{algorithmic}[1]
        \STATE {\bfseries Input:} Initial set $\mathcal{D}_{tr}^{(0)}$; Pool $\mathcal{U}^{(0)}$; Budget $B$; 
        \STATE \quad \quad Rashomon threshold $\epsilon$; Weighting mode $\omega \in \{\text{Gibbs, Uniform}\}$
        \STATE {\bfseries Initialize:} Selector parameters $(d, \lambda, \beta)$ via multi-stage calibration on $\mathcal{D}_{tr}^{(0)}$
        \FOR{$b = 0$ {\bfseries to} $B-1$}
            \algheader{\textbf{1. Version-Space Characterization}}
            \STATE $h^* \leftarrow \text{OptimalTree}(\mathcal{D}_{tr}^{(b)}, d, \lambda)$ 
            \STATE $\hat{\mathcal{R}}(\epsilon) \leftarrow \text{SORTD\_Enumerate}(h^*, \epsilon, \mathcal{D}_{tr}^{(b)})$ 
            \STATE
            
            \algheader{\textbf{2. PAC-Bayesian Weighting}}
            \FOR{{\bfseries each} $h_i \in \hat{\mathcal{R}}(\epsilon)$}
                \IF{$\omega = \text{Gibbs}$}
                    \STATE $w_i \leftarrow \frac{\exp(-\beta \cdot \mathcal{L}(h_i, \mathcal{D}_{tr}^{(b)}))}{\sum_{j} \exp(-\beta \cdot \mathcal{L}(h_j, \mathcal{D}_{tr}^{(b)}))}$ \COMMENT{\textbf{B-REAL} configuration}
                \ELSE
                    \STATE $w_i \leftarrow 1 / |\hat{\mathcal{R}}(\epsilon)|$ \COMMENT{\textbf{UNREAL} configuration}
                \ENDIF
            \ENDFOR
            \STATE $\mathrm{ECS}^{(b)} \leftarrow \exp(-\sum w_i \ln w_i)$ 
            \STATE

            \algheader{\textbf{3. Epistemic Uncertainty Quantification}}
            \FOR{{\bfseries each} candidate $\mathbf{x}_n \in \mathcal{U}^{(b)}$} 
                \STATE $\hat{P}(y| \mathbf{x}_n) \leftarrow \sum_{h_i \in \hat{\mathcal{R}}} w_i \cdot \mathbb{I}(h_i(\mathbf{x}_n) = y)$ 
                \STATE $H_n \leftarrow - \sum_{y \in \mathcal{Y}} \hat{P}(y| \mathbf{x}_n) \log \hat{P}(y| \mathbf{x}_n)$ 
            \ENDFOR
            \STATE

            \algheader{\textbf{4. Label Acquisition and Update}}
            \STATE $\mathbf{x}^* \leftarrow \arg\max_{\mathbf{x}_n \in \mathcal{U}^{(b)}} H_n$
            \STATE $y^* \leftarrow \mathrm{QueryOracle}(\mathbf{x}^*)$
            \STATE $\mathcal{D}_{tr}^{(b+1)} \leftarrow \mathcal{D}_{tr}^{(b)} \cup \{(\mathbf{x}^*, y^*)\}$; $\mathcal{U}^{(b+1)} \leftarrow \mathcal{U}^{(b)} \setminus \{\mathbf{x}^*\}$
        \ENDFOR
        \STATE $f \leftarrow \text{OptimalTree}(\mathcal{D}_{tr}^{(B)}, d_{\text{pred}}, \lambda_{\text{pred}})$ 
        \STATE {\bfseries Return:} Predictor $f$
    \end{algorithmic}
\end{algorithm}

\begin{table}[htbp]
    \centering
    \small
    \begin{tabular}{@{}llll@{}}
        \toprule
        \textbf{Label} & \textbf{Committee Foundation} & \textbf{Weighting} & \textbf{Diversity Mechanism} \\
        \midrule
        Random & Random Forest & Uniform & Passive (Random Baseline) \\
        \midrule
        QBC-RF ($m=\sqrt{d}$) & Random Forest & Uniform & Artificial (Feature Blinding) \\
        B-QBC-RF ($m=\sqrt{d}$) & Random Forest & Gibbs & Artificial + PAC-Bayesian \\
        QBC-RF ($m=d$) & Random Forest & Uniform & Artificial (Bootstrap Only) \\
        B-QBC-RF ($m=d$) & Random Forest & Gibbs & Artificial + PAC-Bayesian \\
        \midrule
        Uncertainty & Optimal Sparse Tree & --- & Greedy \\
        Coreset & Optimal Sparse Tree & --- & Geometric ($k$-Center) \\
        \midrule
        \textbf{UNREAL} & Rashomon Set & Uniform & Version-Space\\
        \textbf{BREAL} & Rashomon Set & Gibbs & Version-Space \\
        \bottomrule
    \end{tabular}
    \caption{Overview of evaluated active learning configurations. We categorize strategies by their committee foundation and weighting scheme to isolate the performance gains of structural version-space characterization from those of PAC-Bayesian posterior weighting.}
    \label{tab:strategy_reference}
\end{table}
\clearpage


\section{Benchmark Details}
\label{Appendix:BenchmarkDetails}
\begin{table*}[htbp]
    \centering
    \scriptsize
    \begin{tabular}{rllrrrrrrrrr}
        \toprule
        \textbf{No.} & \textbf{Dataset} & \textbf{Src} & $N$ & $d$ & \textbf{Maj\%} & \textbf{Lin\%} & \textbf{GBM\%} & \textbf{Orc\%} & $|\hat{\mathcal{R}}|$ & \textbf{ECS} \\ 
        \midrule
        1 & Anneal & UCI & 812 & 44 & 96.2 & 100.0 & 100.0 & 100.0 & 2 & 2.0 \\
        2 & Bank Marketing & UCI & 4,521 & 23 & 97.1 & 97.7 & 98.1 & 97.7 & 19,024 & 19,023.9 \\
        4 & Bar-7 & SOR & 1,913 & 14 & 79.0 & 79.0 & 79.0 & 79.0 & 174,331 & 174,328.0 \\
        5 & Breast Cancer WI & UCI & 699 & 10 & 91.3 & 91.4 & 91.8 & 91.3 & 116,263 & 116,260.4 \\
        6 & Car Ealuation & UCI & 1,728 & 15 & 66.7 & 79.3 & 79.3 & 79.3 & 128,546 & 128,544.6 \\
        7 & Cheap Restaurant & SOR & 2,653 & 15 & 81.8 & 96.8 & 96.8 & 95.9 & 572 & 572.0 \\
        8 & Coffee House & SOR & 3,816 & 15 & 69.2 & 83.2 & 85.3 & 84.6 & 100,412 & 100,393.7 \\
        9 & COMPAS & Pro & 6,907 & 12 & 53.7 & 67.1 & 67.6 & 67.0 & 103,922 & 103,899.3 \\
        10 & Expensive Restaurant & SOR & 1,417 & 15 & 72.5 & 93.6 & 94.6 & 93.4 & 223,876 & 223,874.8 \\
        11 & FICO (HELOC) & FIC & 10,459 & 17 & 52.2 & 72.1 & 73.0 & 71.6 & 104,671 & 104,667.5 \\
        12 & Haberman & UCI & 306 & 92 & 99.7 & 99.7 & 100.0 & 100.0 & 7 & 7.0 \\
        13 & Hepatitis & UCI & 137 & 34 & 54.7 & 75.9 & 97.1 & 78.8 & 24,798 & 24,786.6 \\
        14 & Hypothyroid & UCI & 3,247 & 39 & 98.9 & 98.9 & 99.2 & 98.9 & 39 & 39.0 \\
        15 & Lymphography & UCI & 148 & 47 & 98.6 & 98.6 & 100.0 & 100.0 & 104 & 104.0 \\
        16 & MONK-2 & UCI & 169 & 11 & 50.3 & 56.2 & 87.0 & 69.8 & 7,101 & 7,082.6 \\
        17 & Primary Tumor & UCI & 336 & 17 & 65.8 & 79.2 & 84.8 & 78.0 & 129,946 & 129,893.0 \\
        18 & Tic-Tac-Toe & UCI & 958 & 18 & 56.4 & 99.1 & 90.7 & 81.2 & 82,063 & 82,051.9 \\
        19 & Congressional Vote & UCI & 435 & 48 & 76.1 & 100.0 & 100.0 & 100.0 & 2 & 2.0 \\
        20 & Yeast & UCI & 1,484 & 46 & 92.9 & 95.6 & 97.5 & 94.7 & 202,309 & 202,308.4 \\
        \midrule
        21 & Synth.\ XOR ($\alpha$=0, $\phi$=0) & Syn & 500 & 20 & 51.8 & 56.0 & 100.0 & 100.0 & 2 & 2.0 \\
        22 & Synth.\ XOR ($\alpha$=0.25) & Syn & 500 & 20 & 52.0 & 61.8 & 92.8 & 87.8 & 170,318 & 170,315.6 \\
        23 & Synth.\ XOR ($\alpha$=0.50) & Syn & 500 & 20 & 51.6 & 71.2 & 90.4 & 76.8 & 197,171 & 197,138.4 \\
        24 & Synth.\ XOR ($\alpha$=0.75) & Syn & 500 & 20 & 51.4 & 82.8 & 91.6 & 77.4 & 10,862 & 10,823.5 \\
        25 & Synth.\ XOR ($\alpha$=1.00) & Syn & 500 & 20 & 53.8 & 98.4 & 99.2 & 84.6 & 4,332 & 4,332.0 \\
        26 & Synth.\ XOR ($\phi$=0.05) & Syn & 500 & 20 & 51.6 & 57.6 & 96.4 & 95.0 & 170,318 & 170,315.6 \\
        27 & Synth.\ XOR ($\phi$=0.10) & Syn & 500 & 20 & 51.2 & 55.2 & 90.4 & 90.2 & 170,318 & 170,315.6 \\
        28 & Synth.\ XOR ($\phi$=0.25) & Syn & 500 & 20 & 51.8 & 57.2 & 82.4 & 73.6 & 170,318 & 170,315.6 \\
        29 & Synth.\ XOR ($\phi$=0.45) & Syn & 500 & 20 & 51.4 & 58.6 & 77.6 & 63.2 & 103,727 & 103,656.2 \\
        \bottomrule
    \end{tabular}
    \caption{Datasets.}
    \label{tab:DatasetTable}
\end{table*}

\clearpage


\begin{figure*}[!t] 
\centering
    \begin{subfigure}[b]{0.19\textwidth}
        \includegraphics[width=\linewidth]{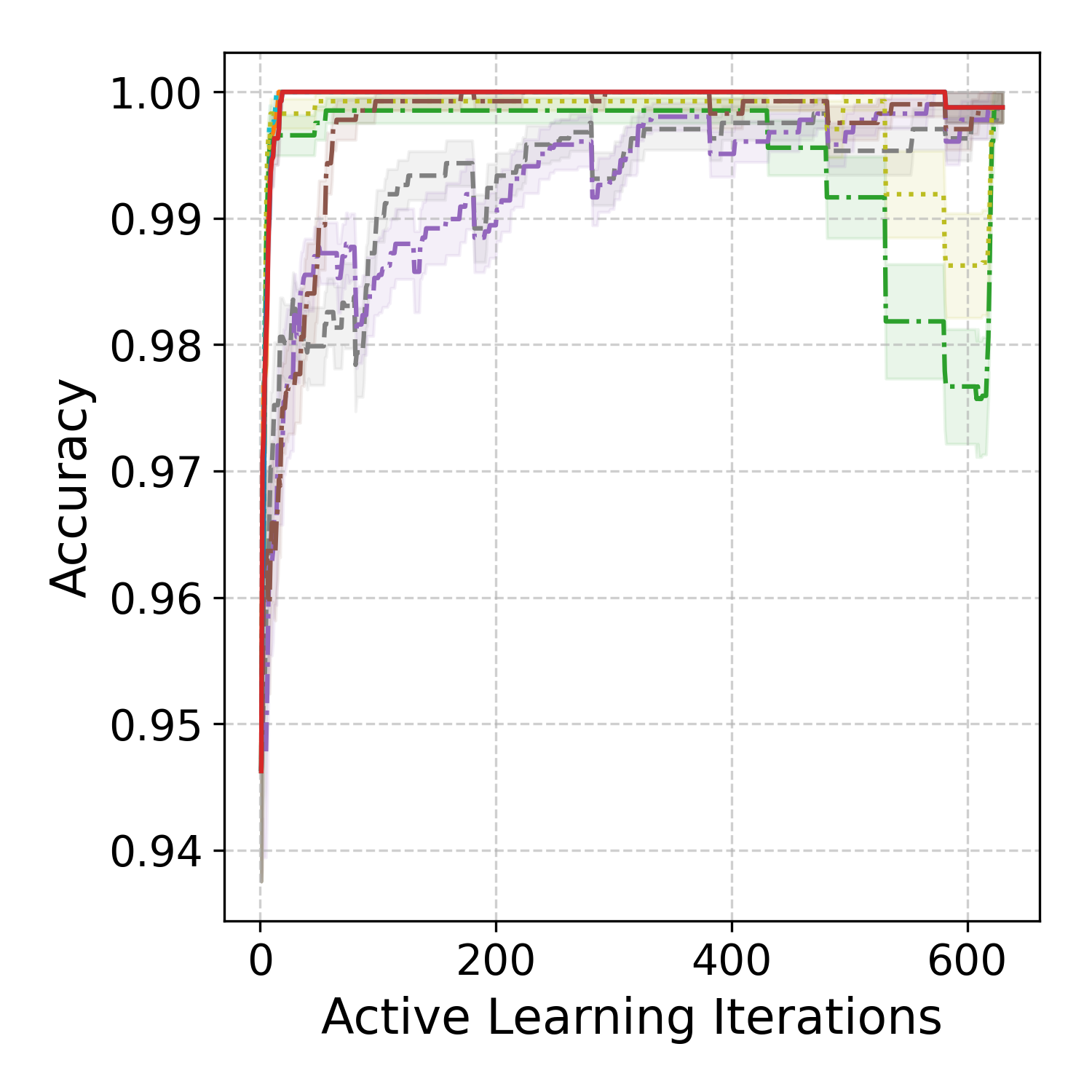}
        \caption{Anneal}
    \end{subfigure}
    \hfill
    \begin{subfigure}[b]{0.19\textwidth}
        \includegraphics[width=\linewidth]{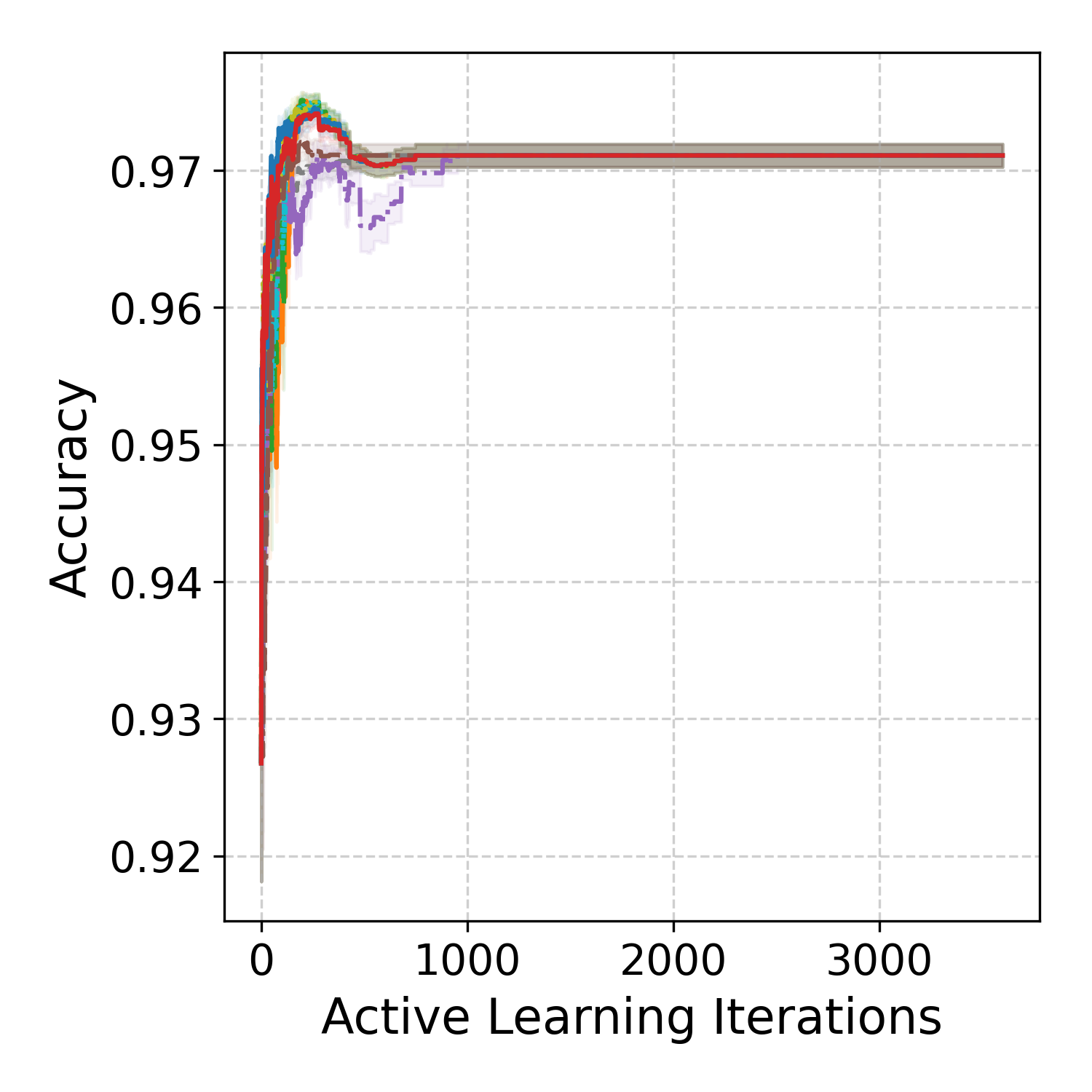}
        \caption{Bank Marketing}
    \end{subfigure}
    \hfill
    \begin{subfigure}[b]{0.19\textwidth}
        \includegraphics[width=\linewidth]{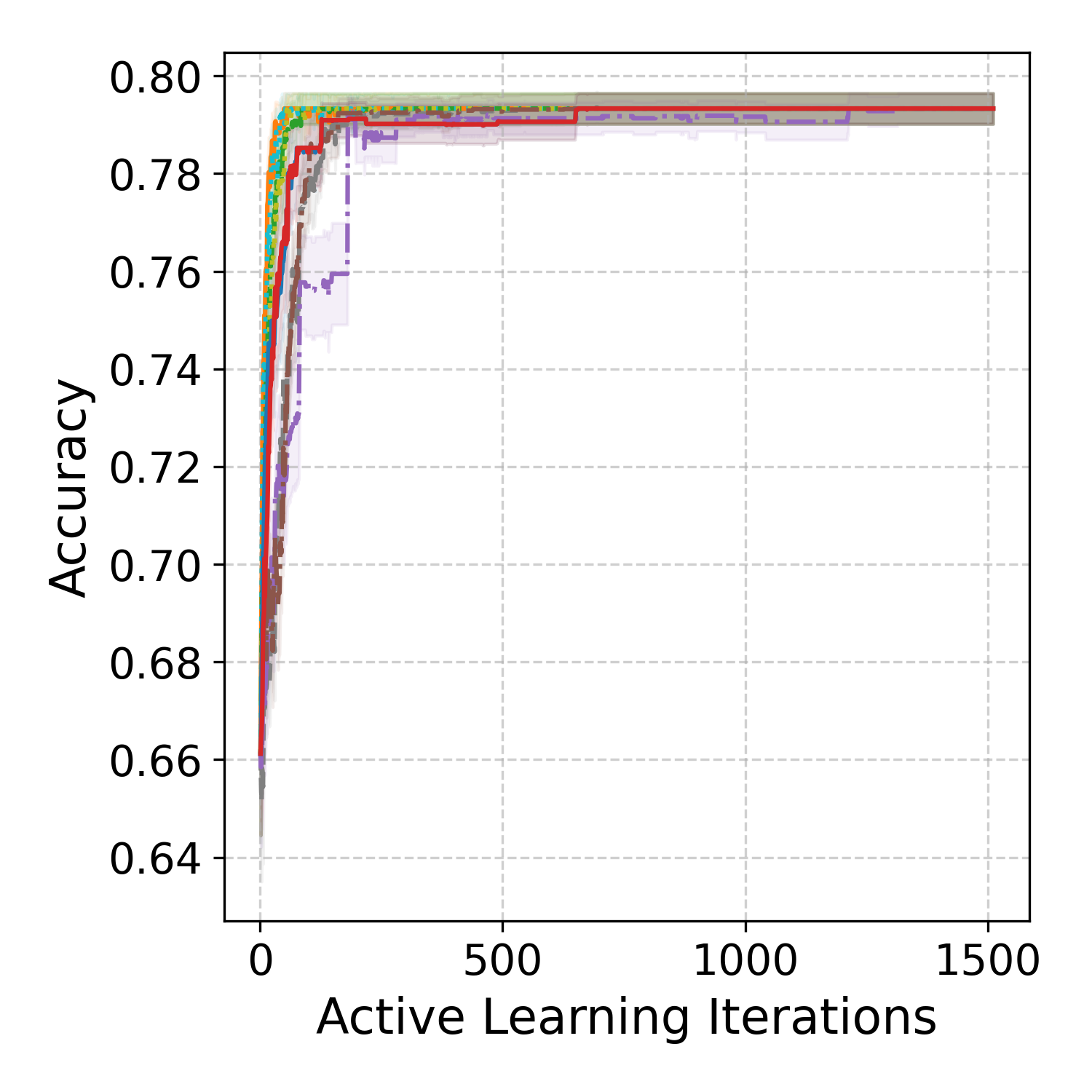}
        \caption{Bar-7}
    \end{subfigure}
    \hfill
    \begin{subfigure}[b]{0.19\textwidth}
        \includegraphics[width=\linewidth]{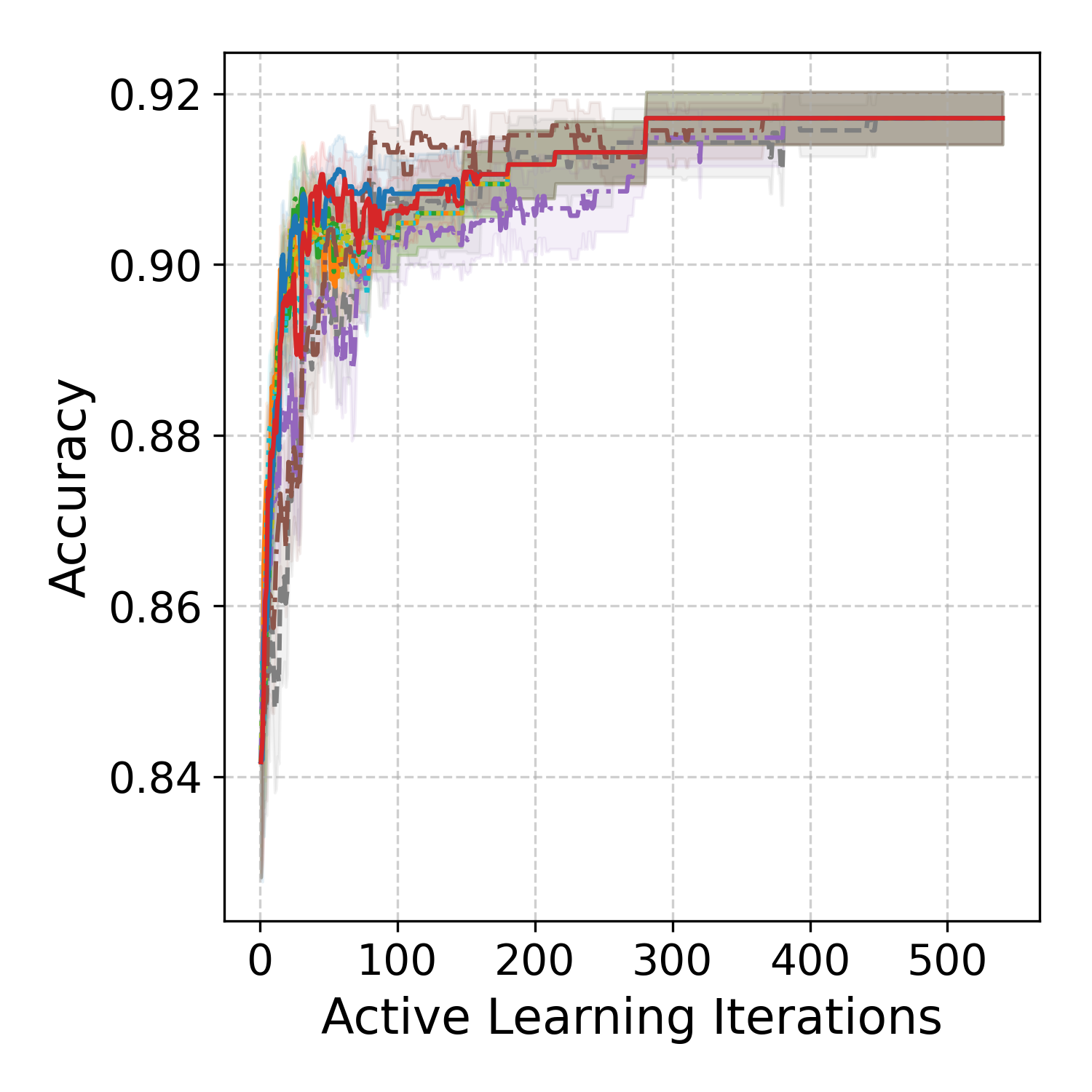}
        \caption{Breast Cancer WI}
    \end{subfigure}
    \hfill
    \begin{subfigure}[b]{0.19\textwidth}
        \includegraphics[width=\linewidth]{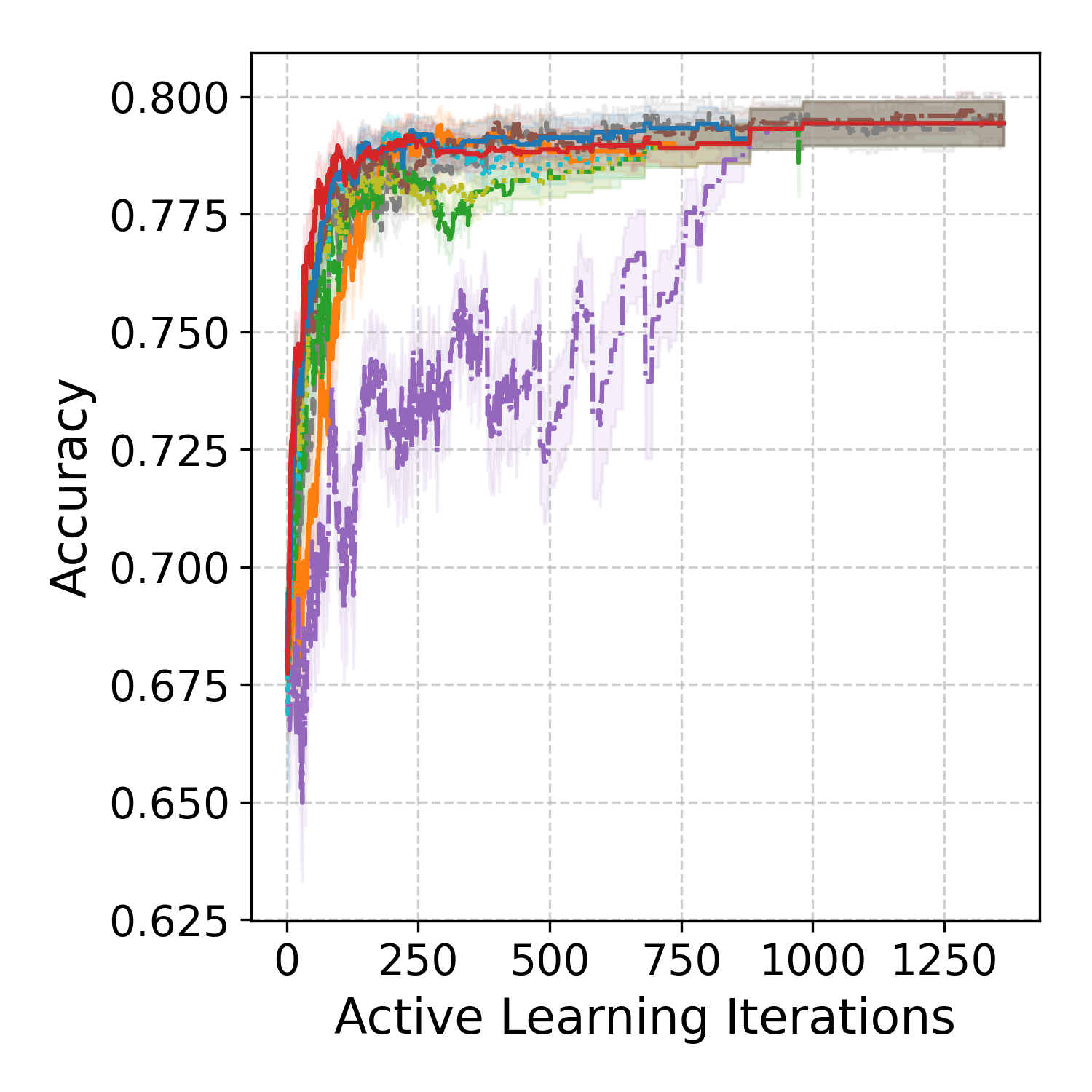}
        \caption{Car Evaluation}
    \end{subfigure}

    \vspace{0.2em}

    \begin{subfigure}[b]{0.19\textwidth}
        \includegraphics[width=\linewidth]{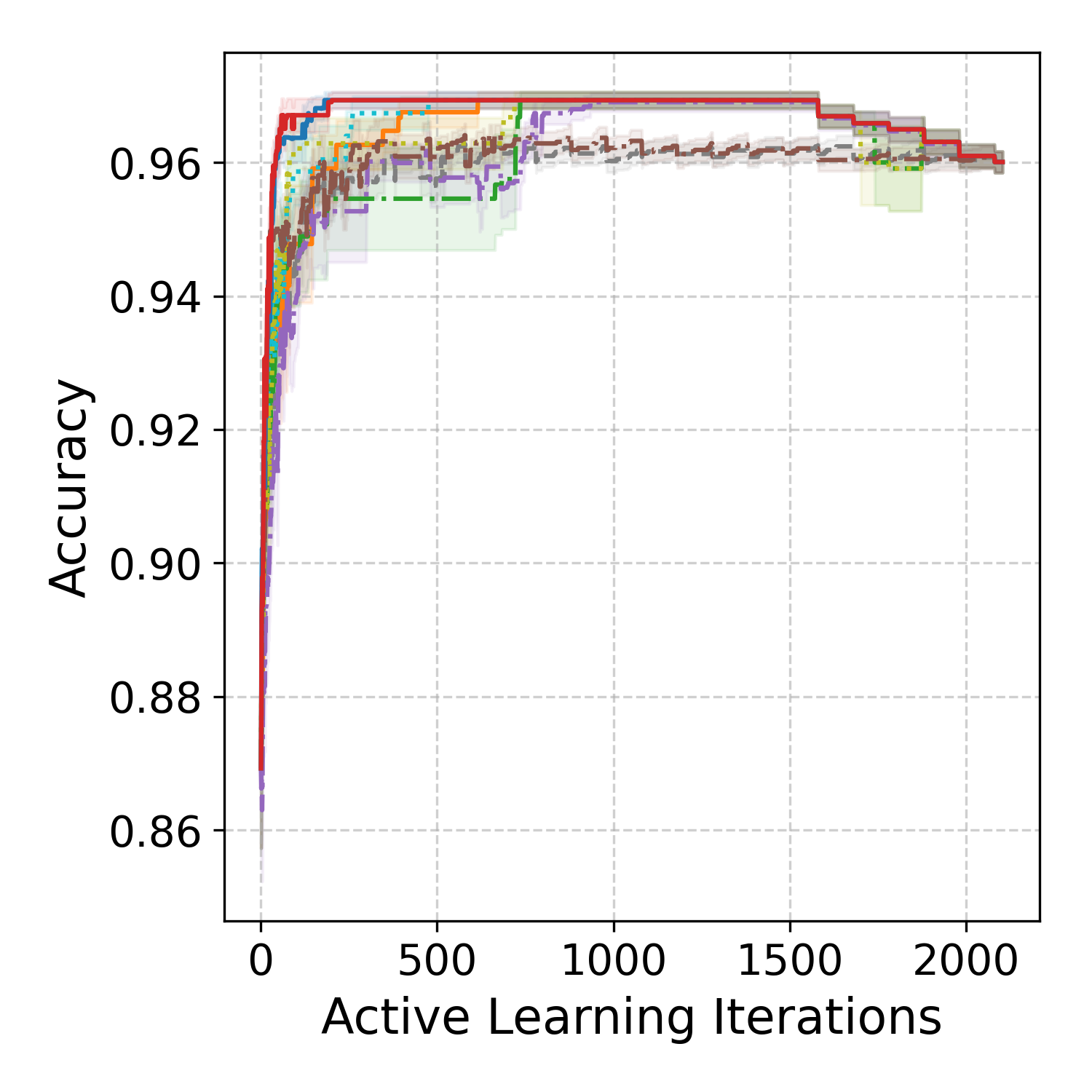}
        \caption{Cheap Restaurant}
    \end{subfigure}
    \hfill
    \begin{subfigure}[b]{0.19\textwidth}
        \includegraphics[width=\linewidth]{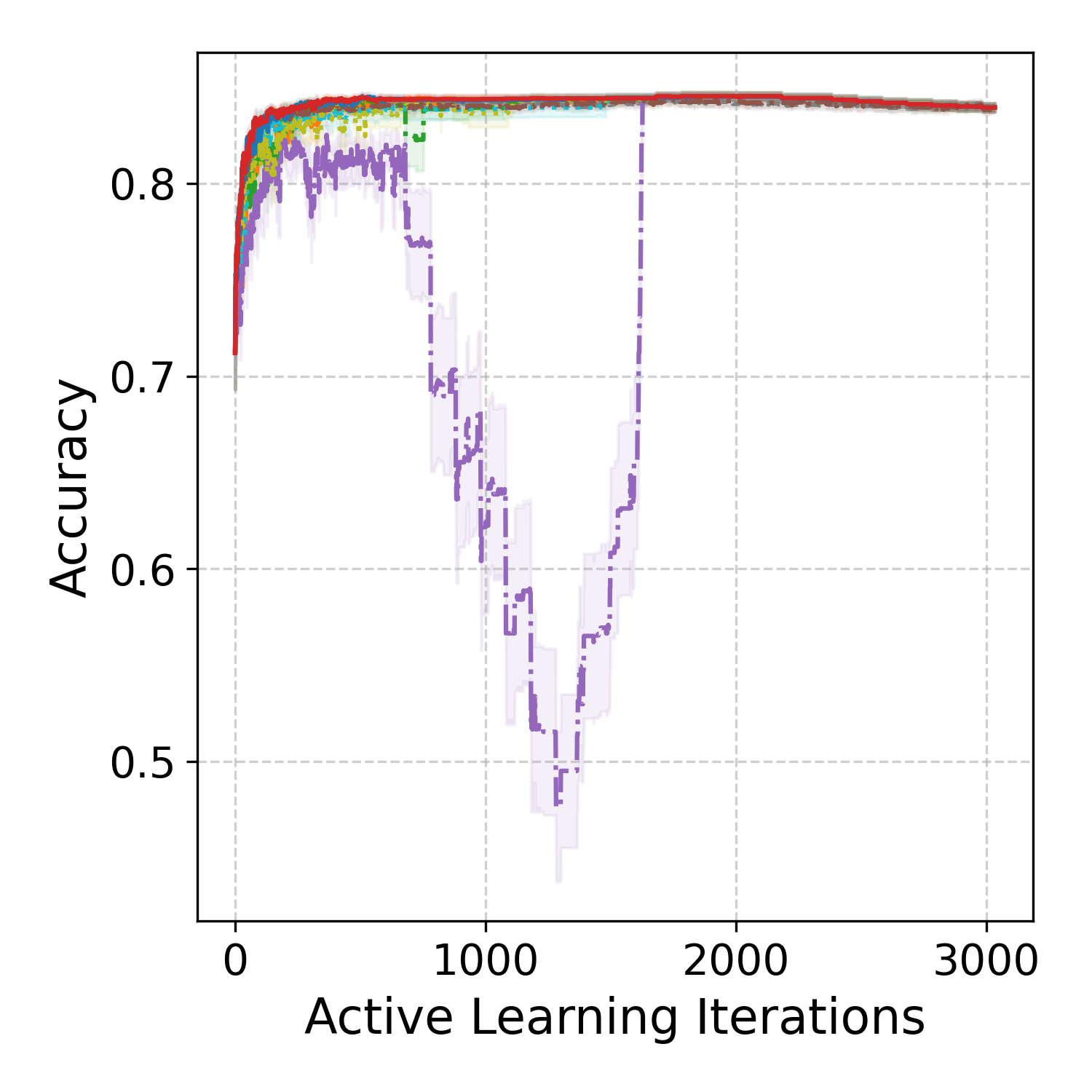}
        \caption{Coffee House}
    \end{subfigure}
    \hfill
    \begin{subfigure}[b]{0.19\textwidth}
        \includegraphics[width=\linewidth]{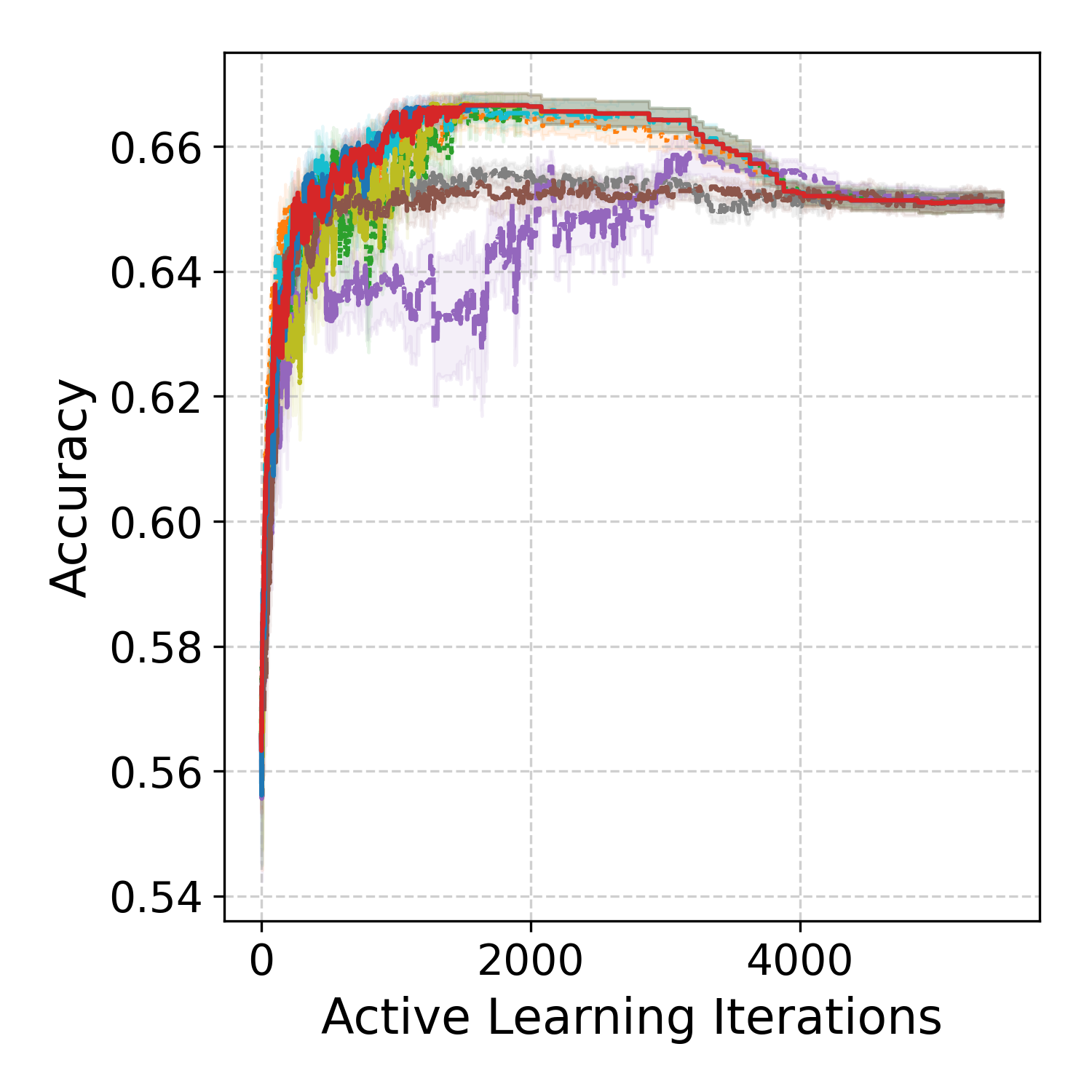}
        \caption{COMPAS}
    \end{subfigure}
    \hfill
    \begin{subfigure}[b]{0.19\textwidth}
        \includegraphics[width=\linewidth]{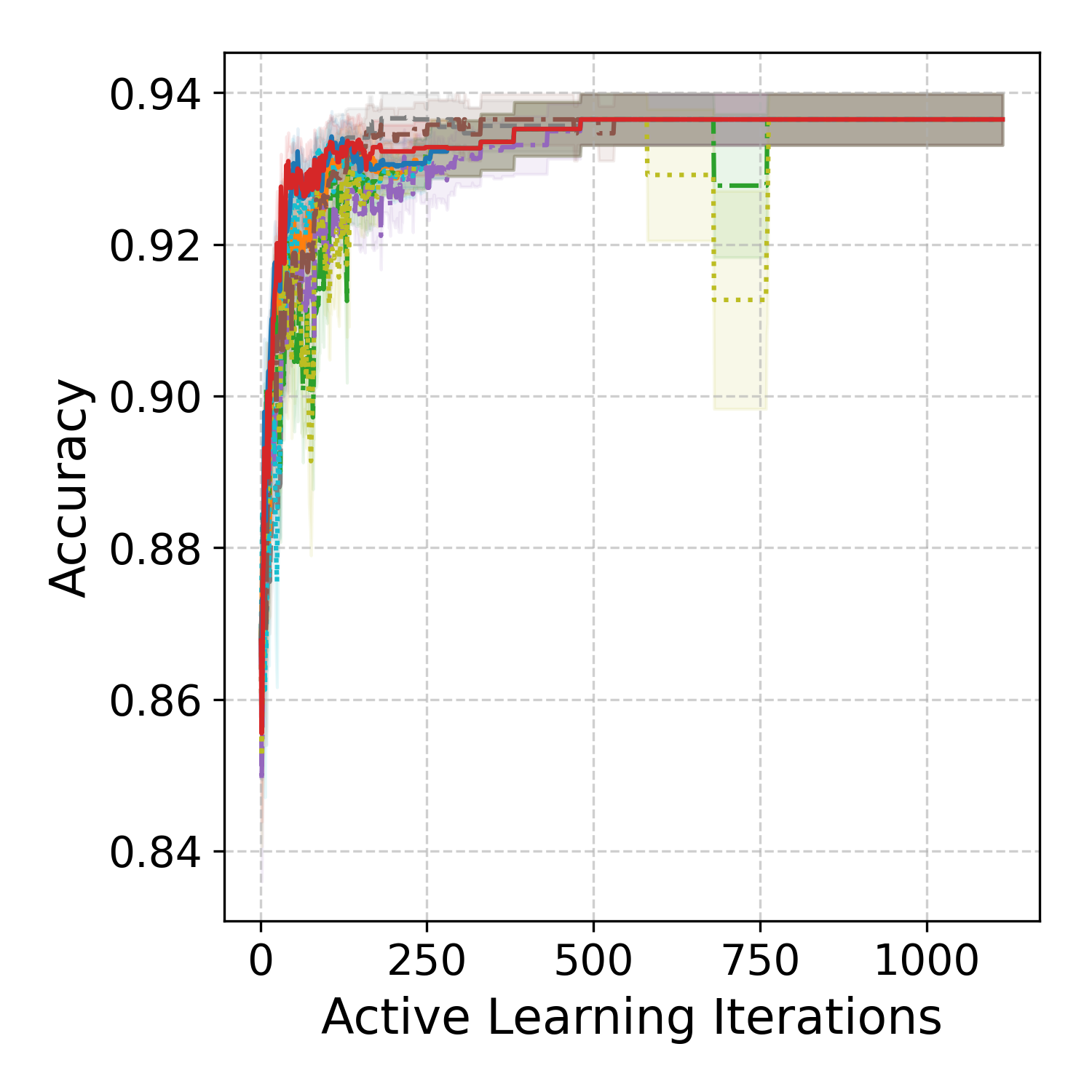}
        \caption{Expensive Restaurant}
    \end{subfigure}
    \hfill
    \begin{subfigure}[b]{0.19\textwidth}
        \includegraphics[width=\linewidth]{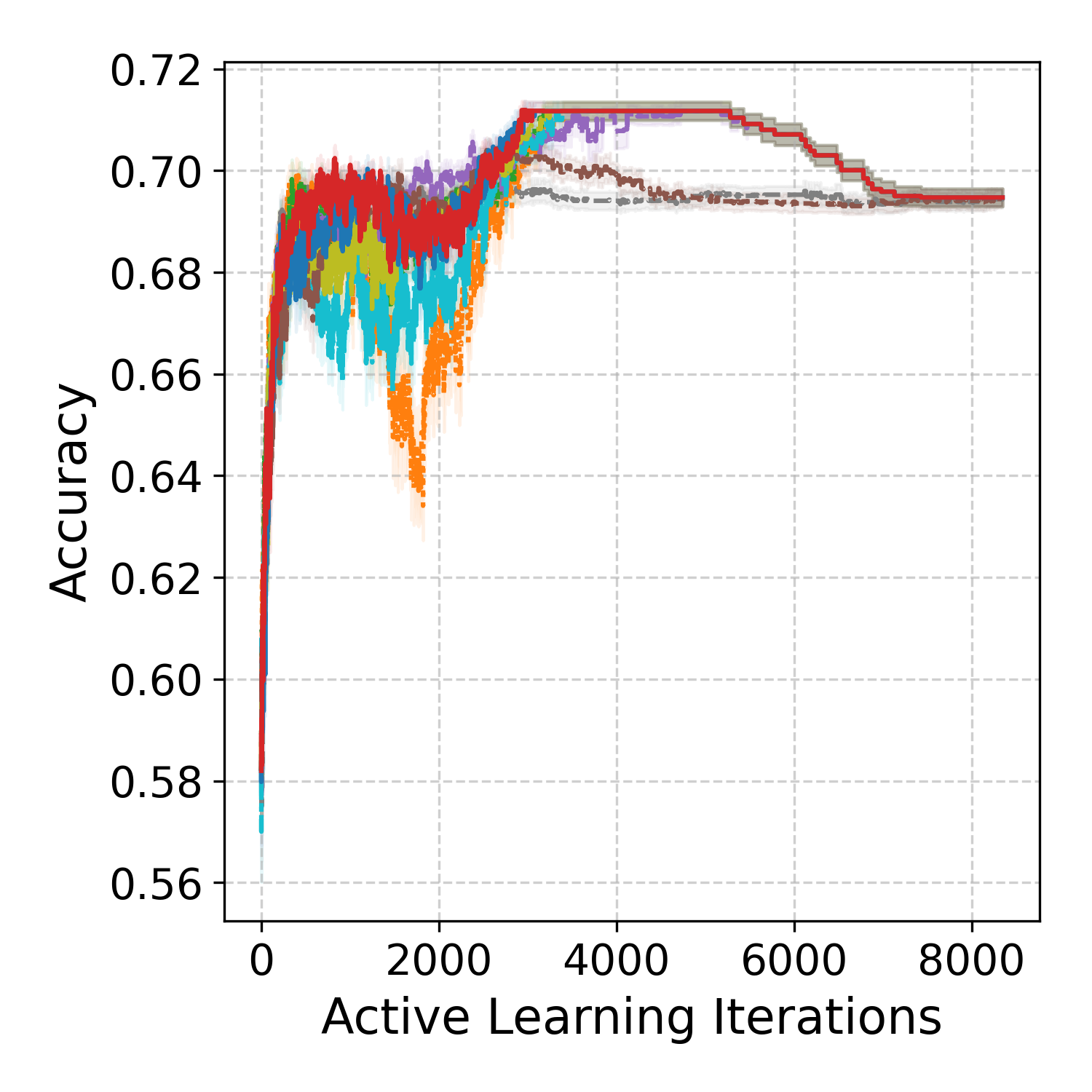}
        \caption{FICO}
    \end{subfigure}

    \vspace{0.2em}

    \begin{subfigure}[b]{0.19\textwidth}
        \includegraphics[width=\linewidth]{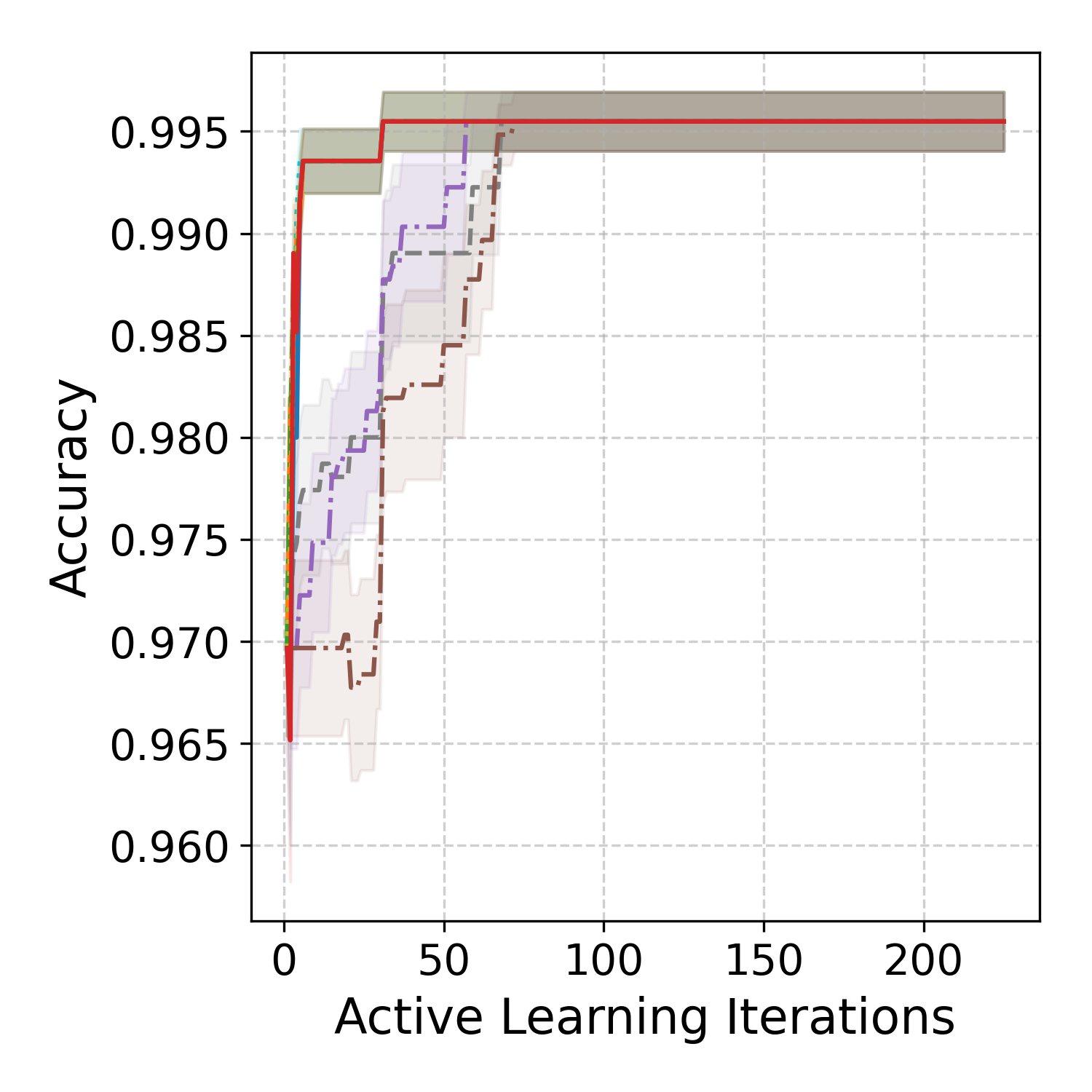}
        \caption{Haberman}
    \end{subfigure}
    \hfill
    \begin{subfigure}[b]{0.19\textwidth}
        \includegraphics[width=\linewidth]{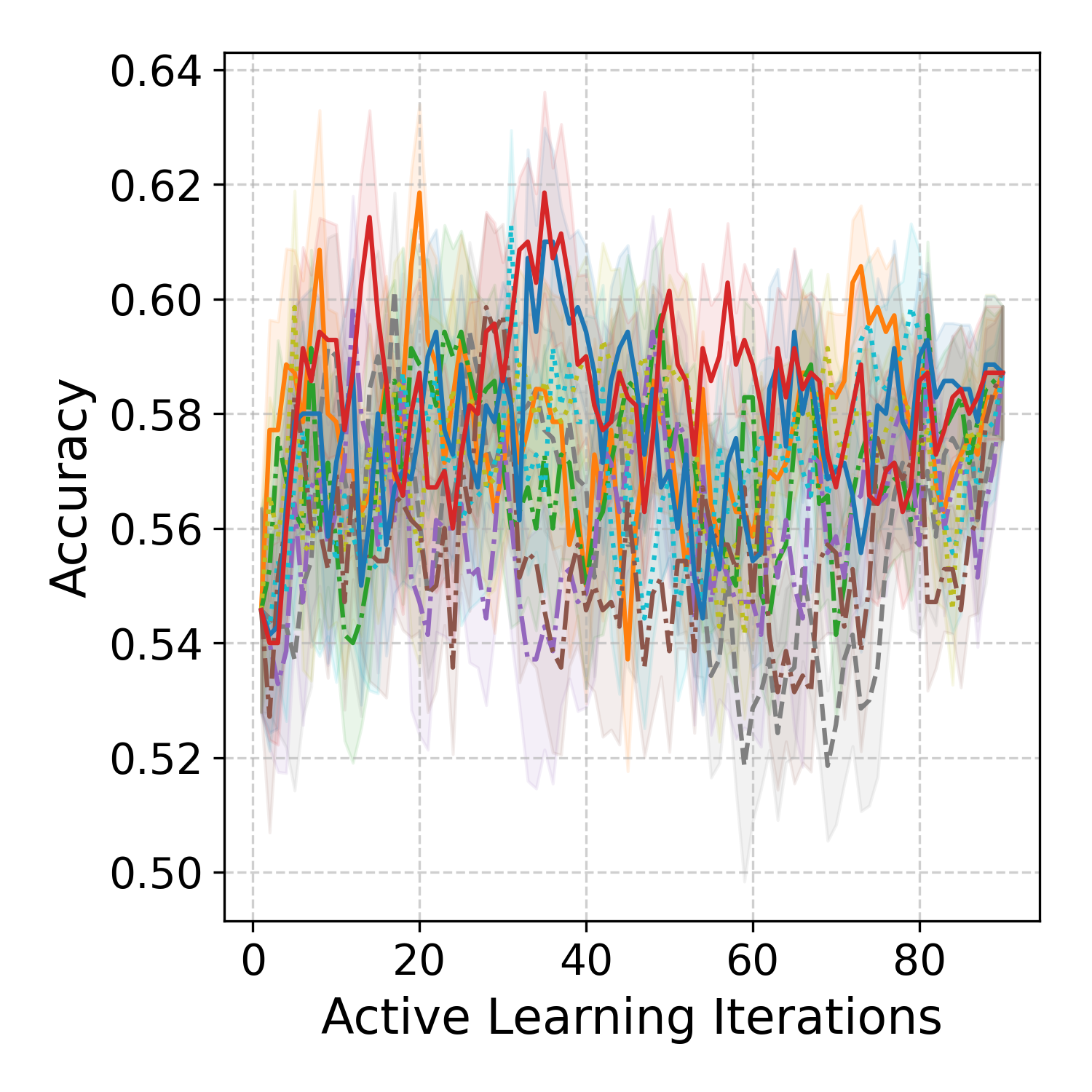}
        \caption{Hepatitis}
    \end{subfigure}
    \hfill
    \begin{subfigure}[b]{0.19\textwidth}
        \includegraphics[width=\linewidth]{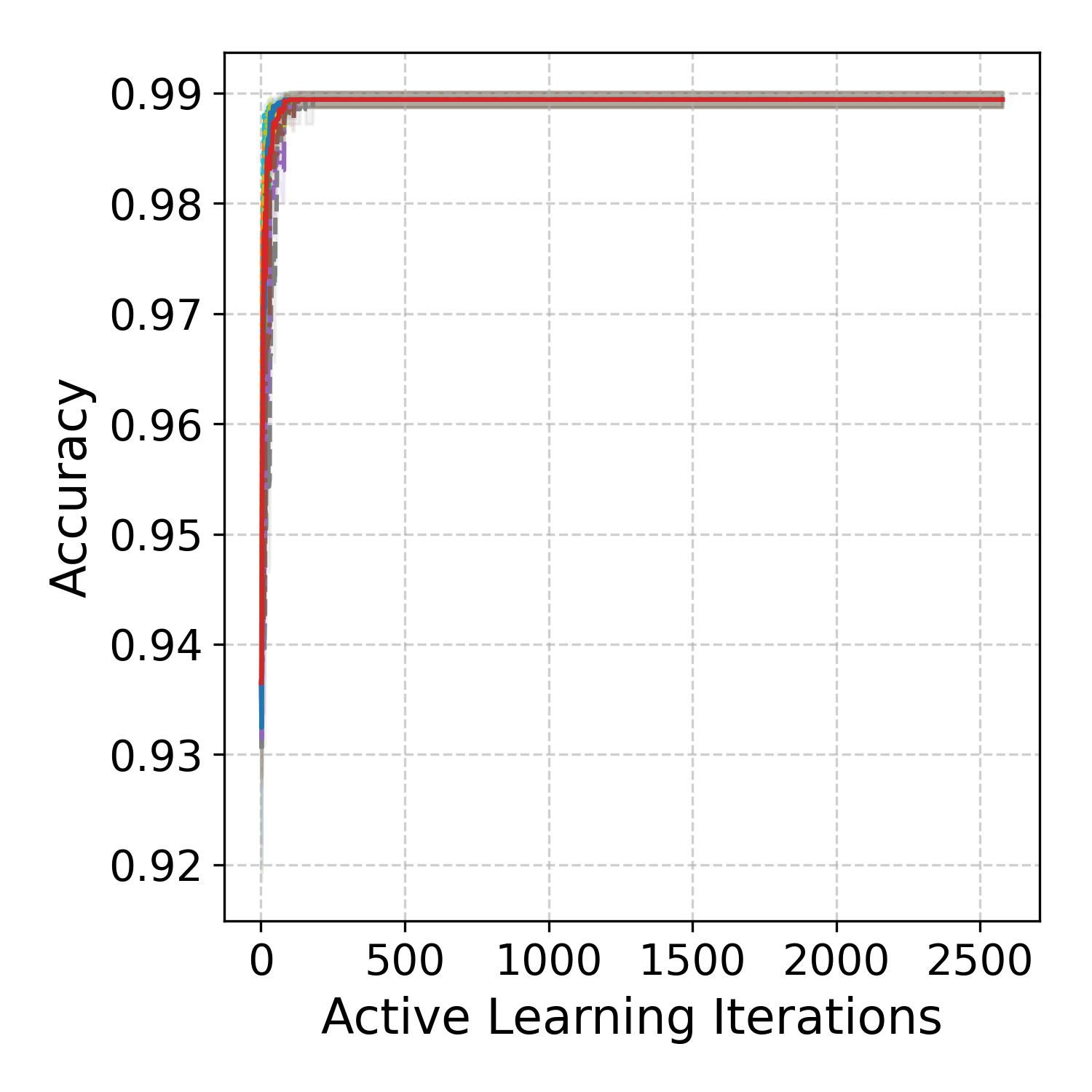}
        \caption{Hypothyroid}
    \end{subfigure}
    \hfill
    \begin{subfigure}[b]{0.19\textwidth}
        \includegraphics[width=\linewidth]{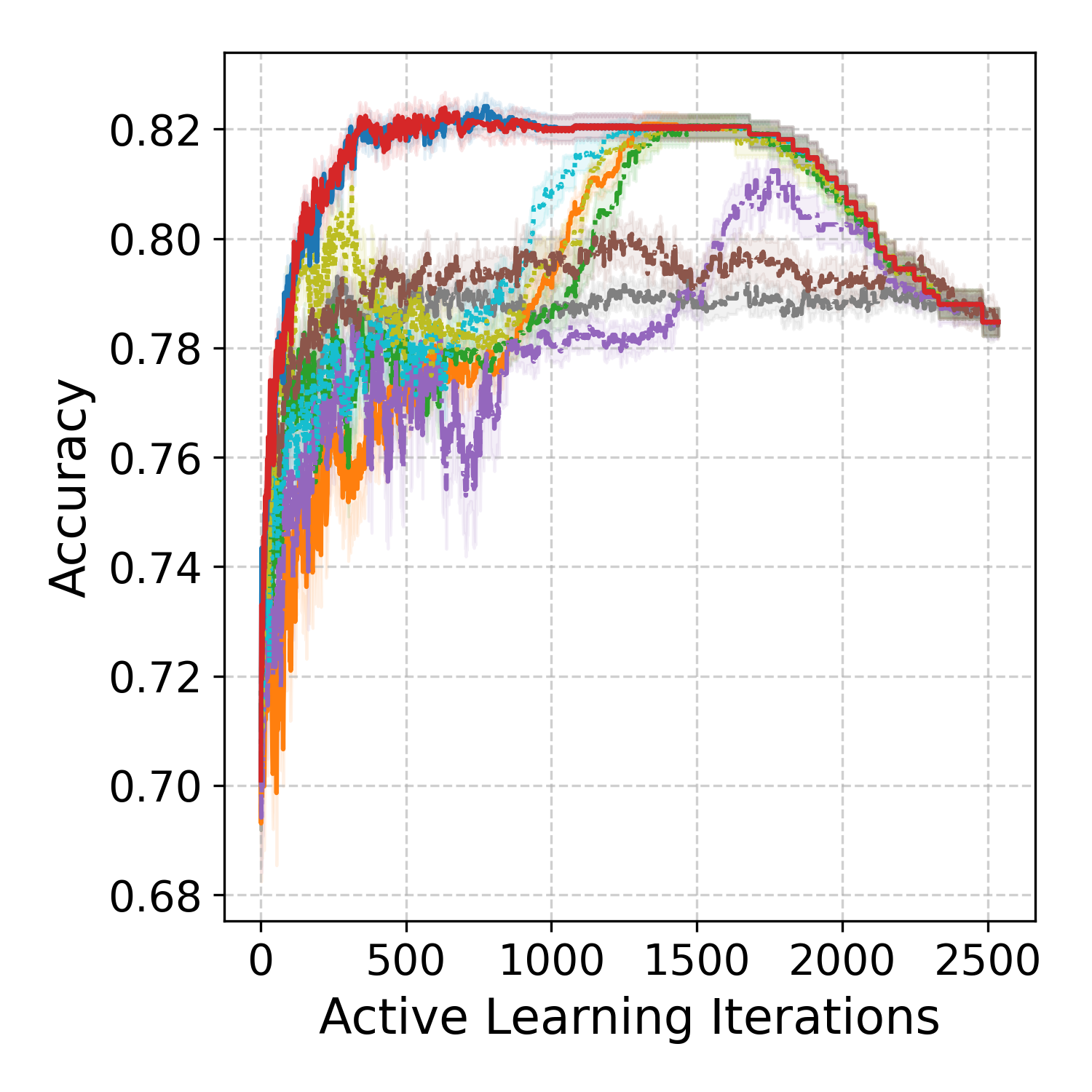}
        \caption{Kr-vs-Kp}
    \end{subfigure}
    \hfill
    \begin{subfigure}[b]{0.19\textwidth}
        \includegraphics[width=\linewidth]{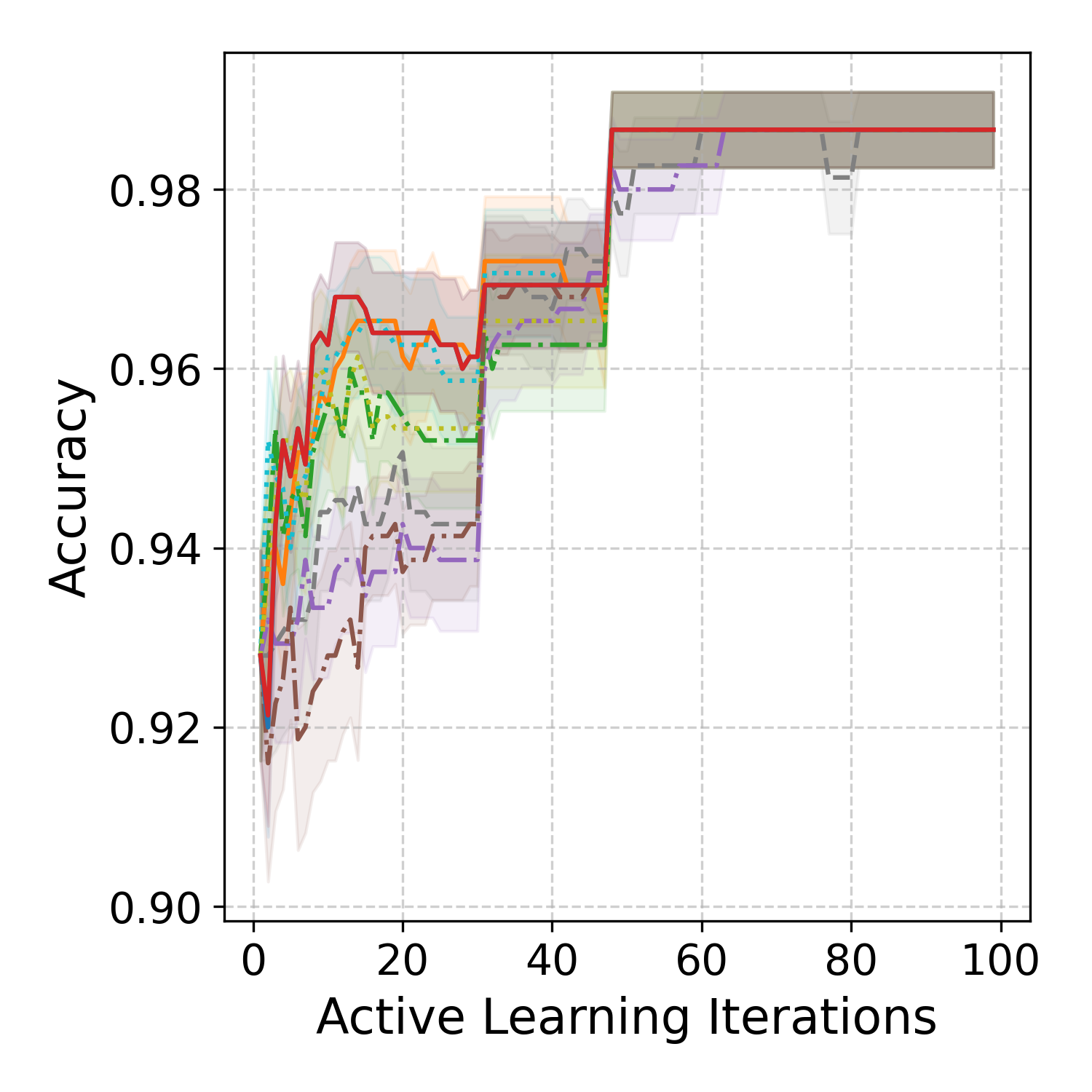}
        \caption{Lymphography}
    \end{subfigure}

    \vspace{0.2em}

    \begin{subfigure}[b]{0.19\textwidth}
        \includegraphics[width=\linewidth]{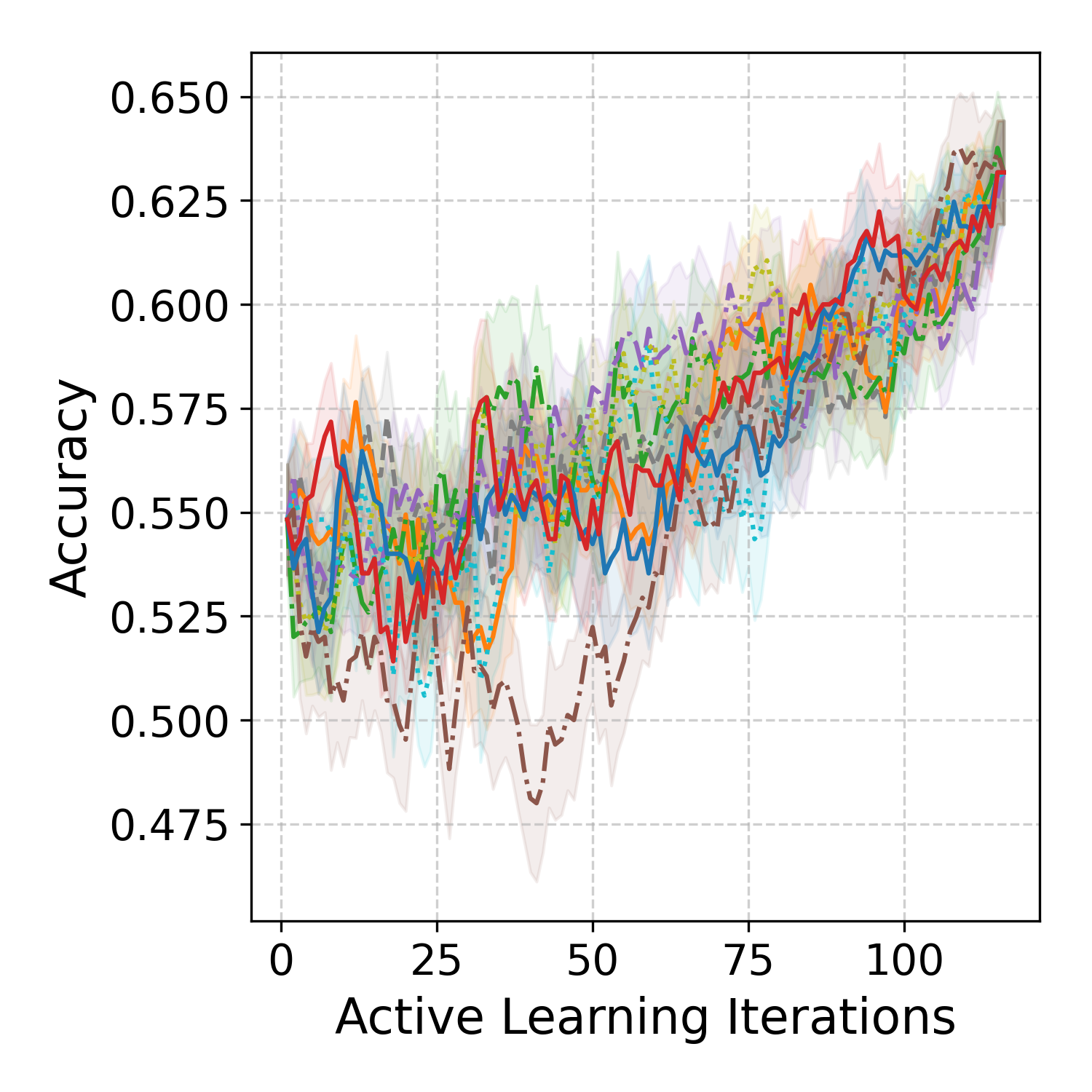}
        \caption{MONK-2}
    \end{subfigure}
    \hfill
    \begin{subfigure}[b]{0.19\textwidth}
        \includegraphics[width=\linewidth]{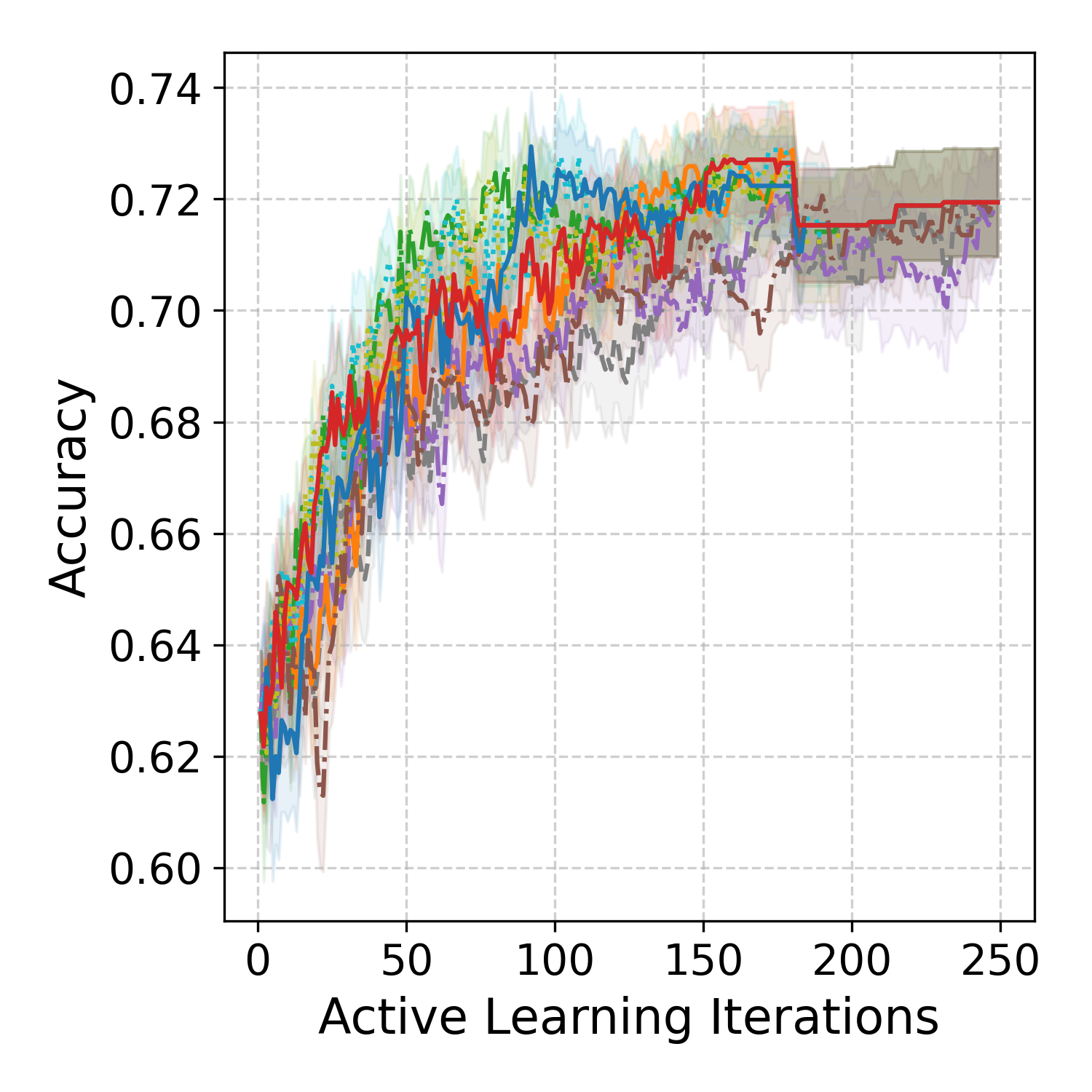}
        \caption{Primary Tumor}
    \end{subfigure}
    \hfill
    \begin{subfigure}[b]{0.19\textwidth}
        \includegraphics[width=\linewidth]{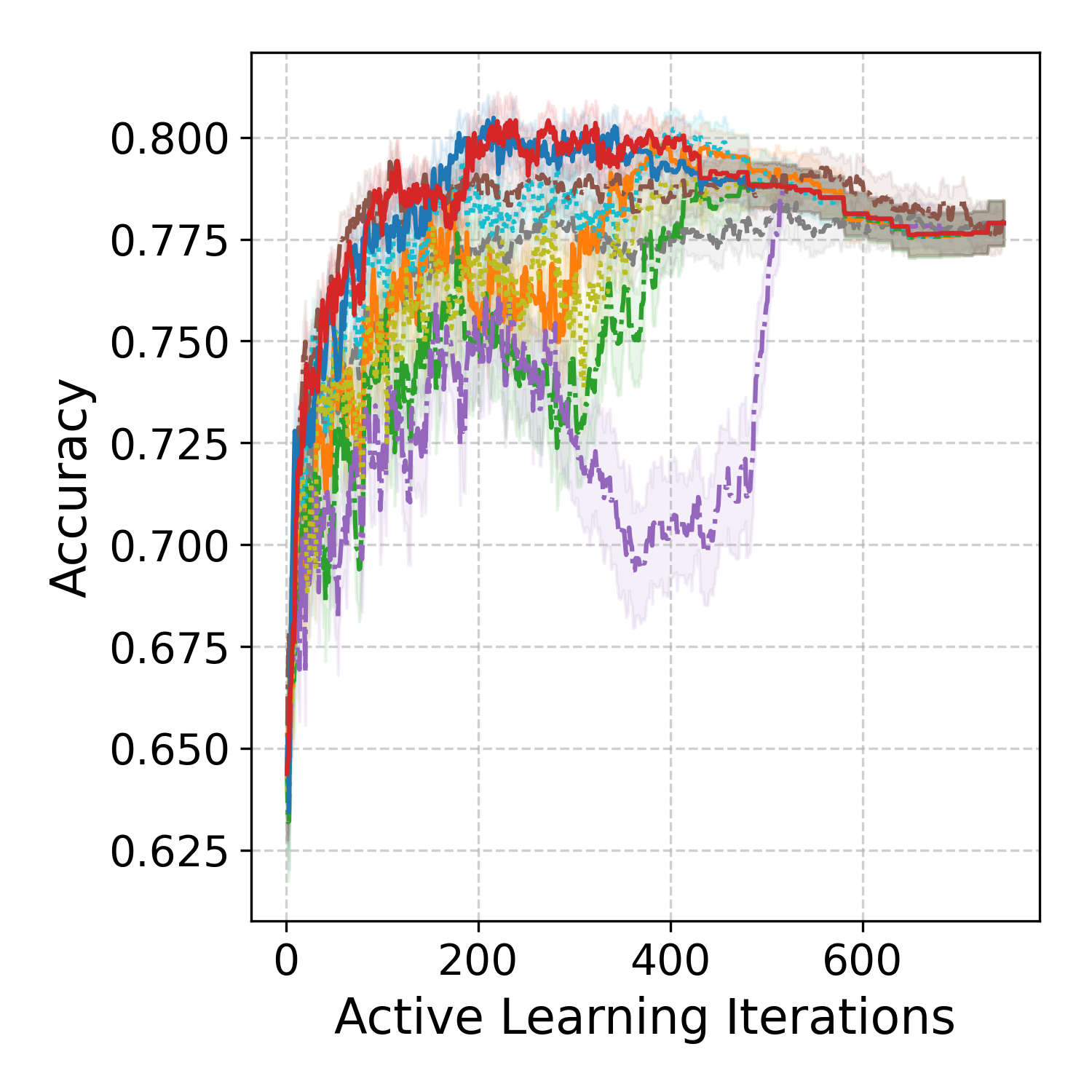}
        \caption{Tic-Tac-Toe}
    \end{subfigure}
    \hfill
    \begin{subfigure}[b]{0.19\textwidth}
        \includegraphics[width=\linewidth]{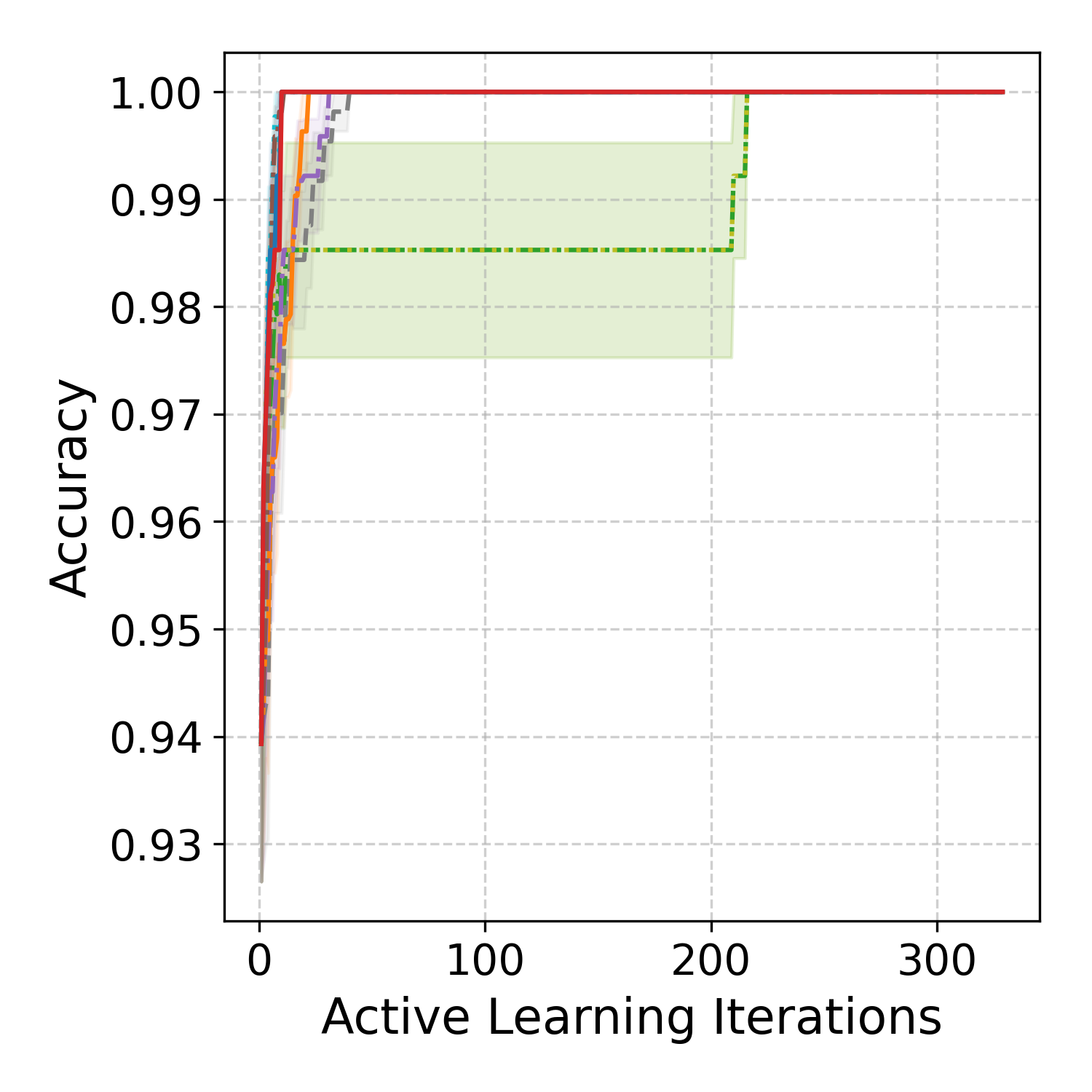}
        \caption{Vote}
    \end{subfigure}
    \hfill
    \begin{subfigure}[b]{0.19\textwidth}
        \includegraphics[width=\linewidth]{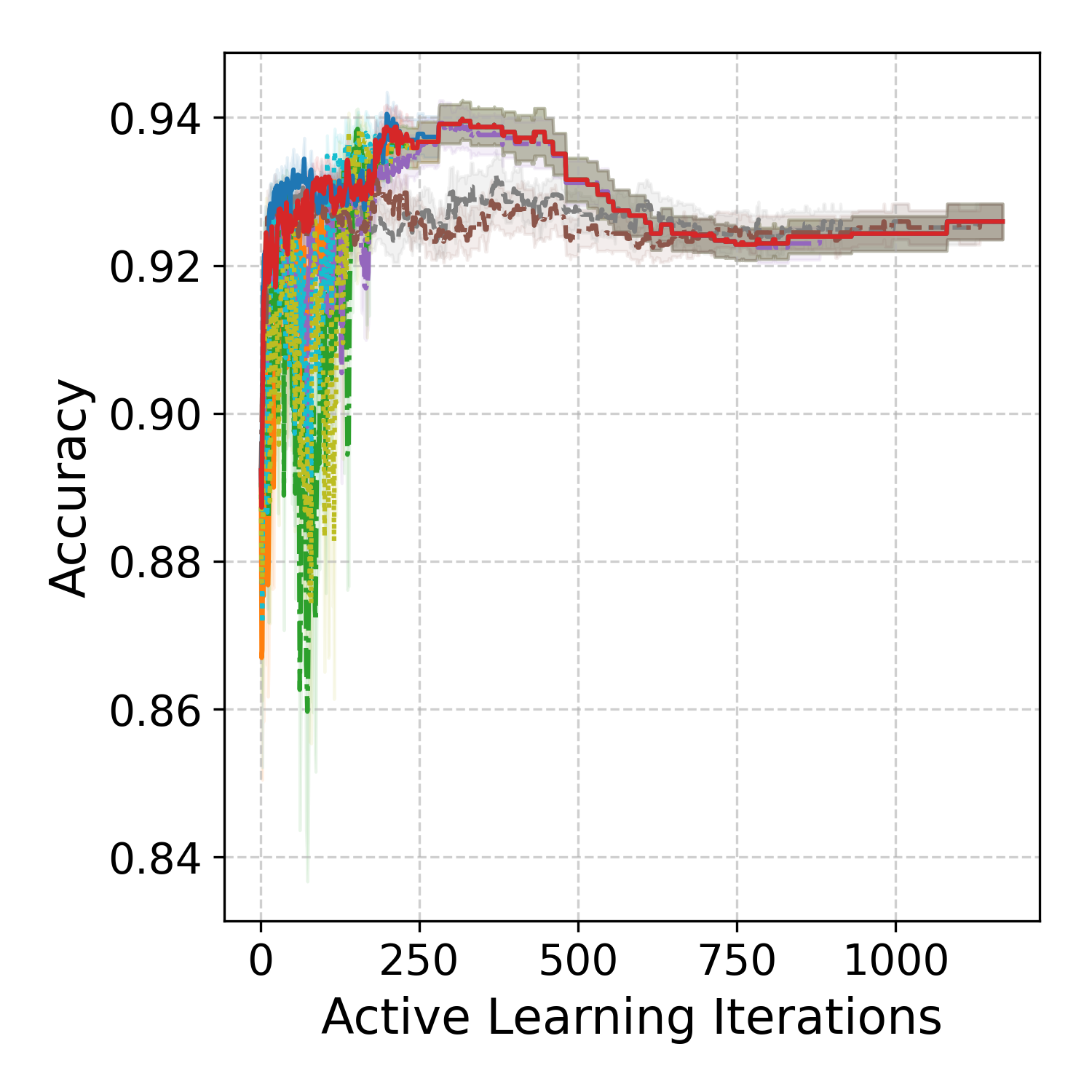}
        \caption{Yeast}
    \end{subfigure}

    \vspace{0.5em}
    \centering
    \includegraphics[width=0.60\linewidth]{upload_all_files/study1_main_AL_results/benchmark_legend.png}

    \caption{\textbf{Active Learning Benchmarks.} Accuracy traces across 20 real-world and structured datasets. UNREAL (Weighted and Uniform) consistently match or exceed baseline performance, demonstrating superior sample efficiency particularly in the early stages of the labeling process.}
    \label{fig:BenchmarkGrid_Accuracy}
\end{figure*}

\begin{table*}[htbp]
    \centering
    \scriptsize
    \setlength{\tabcolsep}{4pt}
    \begin{tabular}{l rrrrrrrrrr}
        \toprule
         & \multicolumn{3}{c}{\textbf{Baselines}} & \multicolumn{3}{c}{\textbf{QBC-RF}} & \multicolumn{2}{c}{\textbf{Weighted RF}} & \multicolumn{2}{c}{\textbf{REAL (Proposed)}} \\
         \cmidrule(lr){2-4} \cmidrule(lr){5-7} \cmidrule(lr){8-9} \cmidrule(lr){10-11}
        \textbf{Dataset} & Random & Uncertainty & Coreset & F=3 & F=Sqrt & F=All & F=Sqrt & F=All & UNREAL & BREAL \\ \midrule
        Synth. Xor Baseline & 137.4 & 80.7 & 98.0 & 162.2 & 164.7 & 159.4 & 137.5 & 140.3 & 2960.0 & 3035.0 \\
        Synth. Xor Phi 05 & 168.2 & 113.2 & 119.7 & 196.0 & 196.0 & 197.4 & 176.3 & 177.1 & 3014.6 & 3050.2 \\
        Synth. Xor Phi 10 & 186.3 & 135.2 & 135.0 & 215.9 & 215.4 & 217.8 & 195.6 & 192.3 & 3052.7 & 3043.3 \\
        Synth. Xor Phi 25 & 259.4 & 203.3 & 210.6 & 286.7 & 284.1 & 287.1 & 262.8 & 266.1 & 2940.8 & 2875.1 \\
        Synth. Xor Phi 45 & 294.1 & 238.5 & 246.7 & 318.3 & 315.3 & 321.2 & 293.2 & 299.3 & 1674.9 & 1661.3 \\
        Synth. Xor Alpha 50 & 245.5 & 195.2 & 197.1 & 278.0 & 277.1 & 286.5 & 263.5 & 265.4 & 1771.7 & 1672.2 \\
        Synth. Xor Alpha 25 & 201.8 & 155.3 & 152.4 & 229.6 & 231.0 & 243.4 & 220.9 & 227.1 & 3048.2 & 2955.7 \\
        Synth. Xor Alpha 75 & 249.9 & 194.7 & 201.2 & 288.2 & 293.9 & 294.3 & 270.7 & 272.7 & 1567.8 & 1588.5 \\
        Synth. Xor Alpha 100 & 225.1 & 176.0 & 177.6 & 279.5 & 281.4 & 286.6 & 257.9 & 263.3 & 1522.2 & 1576.3 \\
        \midrule
        \textbf{MEDIAN (Synth.)} & \textbf{225.1} & \textbf{176.0} & \textbf{177.6} & \textbf{278.0} & \textbf{277.1} & \textbf{286.5} & \textbf{257.9} & \textbf{263.3} & \textbf{2940.8} & \textbf{2875.1} \\
        \midrule
        Anneal & 145.3 & 73.3 & 83.7 & 201.0 & 195.7 & 196.2 & 169.3 & 166.6 & 1851.0 & 1536.5 \\
        Bank Marketing & 1100.6 & 2417.4 & 1922.7 & 3398.2 & 3361.2 & 3773.3 & 3148.9 & 3423.7 & 17800.4 & 13258.9 \\
        Bar-7 & 428.2 & 218.1 & 272.3 & 522.4 & 522.4 & 535.4 & 460.9 & 474.1 & 5423.0 & 5576.8 \\
        Breast Cancer Wisconsin & 106.3 & 32.9 & 33.7 & 149.1 & 149.6 & 151.0 & 123.6 & 123.8 & 1357.8 & 1358.7 \\
        Car Evaluation & 506.6 & 382.0 & 325.7 & 653.6 & 649.5 & 652.0 & 576.0 & 586.4 & 6155.1 & 5711.1 \\
        Cheap Restaurant & 1005.7 & 1116.7 & 794.3 & 1212.0 & 1221.4 & 1243.4 & 1115.6 & 1102.9 & 9185.1 & 8831.7 \\
        Coffee House & 1998.1 & 1740.7 & 1897.1 & 2507.7 & 2546.4 & 2479.2 & 2328.6 & 2349.0 & 22381.6 & 17902.5 \\
        Compas & 4557.0 & 3674.3 & 5095.2 & 4985.1 & 4934.9 & 5127.7 & 4357.3 & 4618.7 & 27084.6 & 27142.3 \\
        Expensive Restaurant & 360.2 & 232.8 & 220.3 & 466.4 & 467.9 & 481.9 & 416.9 & 423.7 & 4301.5 & 4227.2 \\
        Fico & 25019.3 & 24959.6 & 33920.9 & 24983.4 & 25006.9 & 25097.2 & 23517.2 & 24598.6 & 104613.6 & 69284.3 \\
        Haberman & 43.1 & 11.9 & 12.7 & 60.7 & 61.3 & 61.4 & 49.3 & 49.7 & 612.6 & 609.4 \\
        Hepatitis & 122.4 & 109.8 & 115.3 & 135.2 & 135.4 & 135.6 & 127.7 & 130.2 & 478.3 & 488.8 \\
        Hypothyroid & 650.8 & 342.0 & 708.2 & 986.8 & 1087.5 & 1305.1 & 992.5 & 1170.0 & 7461.1 & 7188.5 \\
        Kr-Vs-Kp & 19590.7 & 17263.5 & 18589.2 & 18063.6 & 18646.9 & 18434.5 & 17522.9 & 16722.3 & 30211.2 & 29824.1 \\
        Lymph & 20.3 & 7.7 & 8.0 & 27.7 & 28.5 & 29.2 & 22.5 & 24.6 & 235.2 & 241.2 \\
        Monk2 & 26.6 & 10.7 & 11.1 & 34.9 & 34.6 & 35.2 & 29.4 & 28.7 & 393.1 & 429.3 \\
        Primary-Tumor & 79.7 & 46.0 & 51.3 & 102.4 & 101.6 & 102.9 & 89.7 & 89.8 & 848.5 & 852.5 \\
        Tic-Tac-Toe & 355.4 & 270.1 & 220.4 & 434.1 & 443.6 & 445.9 & 396.7 & 398.9 & 3329.4 & 3673.9 \\
        Vote & 91.4 & 44.7 & 52.6 & 118.1 & 118.8 & 119.6 & 104.6 & 102.9 & 1296.3 & 1326.8 \\
        Yeast & 5286.4 & 6831.6 & 5341.1 & 7043.2 & 7226.2 & 7216.1 & 7021.3 & 7245.5 & 12496.0 & 12705.6 \\
        \midrule
        \textbf{MEDIAN (Real)} & \textbf{394.2} & \textbf{251.5} & \textbf{246.4} & \textbf{494.4} & \textbf{495.2} & \textbf{508.6} & \textbf{438.9} & \textbf{448.9} & \textbf{4862.3} & \textbf{4902.0} \\
        \bottomrule
    \end{tabular}
    \caption{Median experiment runtime (seconds) across 25 simulation seeds. UNREAL and BREAL denote the standard and Bayesian variants of the UNREAL algorithm, respectively.}
    \label{tab:RuntimeComparison}
\end{table*}

Table \ref{tab:RuntimeComparison} presents the median runtime across 25 independent simulations for each dataset. As expected, the REAL variants (UNREAL and BREAL) incur a higher computational cost compared to randomized ensembles. This overhead is a direct consequence of the exhaustive version-space characterization performed by the SORTD algorithm at each iteration.

However, this increased computation is justified by the fundamental objective of active learning: minimizing human labeling cost. In high-stakes applications like medical diagnosis or legal discovery, the cost of a single expert-provided label far exceeds the cost of CPU minutes. In such contexts, the computational delay is negligible compared to the substantial gains in sample efficiency and the robustness to noise achieved by resolving the true epistemic uncertainty of the version space.

\begin{figure*}[!t] 
\centering
    \begin{subfigure}[b]{0.19\textwidth}
        \includegraphics[width=\linewidth]{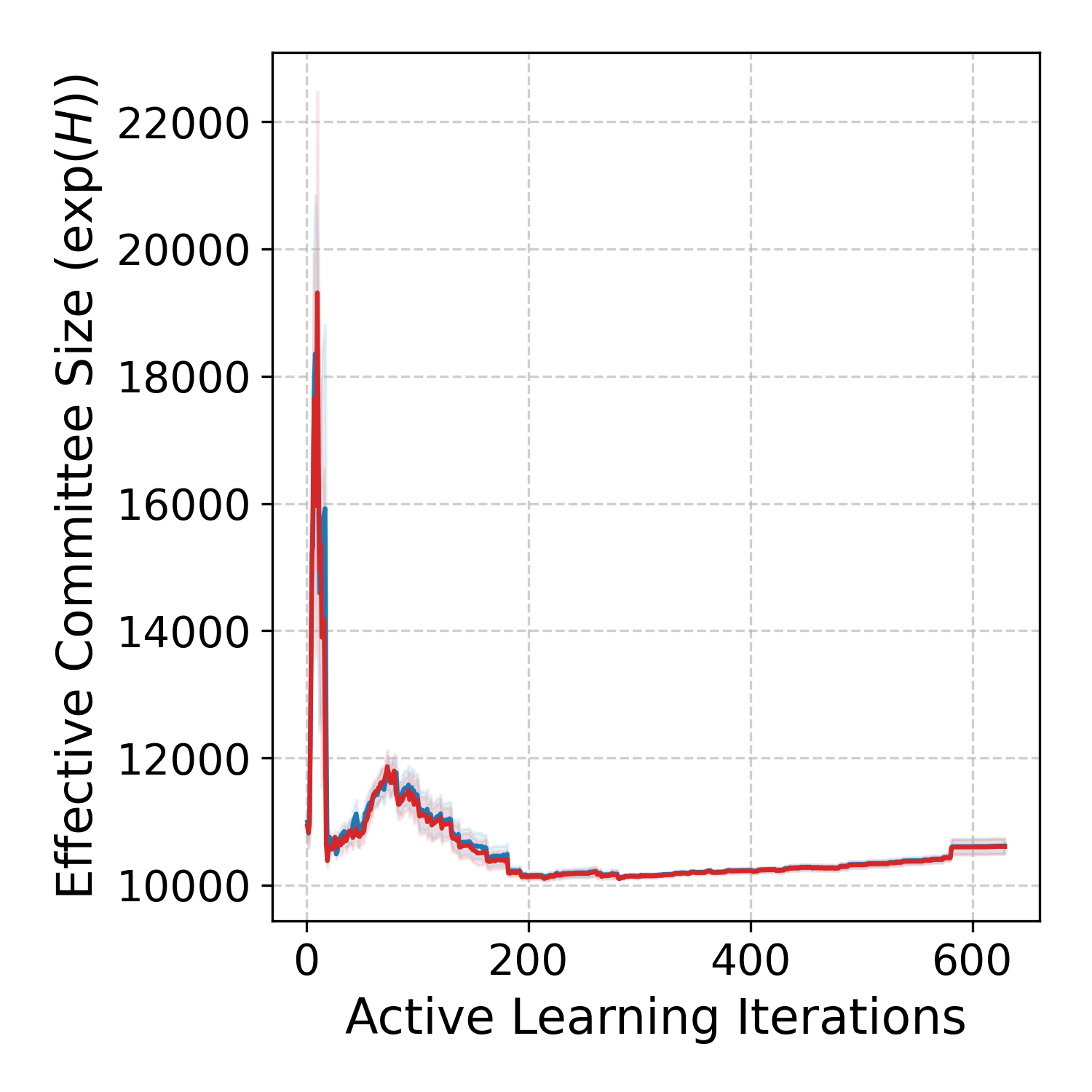}
        \caption{Anneal}
    \end{subfigure}
    \hfill
    \begin{subfigure}[b]{0.19\textwidth}
        \includegraphics[width=\linewidth]{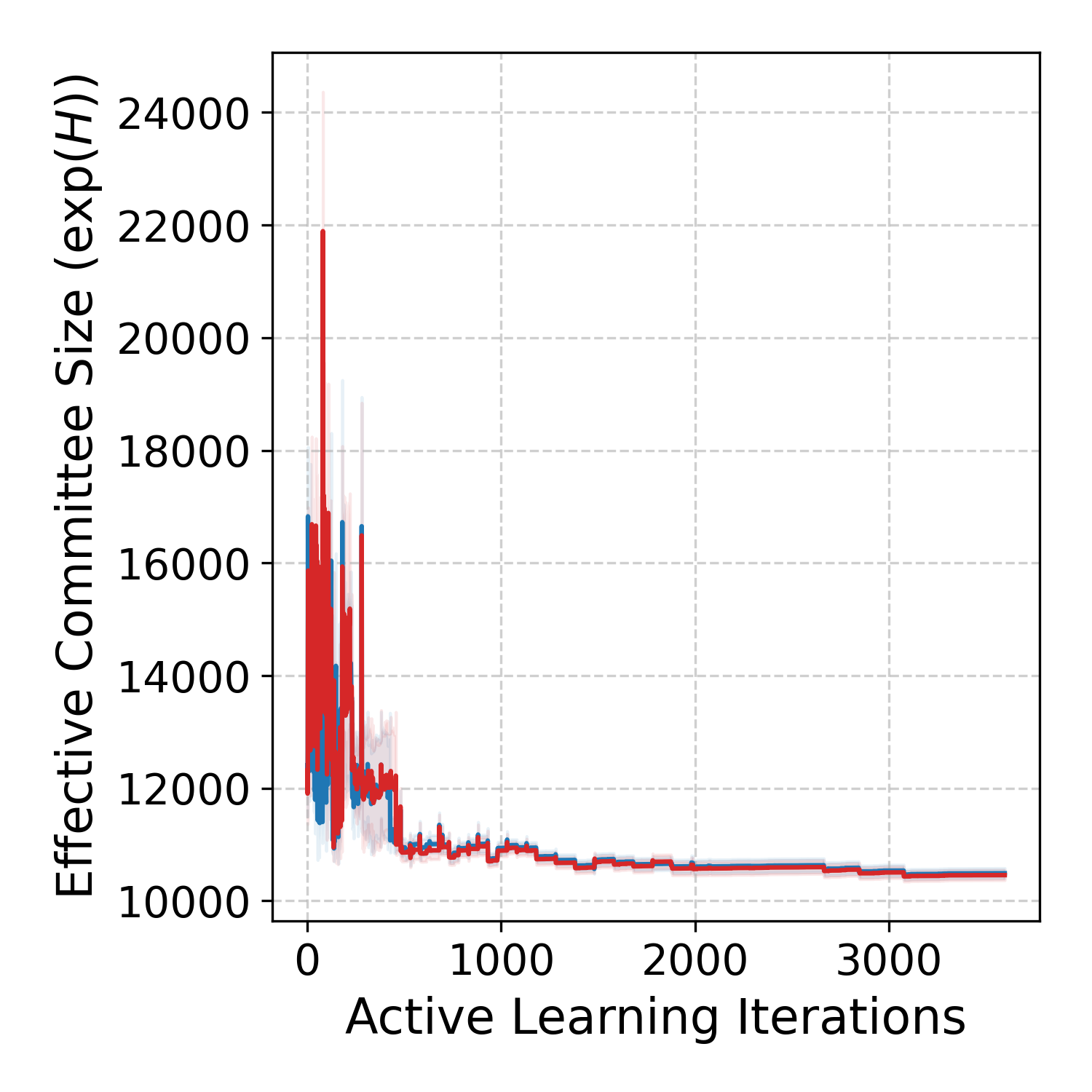}
        \caption{Bank Marketing}
    \end{subfigure}
    \hfill
    \begin{subfigure}[b]{0.19\textwidth}
        \includegraphics[width=\linewidth]{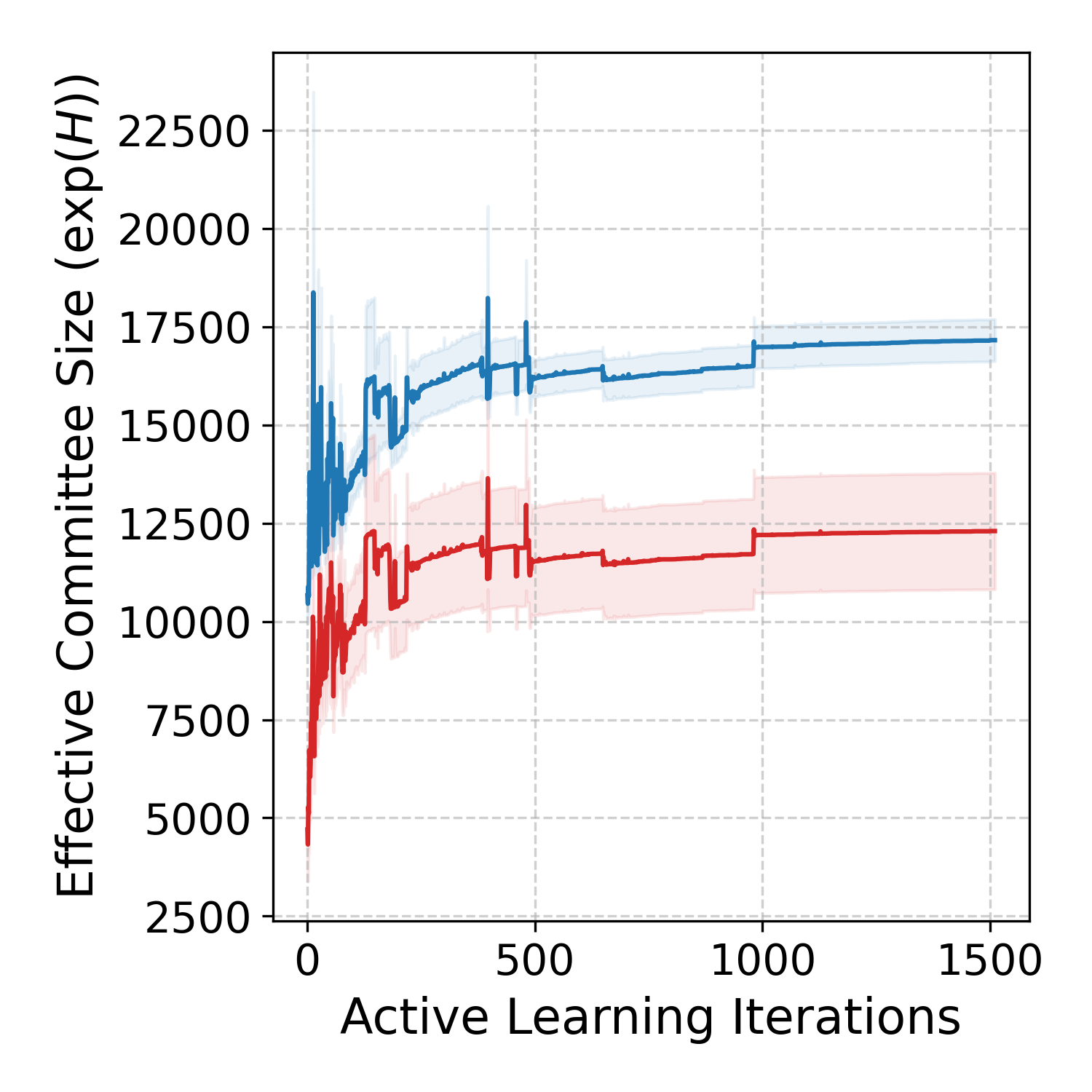}
        \caption{Bar-7}
    \end{subfigure}
    \hfill
    \begin{subfigure}[b]{0.19\textwidth}
        \includegraphics[width=\linewidth]{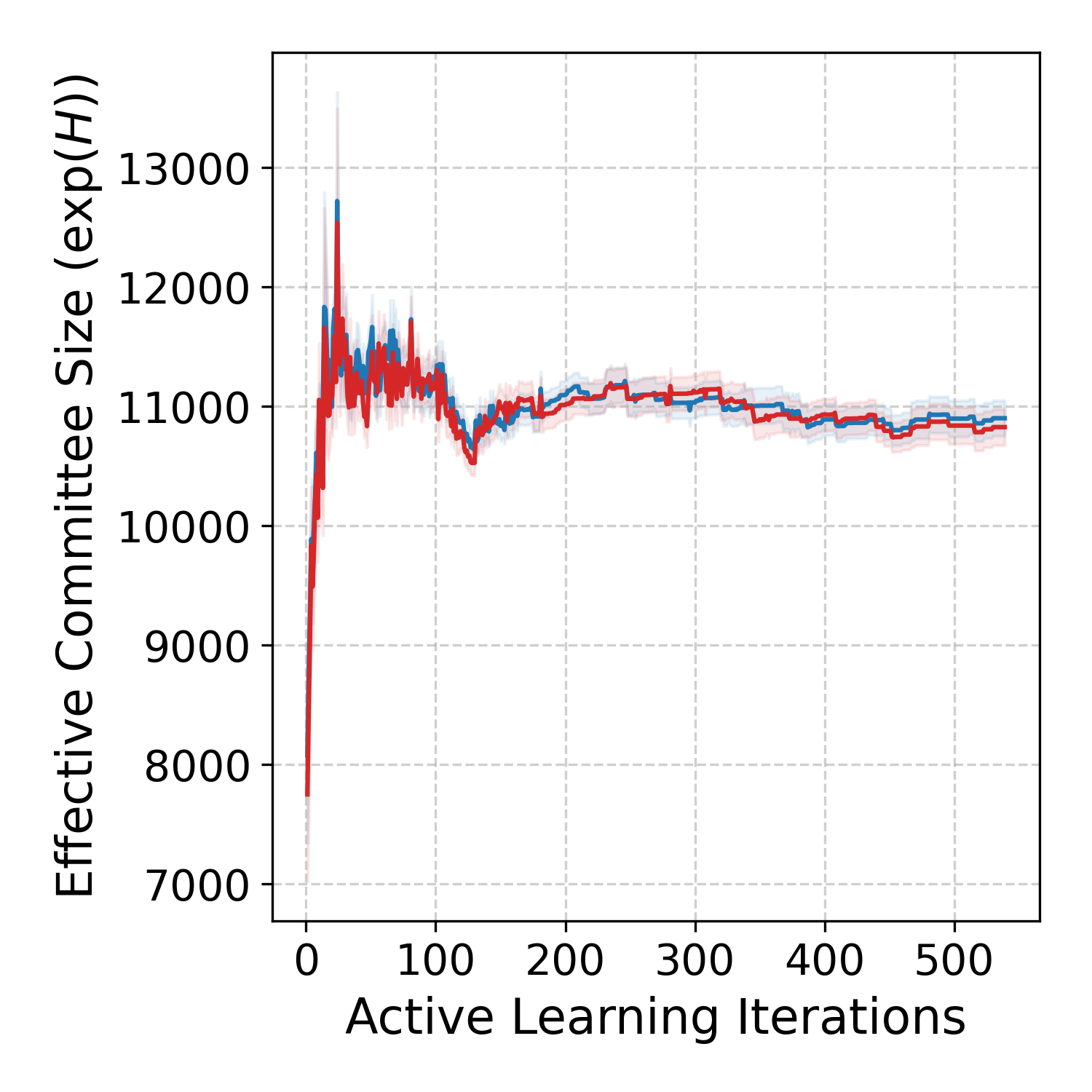}
        \caption{Breast Cancer WI}
    \end{subfigure}
    \hfill
    \begin{subfigure}[b]{0.19\textwidth}
        \includegraphics[width=\linewidth]{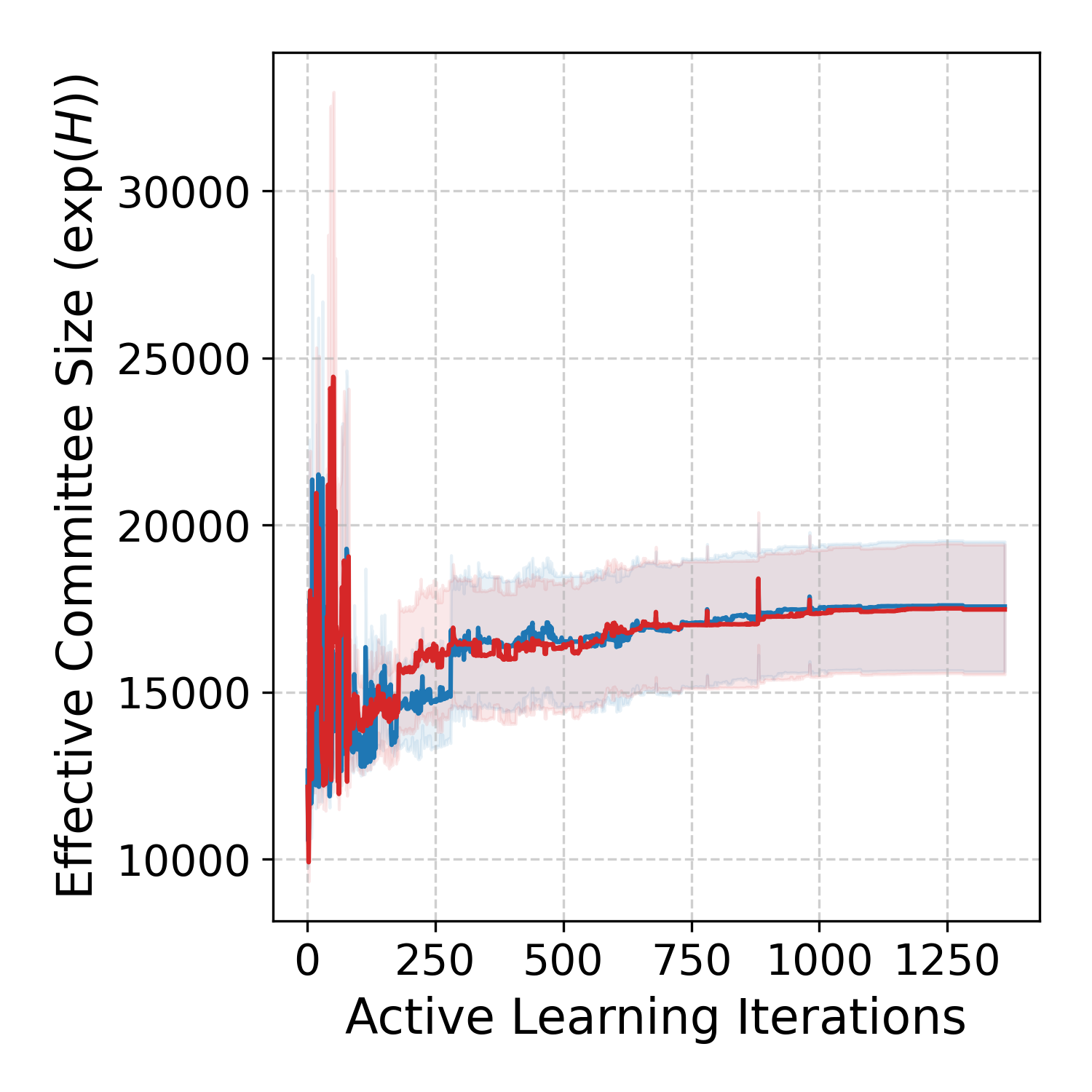}
        \caption{Car Evaluation}
    \end{subfigure}

    \vspace{0.2em}

    \begin{subfigure}[b]{0.19\textwidth}
        \includegraphics[width=\linewidth]{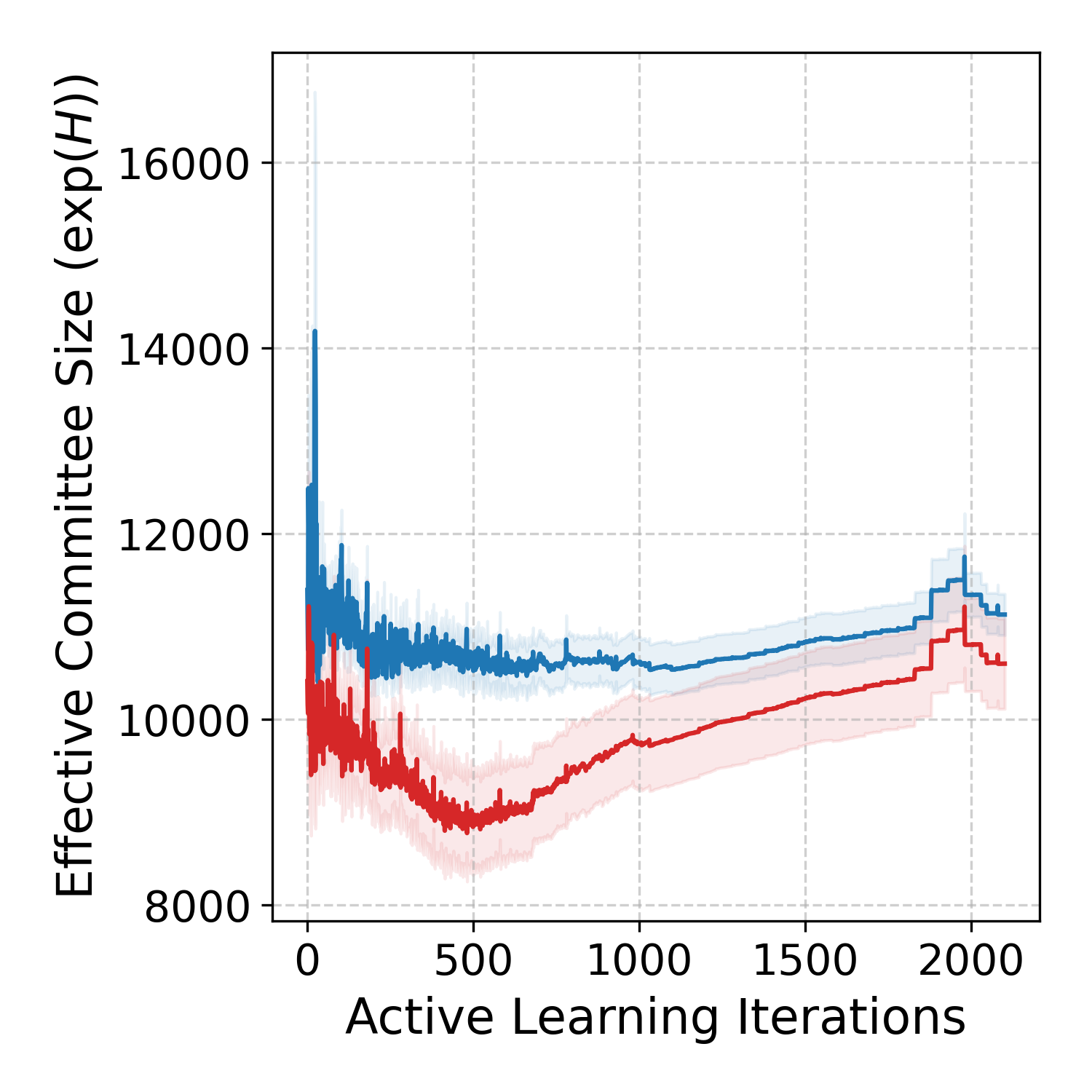}
        \caption{Cheap Restaurant}
    \end{subfigure}
    \hfill
    \begin{subfigure}[b]{0.19\textwidth}
        \includegraphics[width=\linewidth]{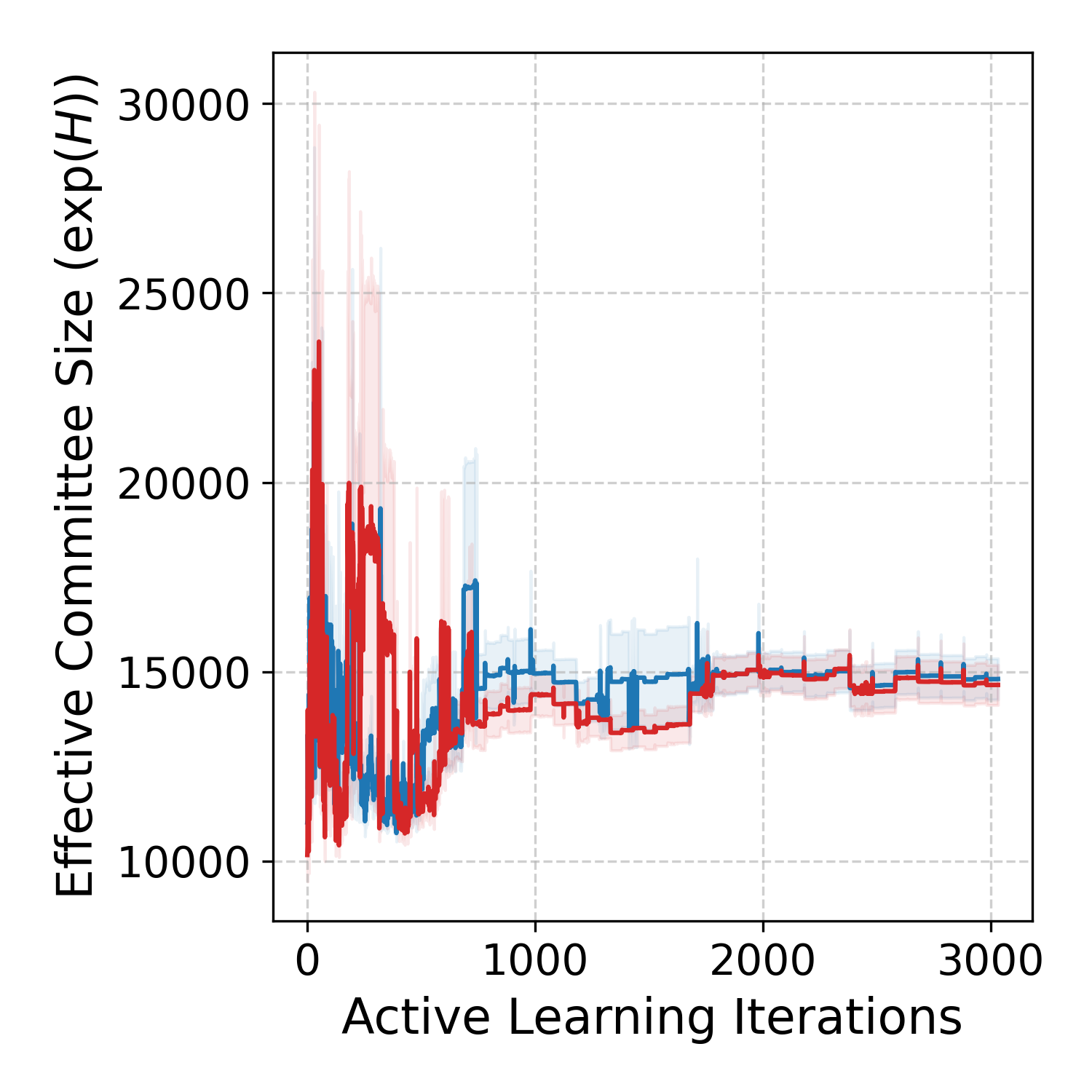}
        \caption{Coffee House}
    \end{subfigure}
    \hfill
    \begin{subfigure}[b]{0.19\textwidth}
        \includegraphics[width=\linewidth]{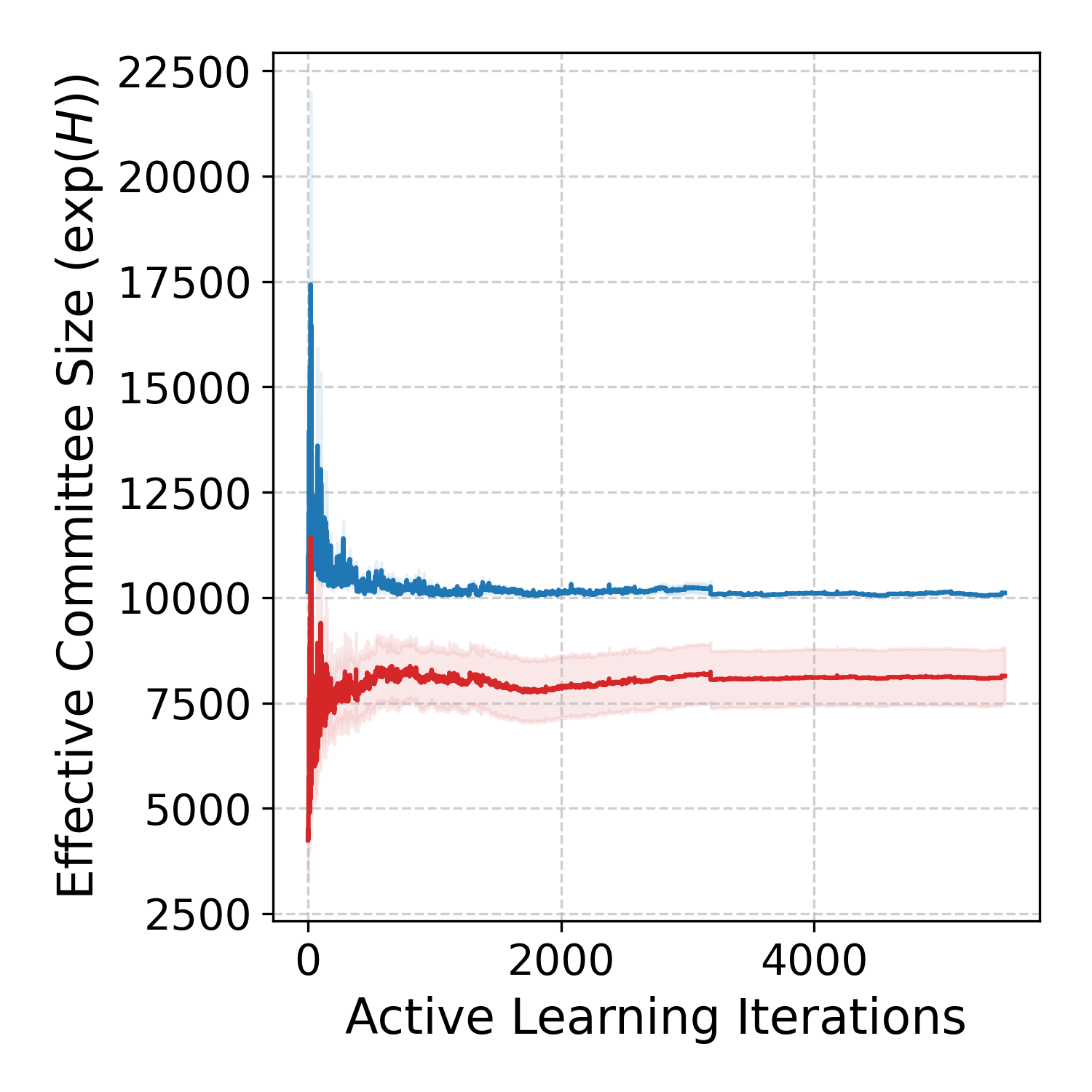}
        \caption{COMPAS}
    \end{subfigure}
    \hfill
    \begin{subfigure}[b]{0.19\textwidth}
        \includegraphics[width=\linewidth]{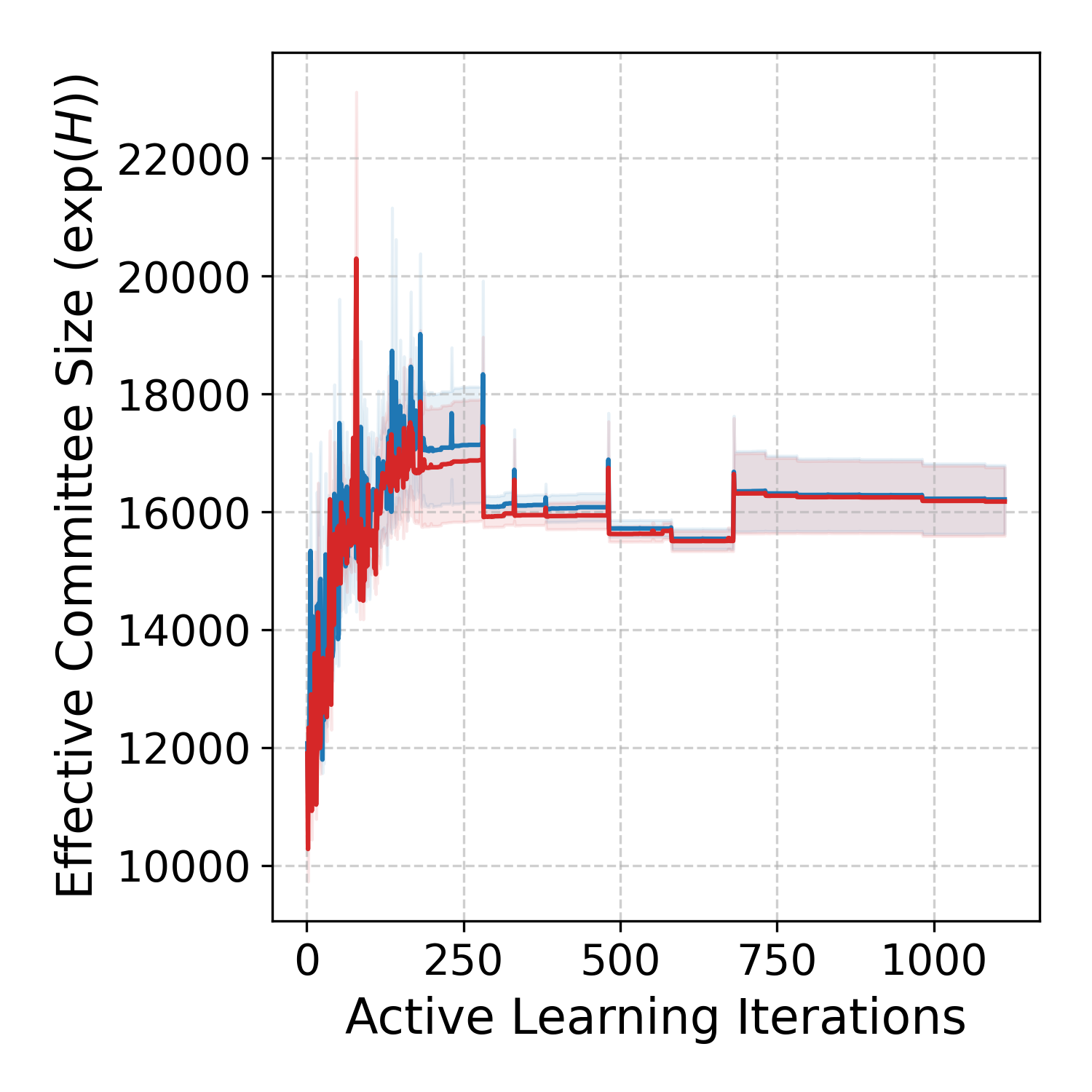}
        \caption{Expensive Restaurant}
    \end{subfigure}
    \hfill
    \begin{subfigure}[b]{0.19\textwidth}
        \includegraphics[width=\linewidth]{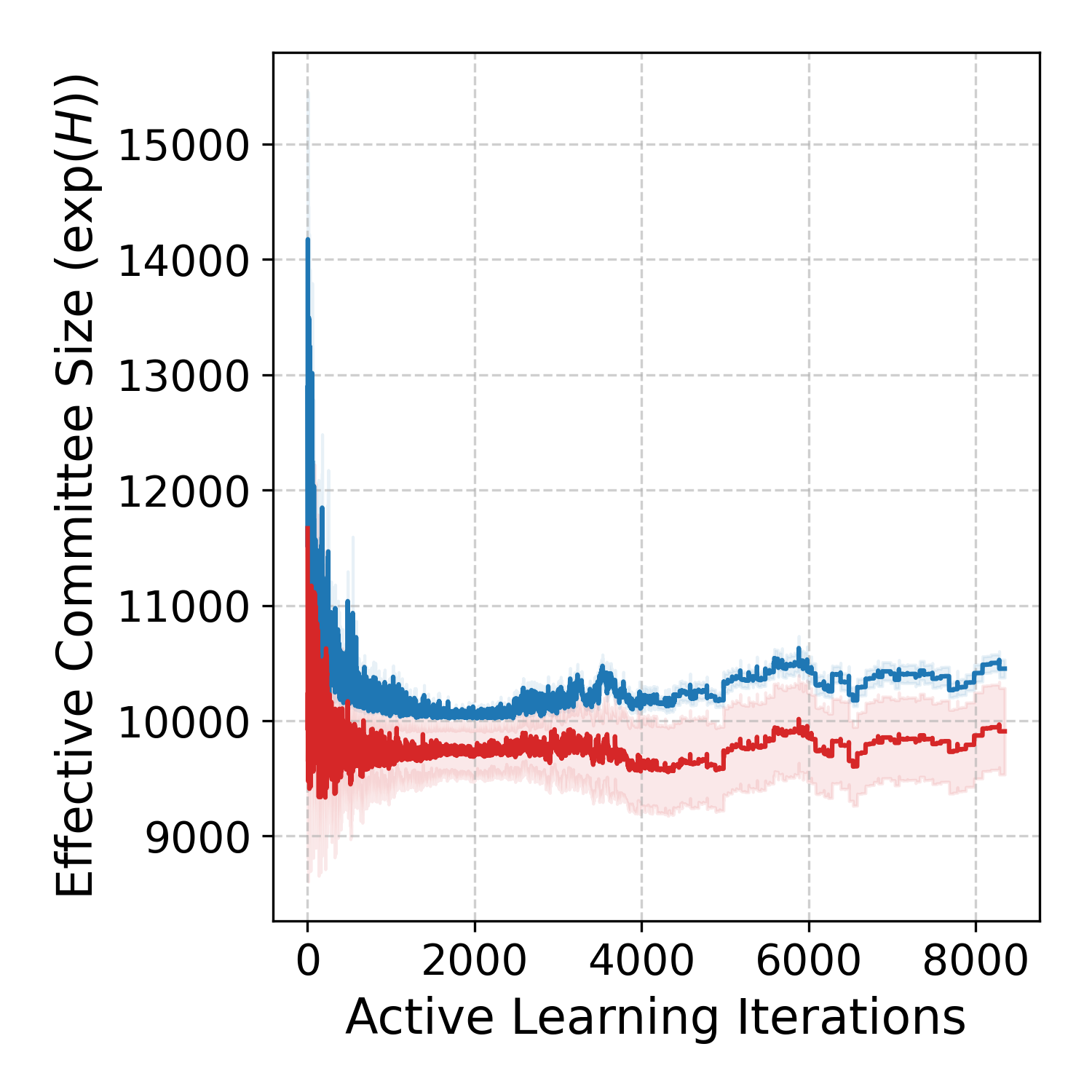}
        \caption{FICO}
    \end{subfigure}

    \vspace{0.2em}

    \begin{subfigure}[b]{0.19\textwidth}
        \includegraphics[width=\linewidth]{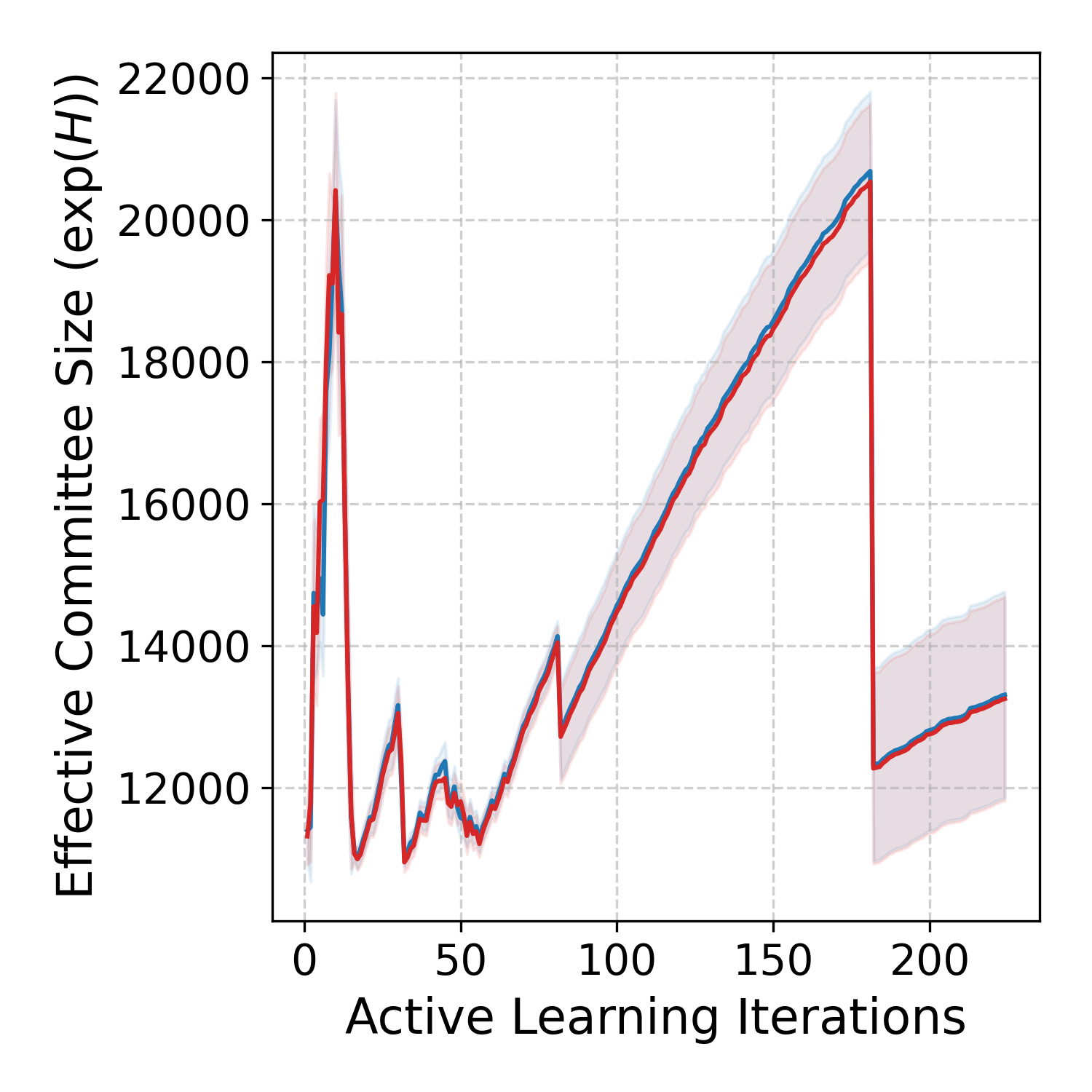}
        \caption{Haberman}
    \end{subfigure}
    \hfill
    \begin{subfigure}[b]{0.19\textwidth}
        \includegraphics[width=\linewidth]{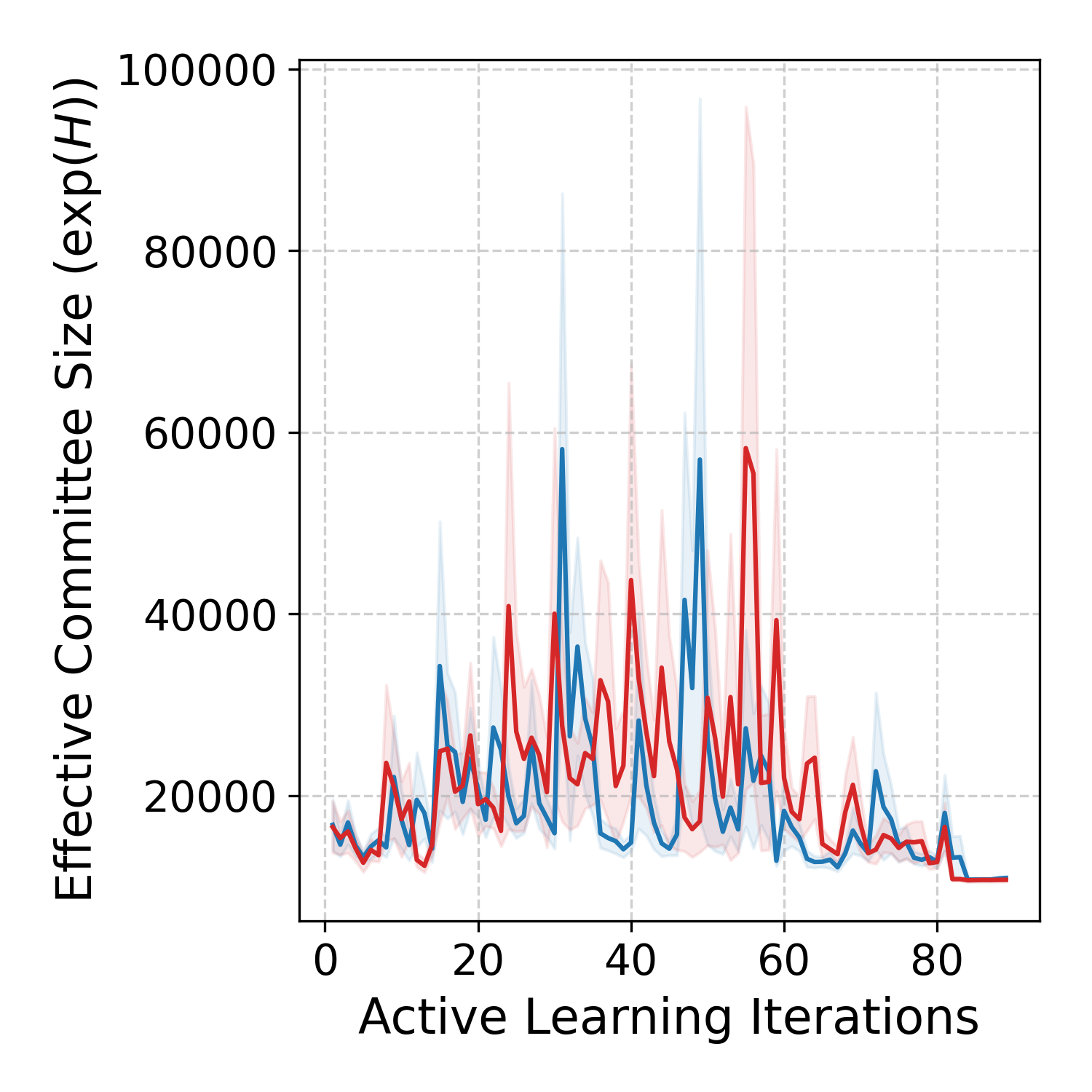}
        \caption{Hepatitis}
    \end{subfigure}
    \hfill
    \begin{subfigure}[b]{0.19\textwidth}
        \includegraphics[width=\linewidth]{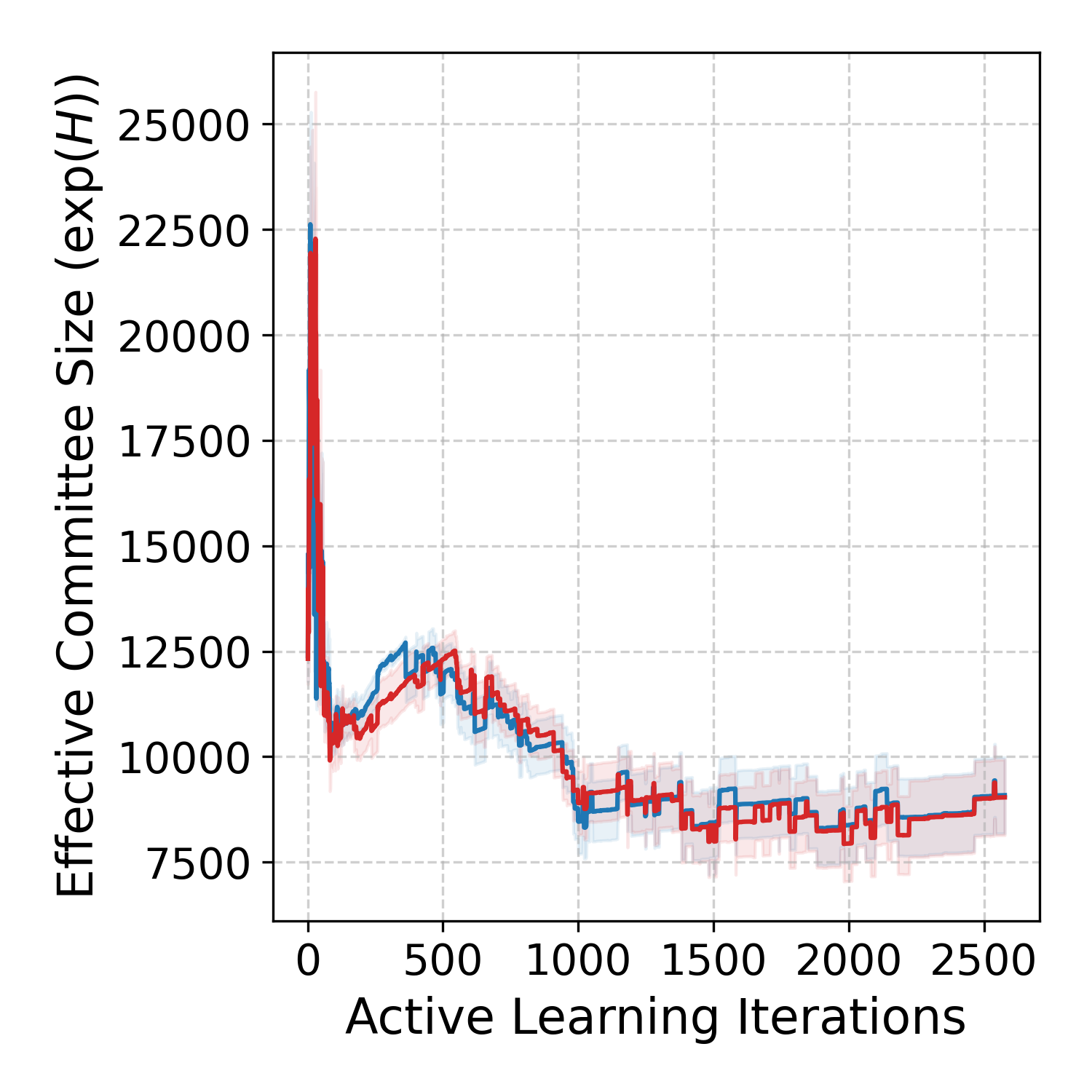}
        \caption{Hypothyroid}
    \end{subfigure}
    \hfill
    \begin{subfigure}[b]{0.19\textwidth}
        \includegraphics[width=\linewidth]{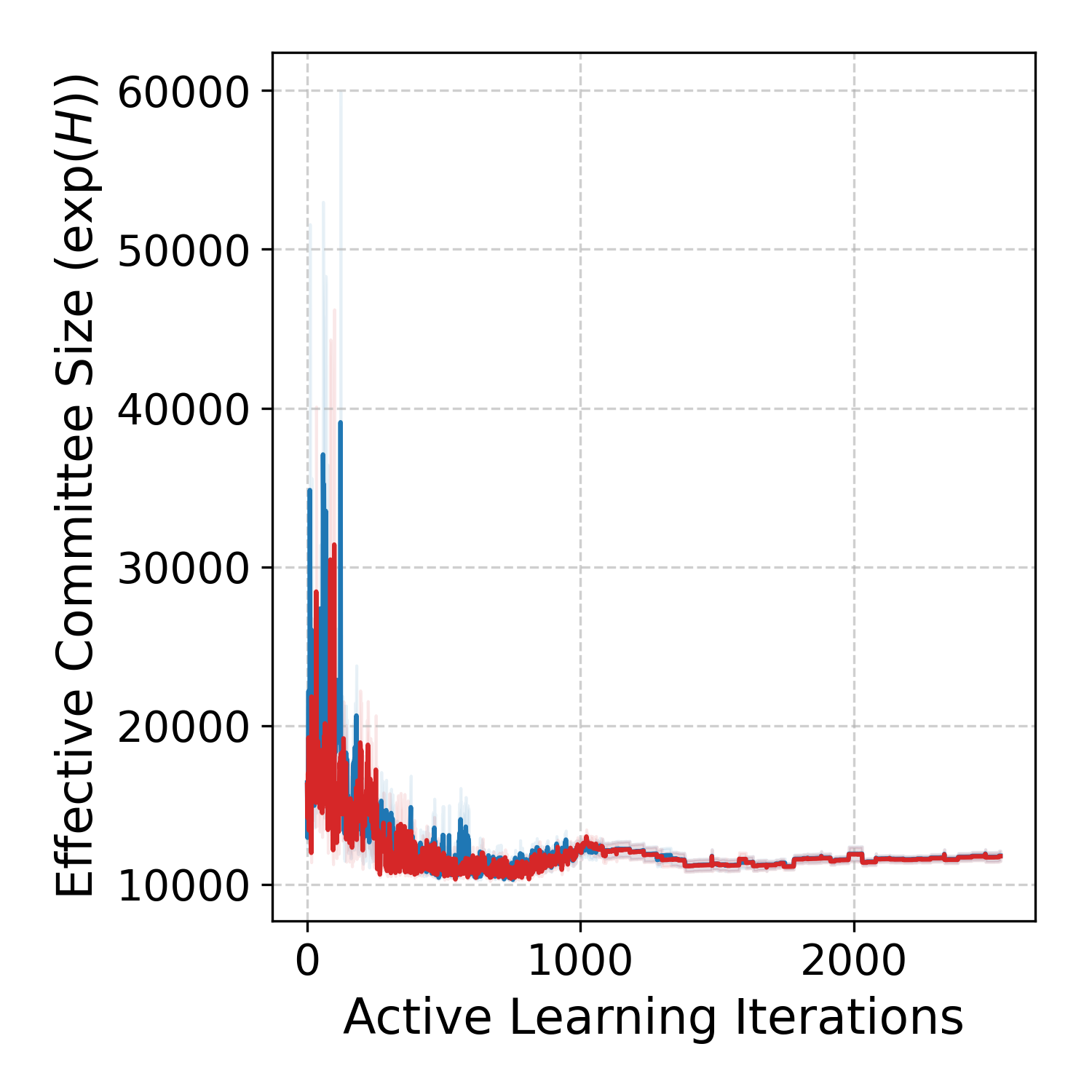}
        \caption{Kr-vs-Kp}
    \end{subfigure}
    \hfill
    \begin{subfigure}[b]{0.19\textwidth}
        \includegraphics[width=\linewidth]{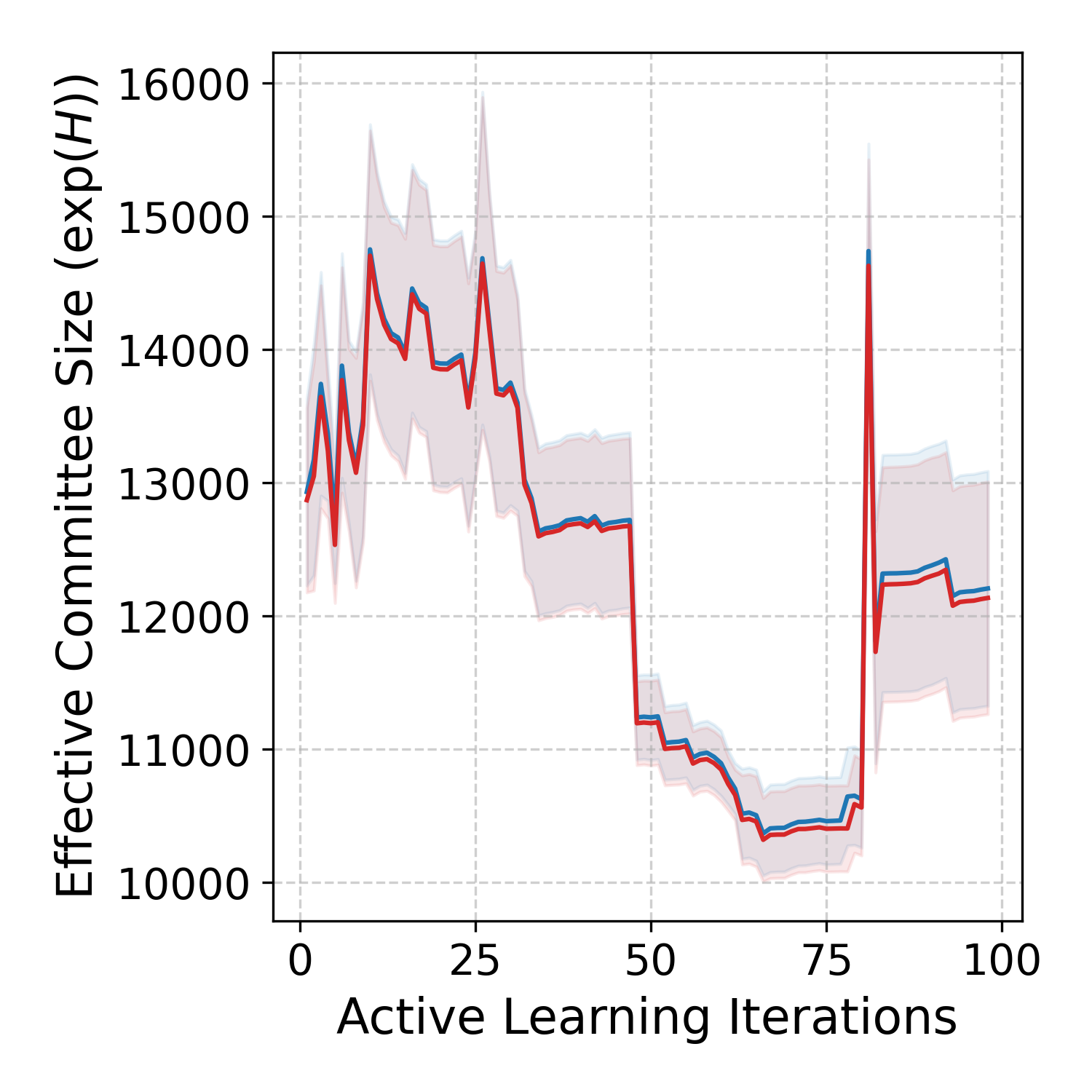}
        \caption{Lymphography}
    \end{subfigure}

    \vspace{0.2em}

    \begin{subfigure}[b]{0.19\textwidth}
        \includegraphics[width=\linewidth]{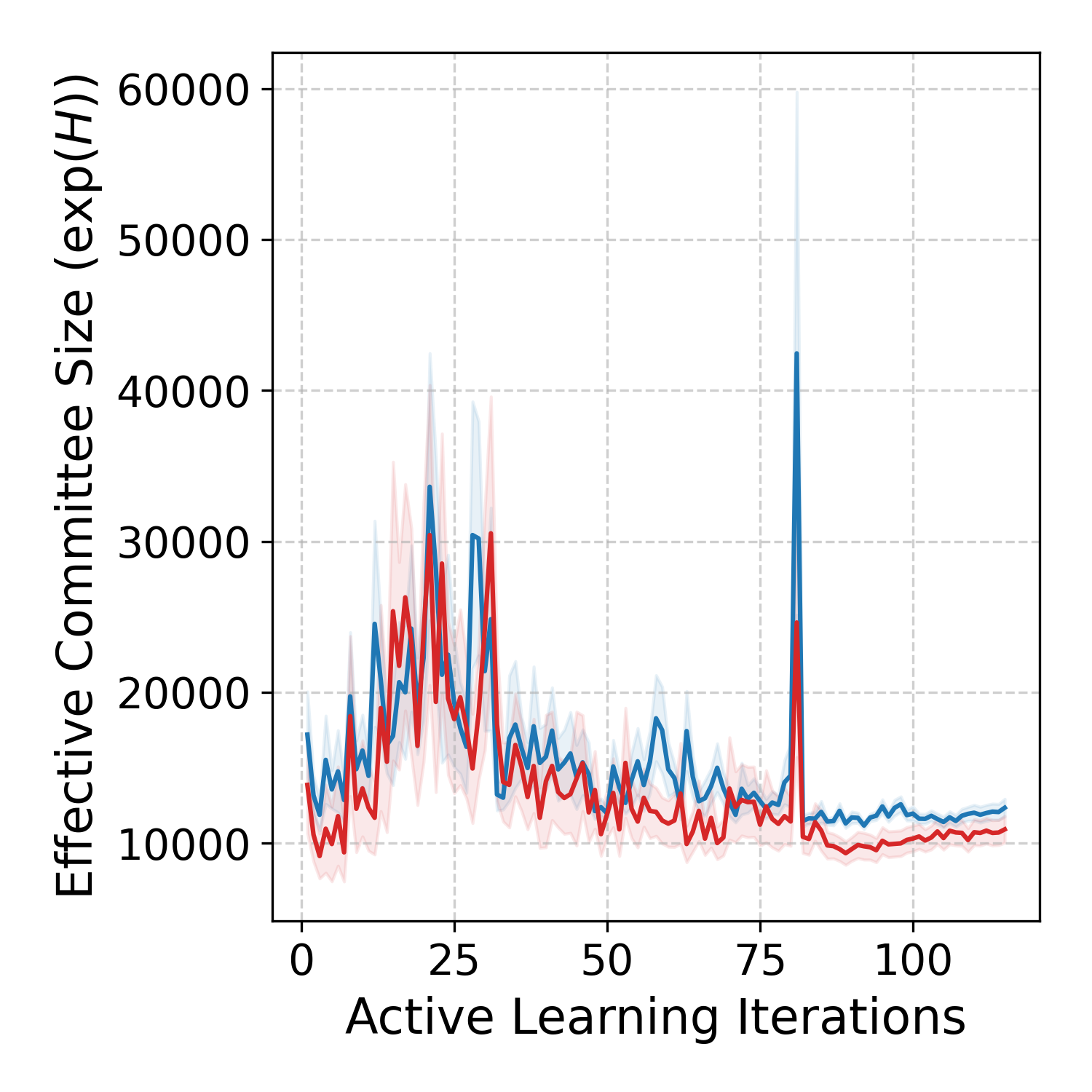}
        \caption{MONK-2}
    \end{subfigure}
    \hfill
    \begin{subfigure}[b]{0.19\textwidth}
        \includegraphics[width=\linewidth]{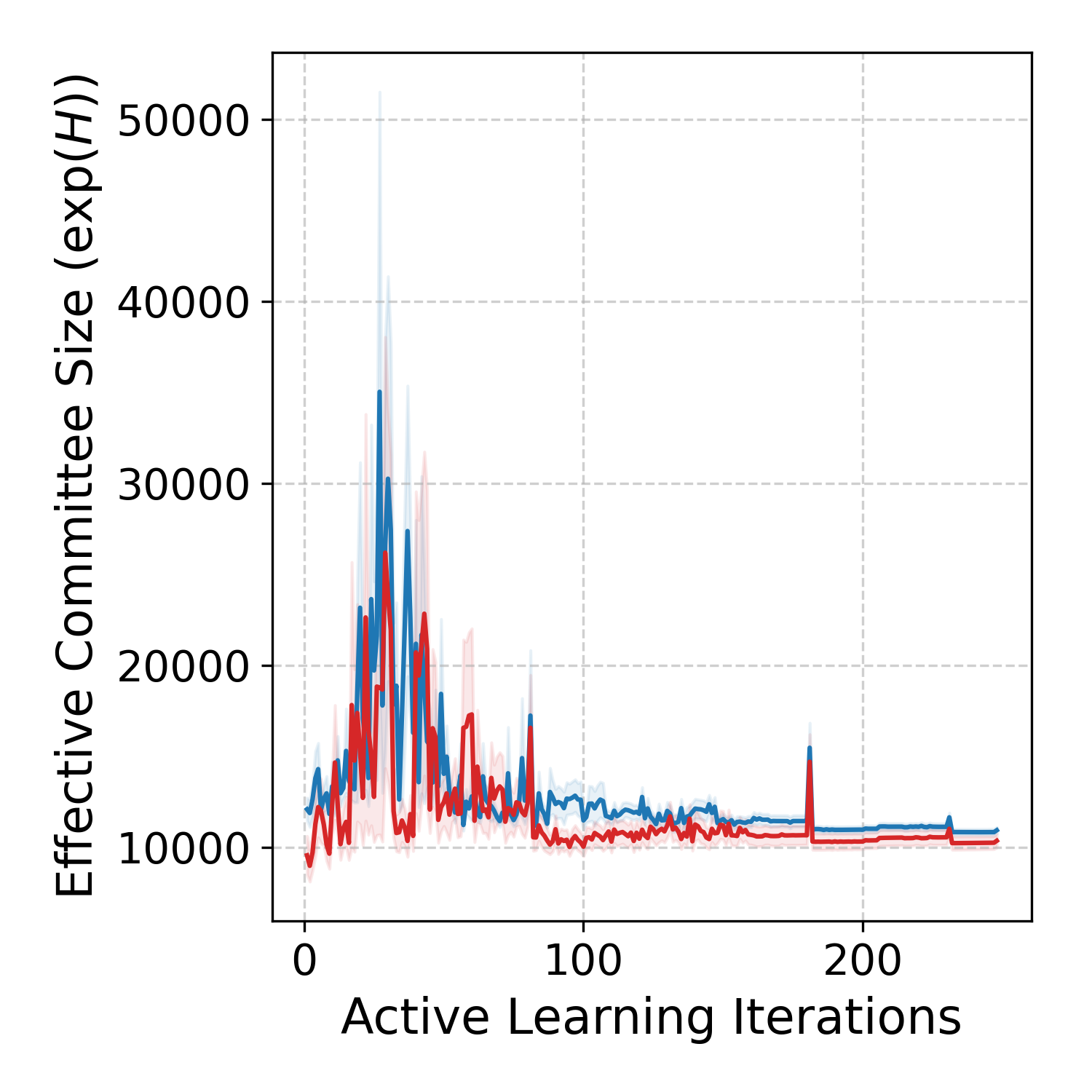}
        \caption{Primary Tumor}
    \end{subfigure}
    \hfill
    \begin{subfigure}[b]{0.19\textwidth}
        \includegraphics[width=\linewidth]{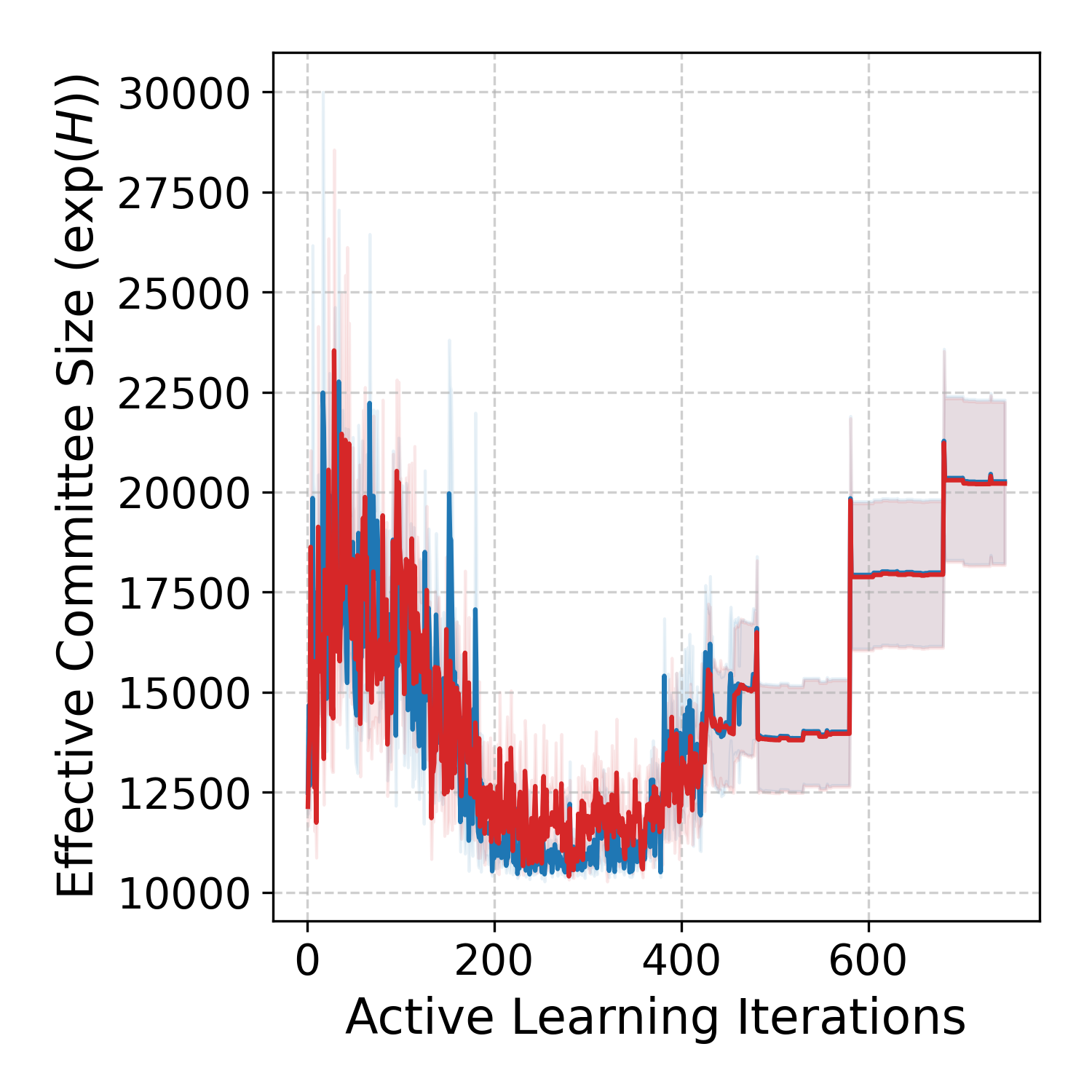}
        \caption{Tic-Tac-Toe}
    \end{subfigure}
    \hfill
    \begin{subfigure}[b]{0.19\textwidth}
        \includegraphics[width=\linewidth]{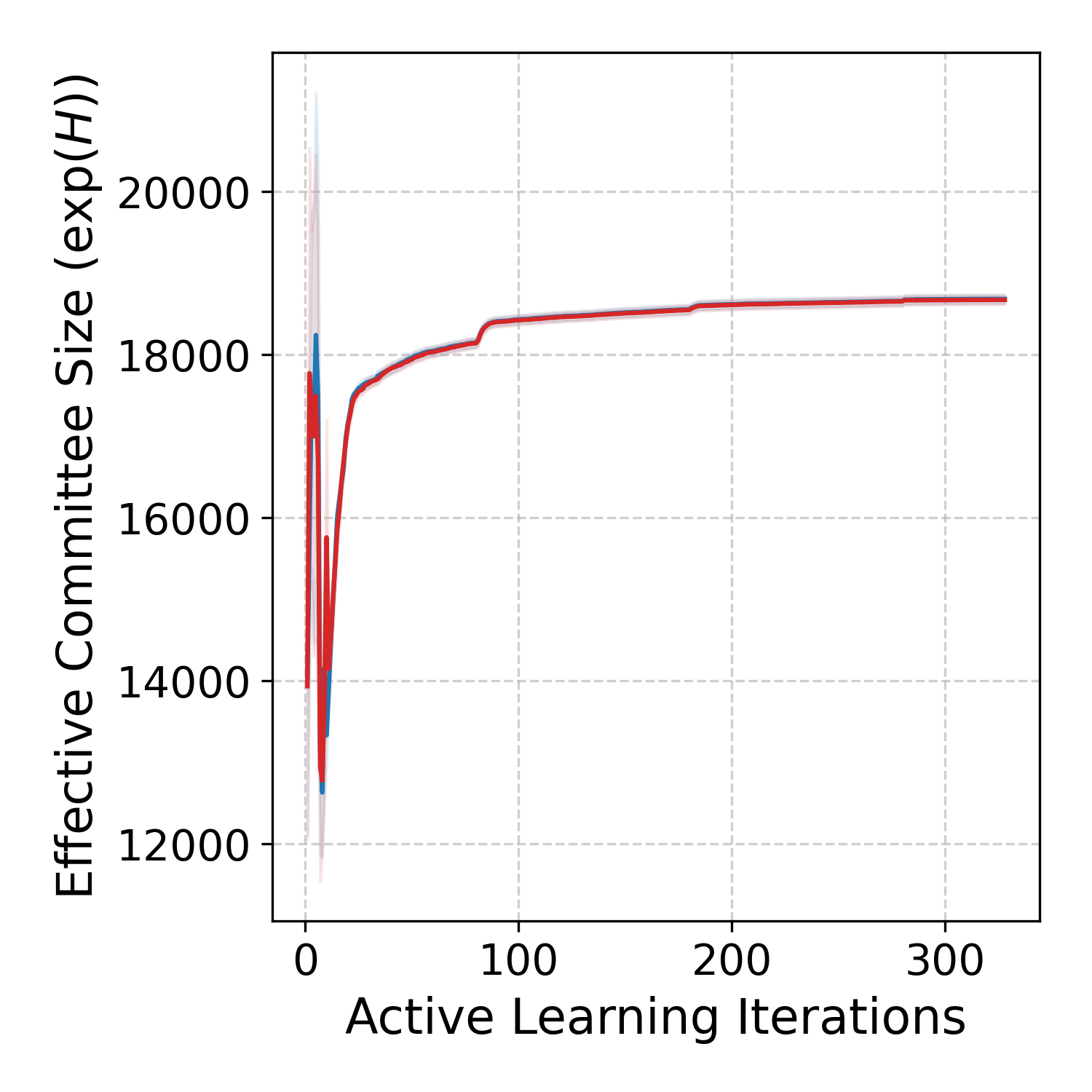}
        \caption{Vote}
    \end{subfigure}
    \hfill
    \begin{subfigure}[b]{0.19\textwidth}
        \includegraphics[width=\linewidth]{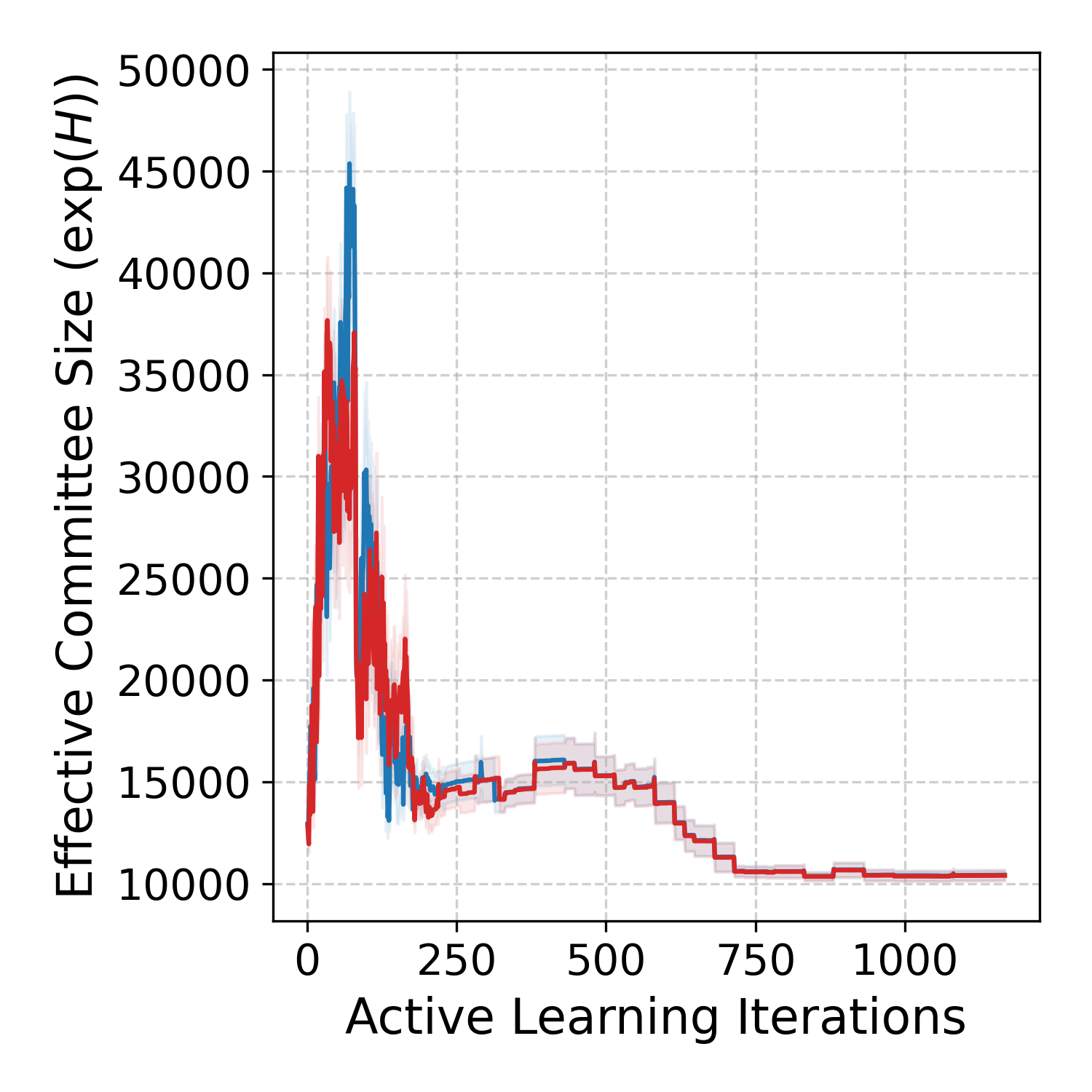}
        \caption{Yeast}
    \end{subfigure}

    \vspace{0.5em}
    \centering
    \includegraphics[width=0.60\linewidth]{upload_all_files/study1_main_AL_results/benchmark_legend.png}

    \caption{\textbf{Committee Size History.} Evolution of the Rashomon Effective Committee Size over active learning iterations across 20 benchmark datasets. The ECS serves as a metric of model certainty, quantifying the diversity of plausible theories within the version space as the learner explores the data manifold. A high ECS reflects a broad set of competing hypotheses, while values approaching $1.0$ indicate the emergence of a single dominant structural explanation.}
    \label{fig:BenchmarkGrid_CommitteeSize}
\end{figure*}

\begin{figure*}[!t] 
\centering
    \begin{subfigure}[b]{0.19\textwidth}
        \includegraphics[width=\linewidth]{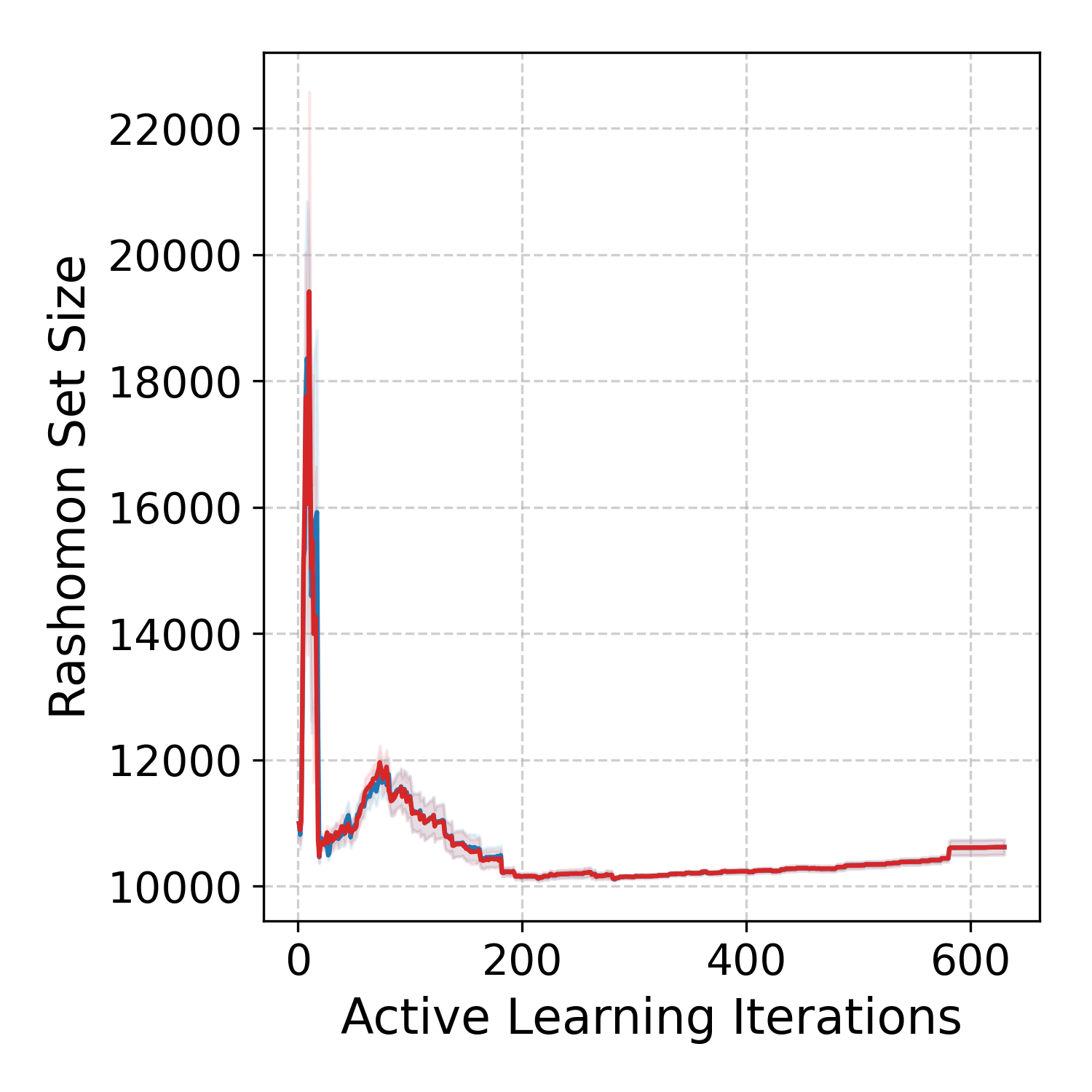}
        \caption{Anneal}
    \end{subfigure}
    \hfill
    \begin{subfigure}[b]{0.19\textwidth}
        \includegraphics[width=\linewidth]{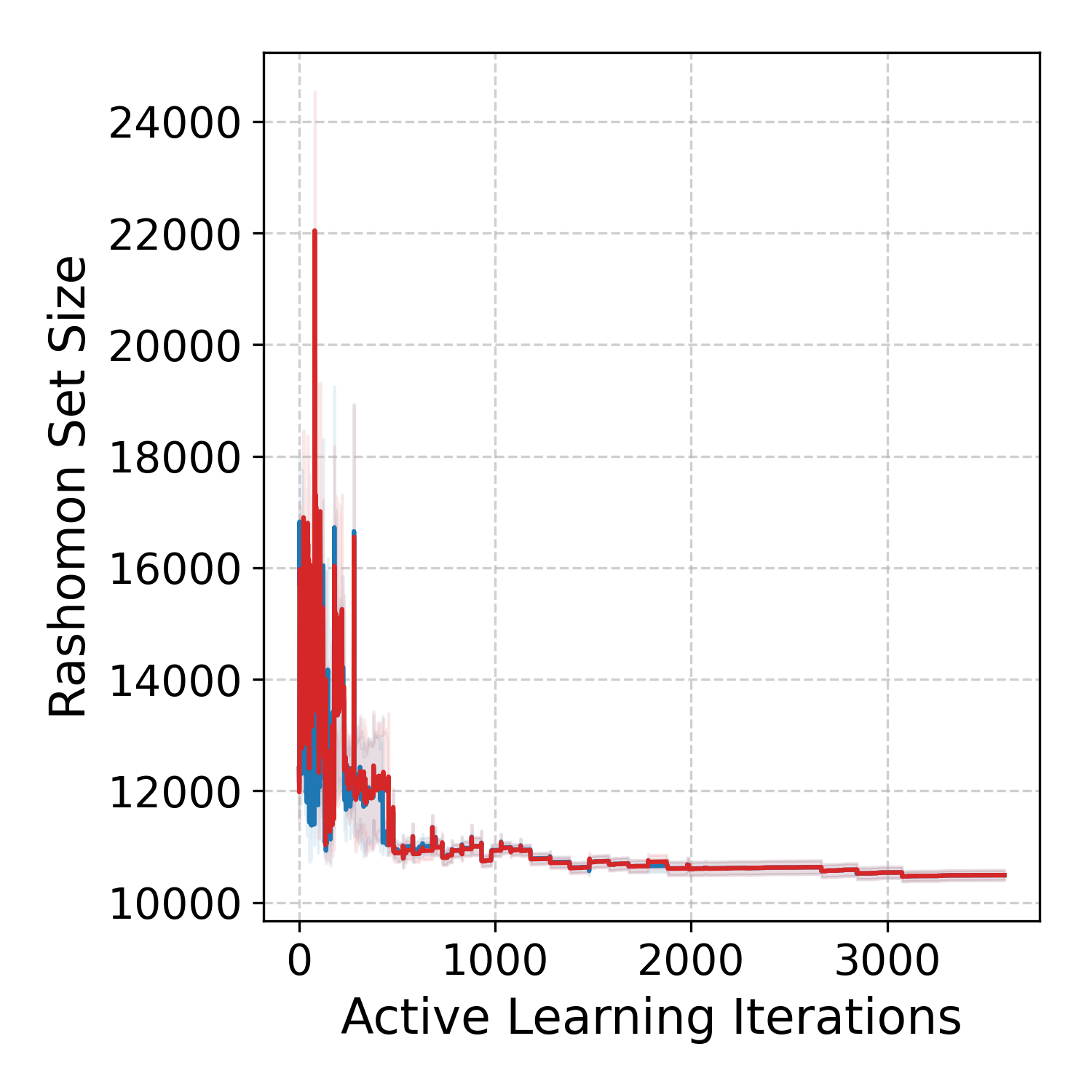}
        \caption{Bank Marketing}
    \end{subfigure}
    \hfill
    \begin{subfigure}[b]{0.19\textwidth}
        \includegraphics[width=\linewidth]{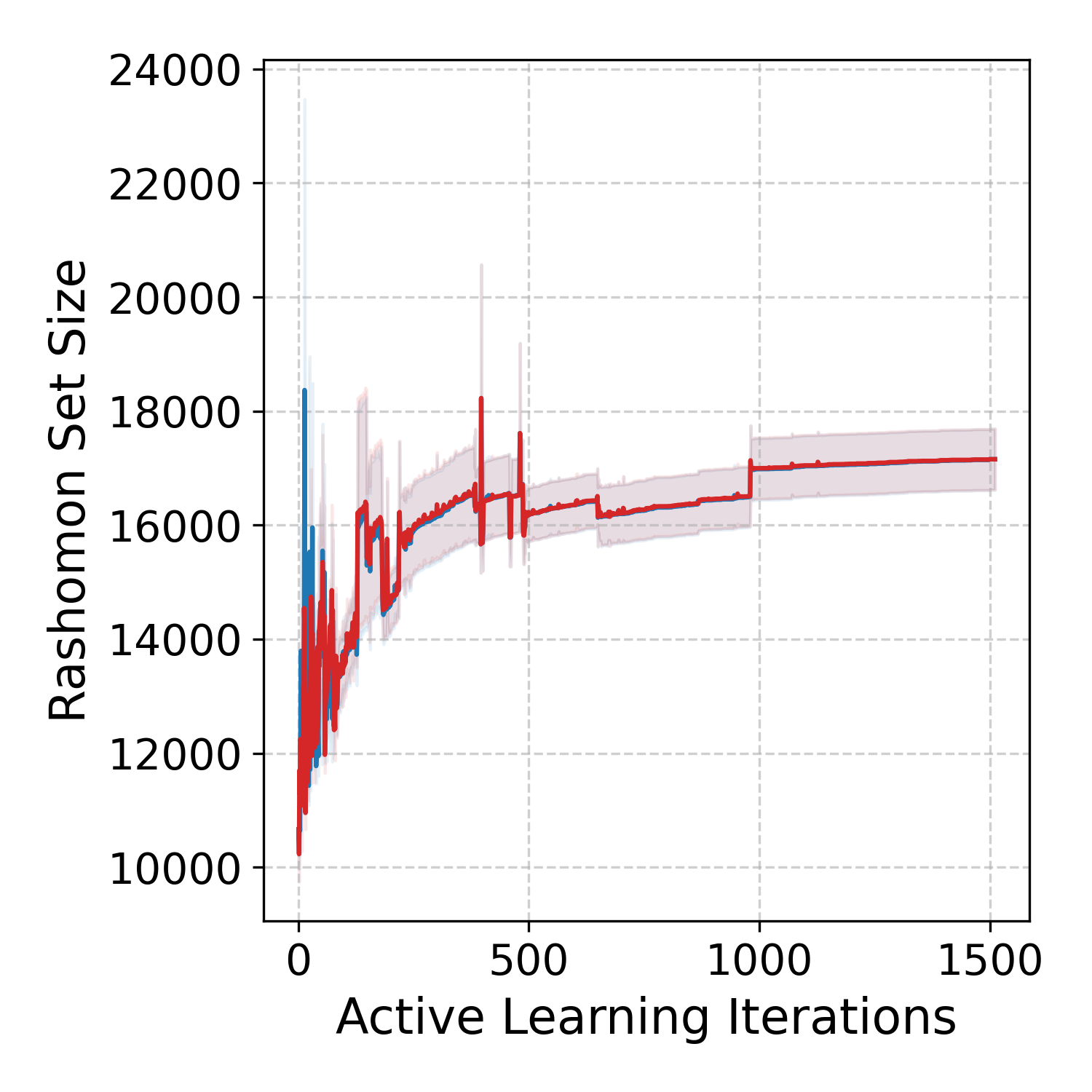}
        \caption{Bar-7}
    \end{subfigure}
    \hfill
    \begin{subfigure}[b]{0.19\textwidth}
        \includegraphics[width=\linewidth]{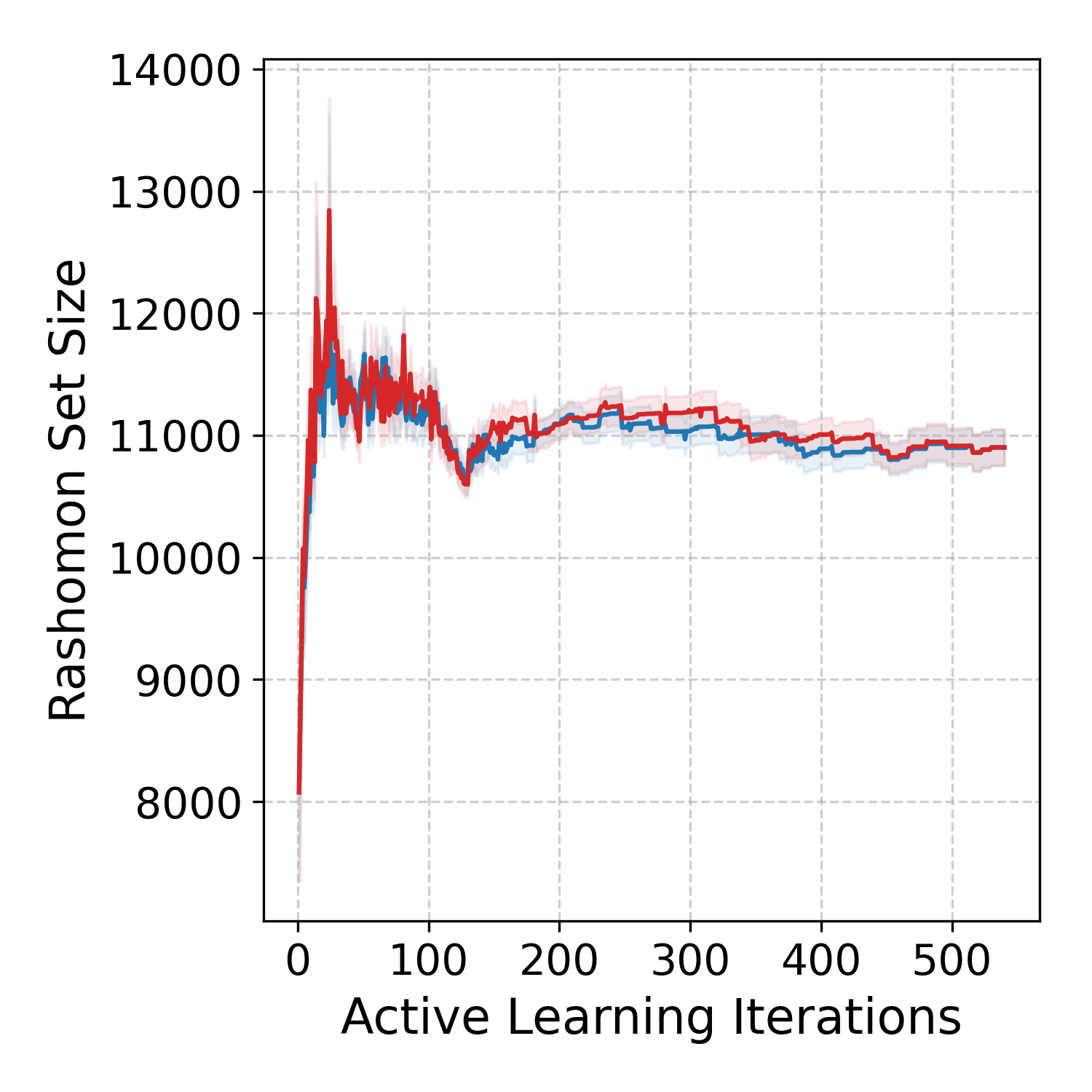}
        \caption{Breast Cancer WI}
    \end{subfigure}
    \hfill
    \begin{subfigure}[b]{0.19\textwidth}
        \includegraphics[width=\linewidth]{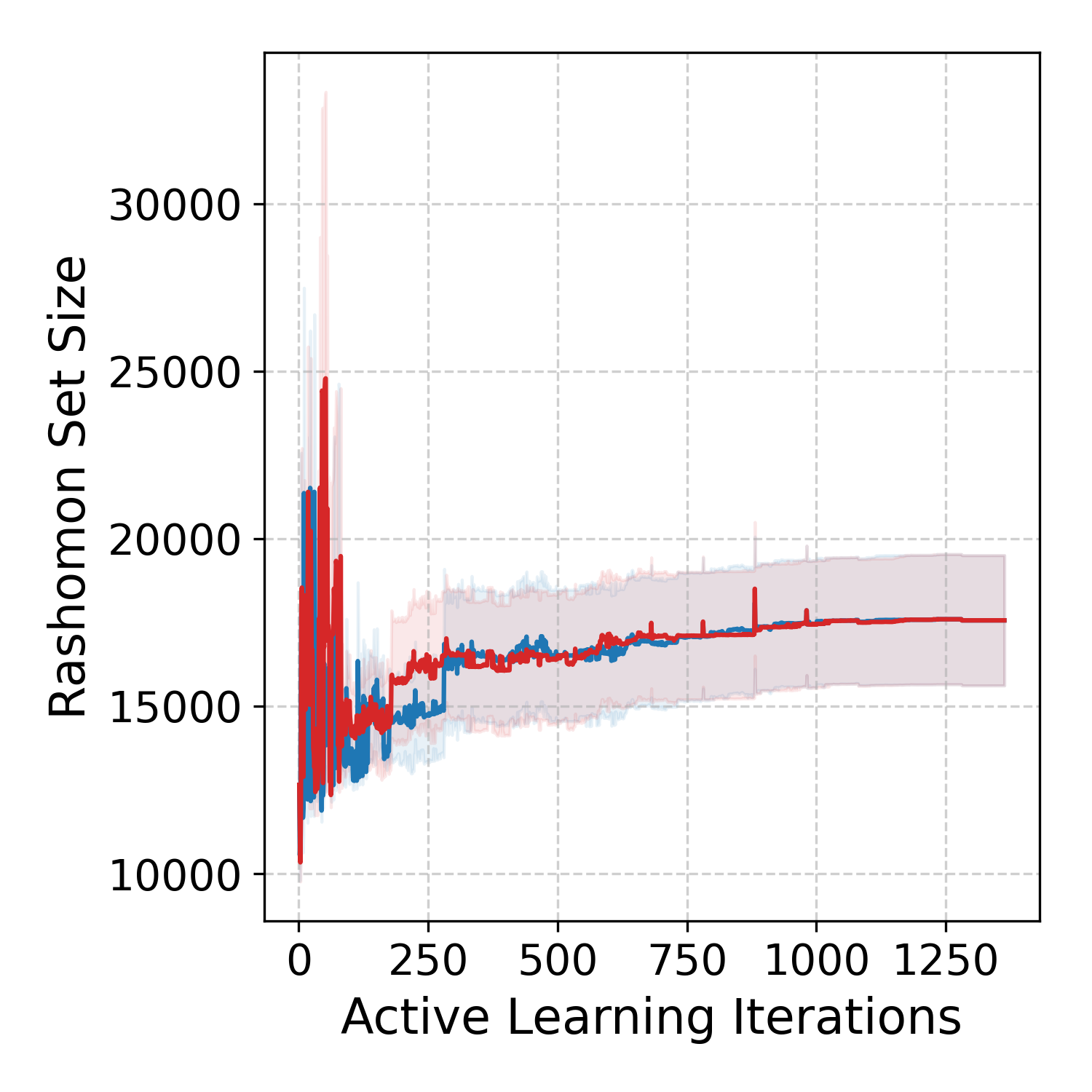}
        \caption{Car Evaluation}
    \end{subfigure}

    \vspace{0.2em}

    \begin{subfigure}[b]{0.19\textwidth}
        \includegraphics[width=\linewidth]{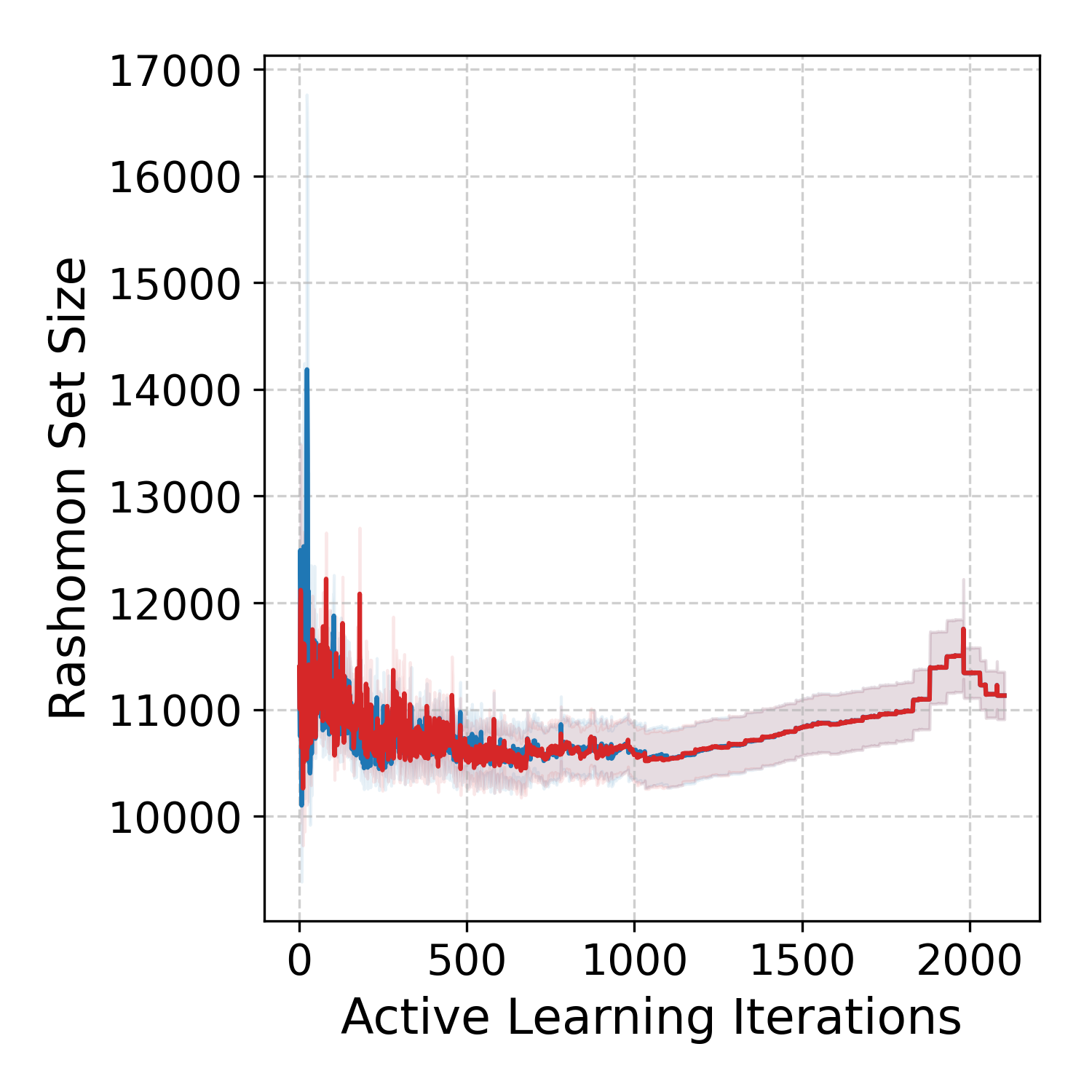}
        \caption{Cheap Restaurant}
    \end{subfigure}
    \hfill
    \begin{subfigure}[b]{0.19\textwidth}
        \includegraphics[width=\linewidth]{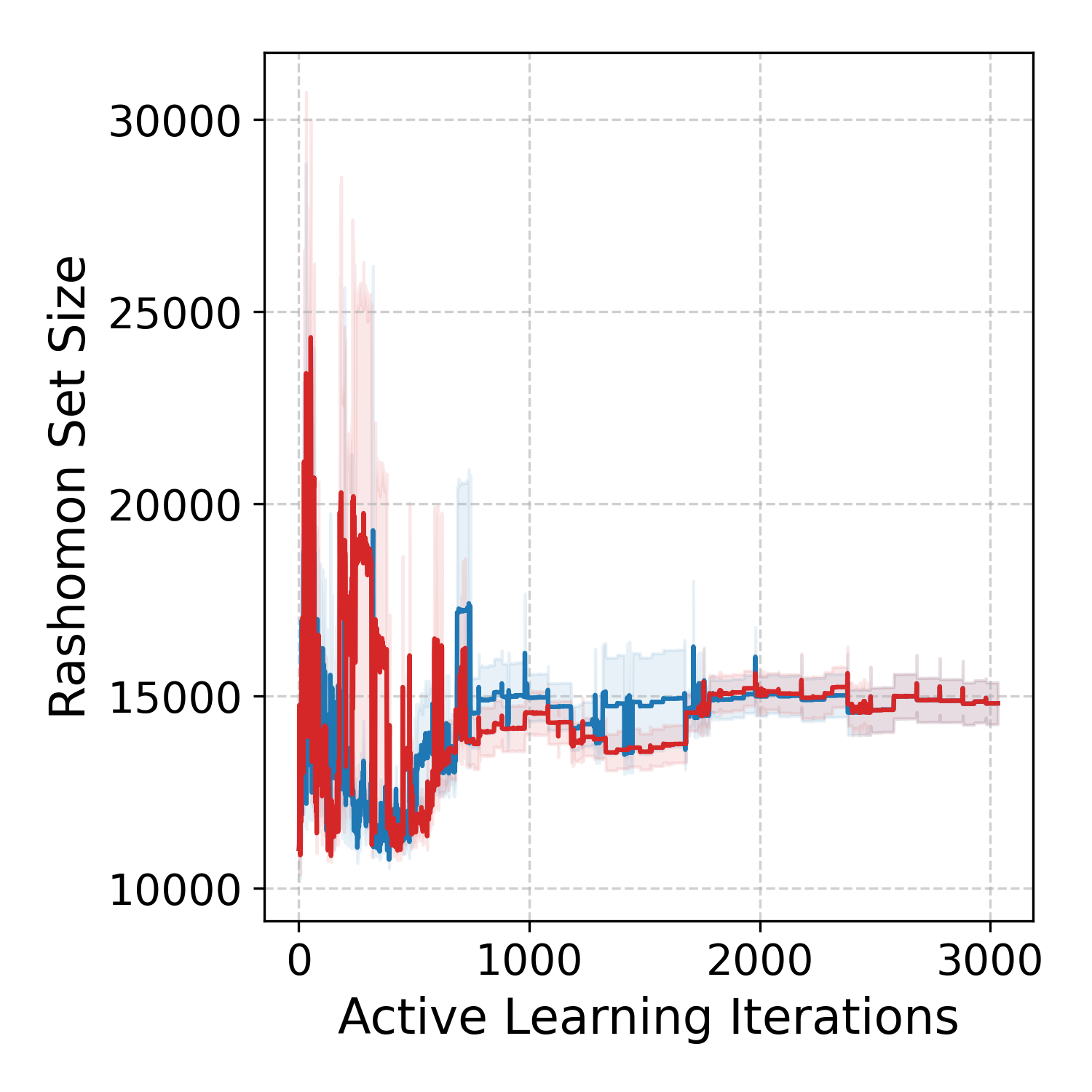}
        \caption{Coffee House}
    \end{subfigure}
    \hfill
    \begin{subfigure}[b]{0.19\textwidth}
        \includegraphics[width=\linewidth]{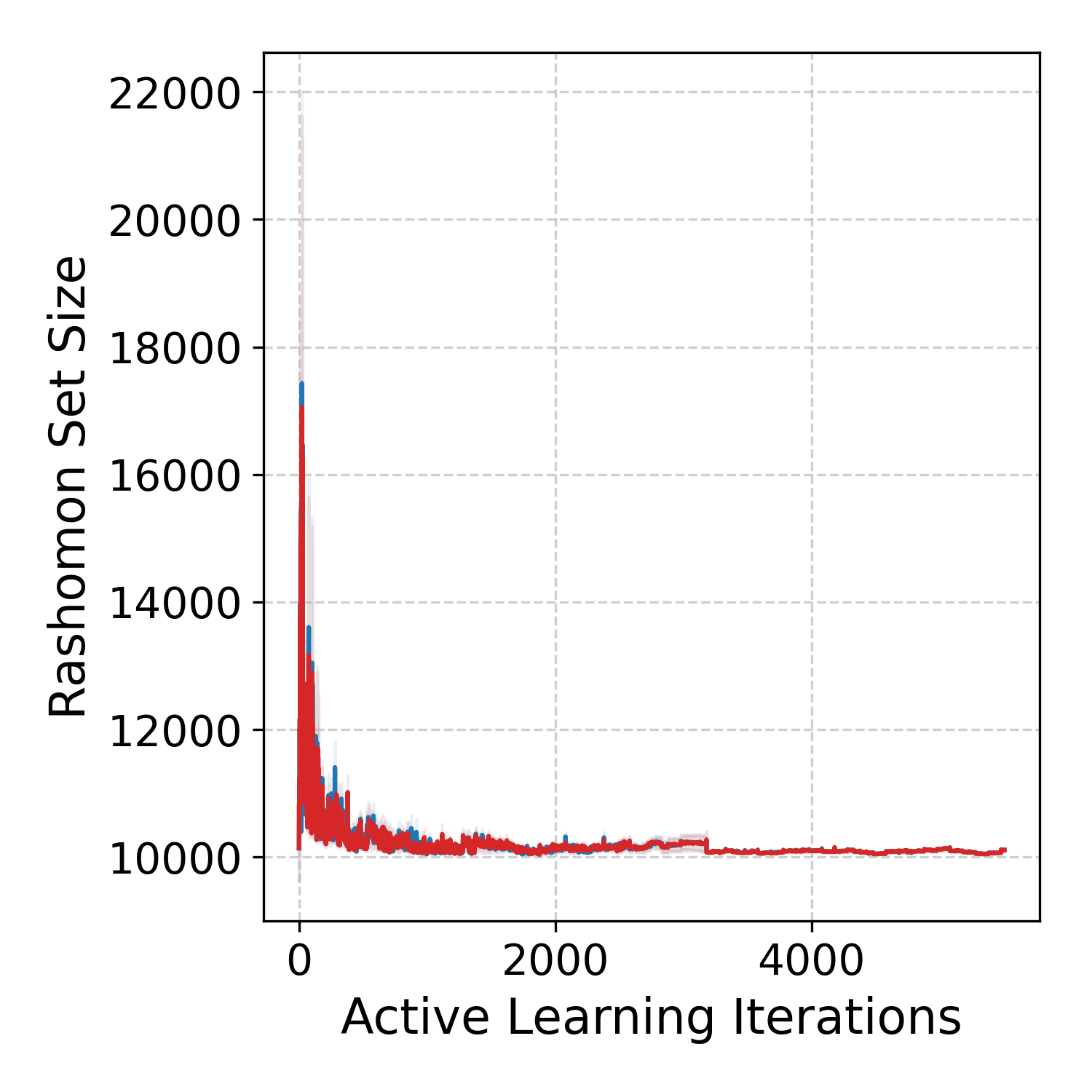}
        \caption{COMPAS}
    \end{subfigure}
    \hfill
    \begin{subfigure}[b]{0.19\textwidth}
        \includegraphics[width=\linewidth]{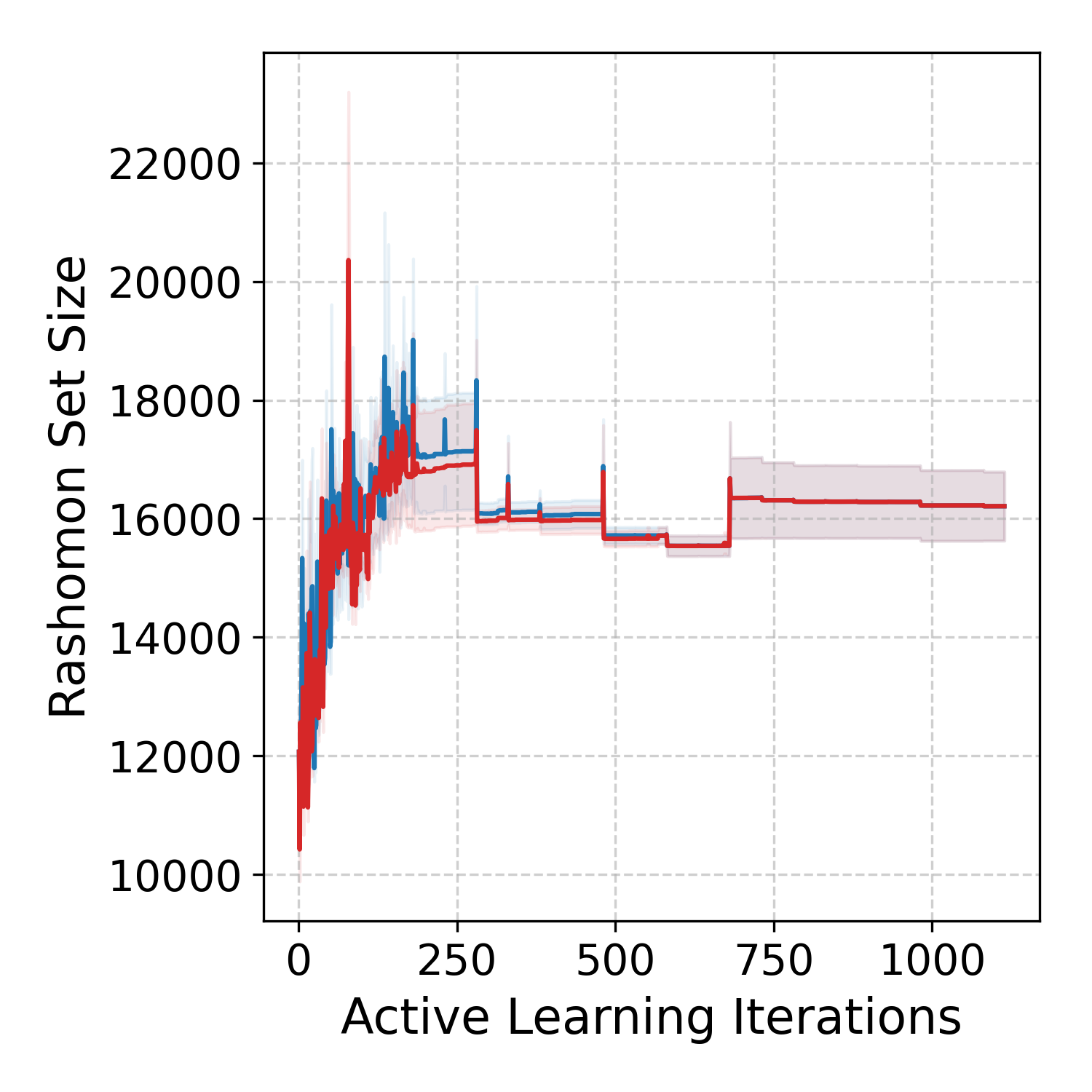}
        \caption{Expensive Restaurant}
    \end{subfigure}
    \hfill
    \begin{subfigure}[b]{0.19\textwidth}
        \includegraphics[width=\linewidth]{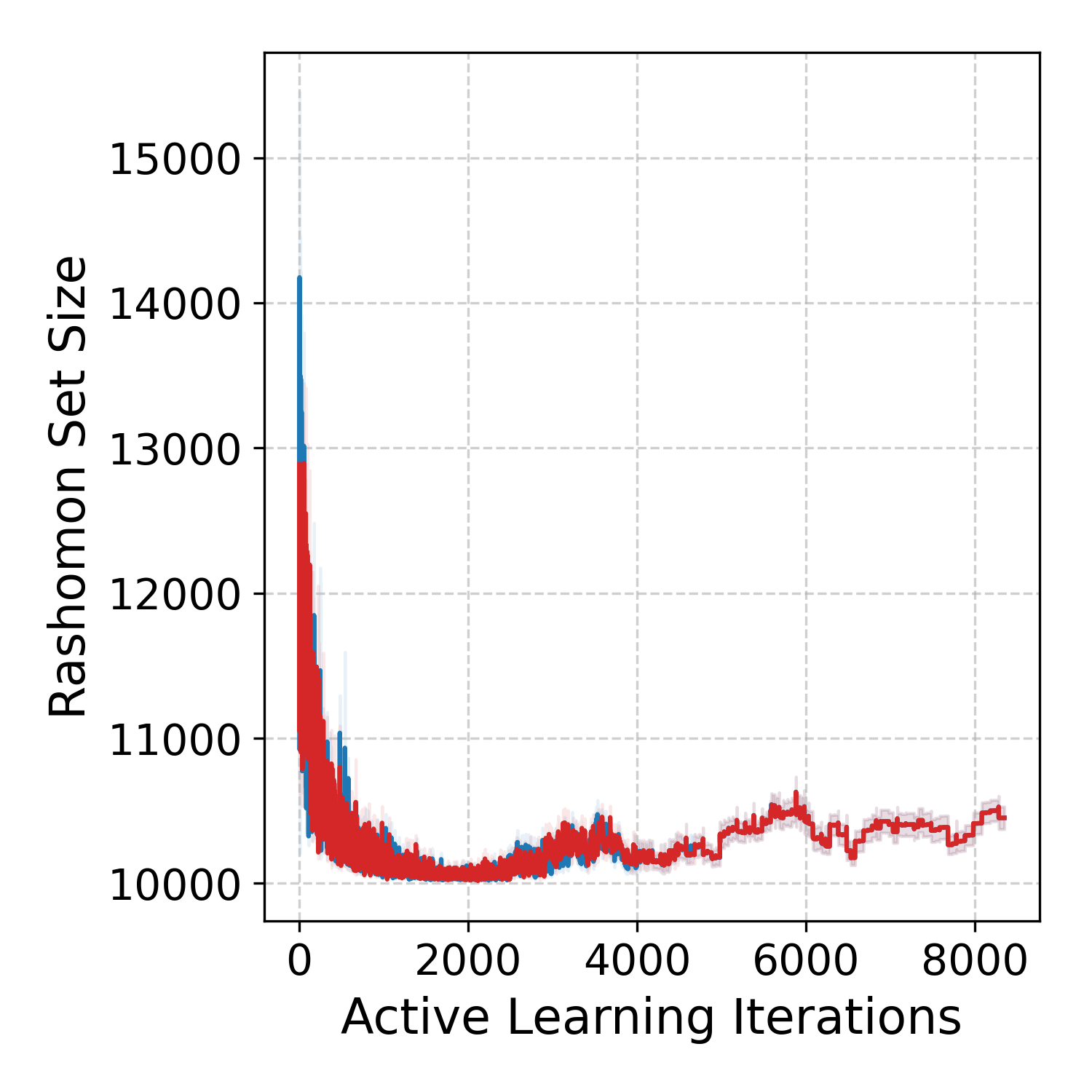}
        \caption{FICO}
    \end{subfigure}

    \vspace{0.2em}

    \begin{subfigure}[b]{0.19\textwidth}
        \includegraphics[width=\linewidth]{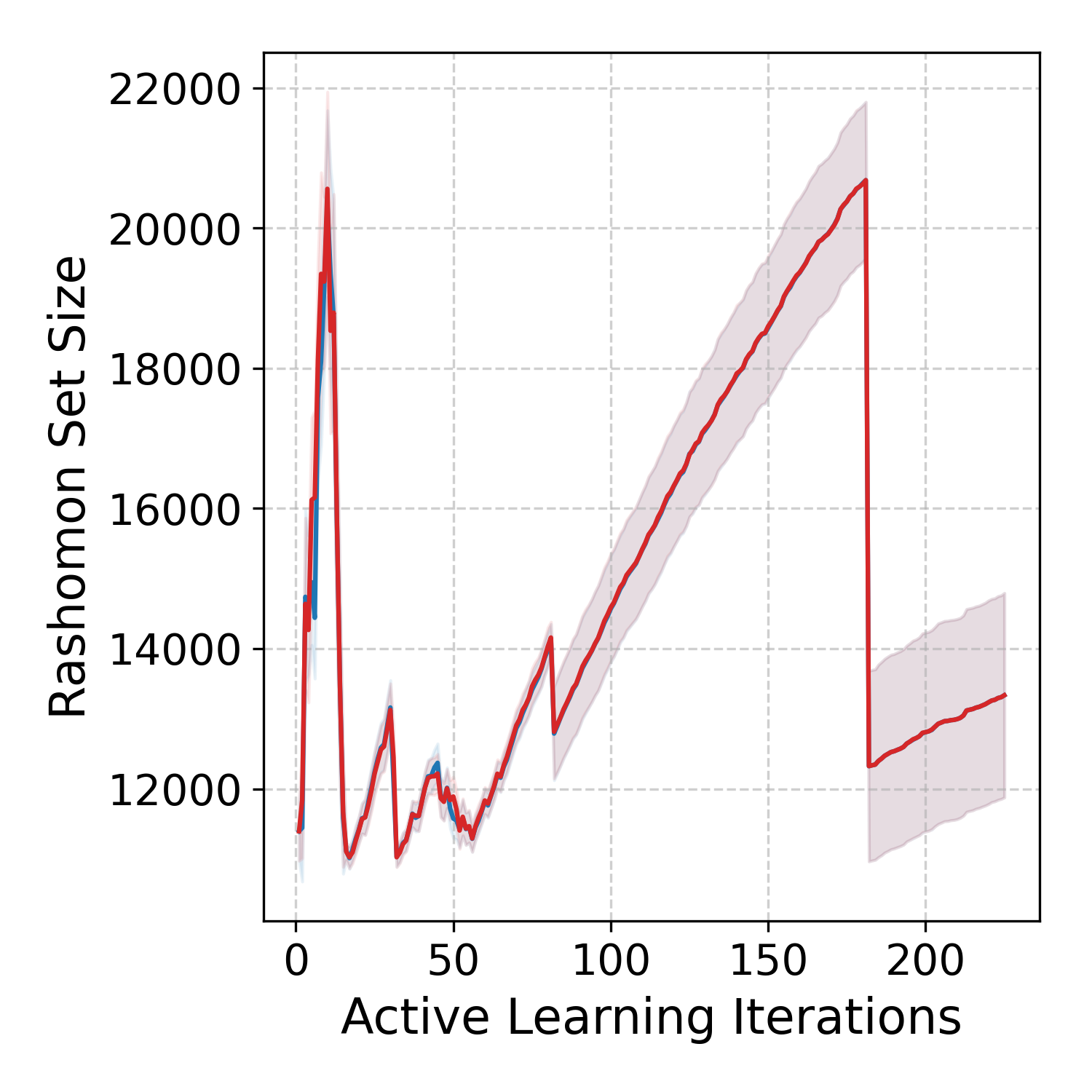}
        \caption{Haberman}
    \end{subfigure}
    \hfill
    \begin{subfigure}[b]{0.19\textwidth}
        \includegraphics[width=\linewidth]{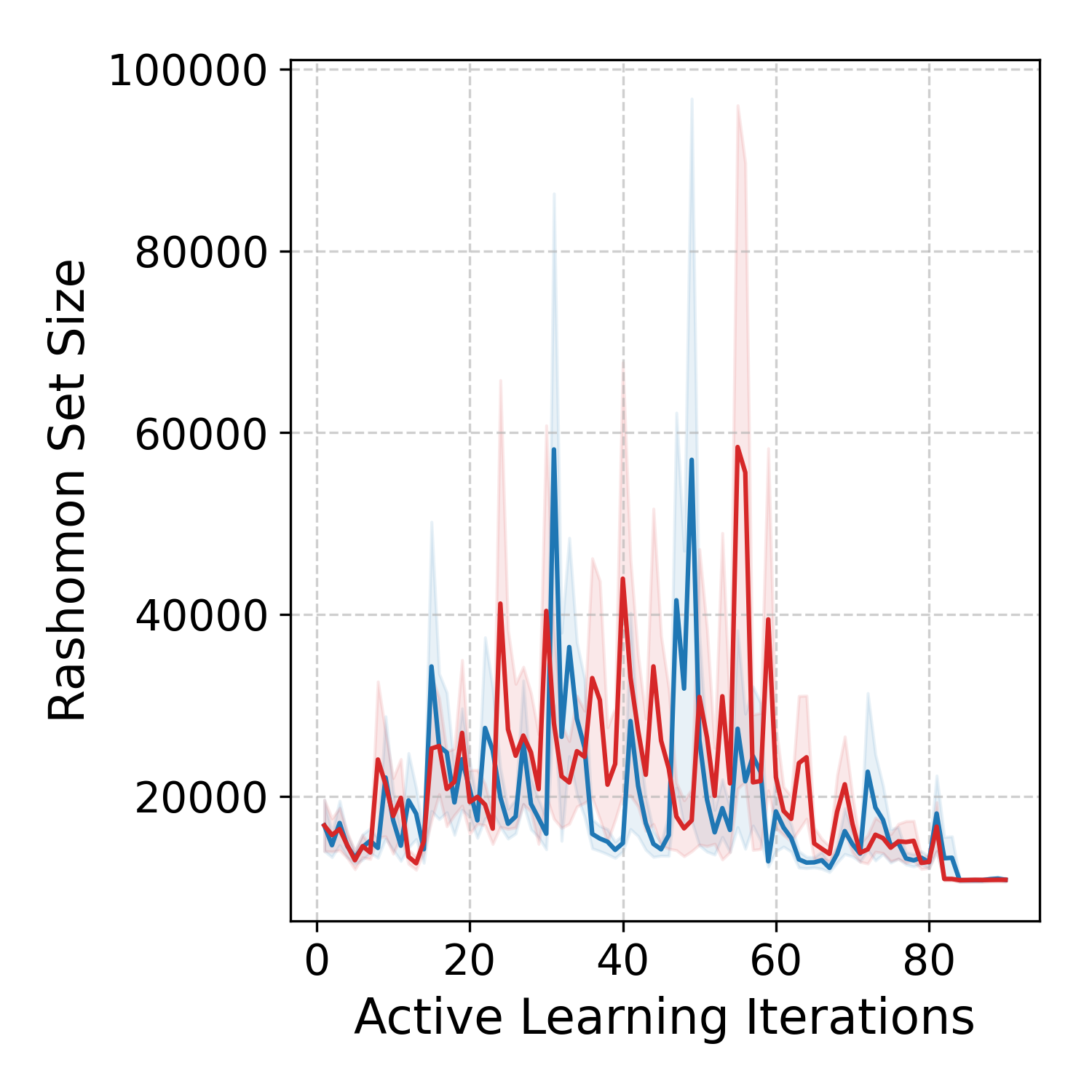}
        \caption{Hepatitis}
    \end{subfigure}
    \hfill
    \begin{subfigure}[b]{0.19\textwidth}
        \includegraphics[width=\linewidth]{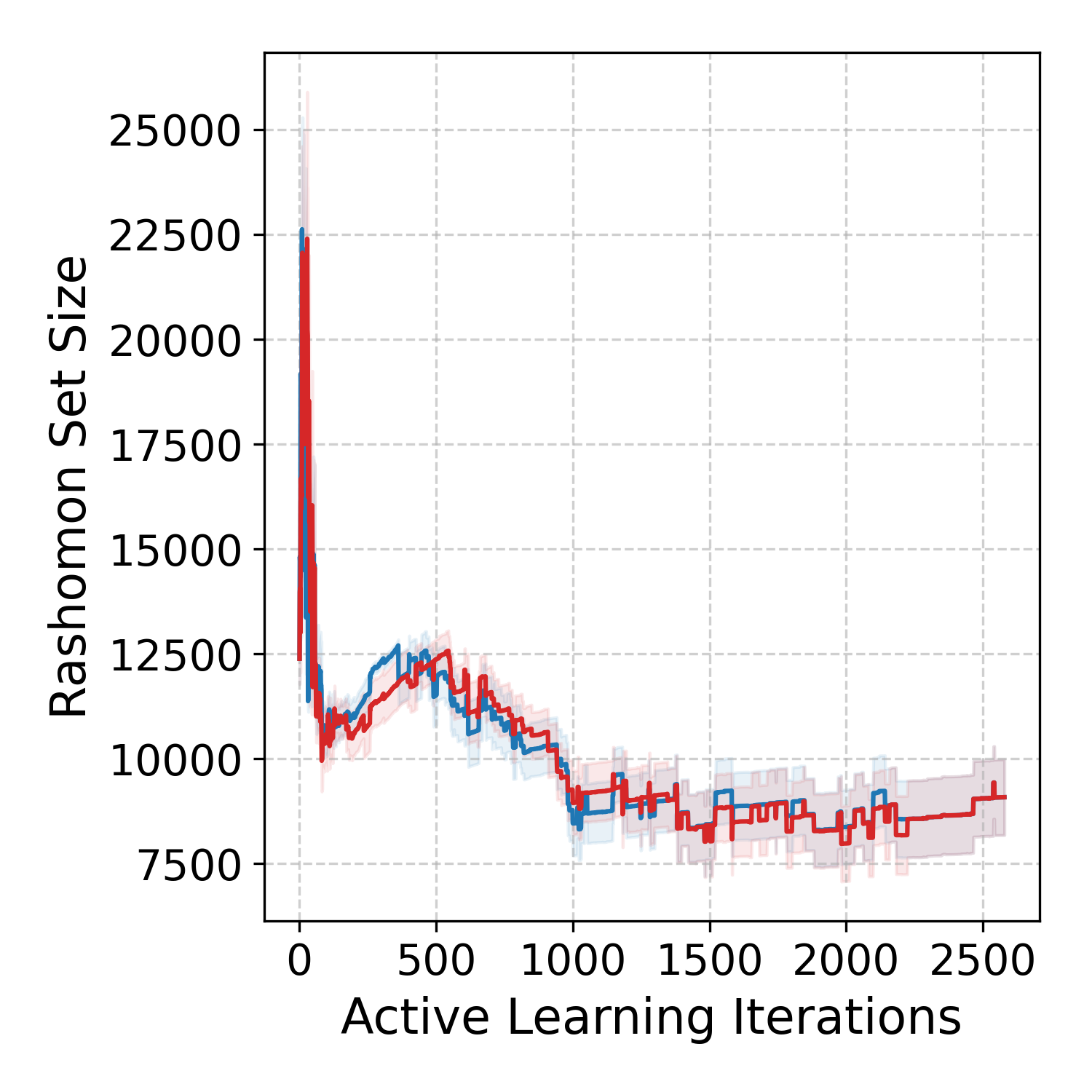}
        \caption{Hypothyroid}
    \end{subfigure}
    \hfill
    \begin{subfigure}[b]{0.19\textwidth}
        \includegraphics[width=\linewidth]{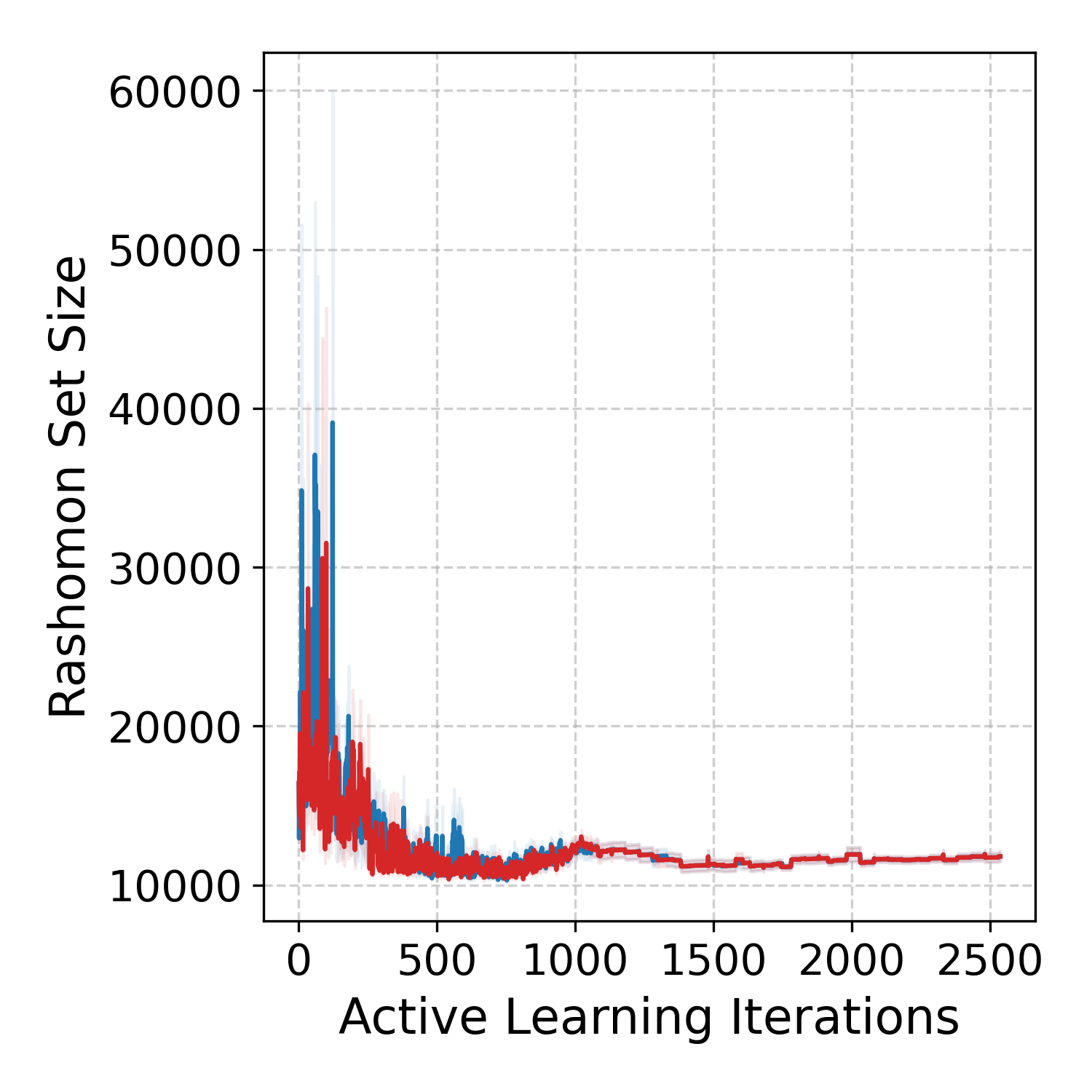}
        \caption{Kr-vs-Kp}
    \end{subfigure}
    \hfill
    \begin{subfigure}[b]{0.19\textwidth}
        \includegraphics[width=\linewidth]{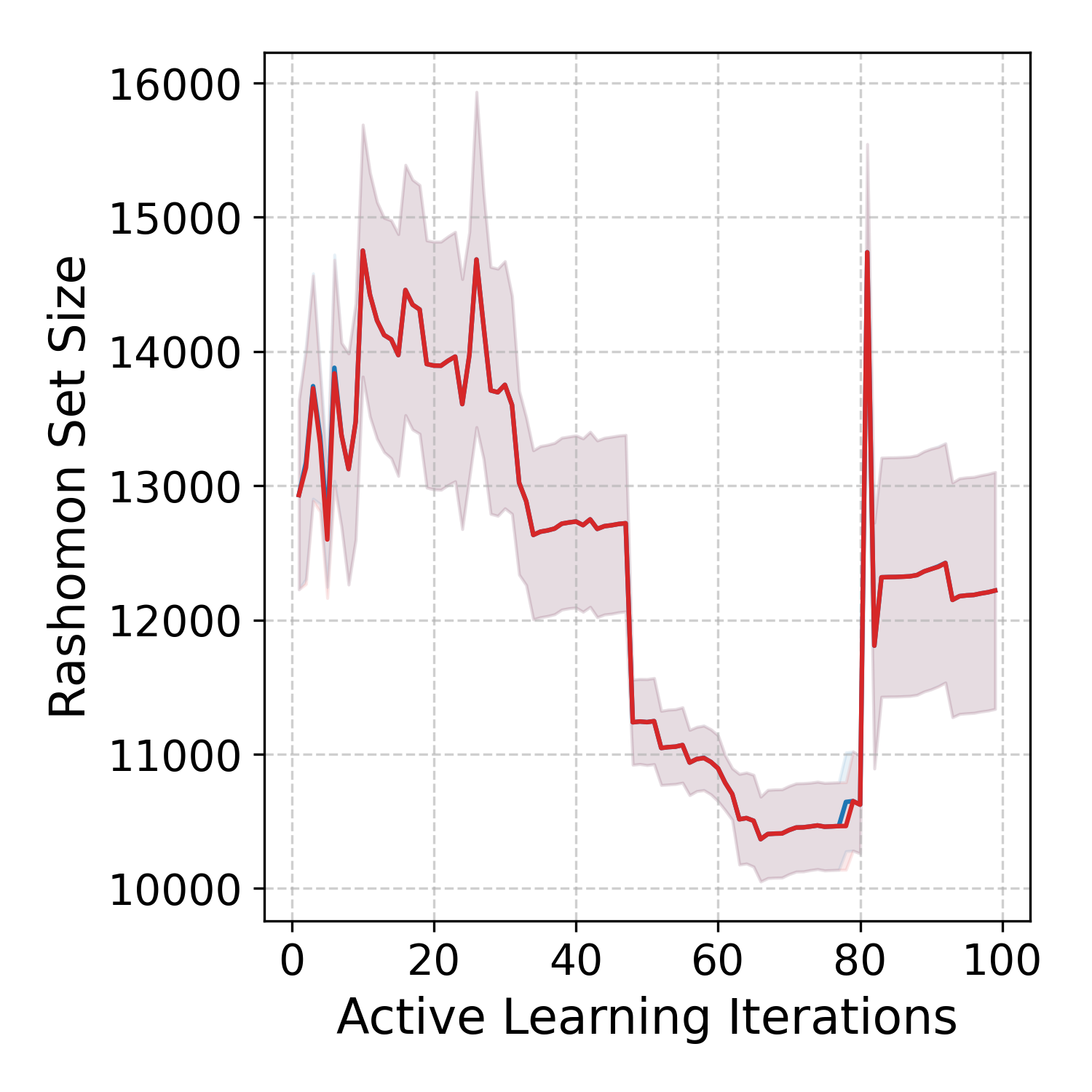}
        \caption{Lymphography}
    \end{subfigure}

    \vspace{0.2em}

    \begin{subfigure}[b]{0.19\textwidth}
        \includegraphics[width=\linewidth]{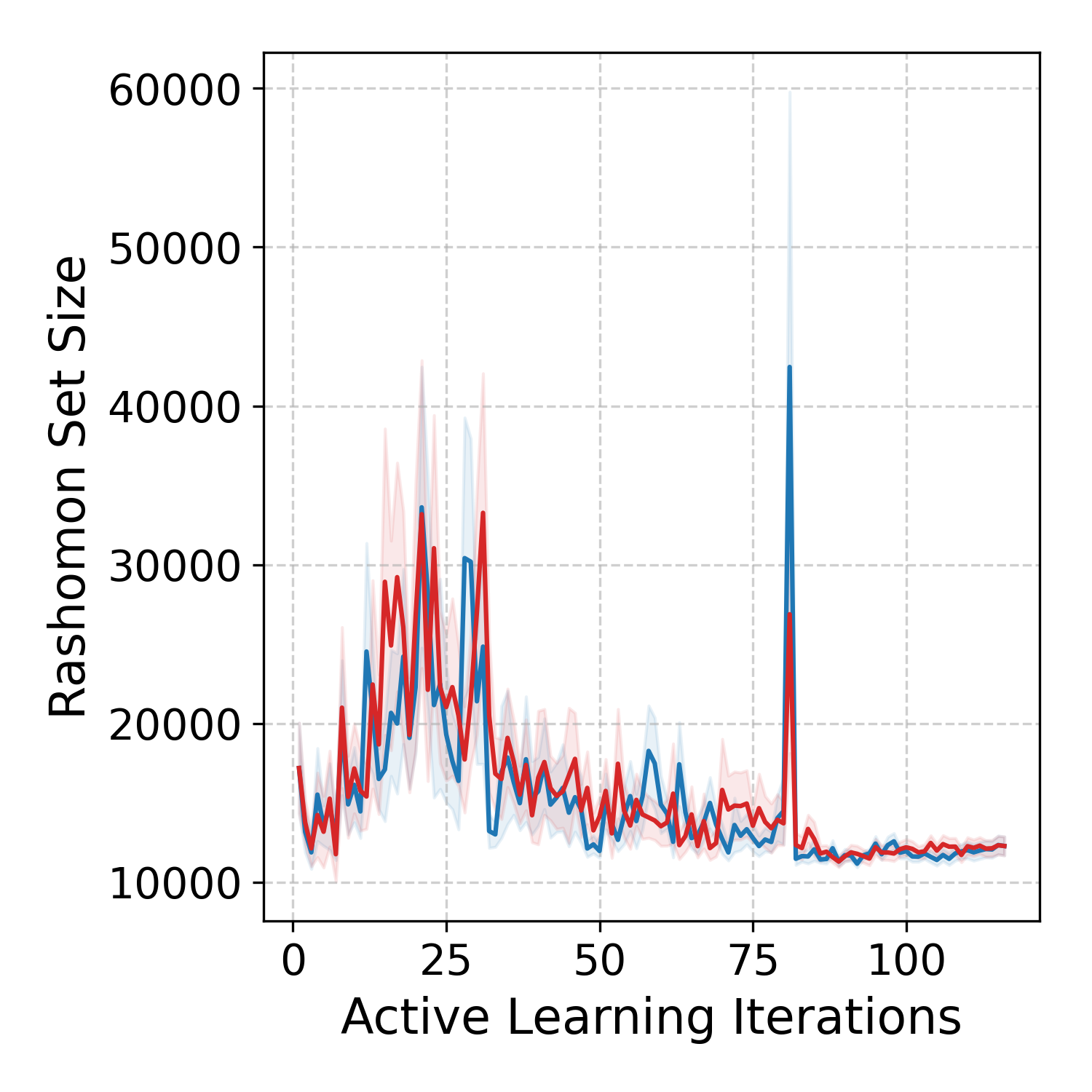}
        \caption{MONK-2}
    \end{subfigure}
    \hfill
    \begin{subfigure}[b]{0.19\textwidth}
        \includegraphics[width=\linewidth]{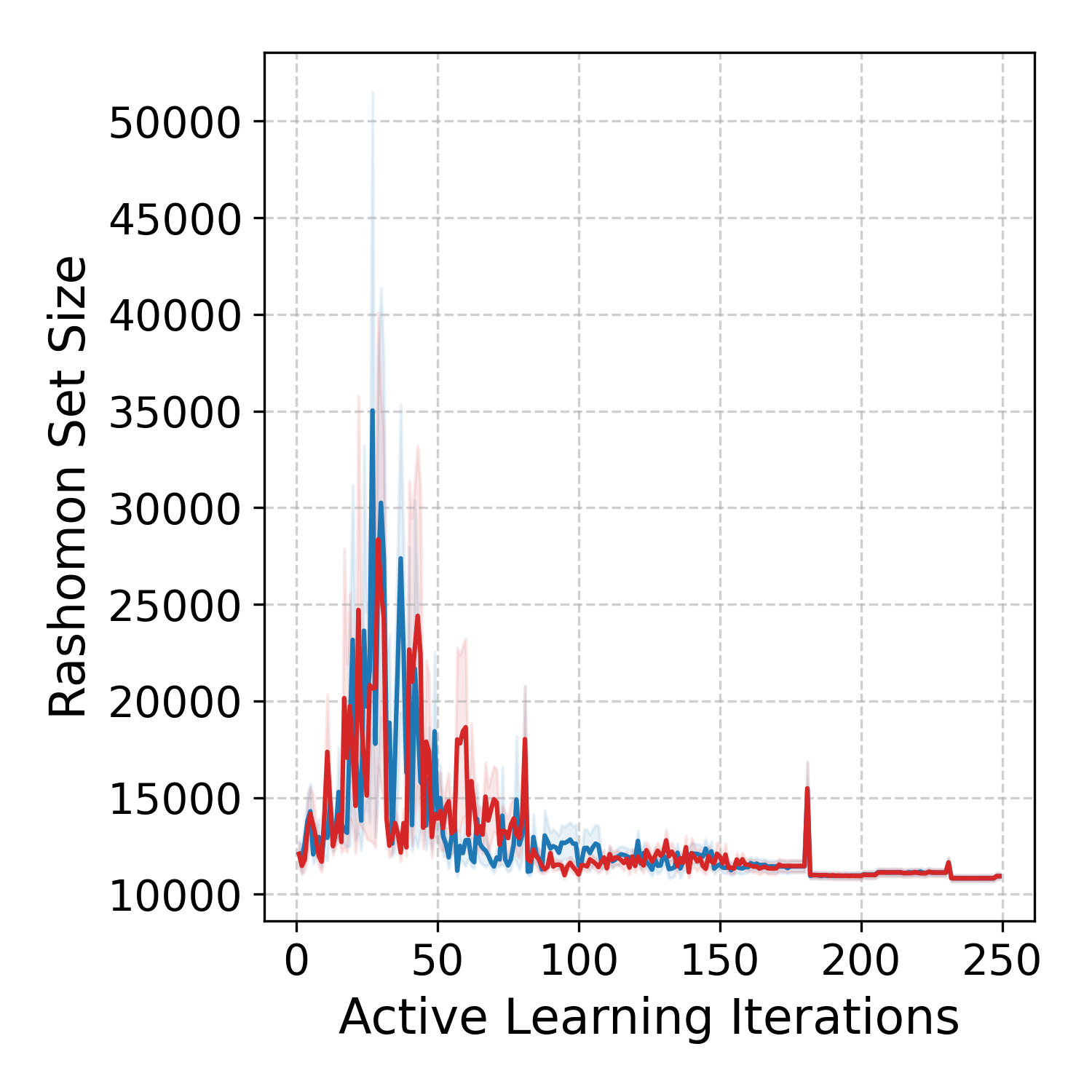}
        \caption{Primary Tumor}
    \end{subfigure}
    \hfill
    \begin{subfigure}[b]{0.19\textwidth}
        \includegraphics[width=\linewidth]{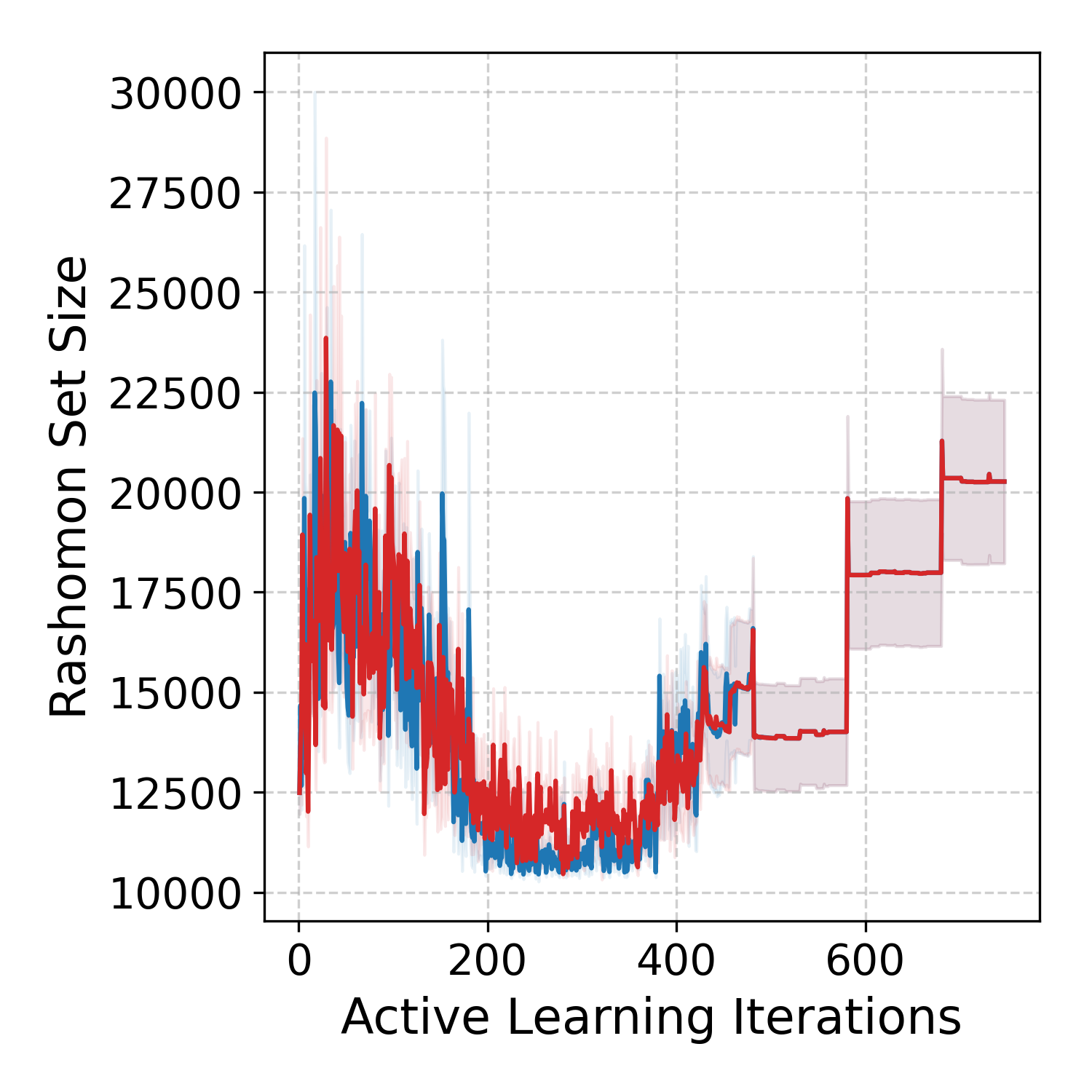}
        \caption{Tic-Tac-Toe}
    \end{subfigure}
    \hfill
    \begin{subfigure}[b]{0.19\textwidth}
        \includegraphics[width=\linewidth]{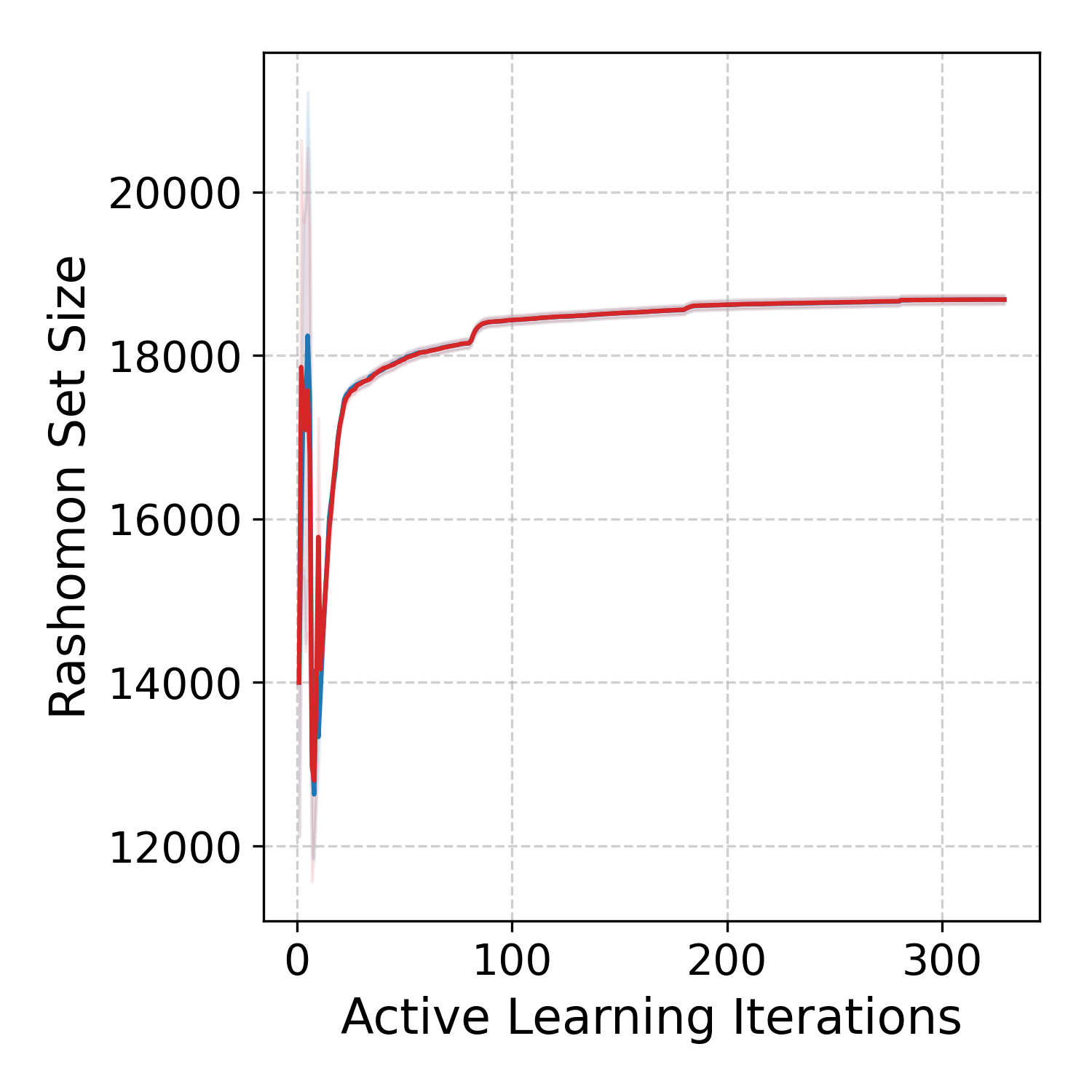}
        \caption{Vote}
    \end{subfigure}
    \hfill
    \begin{subfigure}[b]{0.19\textwidth}
        \includegraphics[width=\linewidth]{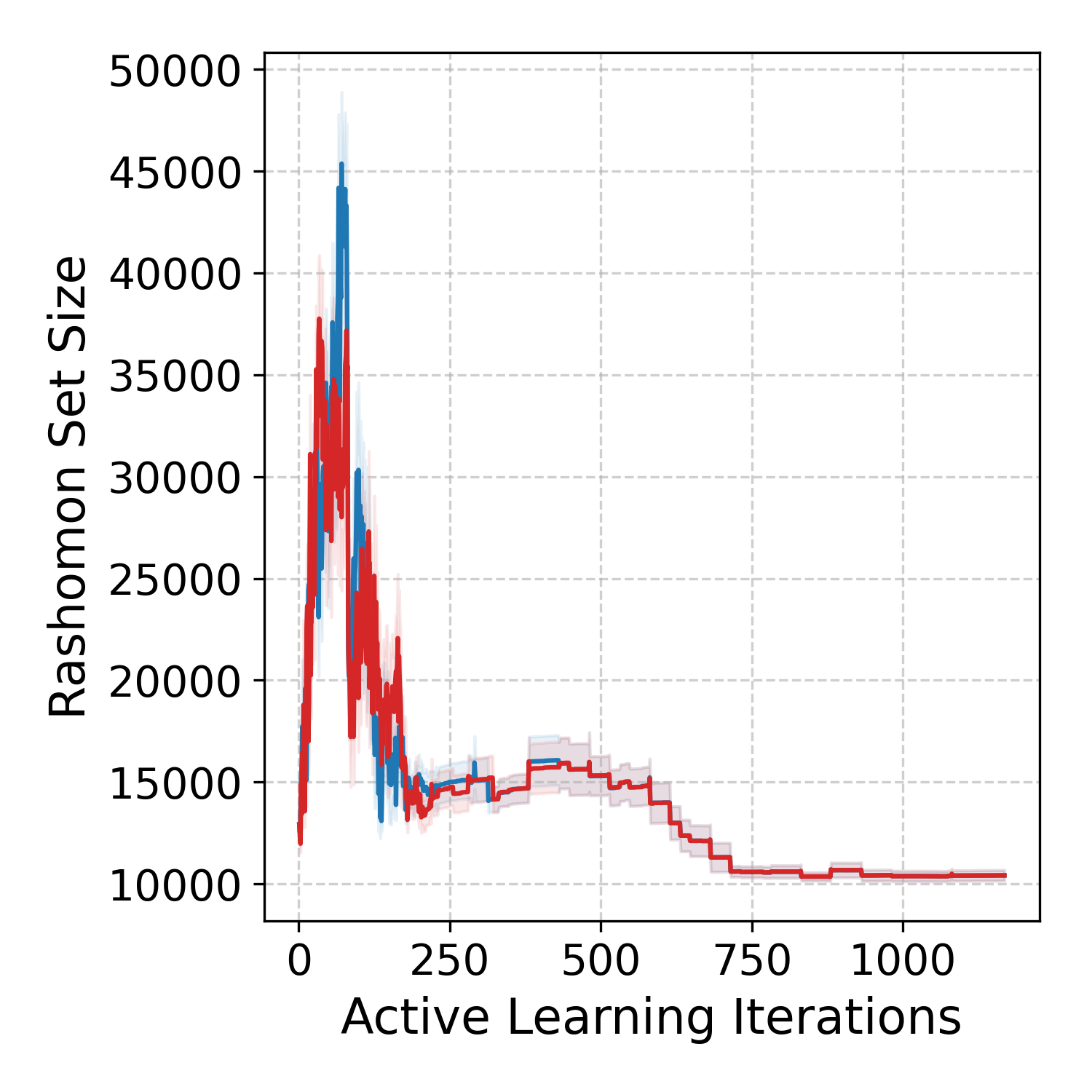}
        \caption{Yeast}
    \end{subfigure}

    \vspace{0.5em}
    \centering
    \includegraphics[width=0.60\linewidth]{upload_all_files/study1_main_AL_results/benchmark_legend.png}

    \caption{\textbf{Rashomon Set Size Evolution.} The total count of near-optimal models identifying the $\epsilon$-version space during each iteration. These trajectories illustrate how acquiring specific labels prunes the hypothesis space, revealing the sensitivity of model multiplicity to new evidence across diverse dataset geometries.}
    \label{fig:BenchmarkGrid_RashomonSize}
\end{figure*}

\begin{figure*}[!t] 
\centering
    \begin{subfigure}[b]{0.19\textwidth}
        \includegraphics[width=\linewidth]{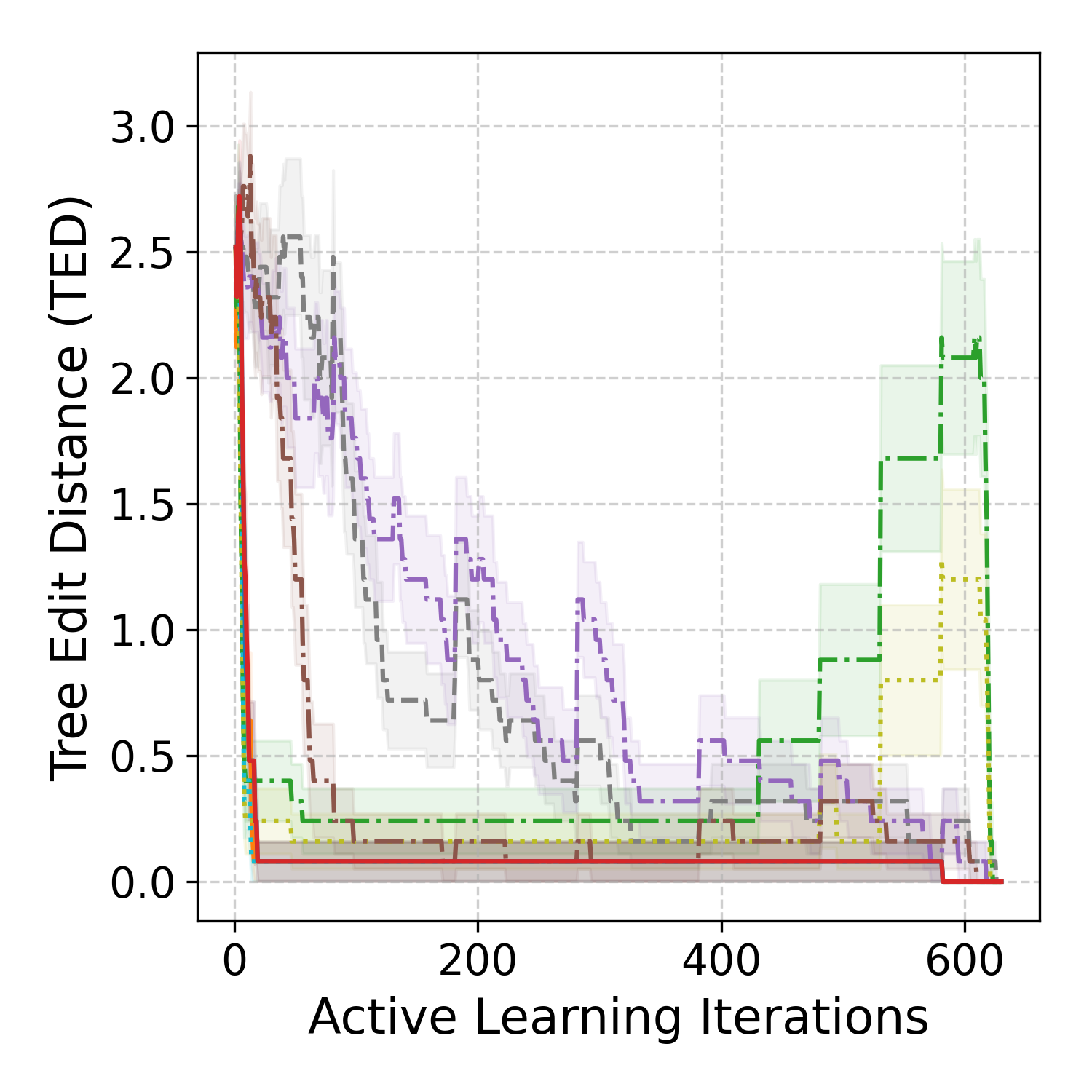}
        \caption{Anneal}
    \end{subfigure}
    \hfill
    \begin{subfigure}[b]{0.19\textwidth}
        \includegraphics[width=\linewidth]{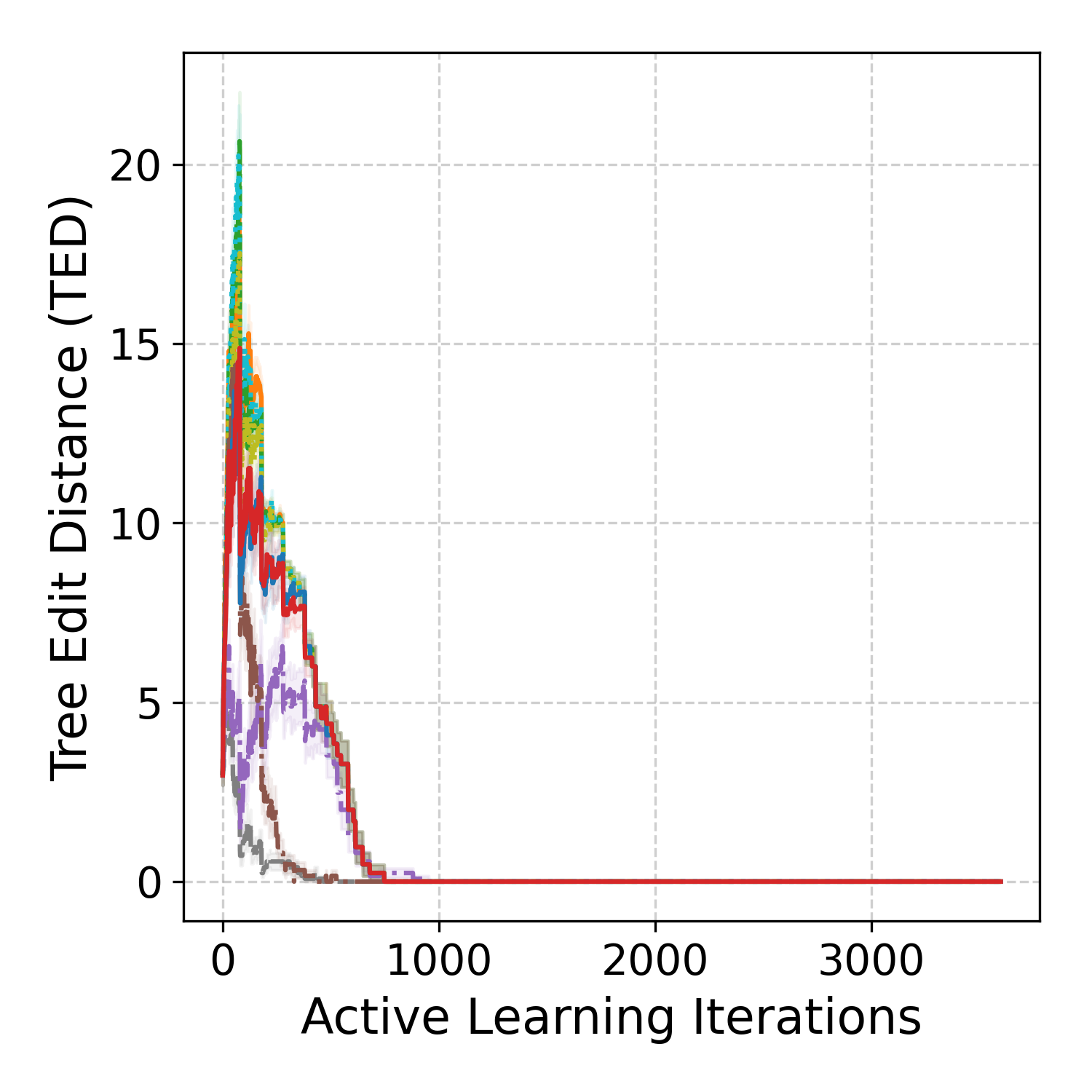}
        \caption{Bank Marketing}
    \end{subfigure}
    \hfill
    \begin{subfigure}[b]{0.19\textwidth}
        \includegraphics[width=\linewidth]{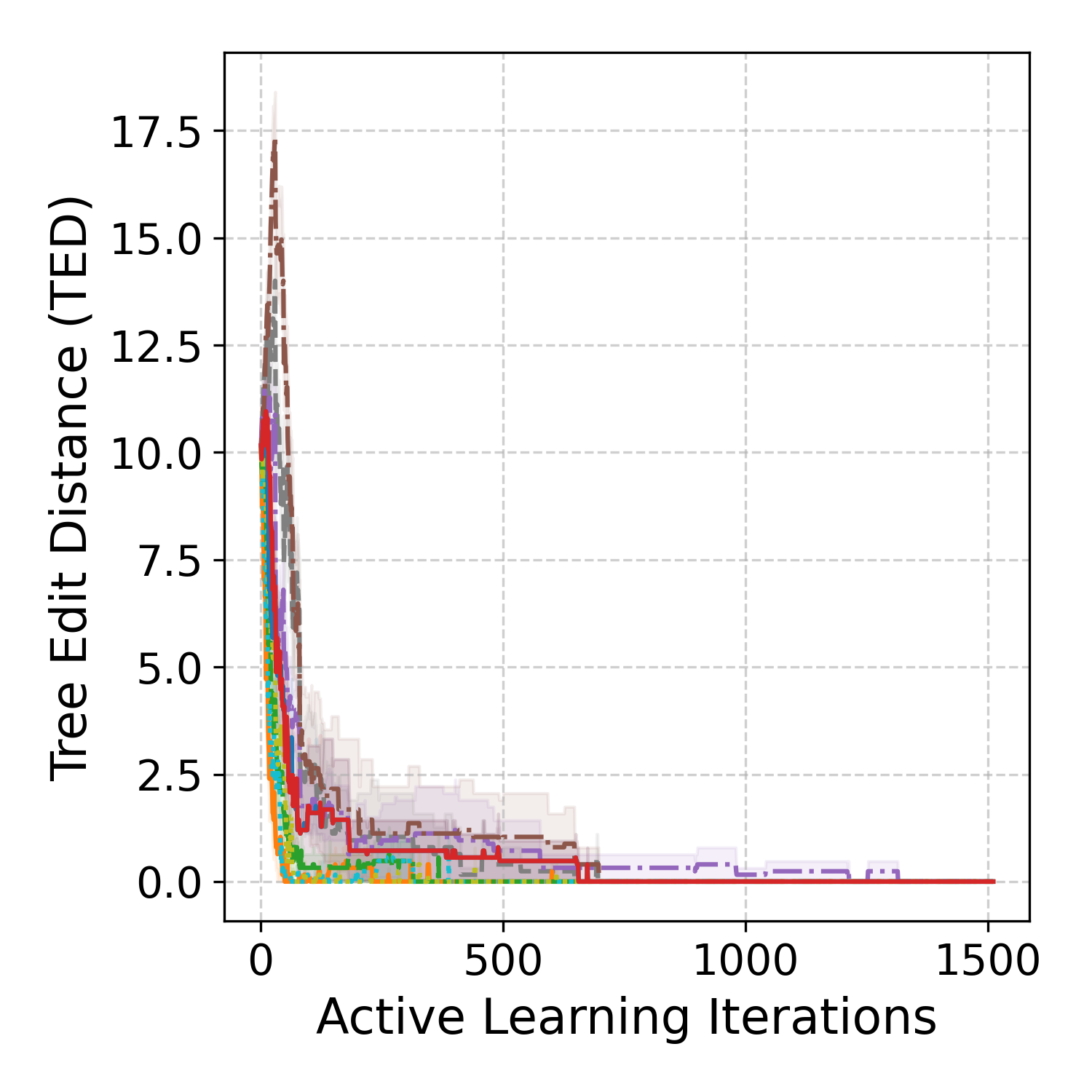}
        \caption{Bar-7}
    \end{subfigure}
    \hfill
    \begin{subfigure}[b]{0.19\textwidth}
        \includegraphics[width=\linewidth]{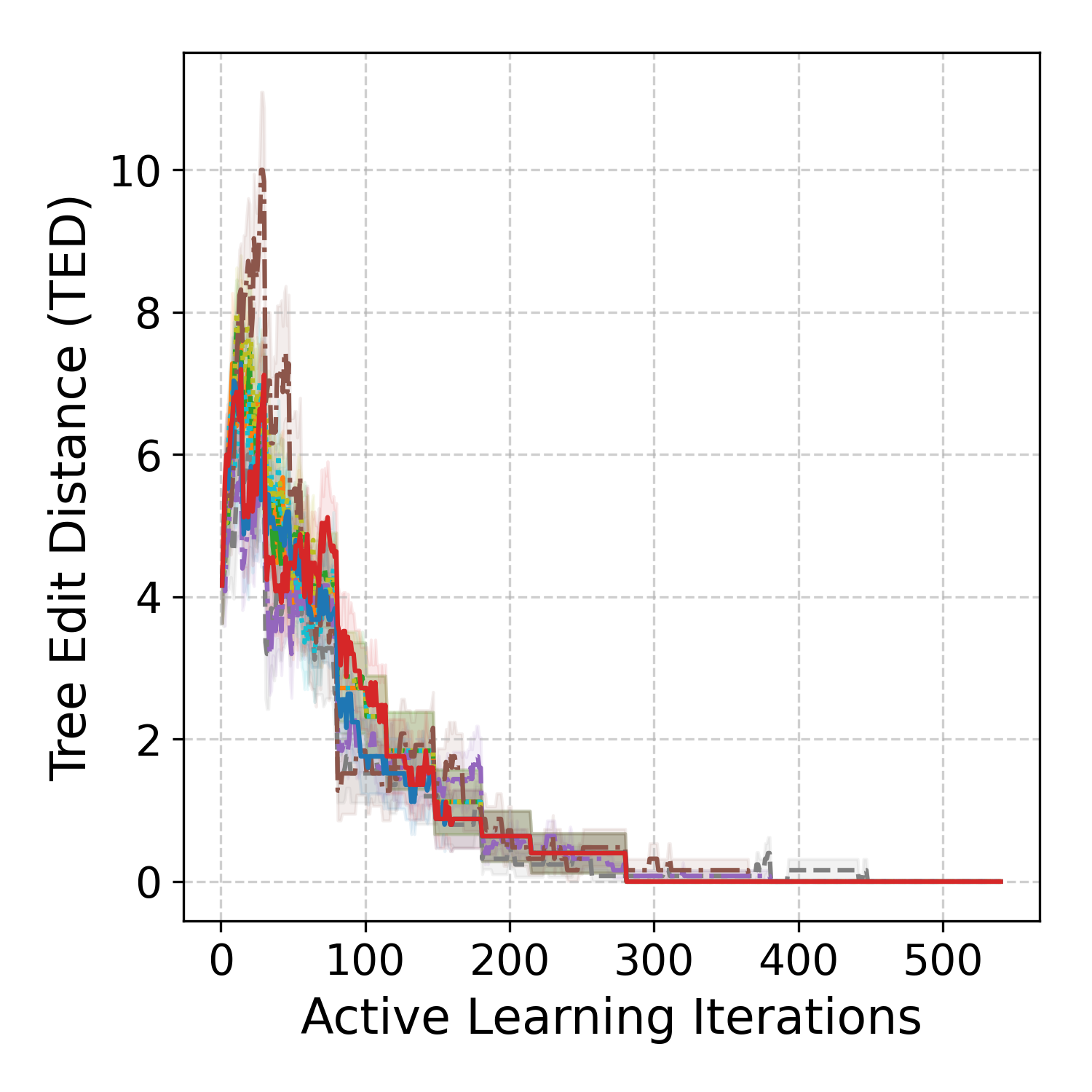}
        \caption{Breast Cancer WI}
    \end{subfigure}
    \hfill
    \begin{subfigure}[b]{0.19\textwidth}
        \includegraphics[width=\linewidth]{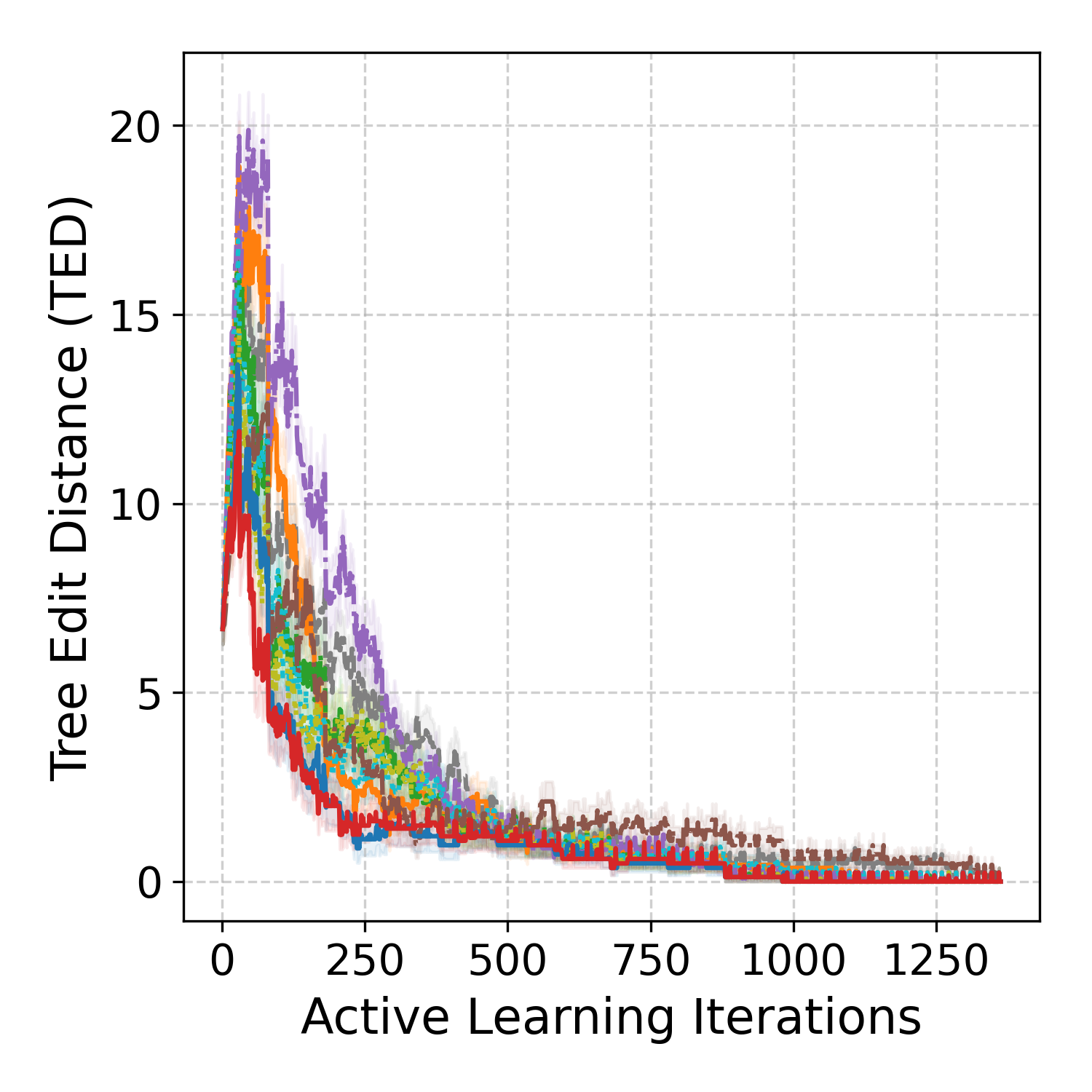}
        \caption{Car Evaluation}
    \end{subfigure}

    \vspace{0.2em}

    \begin{subfigure}[b]{0.19\textwidth}
        \includegraphics[width=\linewidth]{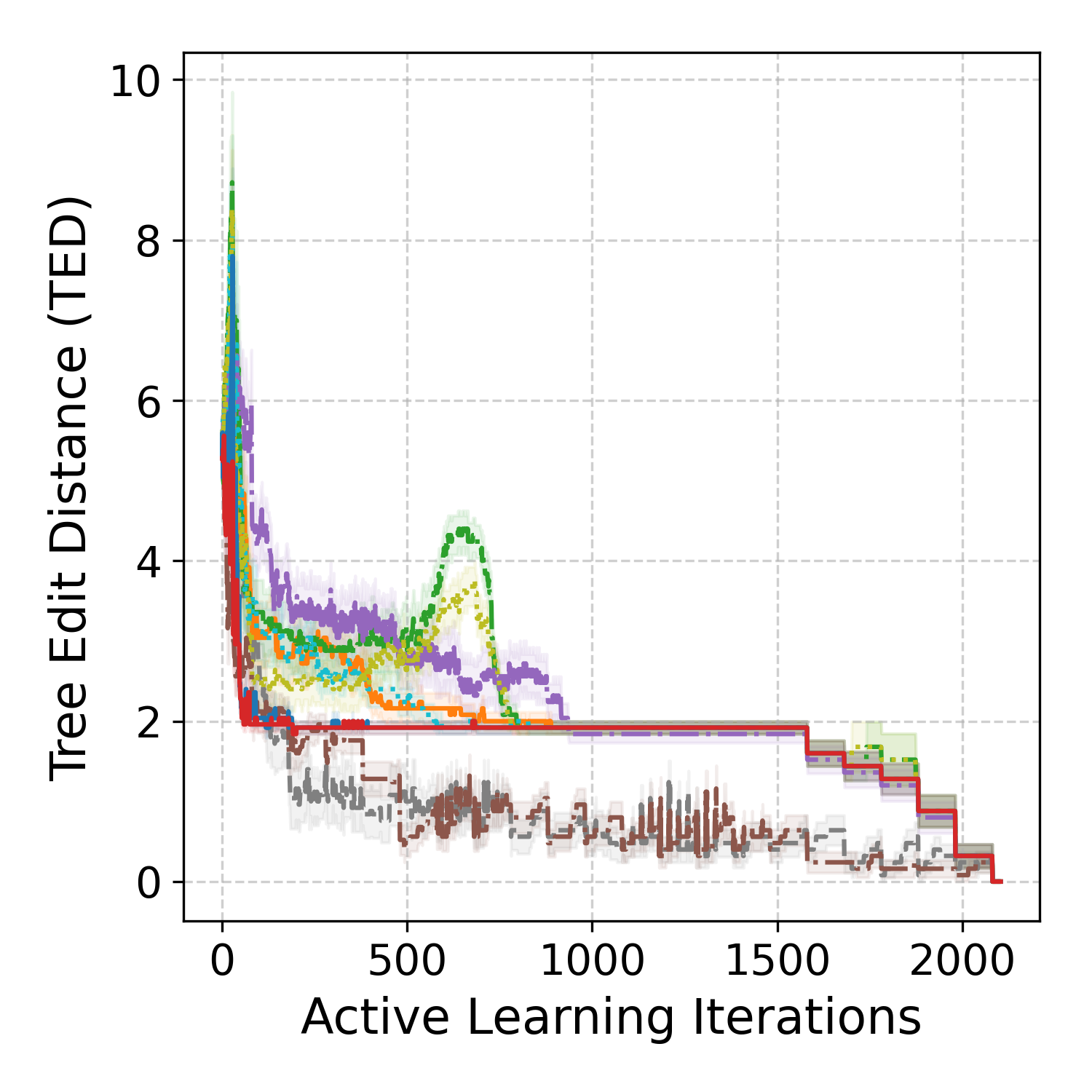}
        \caption{Cheap Restaurant}
    \end{subfigure}
    \hfill
    \begin{subfigure}[b]{0.19\textwidth}
        \includegraphics[width=\linewidth]{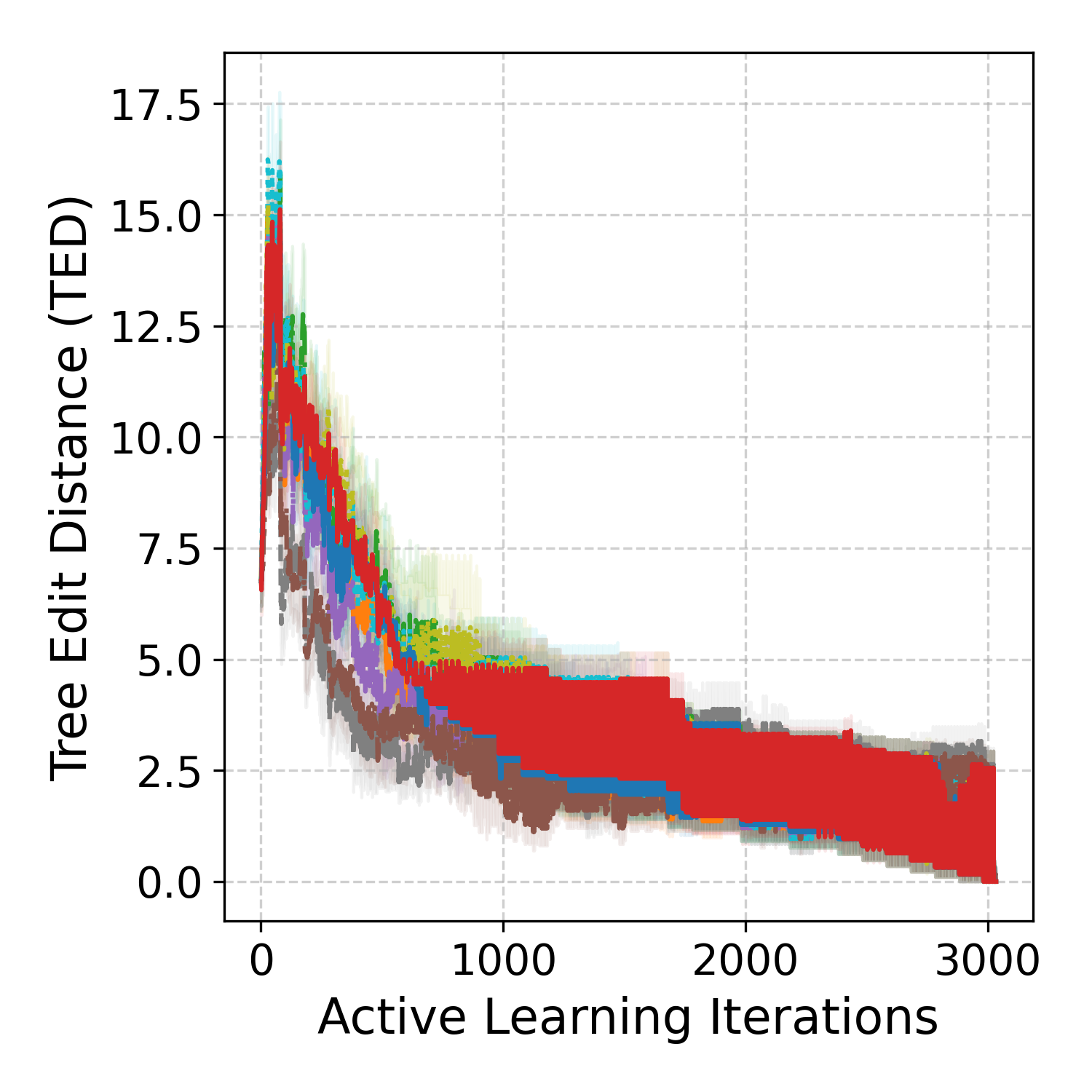}
        \caption{Coffee House}
    \end{subfigure}
    \hfill
    \begin{subfigure}[b]{0.19\textwidth}
        \includegraphics[width=\linewidth]{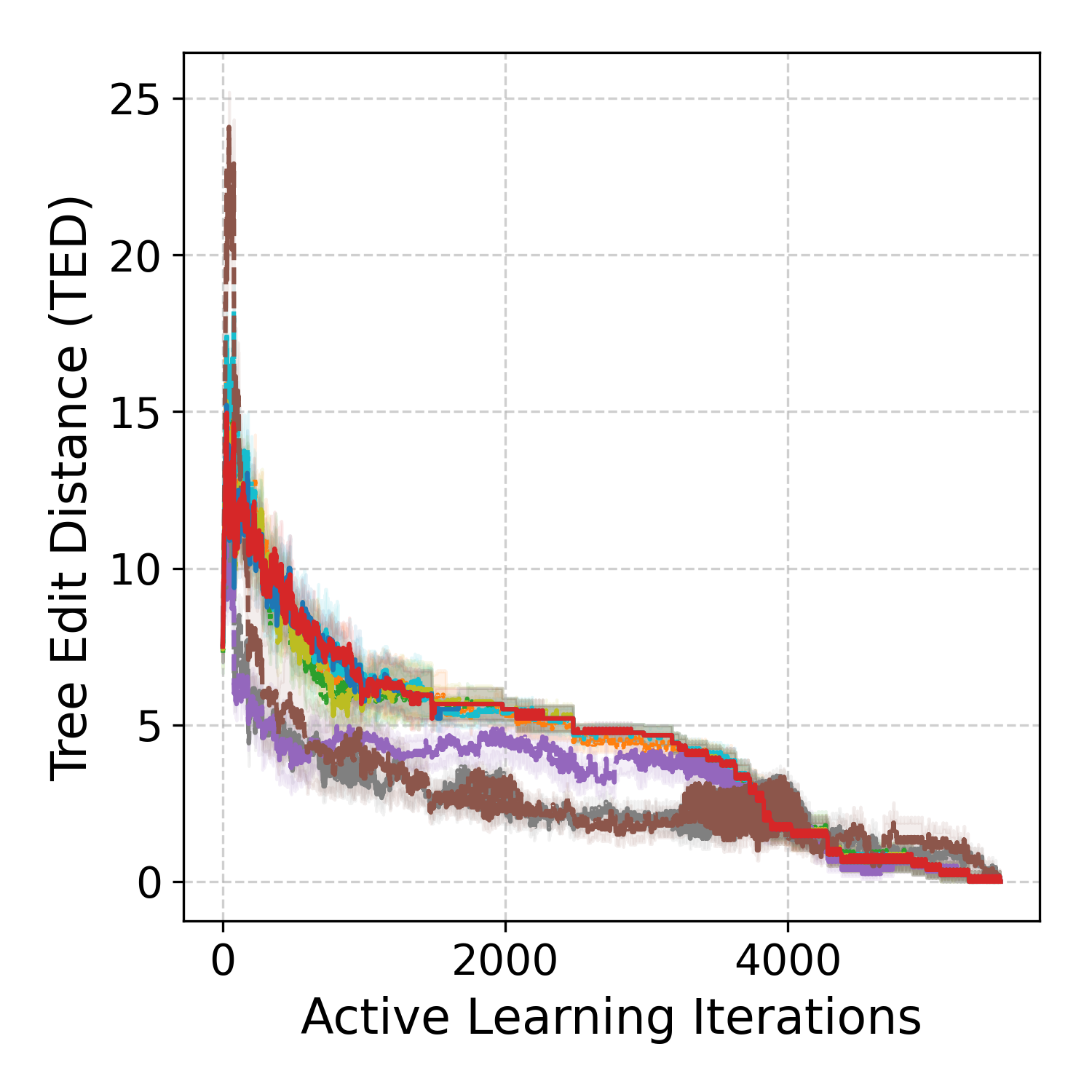}
        \caption{COMPAS}
    \end{subfigure}
    \hfill
    \begin{subfigure}[b]{0.19\textwidth}
        \includegraphics[width=\linewidth]{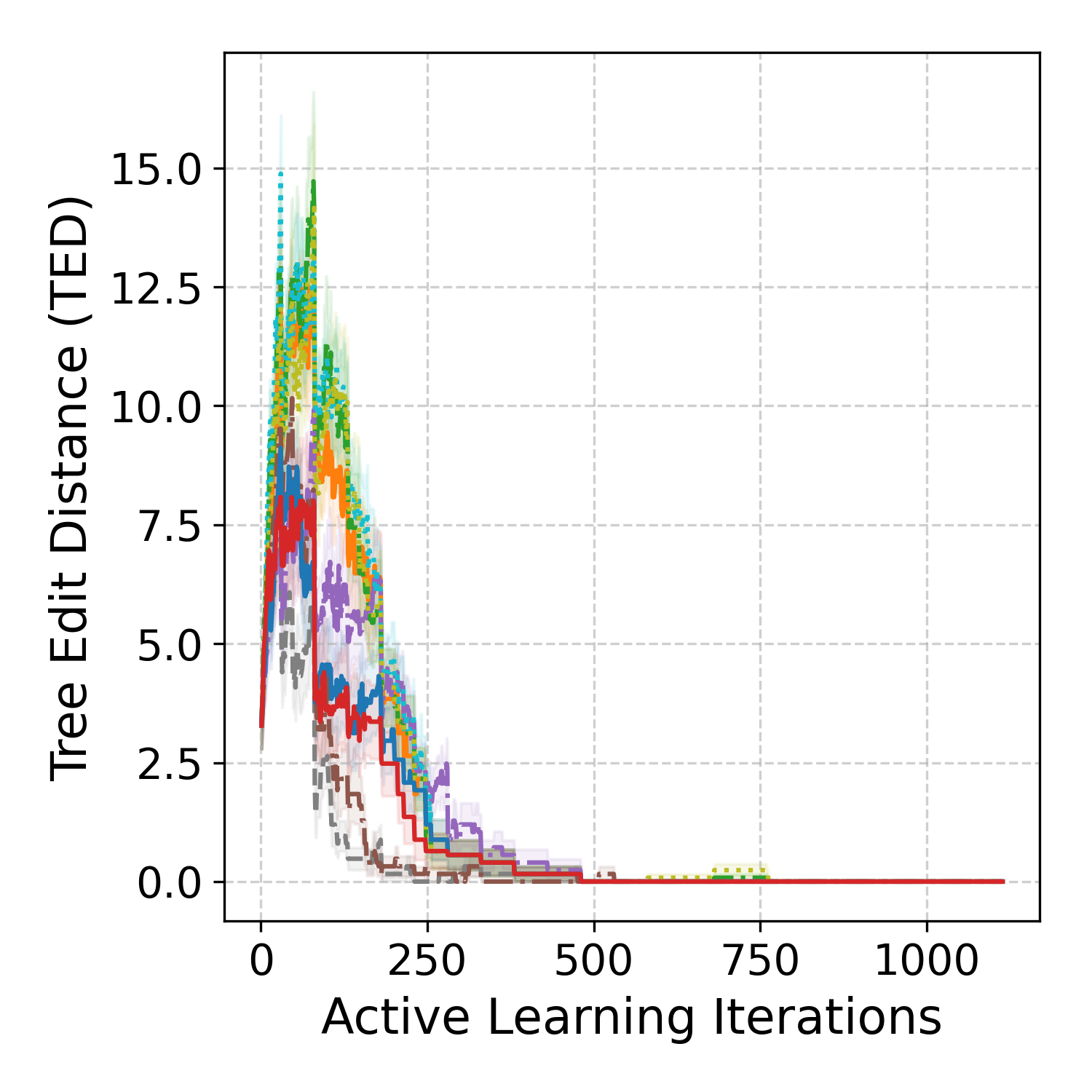}
        \caption{Expensive Restaurant}
    \end{subfigure}
    \hfill
    \begin{subfigure}[b]{0.19\textwidth}
        \includegraphics[width=\linewidth]{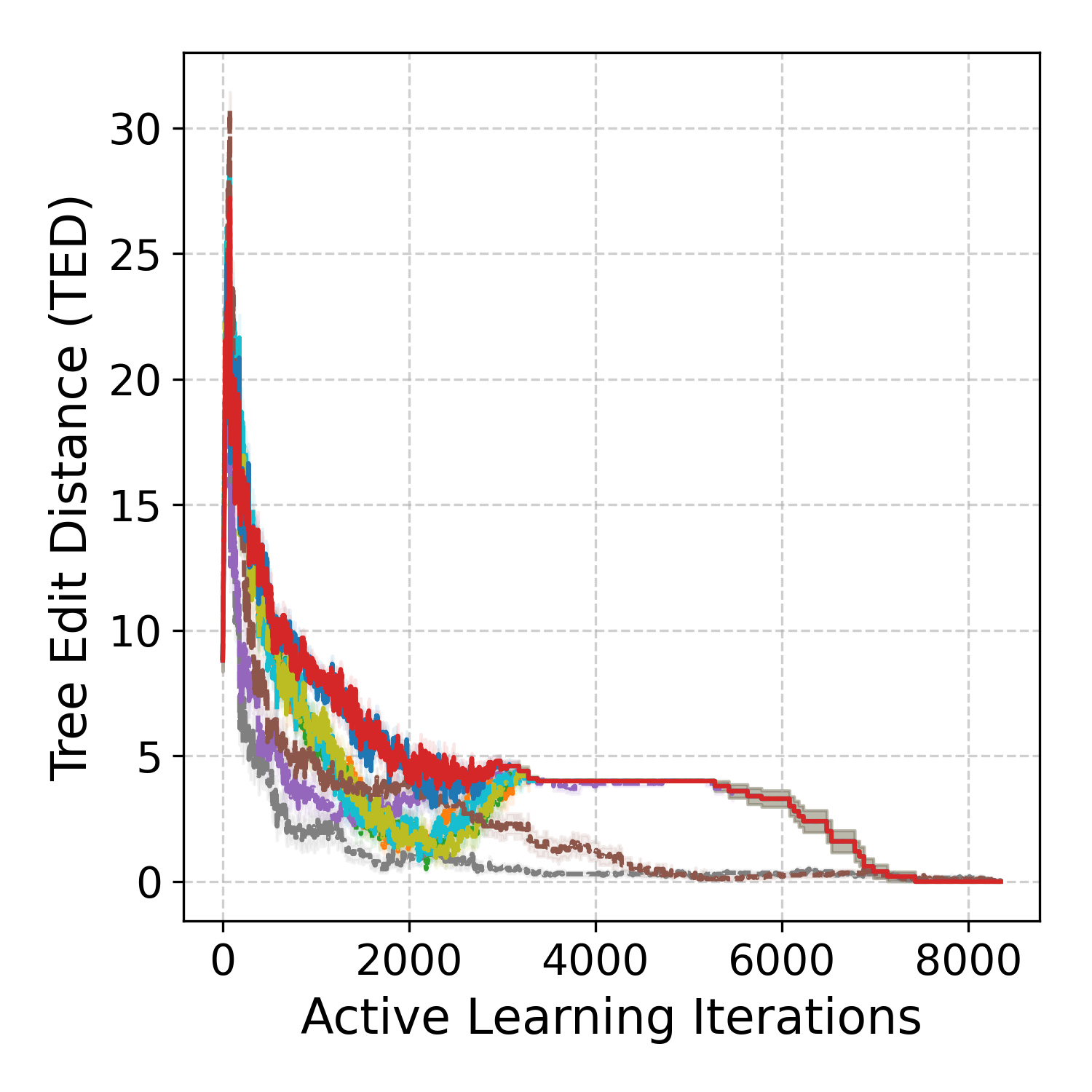}
        \caption{FICO}
    \end{subfigure}

    \vspace{0.2em}

    \begin{subfigure}[b]{0.19\textwidth}
        \includegraphics[width=\linewidth]{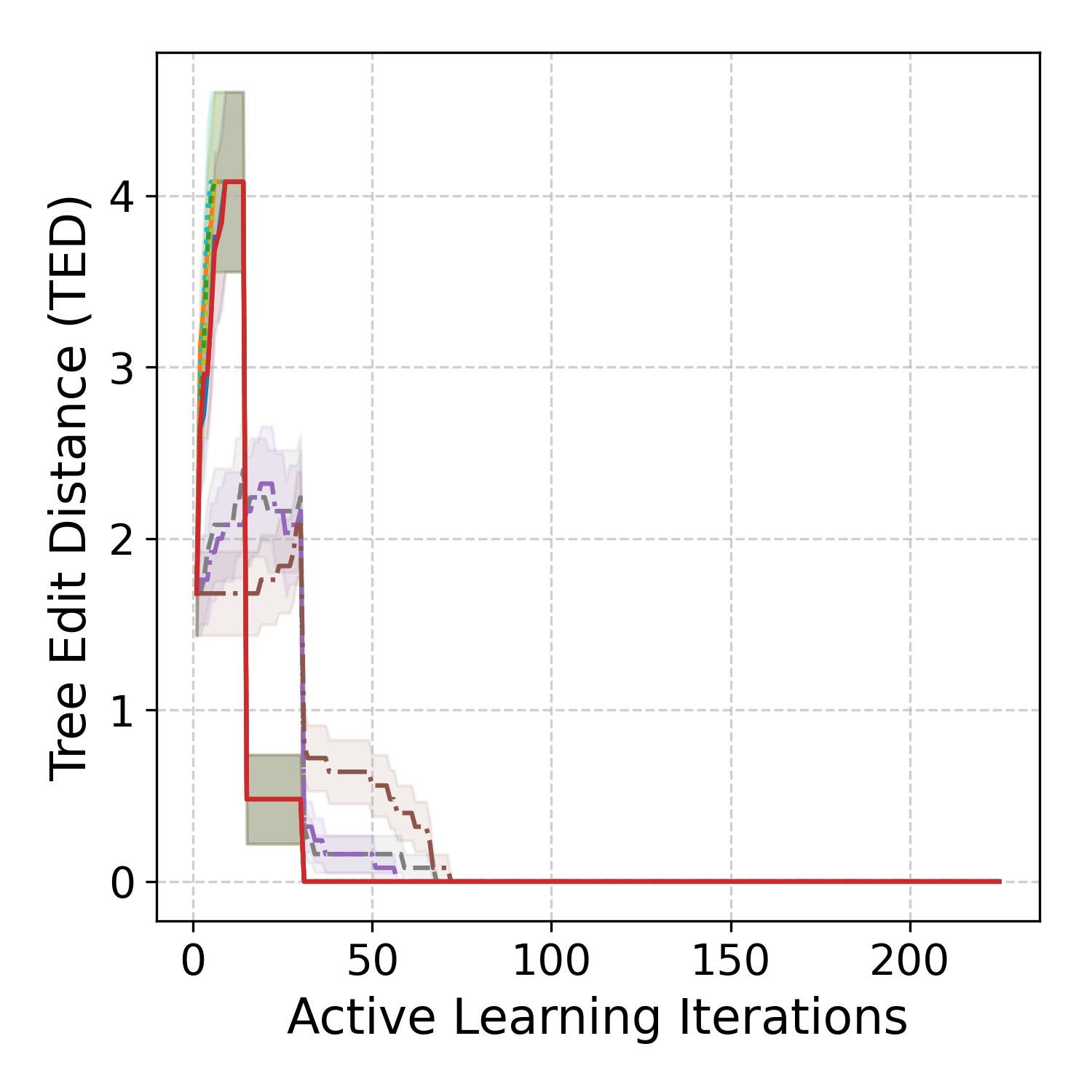}
        \caption{Haberman}
    \end{subfigure}
    \hfill
    \begin{subfigure}[b]{0.19\textwidth}
        \includegraphics[width=\linewidth]{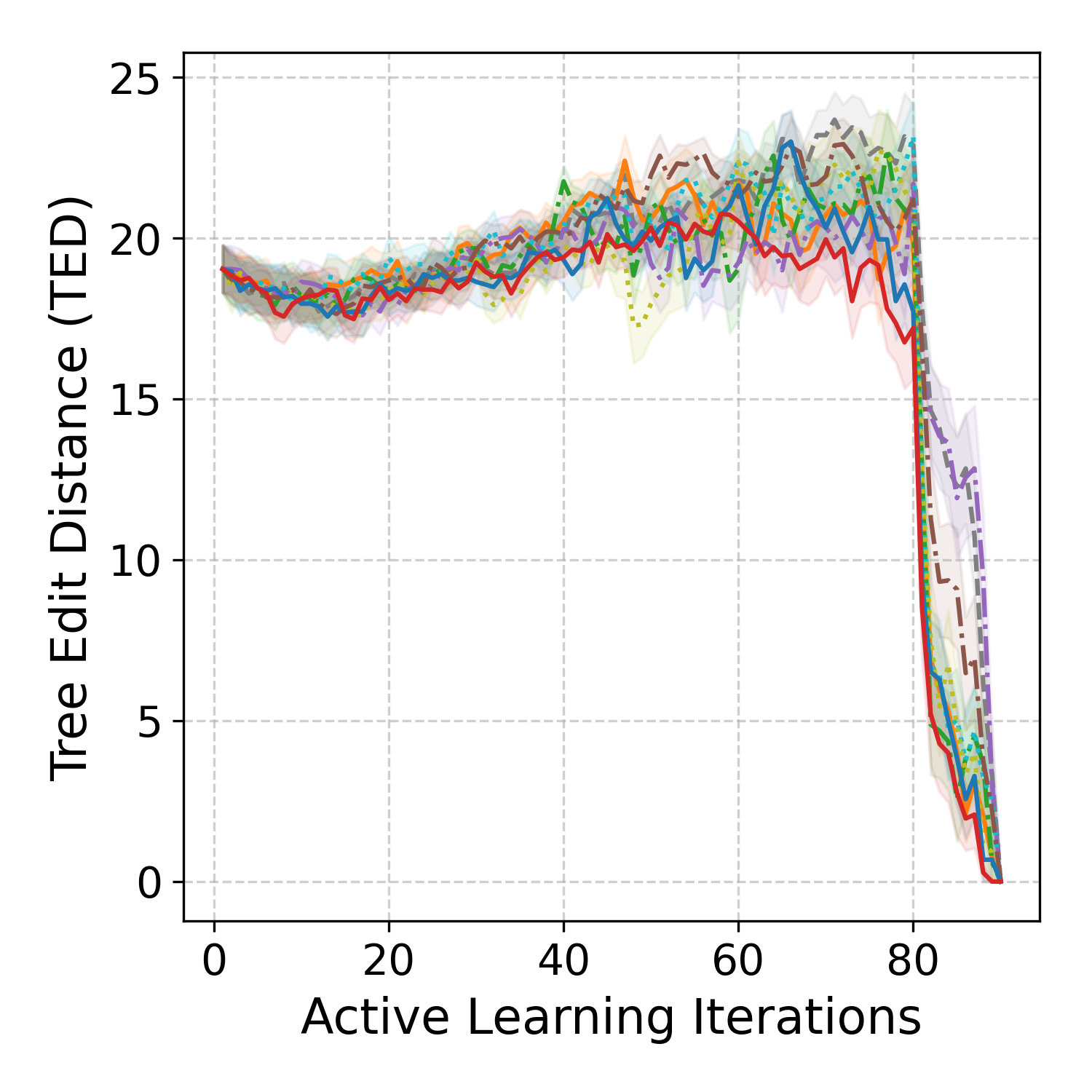}
        \caption{Hepatitis}
    \end{subfigure}
    \hfill
    \begin{subfigure}[b]{0.19\textwidth}
        \includegraphics[width=\linewidth]{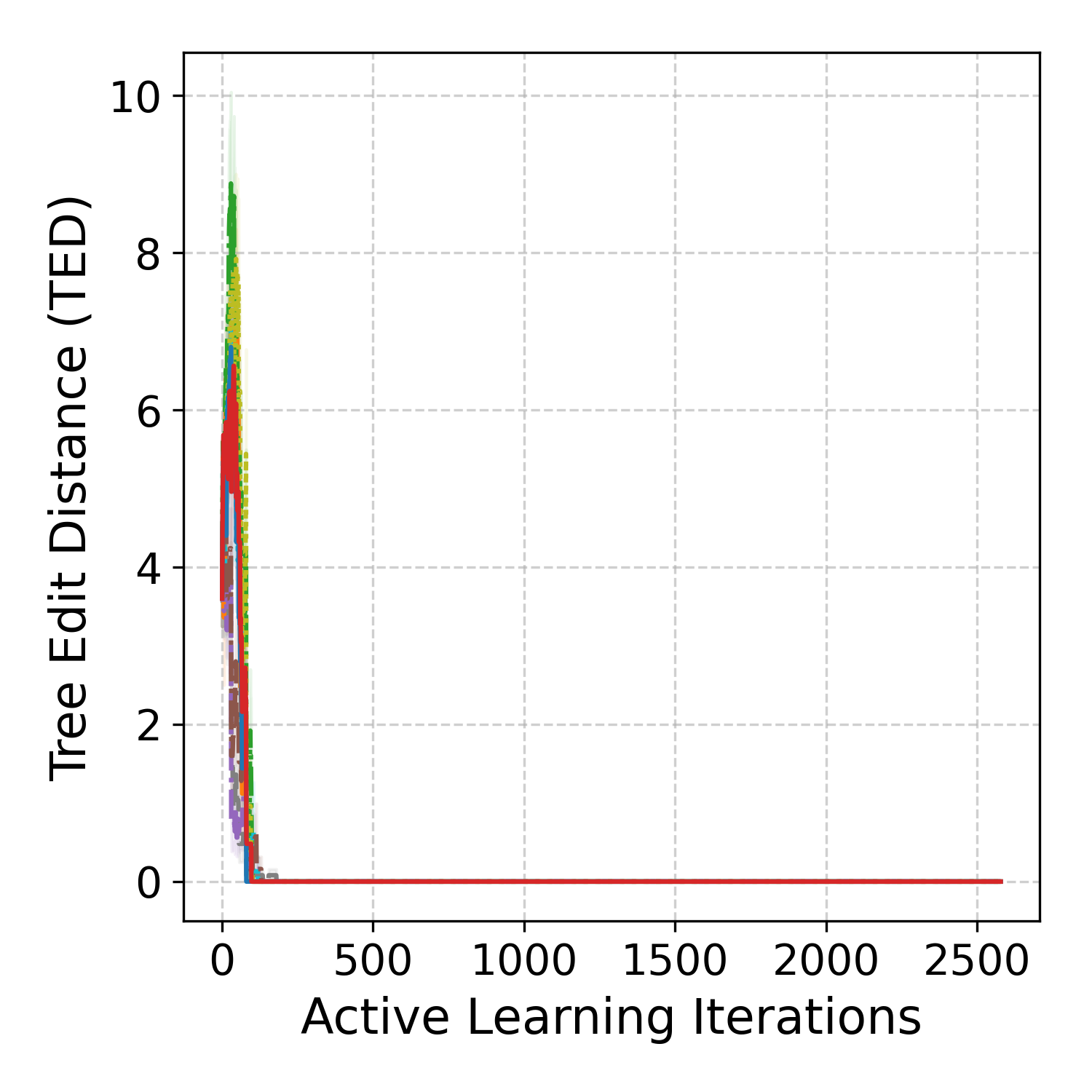}
        \caption{Hypothyroid}
    \end{subfigure}
    \hfill
    \begin{subfigure}[b]{0.19\textwidth}
        \includegraphics[width=\linewidth]{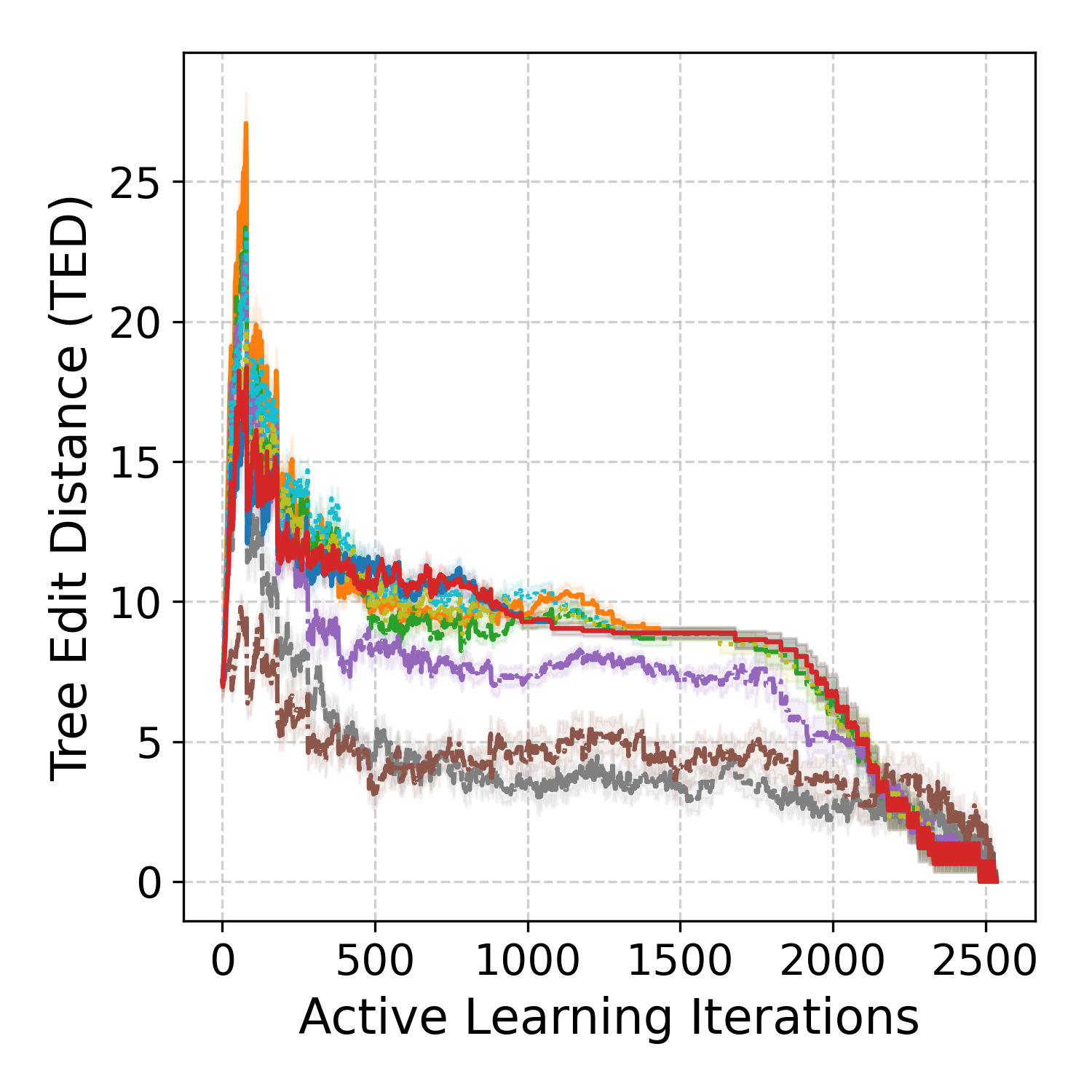}
        \caption{Kr-vs-Kp}
    \end{subfigure}
    \hfill
    \begin{subfigure}[b]{0.19\textwidth}
        \includegraphics[width=\linewidth]{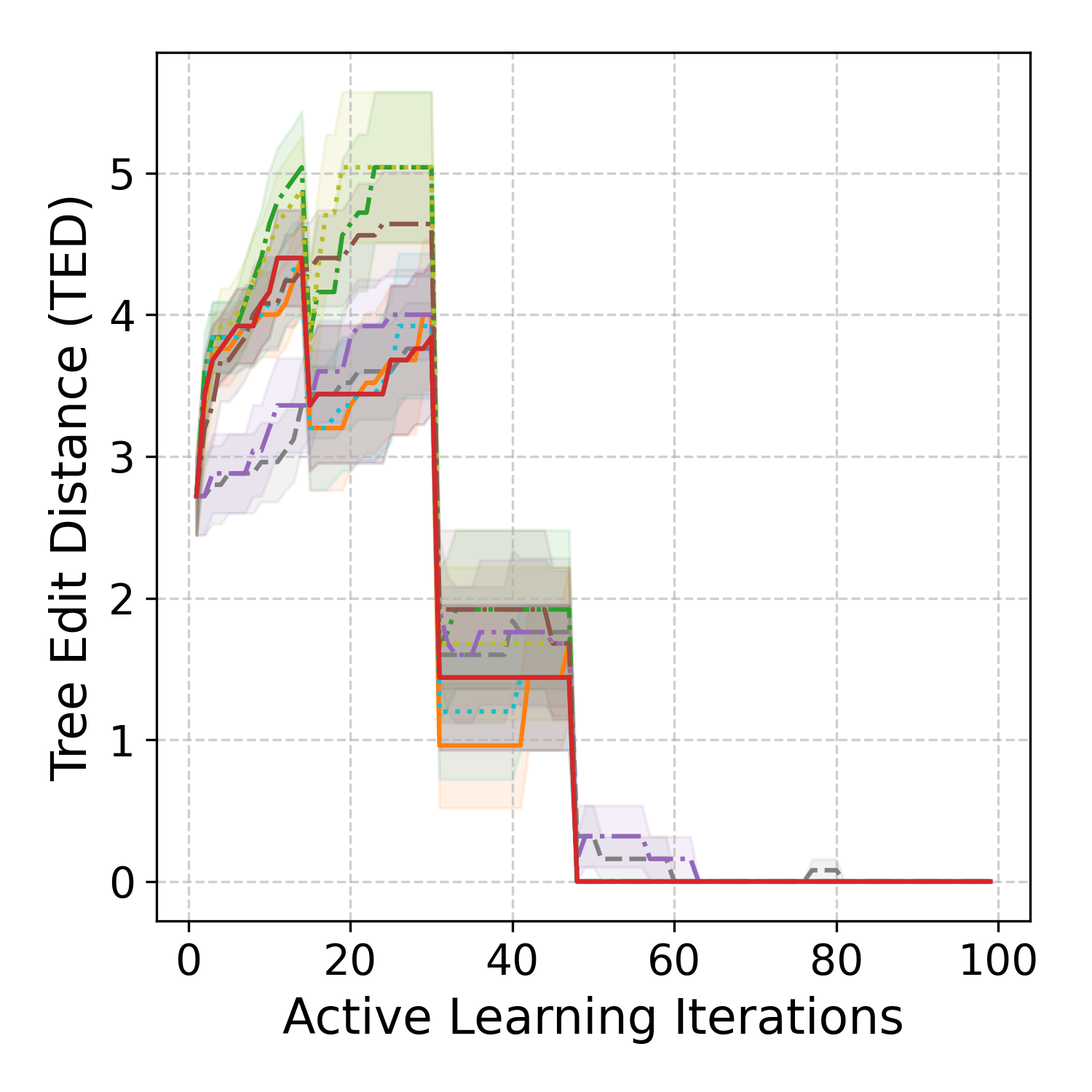}
        \caption{Lymphography}
    \end{subfigure}

    \vspace{0.2em}

    \begin{subfigure}[b]{0.19\textwidth}
        \includegraphics[width=\linewidth]{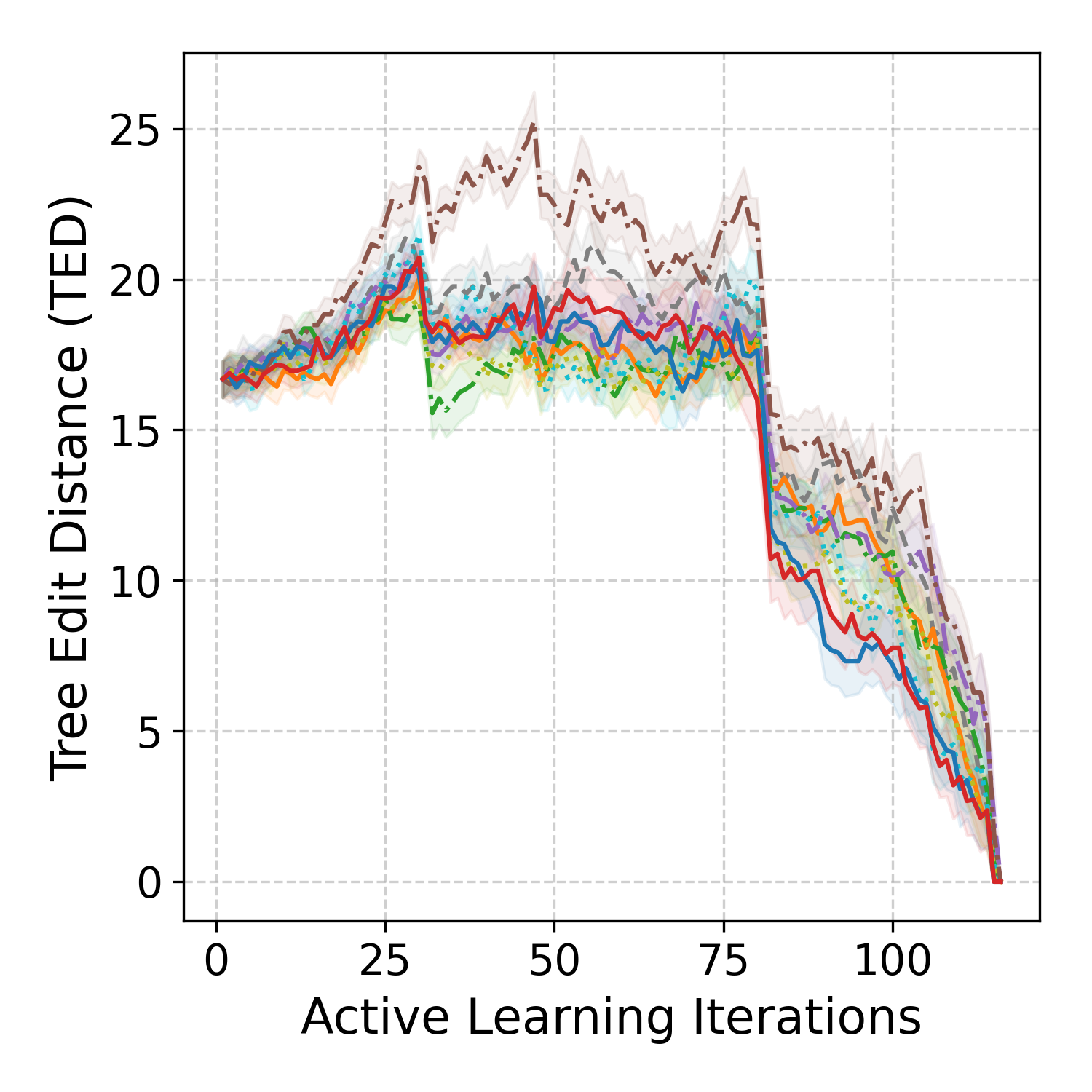}
        \caption{MONK-2}
    \end{subfigure}
    \hfill
    \begin{subfigure}[b]{0.19\textwidth}
        \includegraphics[width=\linewidth]{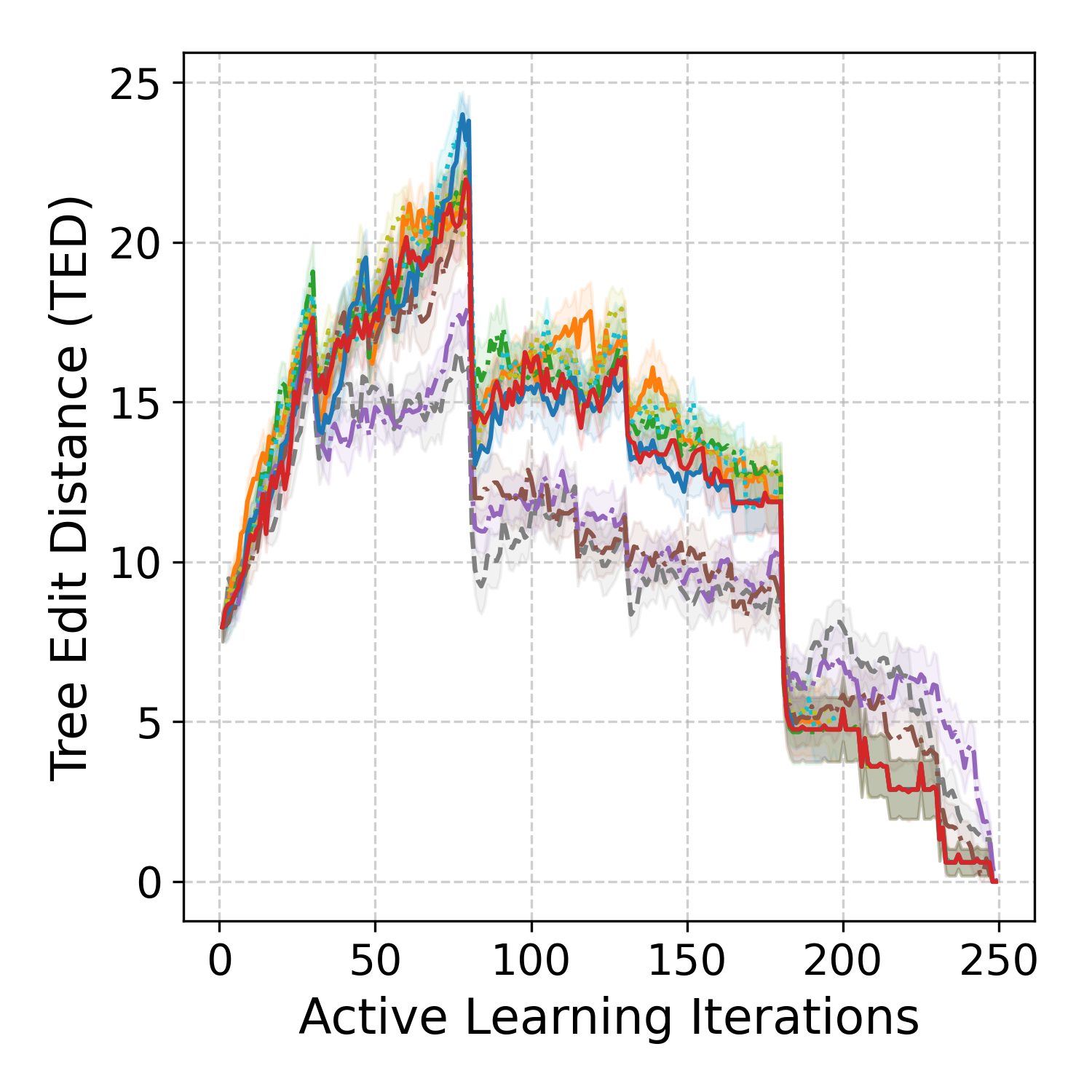}
        \caption{Primary Tumor}
    \end{subfigure}
    \hfill
    \begin{subfigure}[b]{0.19\textwidth}
        \includegraphics[width=\linewidth]{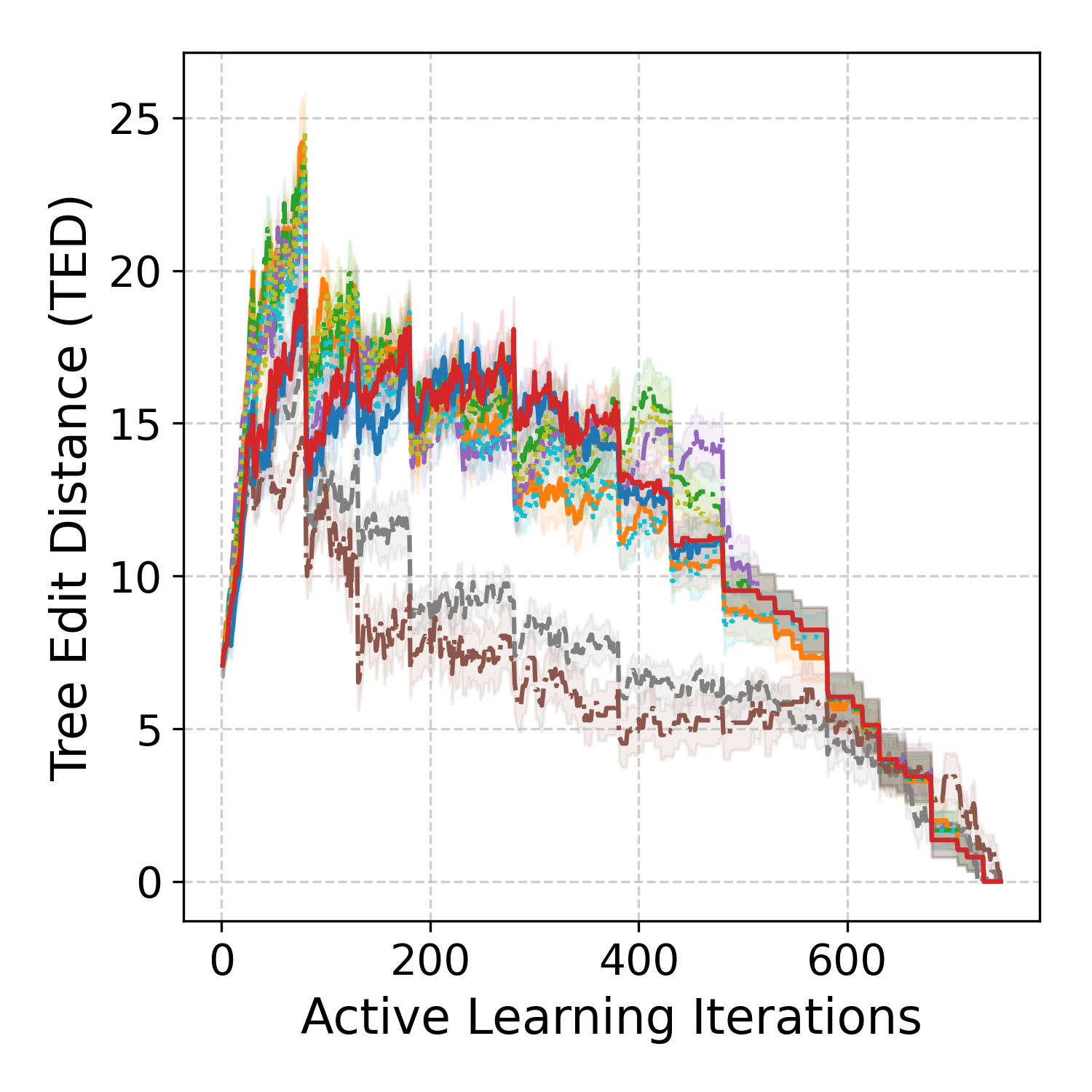}
        \caption{Tic-Tac-Toe}
    \end{subfigure}
    \hfill
    \begin{subfigure}[b]{0.19\textwidth}
        \includegraphics[width=\linewidth]{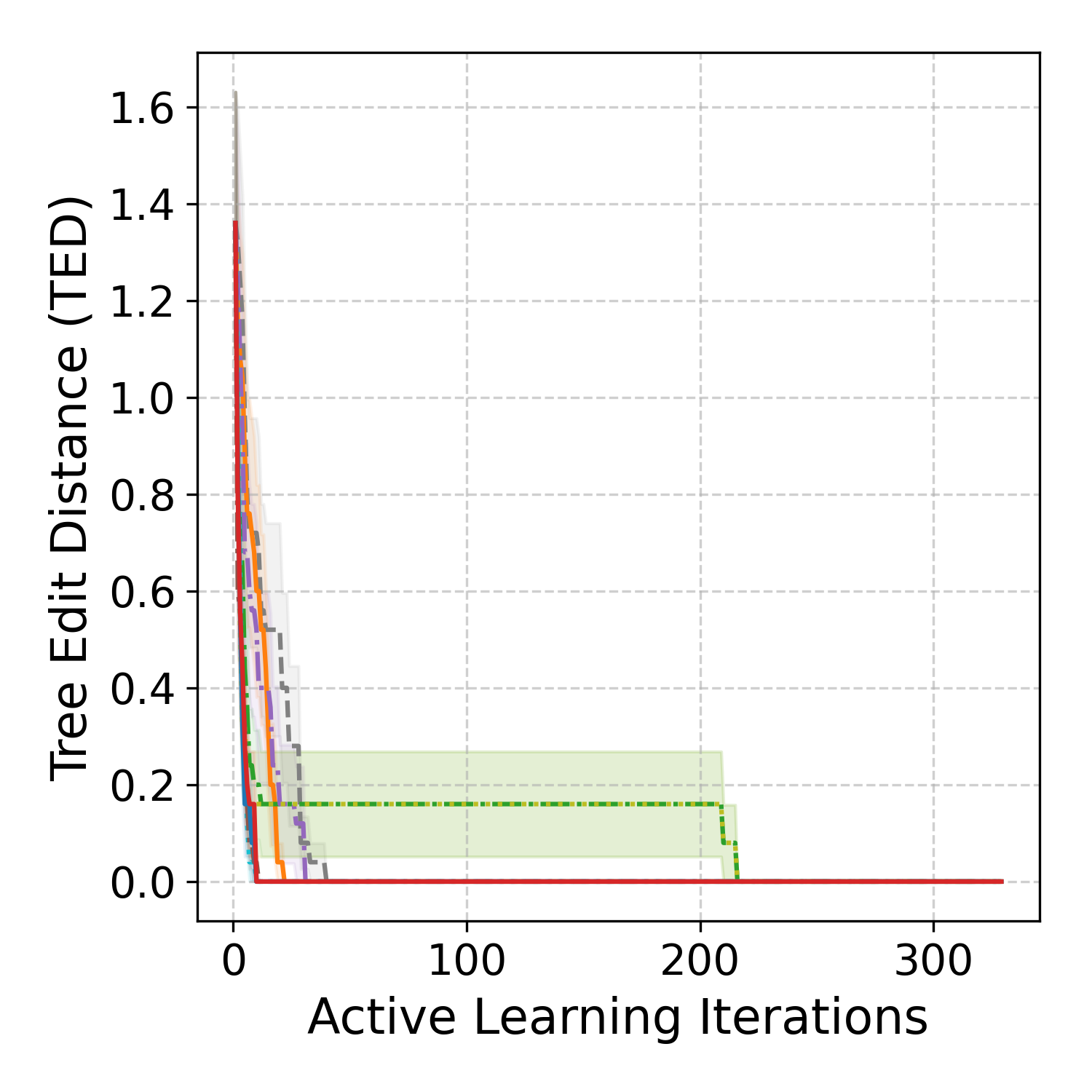}
        \caption{Vote}
    \end{subfigure}
    \hfill
    \begin{subfigure}[b]{0.19\textwidth}
        \includegraphics[width=\linewidth]{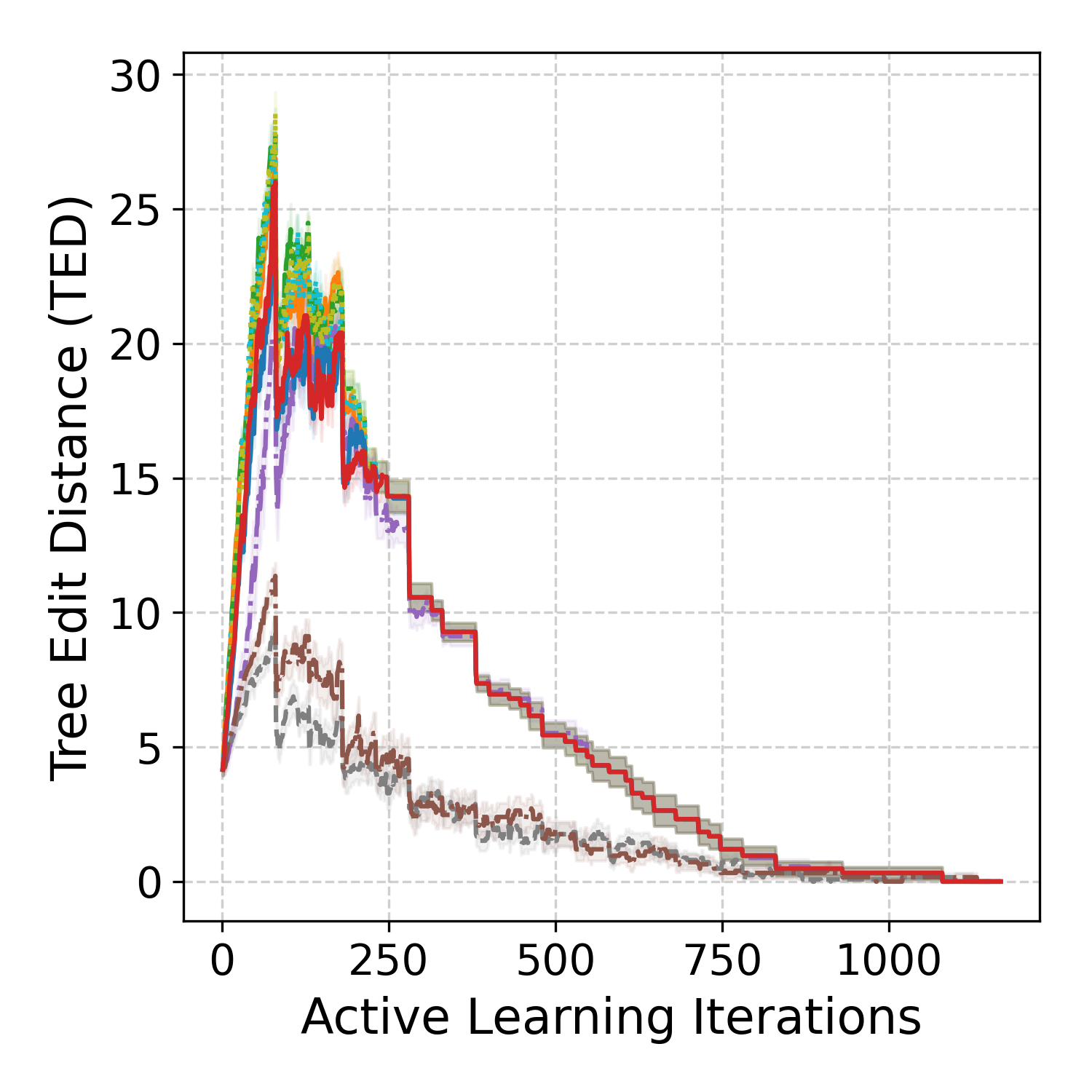}
        \caption{Yeast}
    \end{subfigure}

    \vspace{0.5em}
    \centering
    \includegraphics[width=0.60\linewidth]{upload_all_files/study1_main_AL_results/benchmark_legend.png}

    \caption{\textbf{Tree Edit Distance Benchmarks.} Evaluation of the alignment between the decision logic of the learned hypotheses and the ground-truth Oracle tree. A lower distance indicates that BREAL is not merely achieving predictive success through spurious correlations, but is successfully recovering the true underlying structural logic of the task.}
    \label{fig:BenchmarkGrid_TreeEditDistance}
\end{figure*}
\clearpage

\begin{figure*}[!t] 
\centering
    \begin{subfigure}[b]{0.19\textwidth}
        \includegraphics[width=\linewidth]{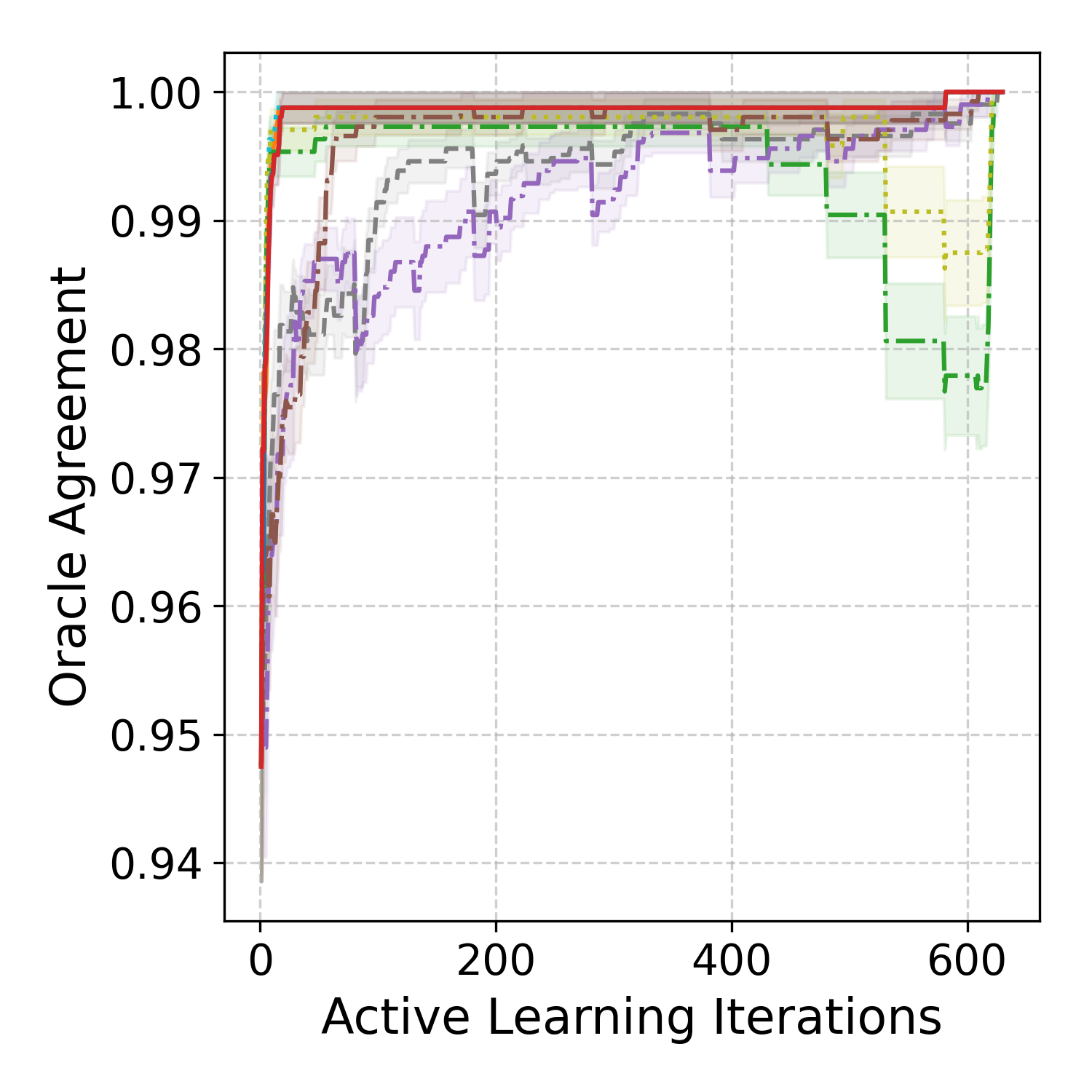}
        \caption{Anneal}
    \end{subfigure}
    \hfill
    \begin{subfigure}[b]{0.19\textwidth}
        \includegraphics[width=\linewidth]{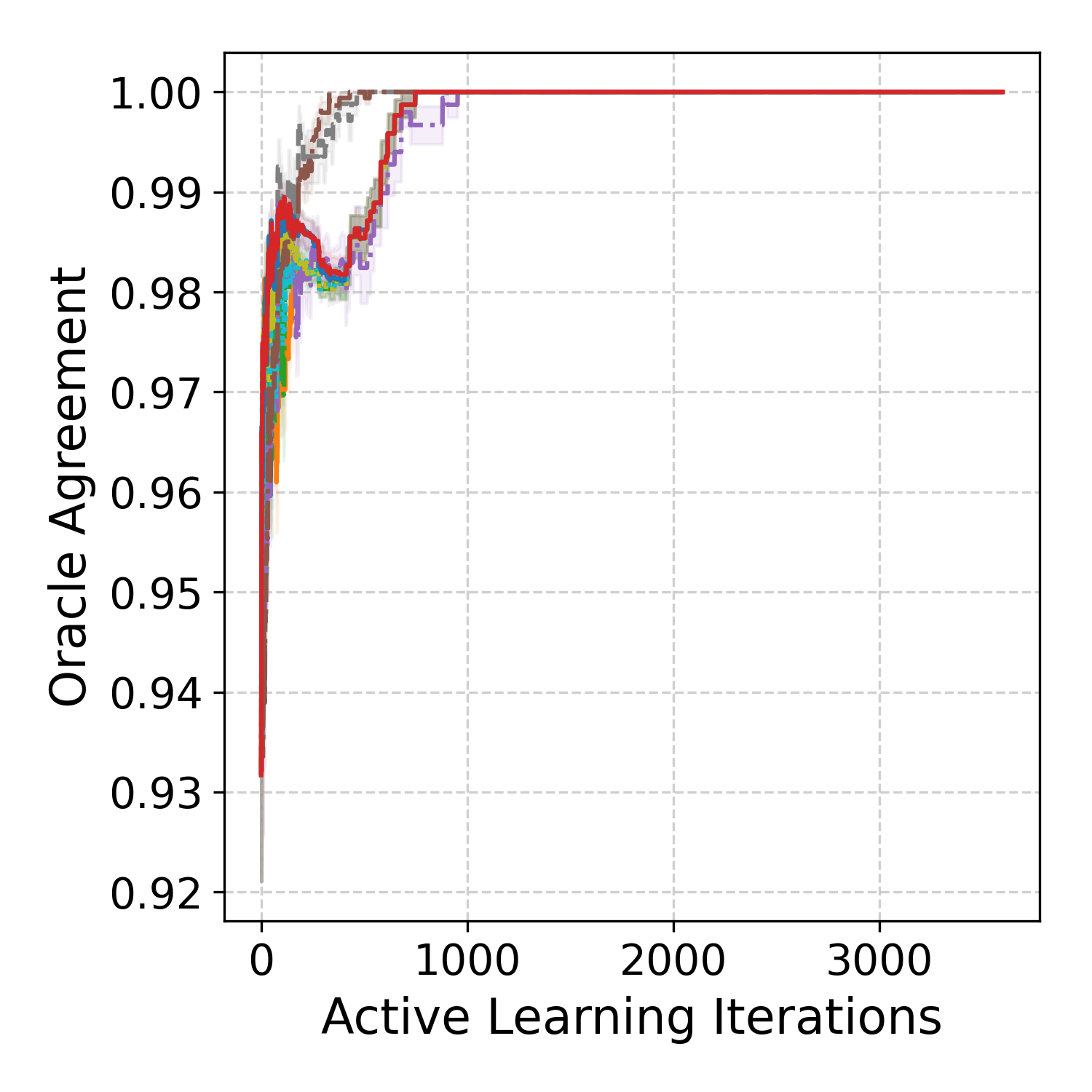}
        \caption{Bank Marketing}
    \end{subfigure}
    \hfill
    \begin{subfigure}[b]{0.19\textwidth}
        \includegraphics[width=\linewidth]{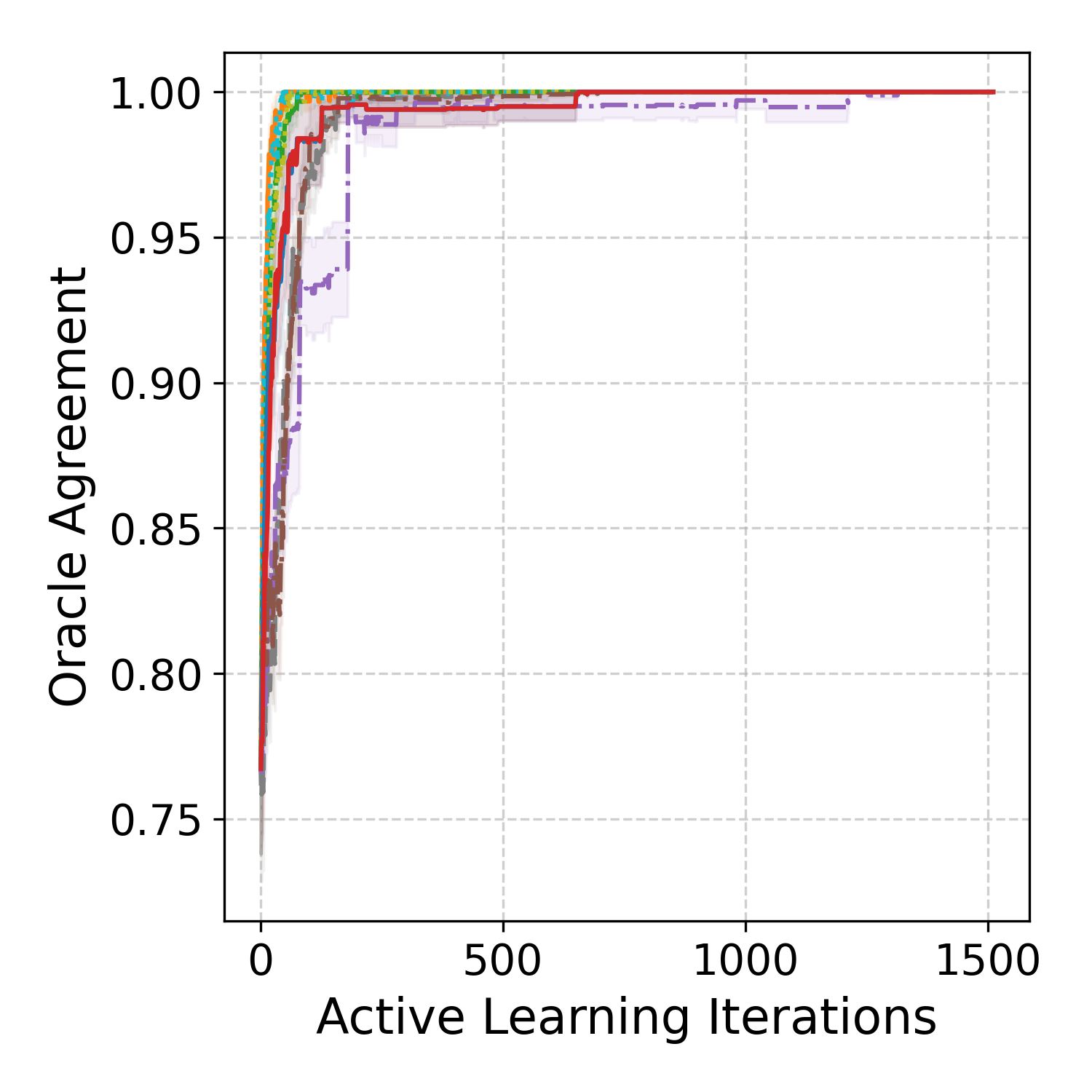}
        \caption{Bar-7}
    \end{subfigure}
    \hfill
    \begin{subfigure}[b]{0.19\textwidth}
        \includegraphics[width=\linewidth]{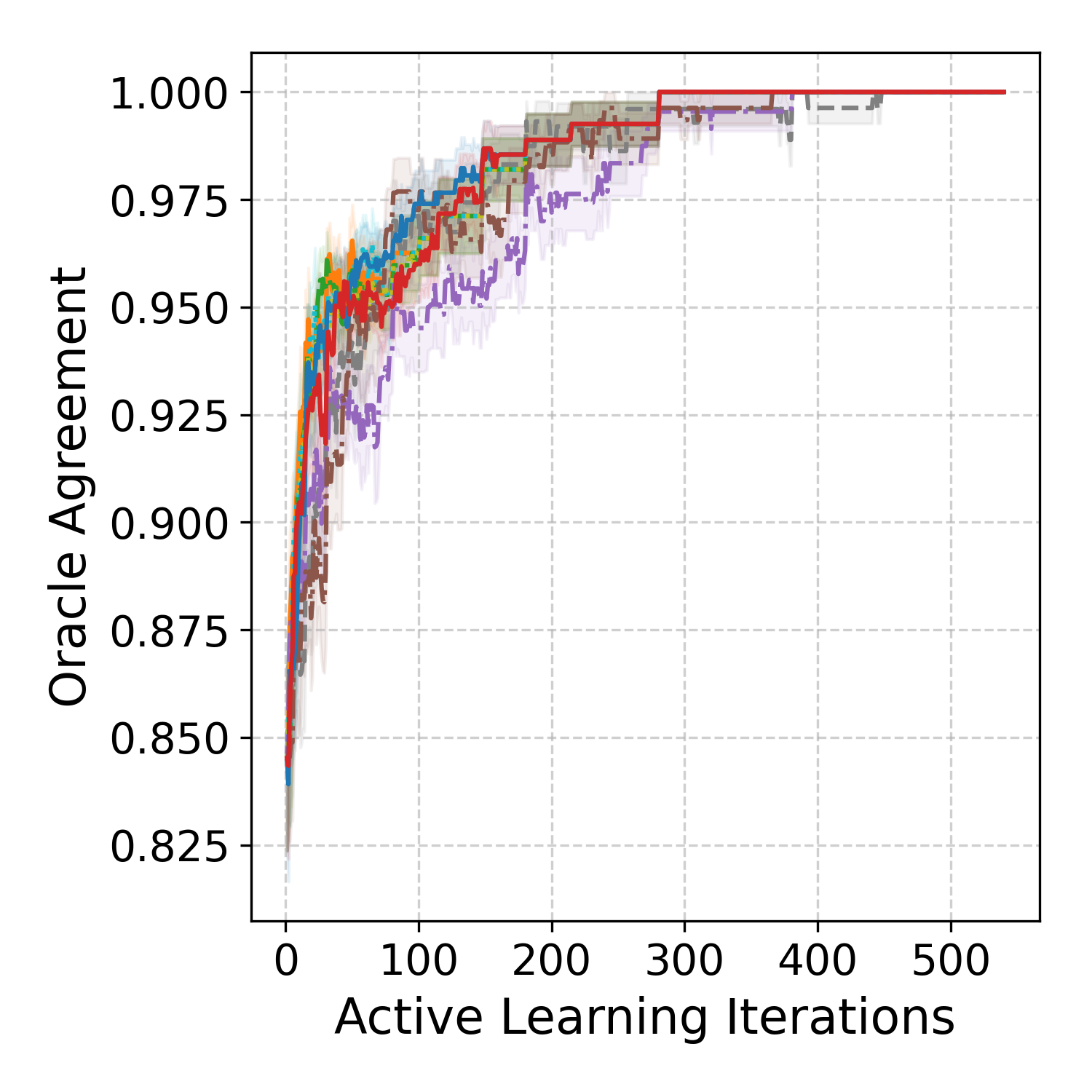}
        \caption{Breast Cancer WI}
    \end{subfigure}
    \hfill
    \begin{subfigure}[b]{0.19\textwidth}
        \includegraphics[width=\linewidth]{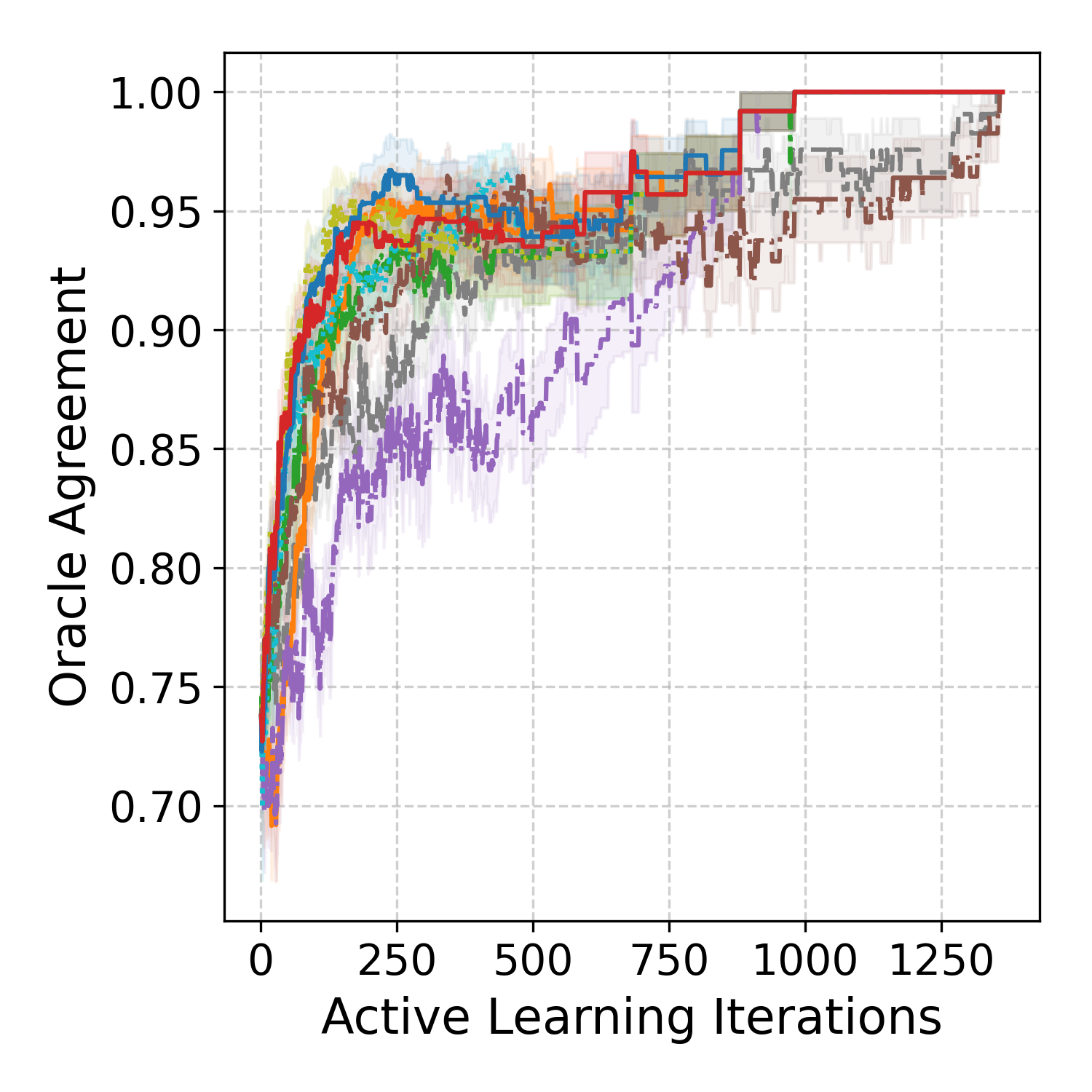}
        \caption{Car Evaluation}
    \end{subfigure}

    \vspace{0.2em}

    \begin{subfigure}[b]{0.19\textwidth}
        \includegraphics[width=\linewidth]{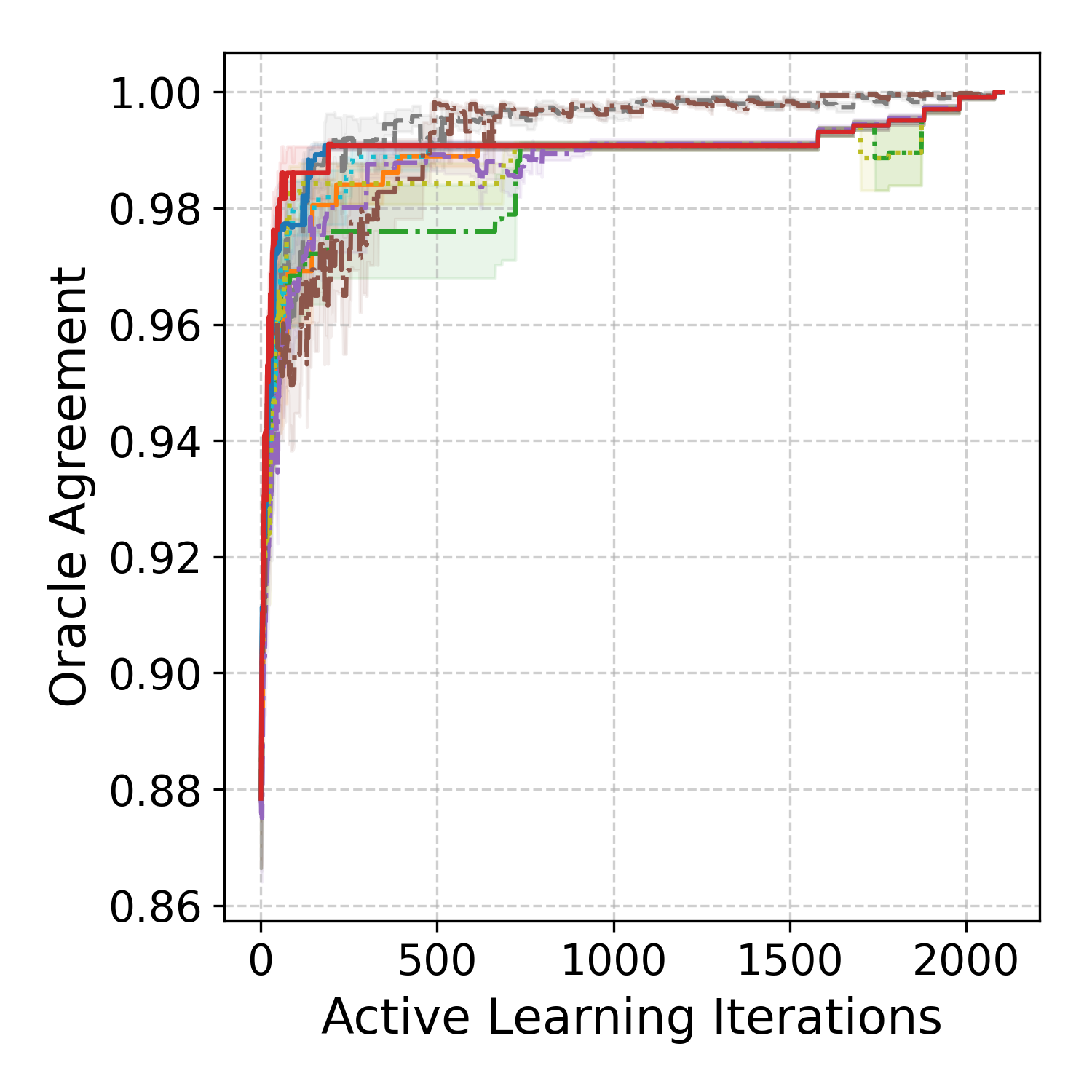}
        \caption{Cheap Restaurant}
    \end{subfigure}
    \hfill
    \begin{subfigure}[b]{0.19\textwidth}
        \includegraphics[width=\linewidth]{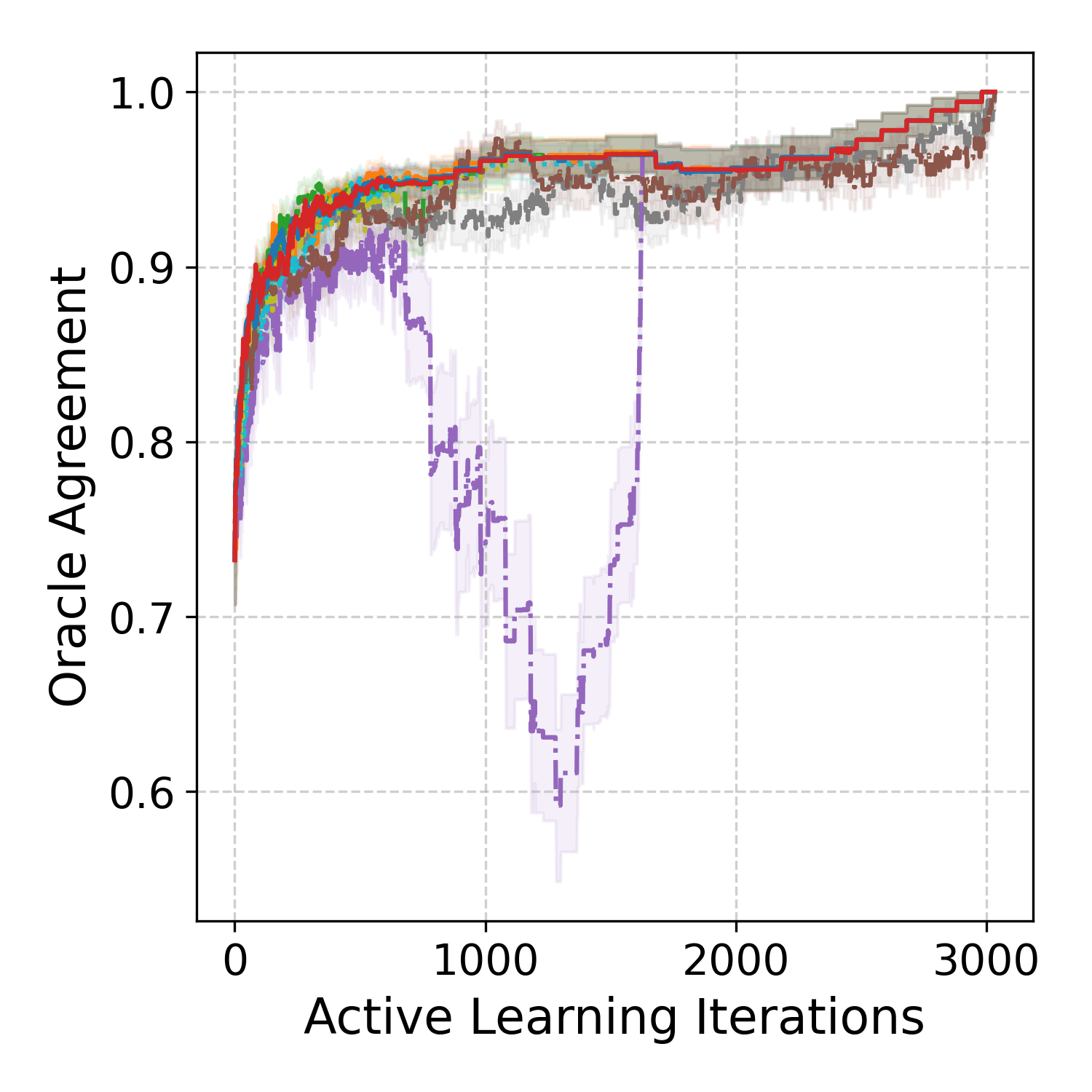}
        \caption{Coffee House}
    \end{subfigure}
    \hfill
    \begin{subfigure}[b]{0.19\textwidth}
        \includegraphics[width=\linewidth]{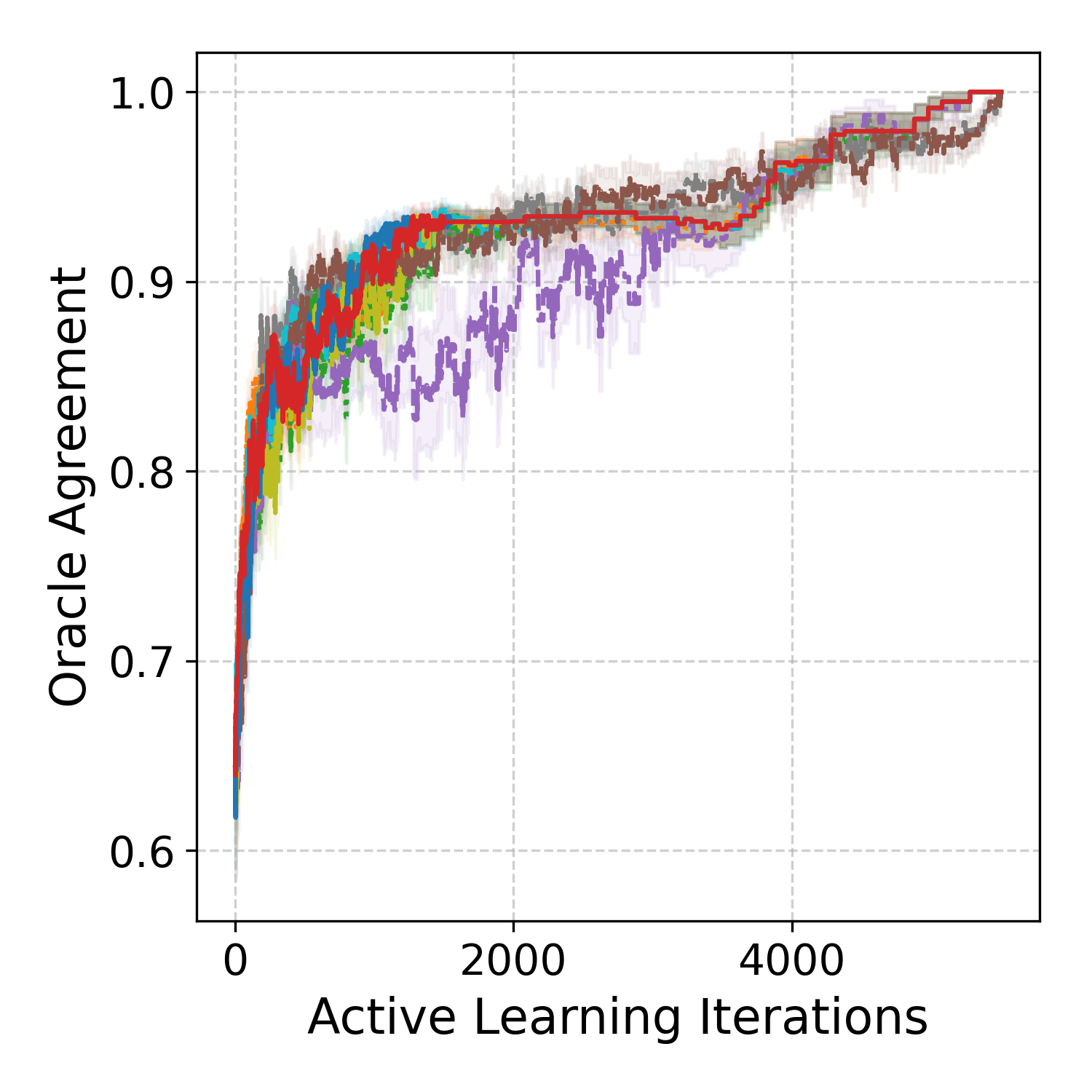}
        \caption{COMPAS}
    \end{subfigure}
    \hfill
    \begin{subfigure}[b]{0.19\textwidth}
        \includegraphics[width=\linewidth]{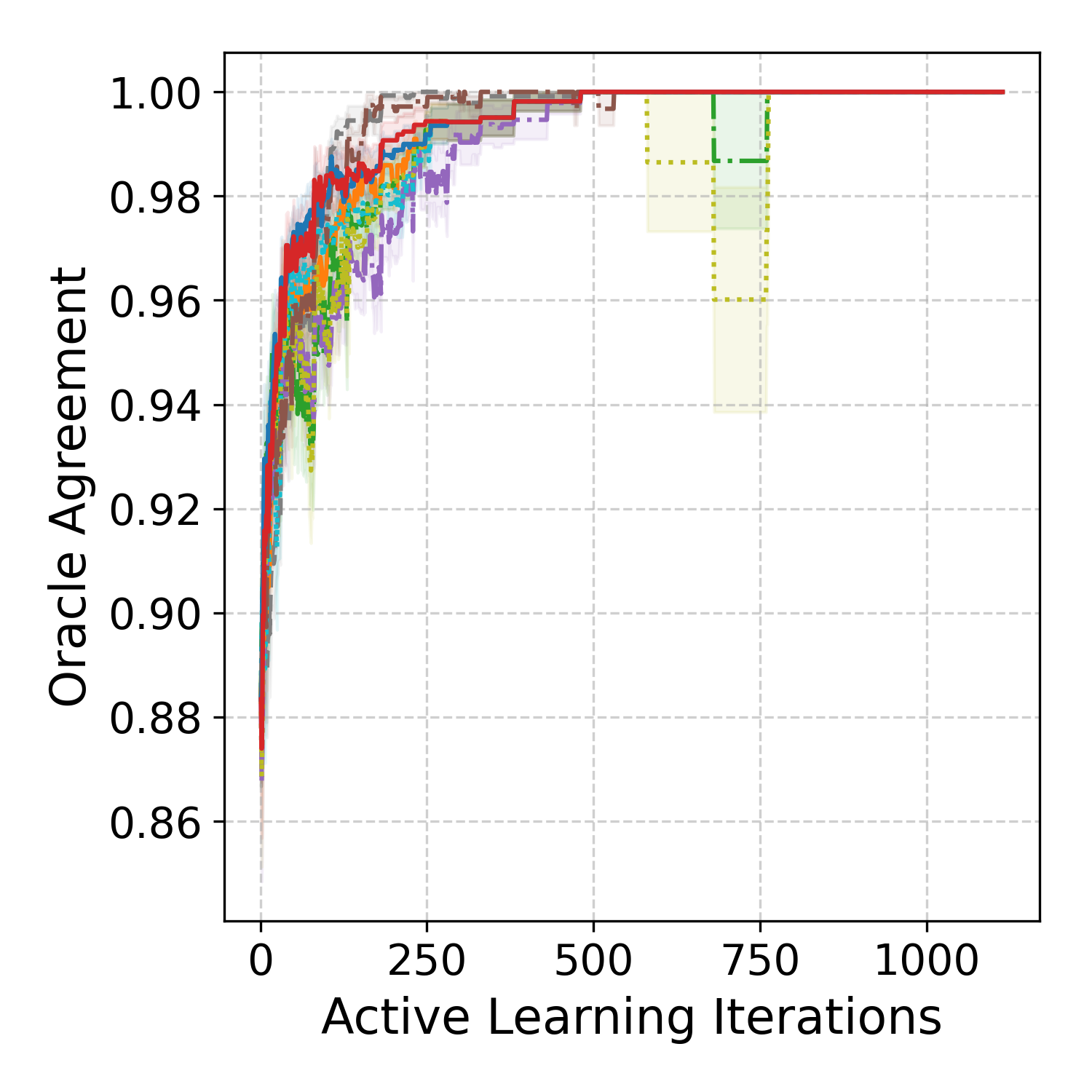}
        \caption{Expensive Restaurant}
    \end{subfigure}
    \hfill
    \begin{subfigure}[b]{0.19\textwidth}
        \includegraphics[width=\linewidth]{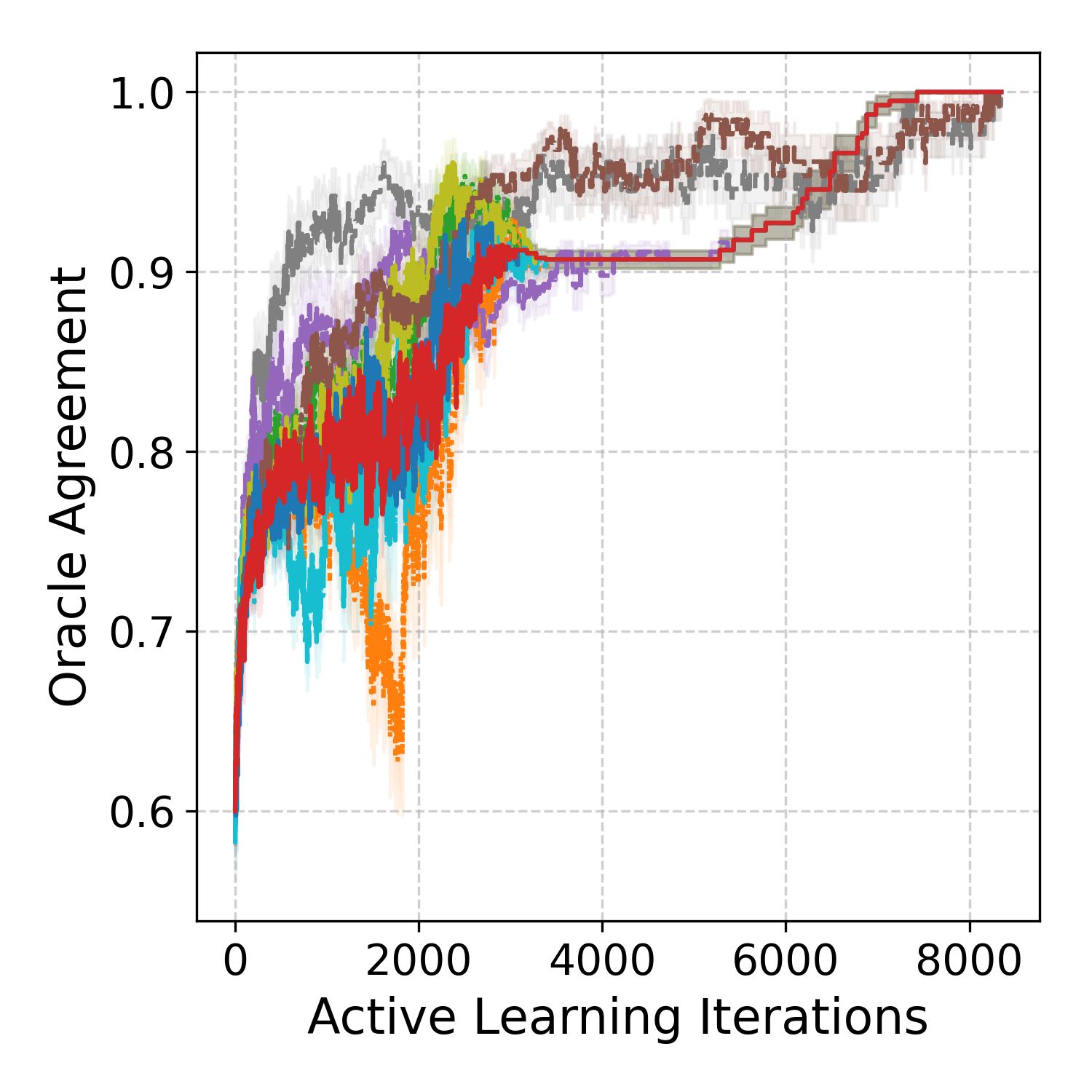}
        \caption{FICO}
    \end{subfigure}

    \vspace{0.2em}

    \begin{subfigure}[b]{0.19\textwidth}
        \includegraphics[width=\linewidth]{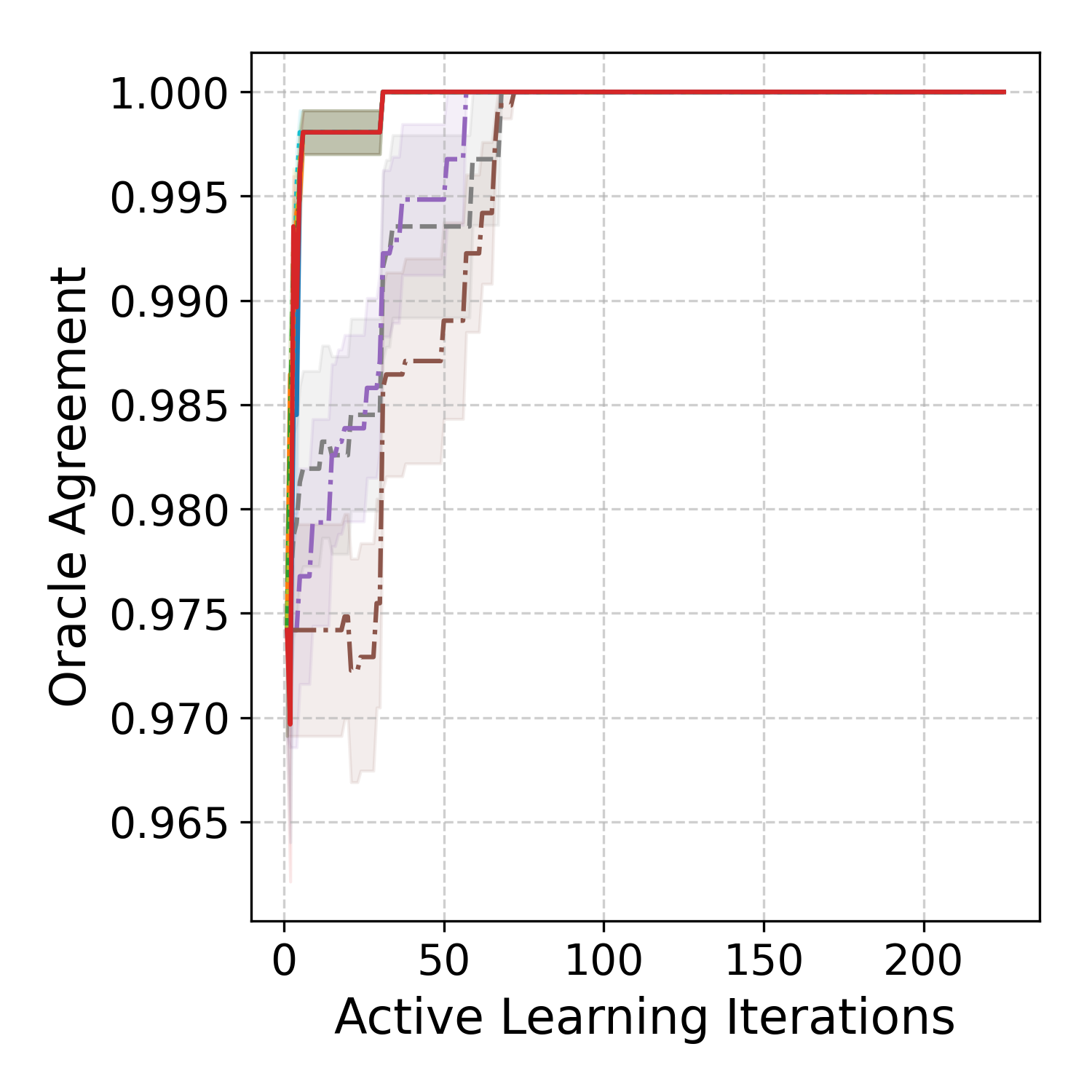}
        \caption{Haberman}
    \end{subfigure}
    \hfill
    \begin{subfigure}[b]{0.19\textwidth}
        \includegraphics[width=\linewidth]{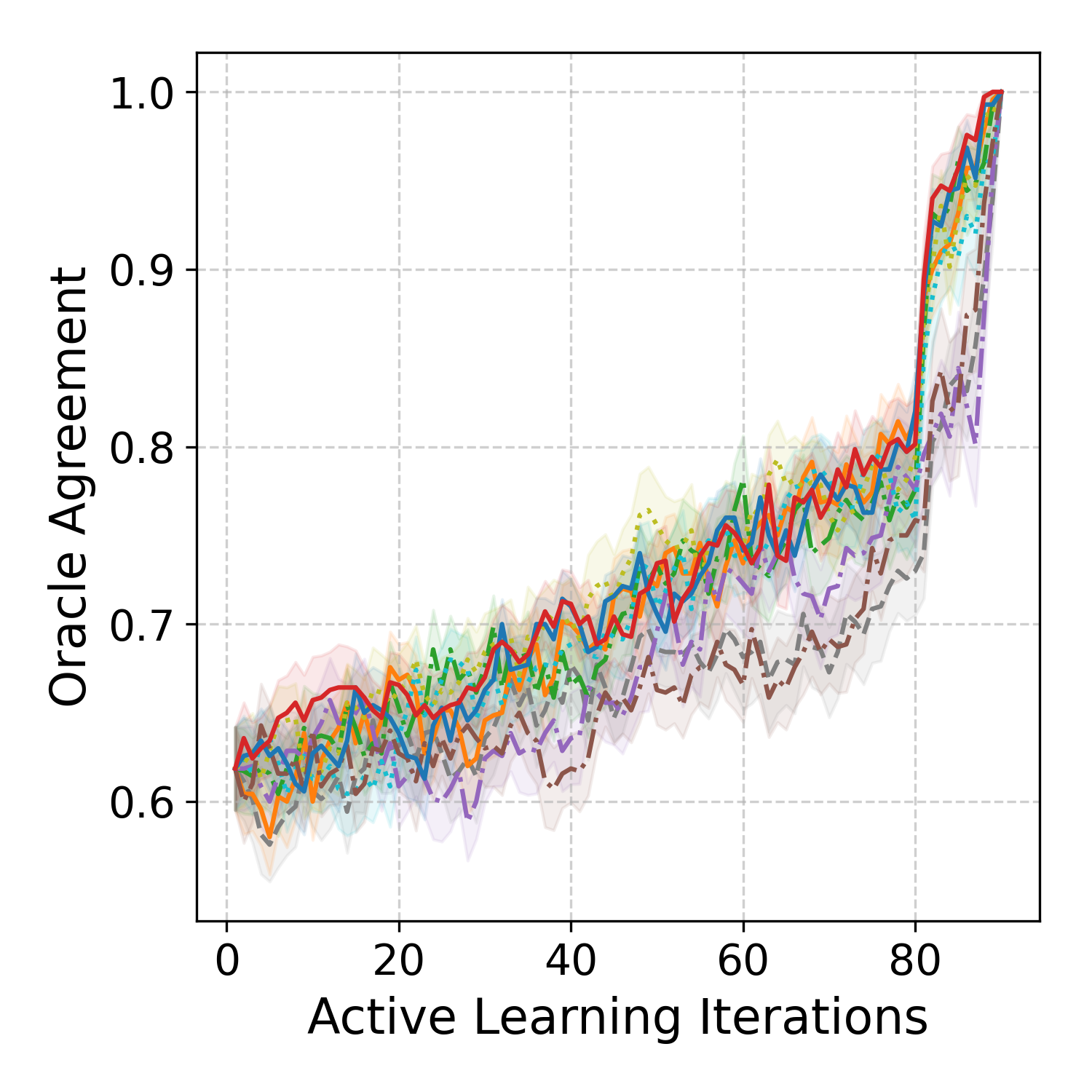}
        \caption{Hepatitis}
    \end{subfigure}
    \hfill
    \begin{subfigure}[b]{0.19\textwidth}
        \includegraphics[width=\linewidth]{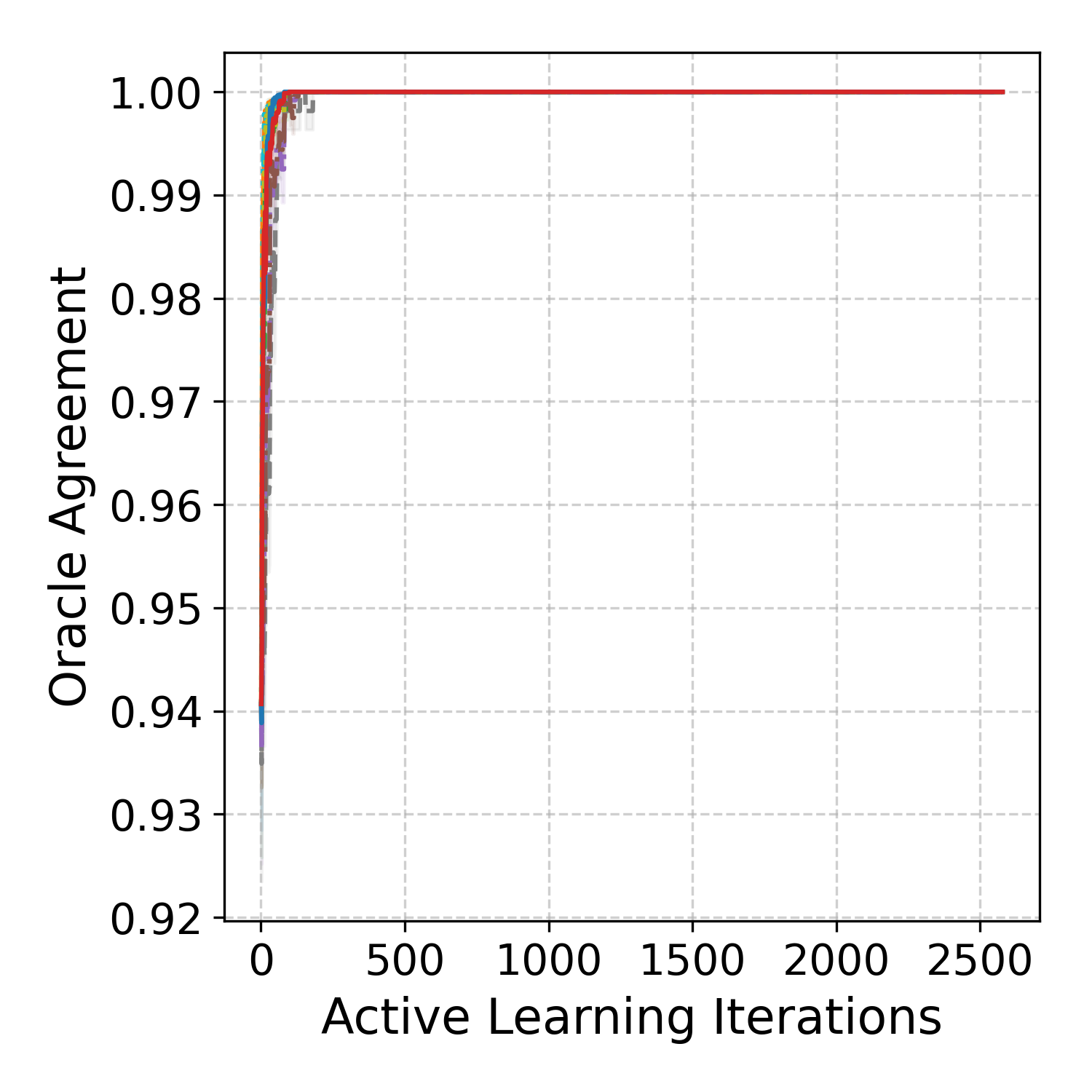}
        \caption{Hypothyroid}
    \end{subfigure}
    \hfill
    \begin{subfigure}[b]{0.19\textwidth}
        \includegraphics[width=\linewidth]{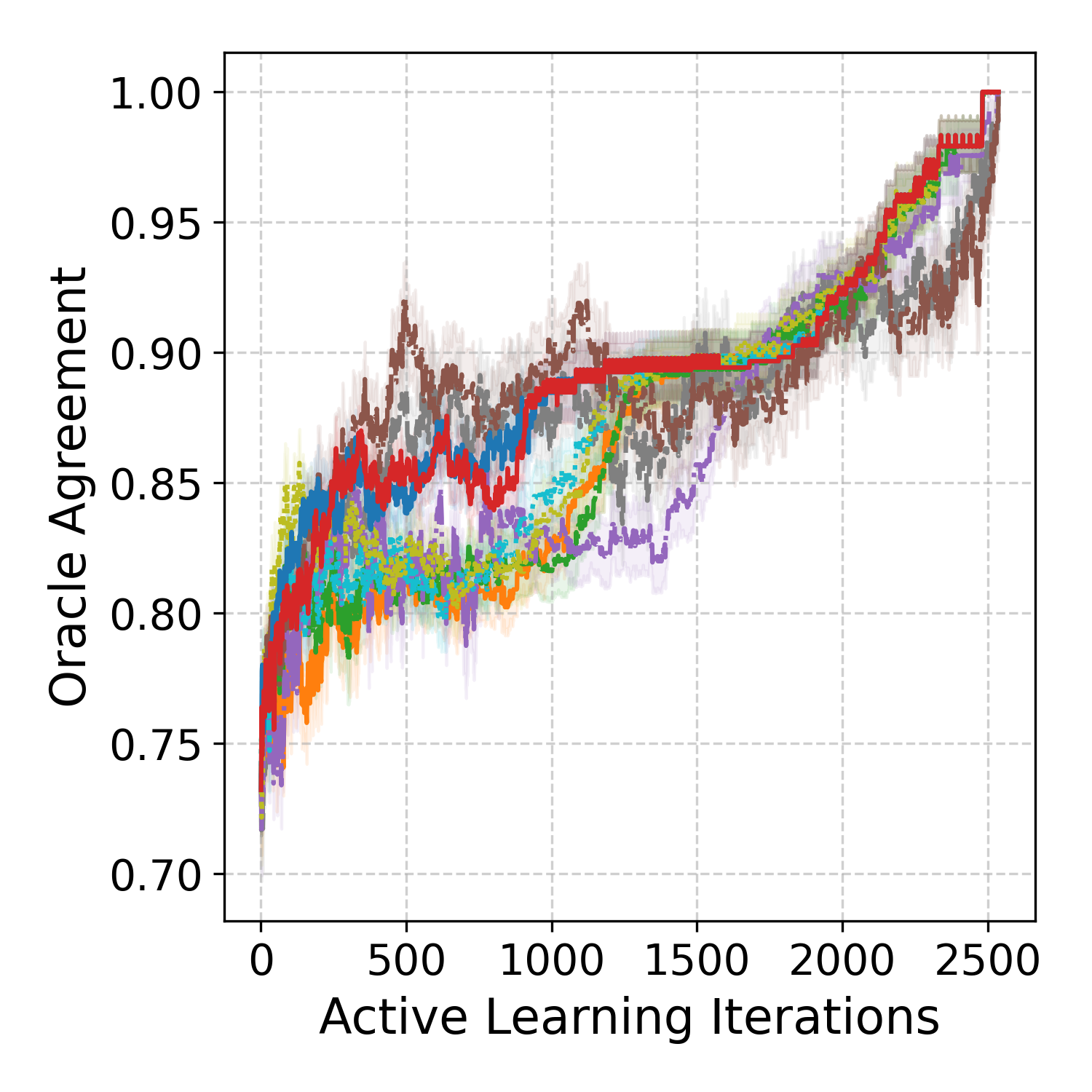}
        \caption{Kr-vs-Kp}
    \end{subfigure}
    \hfill
    \begin{subfigure}[b]{0.19\textwidth}
        \includegraphics[width=\linewidth]{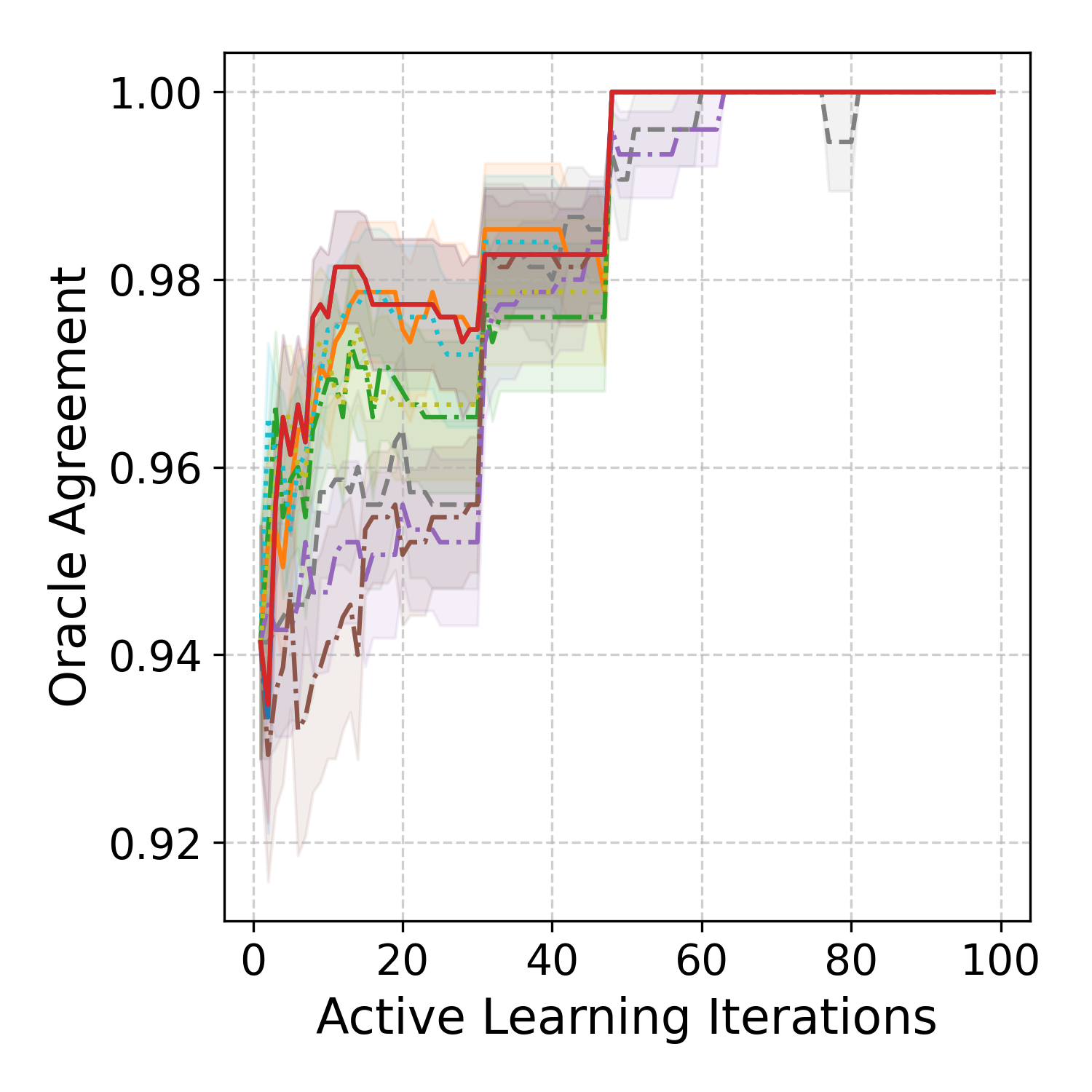}
        \caption{Lymphography}
    \end{subfigure}

    \vspace{0.2em}

    \begin{subfigure}[b]{0.19\textwidth}
        \includegraphics[width=\linewidth]{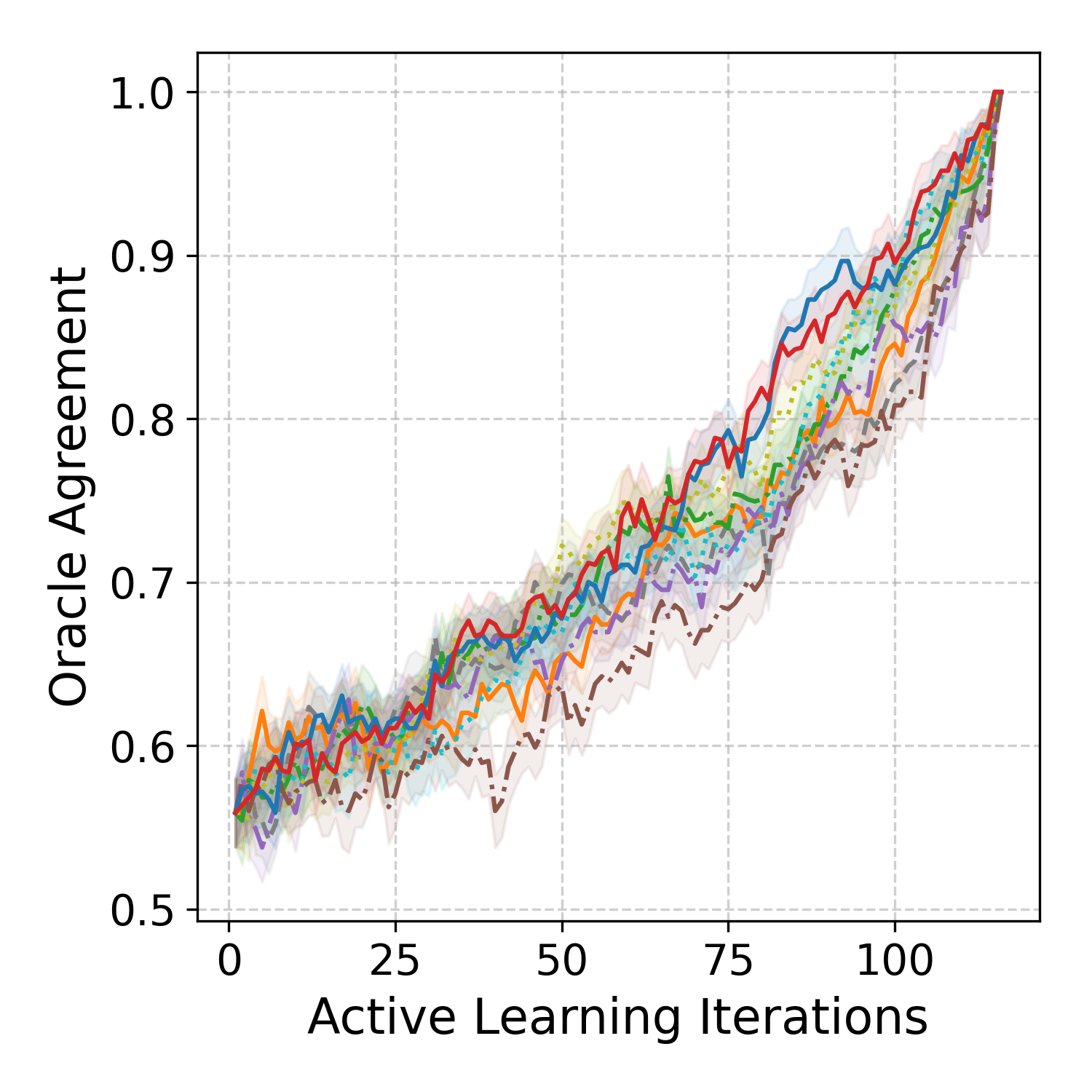}
        \caption{MONK-2}
    \end{subfigure}
    \hfill
    \begin{subfigure}[b]{0.19\textwidth}
        \includegraphics[width=\linewidth]{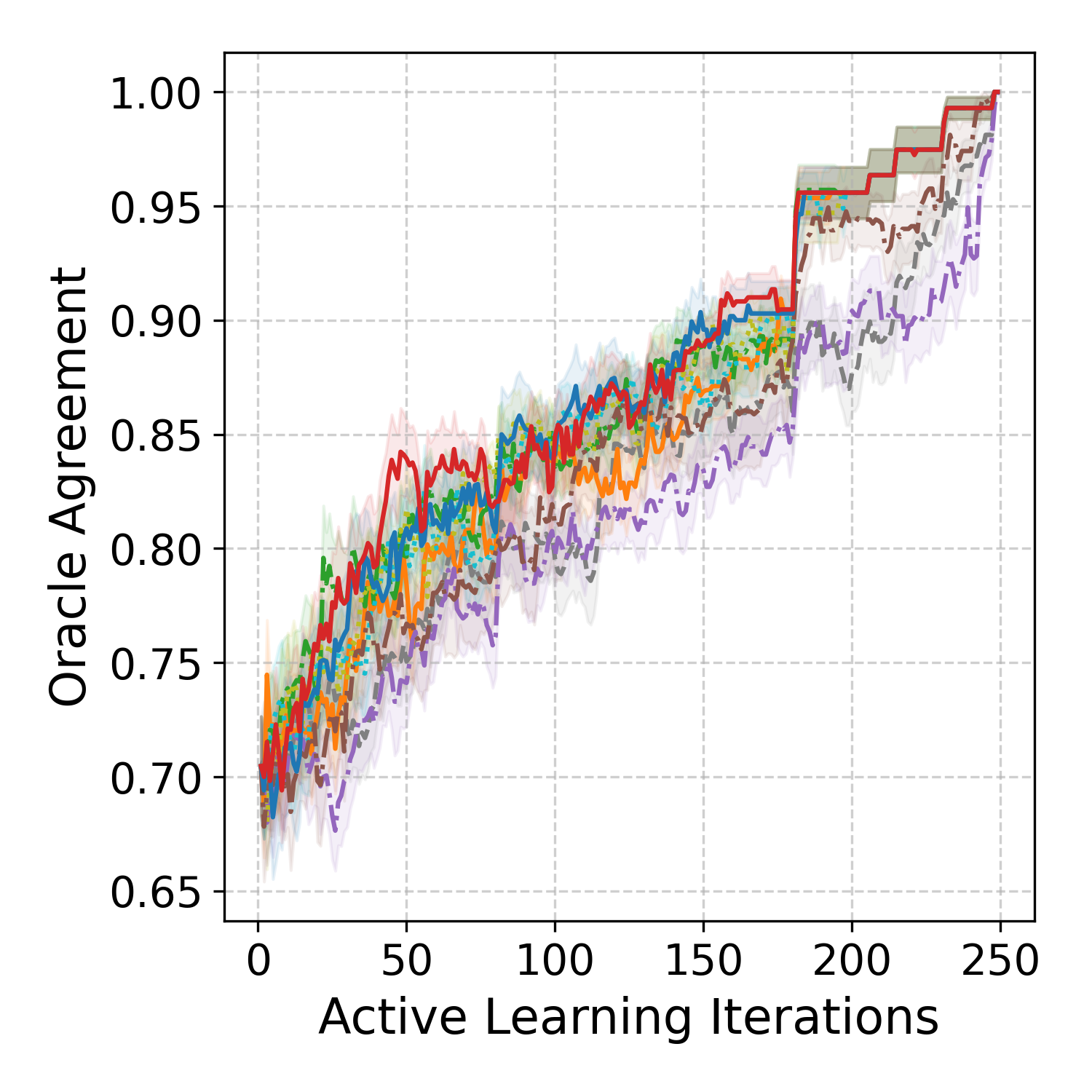}
        \caption{Primary Tumor}
    \end{subfigure}
    \hfill
    \begin{subfigure}[b]{0.19\textwidth}
        \includegraphics[width=\linewidth]{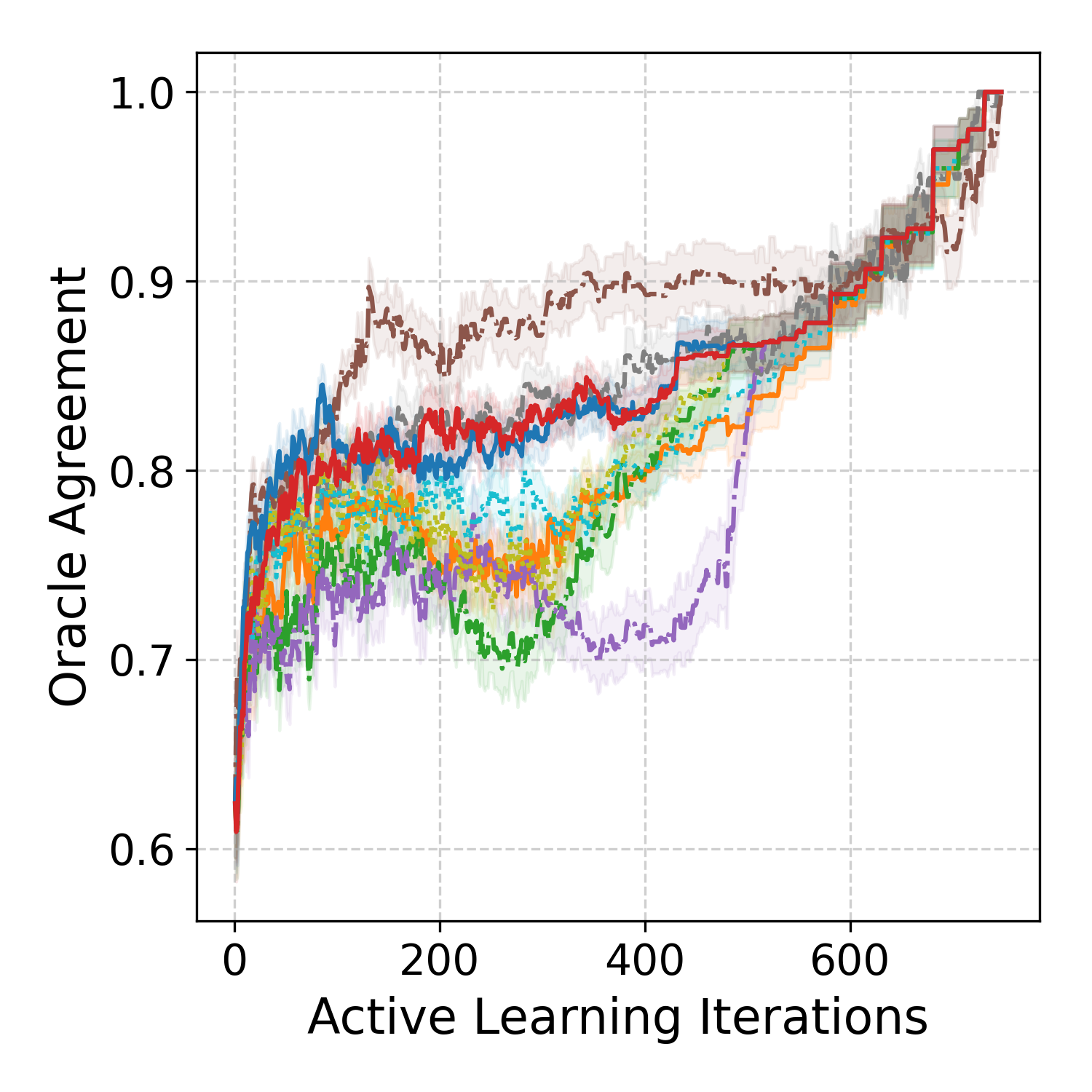}
        \caption{Tic-Tac-Toe}
    \end{subfigure}
    \hfill
    \begin{subfigure}[b]{0.19\textwidth}
        \includegraphics[width=\linewidth]{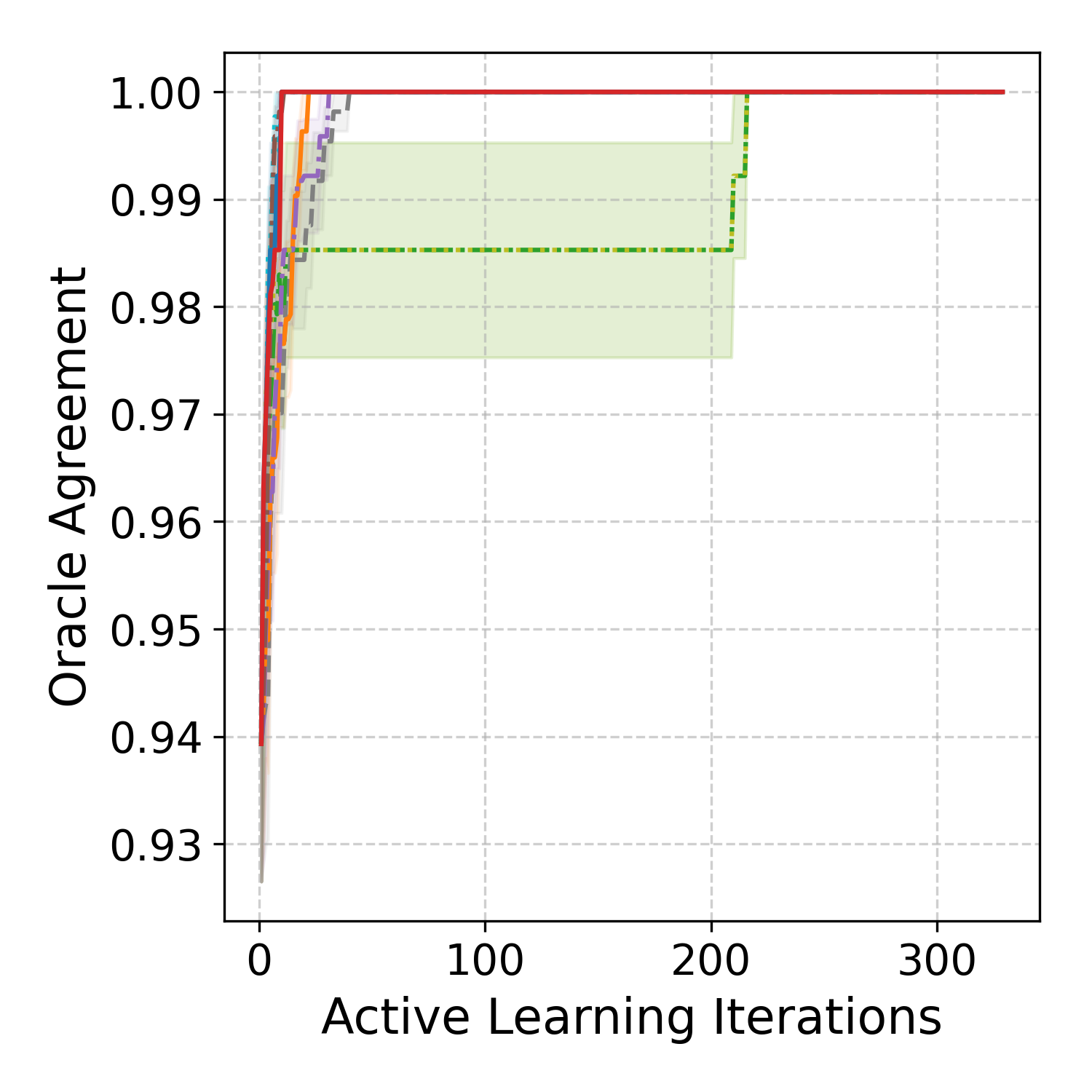}
        \caption{Vote}
    \end{subfigure}
    \hfill
    \begin{subfigure}[b]{0.19\textwidth}
        \includegraphics[width=\linewidth]{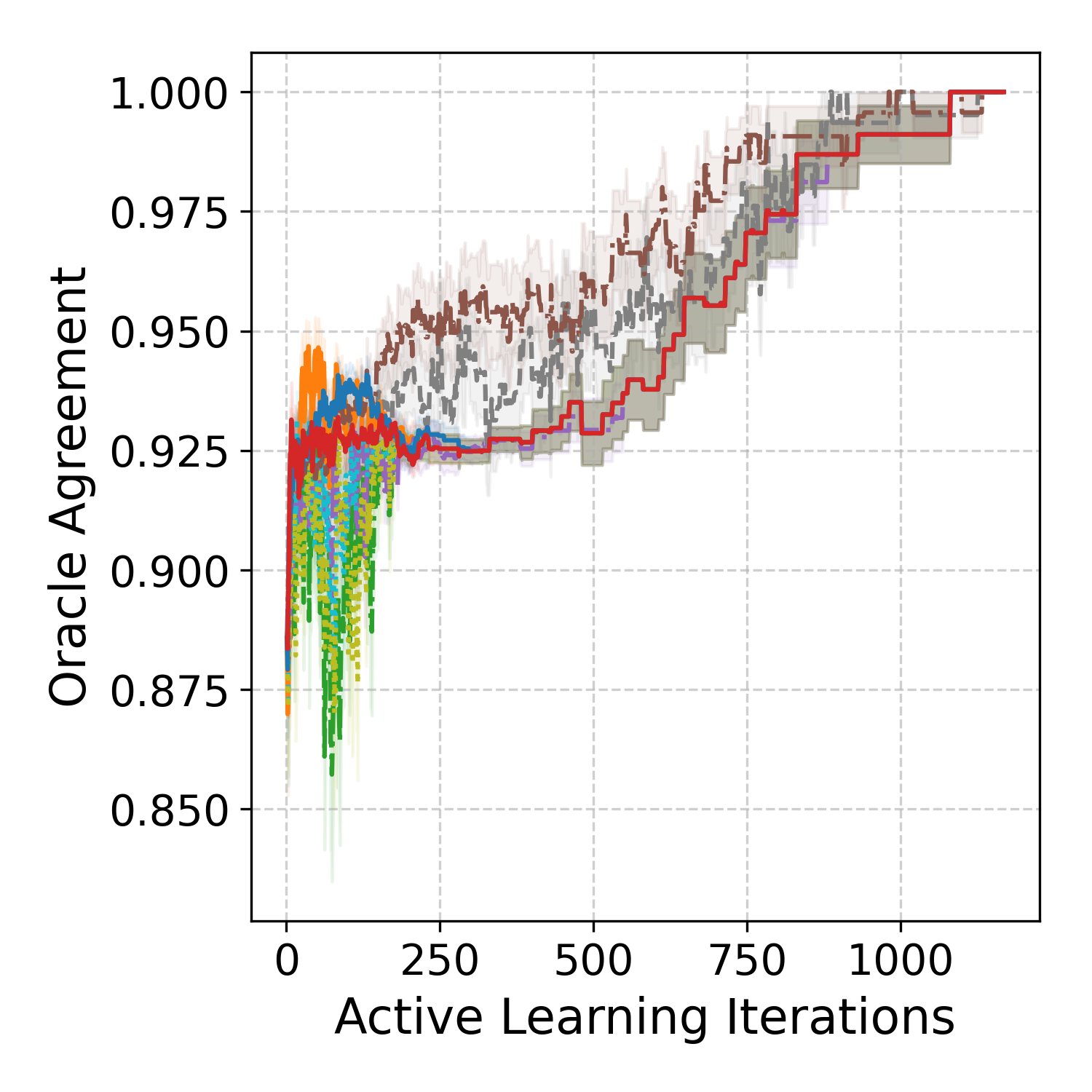}
        \caption{Yeast}
    \end{subfigure}

    \vspace{0.5em}
    \centering
    \includegraphics[width=0.60\linewidth]{upload_all_files/study1_main_AL_results/benchmark_legend.png}

    \caption{\textbf{Oracle Agreement Benchmarks.} Tracking consensus between the active learning committee and a ground-truth "Oracle" (optimal sparse tree fitted to the full dataset). In high-multiplicity environments, Weighted UNREAL achieves higher-fidelity alignment significantly faster than randomized baselines, demonstrating that its structural uncertainty signal successfully identifies the primary decision boundaries defining the task logic.}
    \label{fig:BenchmarkGrid_OracleAgreement}
\end{figure*}

\clearpage
\section{Evaluating REAL on High-Dimensional Sparse Parity Datasets}
\label{Appendix:Parity}
The $n$-bit parity problem serves as a critical benchmark for evaluating decision trees in high-dimensional sparse scenarios. We utilize a 3-bit parity core (where $Y=1$ if the sum of the first 3 features is even) and introduce an increasing number of meaningless Bernoulli noise covariates ($0, 6, 16, 26$). To simulate extreme small-sample constraints, we restrict the total pool to $N=100$ samples.

\begin{figure*}[ht]
    \centering
    \centerline{\small \textbf{3-bit Parity Scalability} ($N=100$, Increasing Noise Dimensions)}
    \vspace{0.2em}
    \begin{subfigure}[b]{0.24\textwidth}
        \centering
        \includegraphics[width=\linewidth]{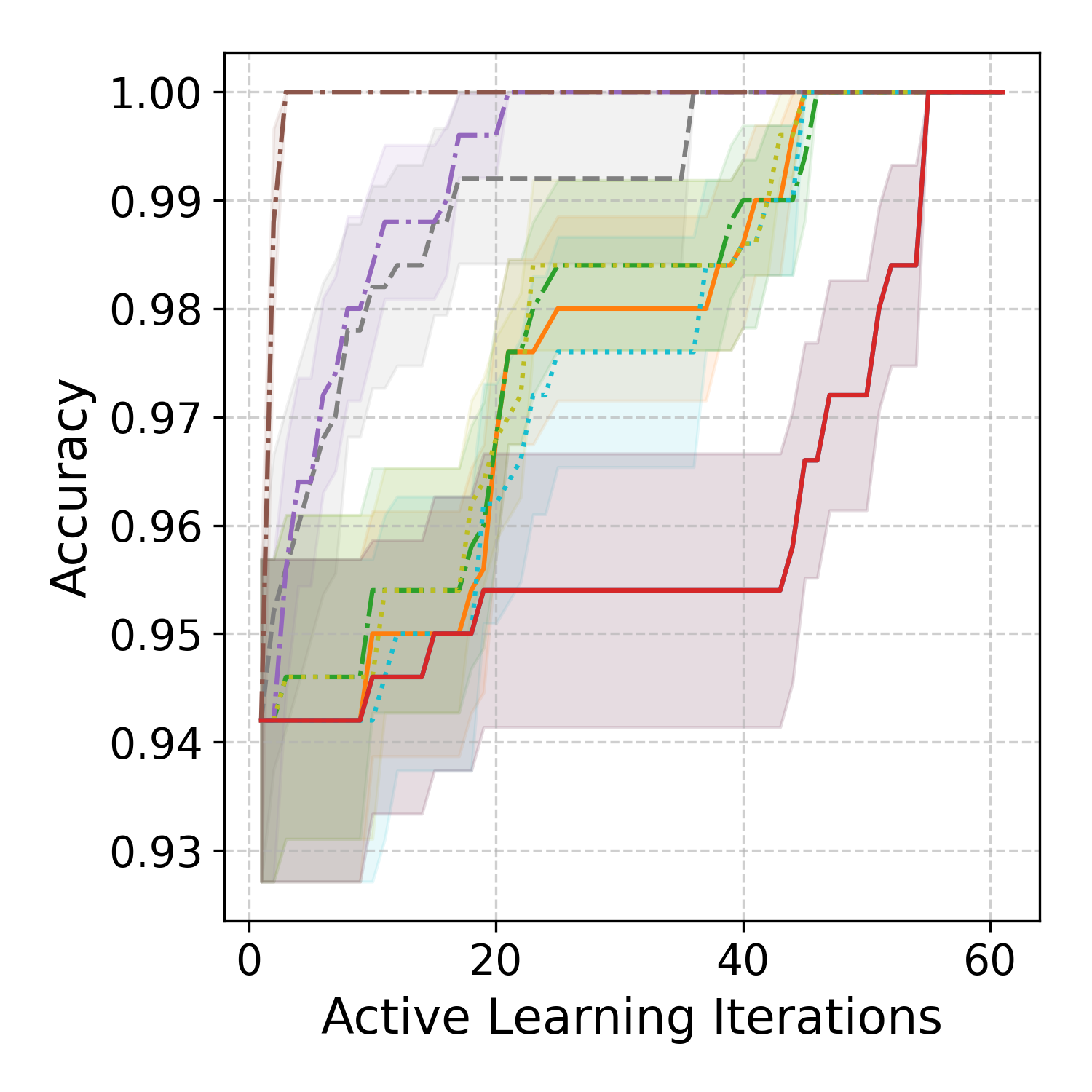}
        \caption{3 Dim (0 Noise)}
    \end{subfigure}
    \hfill
    \begin{subfigure}[b]{0.24\textwidth}
        \centering
        \includegraphics[width=\linewidth]{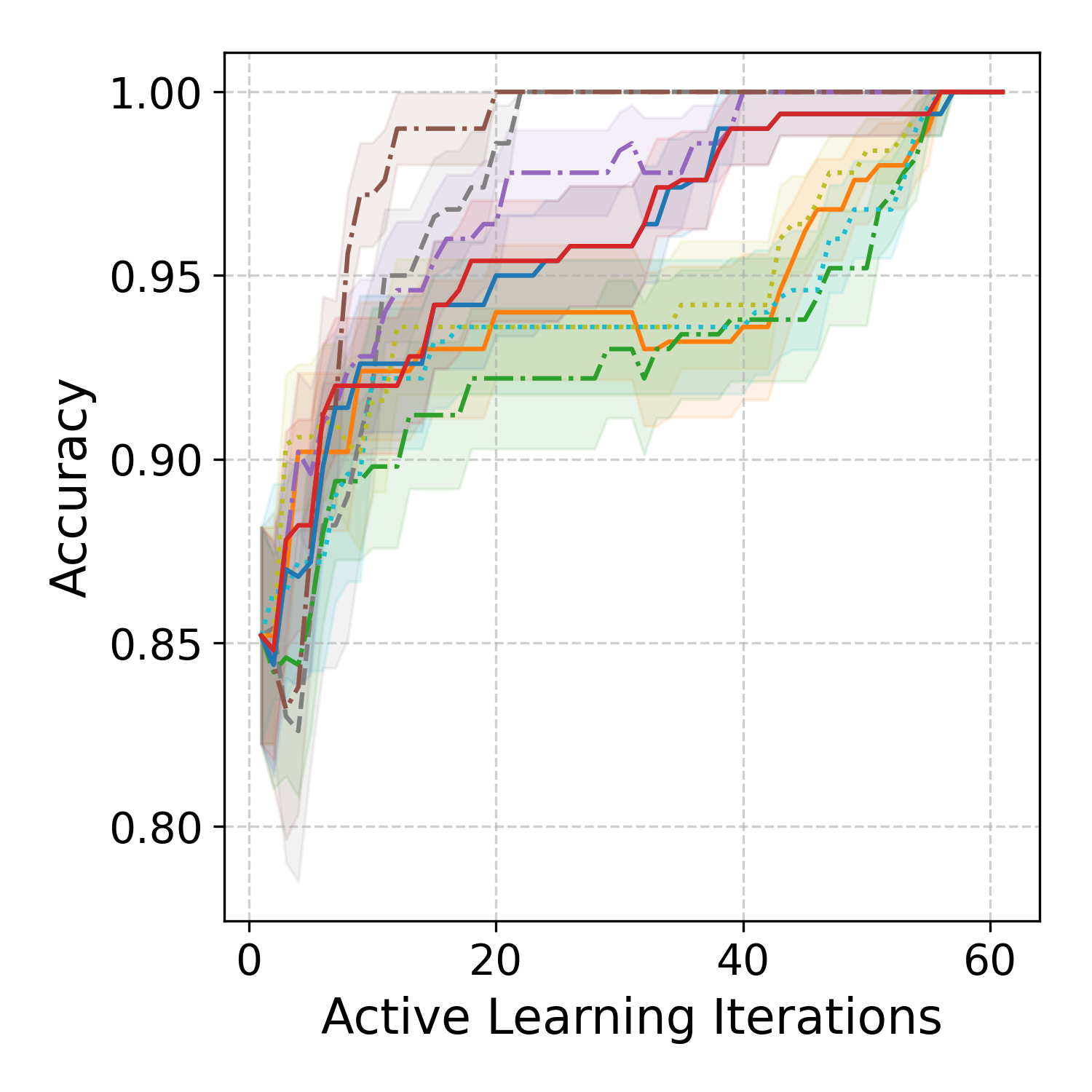}
        \caption{9 Dim (6 Noise)}
    \end{subfigure}
    \hfill
    \begin{subfigure}[b]{0.24\textwidth}
        \centering
        \includegraphics[width=\linewidth]{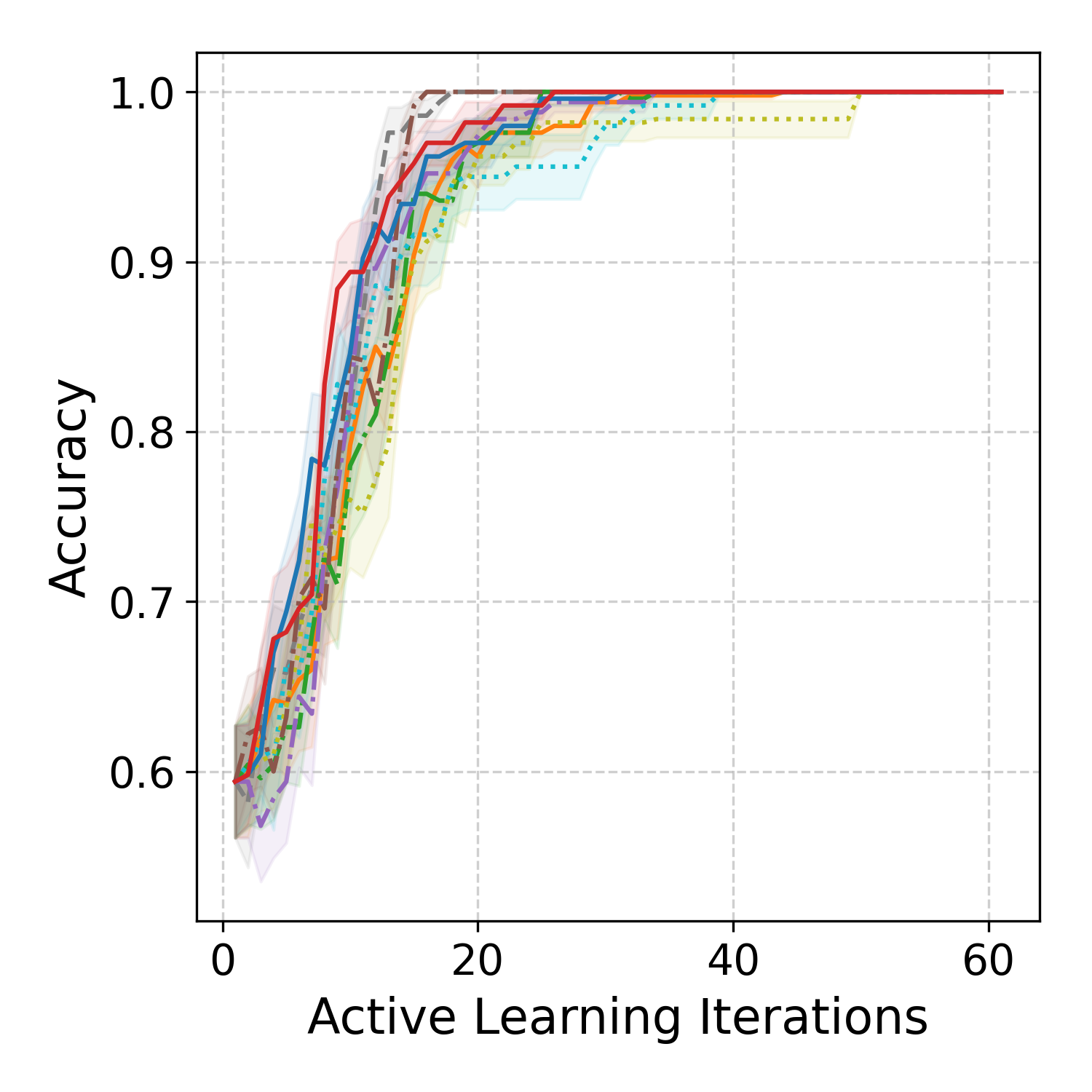}
        \caption{19 Dim (16 Noise)}
    \end{subfigure}
    \hfill
    \begin{subfigure}[b]{0.24\textwidth}
        \centering
        \includegraphics[width=\linewidth]{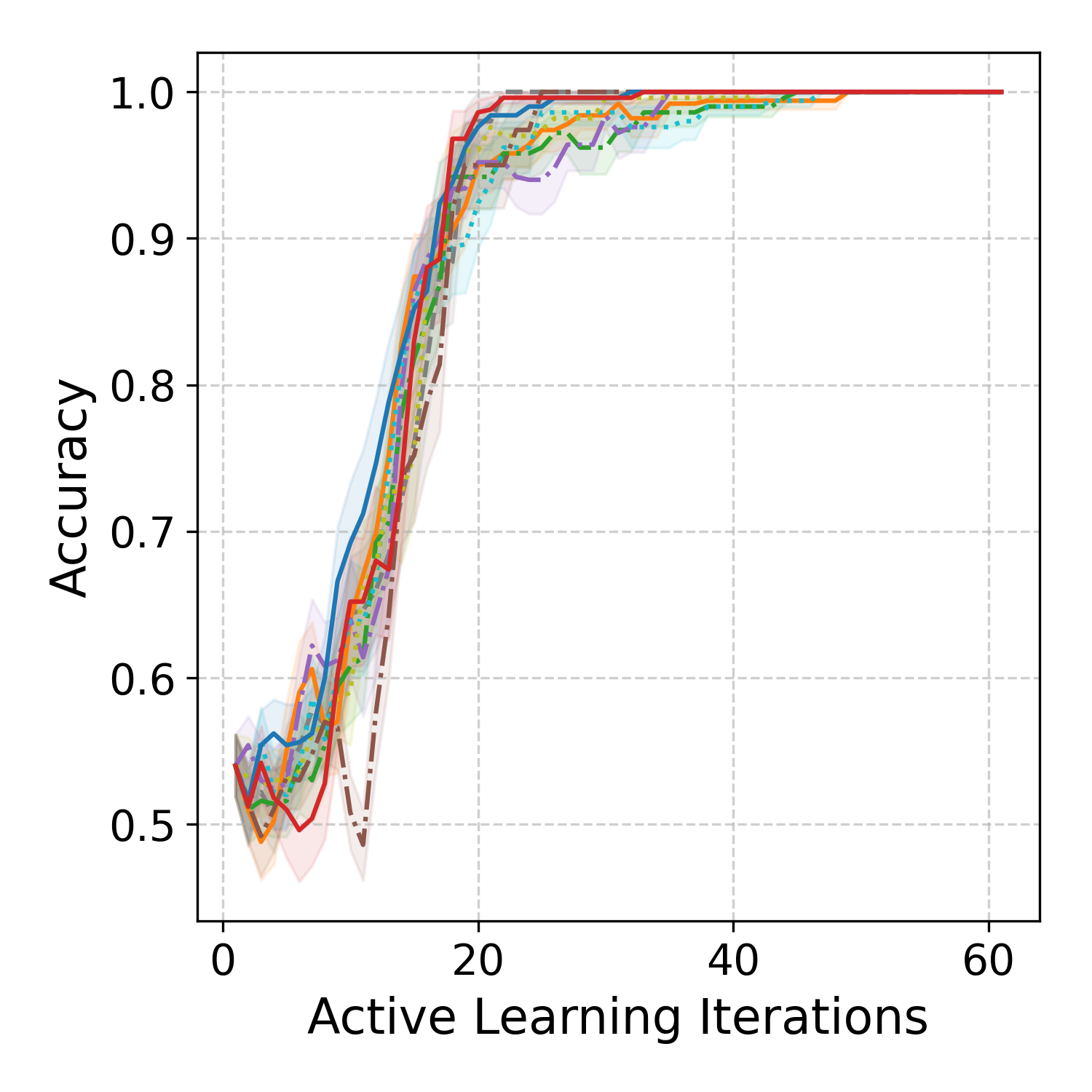}
        \caption{29 Dim (26 Noise)}
    \end{subfigure}
    
    \vspace{0.5em}
    \centering
    \includegraphics[width=0.60\linewidth]{upload_all_files/study1_main_AL_results/benchmark_legend.png}
    
    \caption{\textbf{Trace Plots on Parity Datasets.} In high-dimensional sparse regimes ($N=100$, up to 26 noise covariates), all tree-based active learning strategies perform below Passive Sampling as noise-induced patterns dominate. As we increase the number of noisy, useless covariates, QBC-RF becomes increasingly confused by stochastic variance, whereas REAL variants demonstrate superior recovery. By focusing its query strategy on the effective version space, REAL resolves the sparse 3-bit XOR signal more reliably than QBC-RF, eventually matching the accuracy of the other baselines and exceeding the performance of randomized ensembles.}    
    \label{fig:Parity_Scalability}
\end{figure*}

In these experiments, all tree-based active learning methods initially perform worse than fundamental active learning methods Coreset and Uncertainty Sampling and even Passive Sampling. With only 100 samples available and up to 29 total features, the version space is initially dominated by ``false'' structural patterns induced by noise covariates. This leads to greedy queries that provide less information than a uniform random search of the feature space.

In this low-sample regime, UNREAL and BREAL initially trail their QBC-RF counterparts. We attribute this to the stochastic buffer of randomized ensembles: feature subsetting and bootstrapping allow QBC-RF to smooth over geometric confusion by chance. In contrast, the precision of the REAL framework makes it hypersensitive to local misalignments before the version space has sufficiently collapsed.

As we increase the number of meaningless covariates, a secondary trend emerges: the exhaustive version-space characterization performed by REAL filters out aleatoric noise more effectively than randomized heuristics. While randomized ensembles (QBC-RF) become increasingly confused by the local patterns found in random feature subsets, REAL maintains a global perspective by explicitly identifying all models within the Rashomon set that are consistent with the data.

These Parity results coincide with our findings on model misspecification and label noise in the main manuscript. As the number of noise covariates increases, the Rashomon set expands to include a wider array of potential structural explanations—a phenomenon identical to the expansion seen under label noise ($\phi$). By focusing its query strategy on the intersection of these competing hypotheses, REAL  ensures that it only labels points that resolve true epistemic ambiguity. This allows the framework to navigate the expanded search space with a level of precision that stochastic approximations cannot match, eventually identifying the 3-bit XOR core once a critical mass of labels is acquired.

\end{document}